%% file: main.tex
\documentclass[10pt,twocolumn,letterpaper]{article}

\usepackage[cvprfinal]{cvpr}  
\usepackage{tabularx}

\usepackage{float}
\definecolor{cvprblue}{rgb}{0.21,0.49,0.74}
\usepackage[pagebackref,breaklinks,colorlinks,allcolors=cvprblue]{hyperref}
\usepackage{float}

\def\paperID{*****} 
\def\confName{CVPR}
\def\confYear{2025}
\usepackage{graphicx}
\usepackage{multirow}

\title{Controllable Shadow Generation with \\ Single-Step Diffusion Models from Synthetic Data}

\author{Onur Tasar\\
Jasper Research\\
{\tt\small onur.tasar@jasper.ai}
\and
Clément Chadebec\\
Jasper Research\\
{\tt\small clement.chadebec@jasper.ai}
\and
Benjamin Aubin\\
Jasper Research\\
{\tt\small benjamin.aubin@jasper.ai}
}

\begin{document}

\twocolumn[{
\maketitle
\begin{minipage}{\textwidth}
\begin{figure}[H]
     \centering
    \begin{subfigure}{0.42\textwidth}
    \centering
    \includegraphics[width=\textwidth]{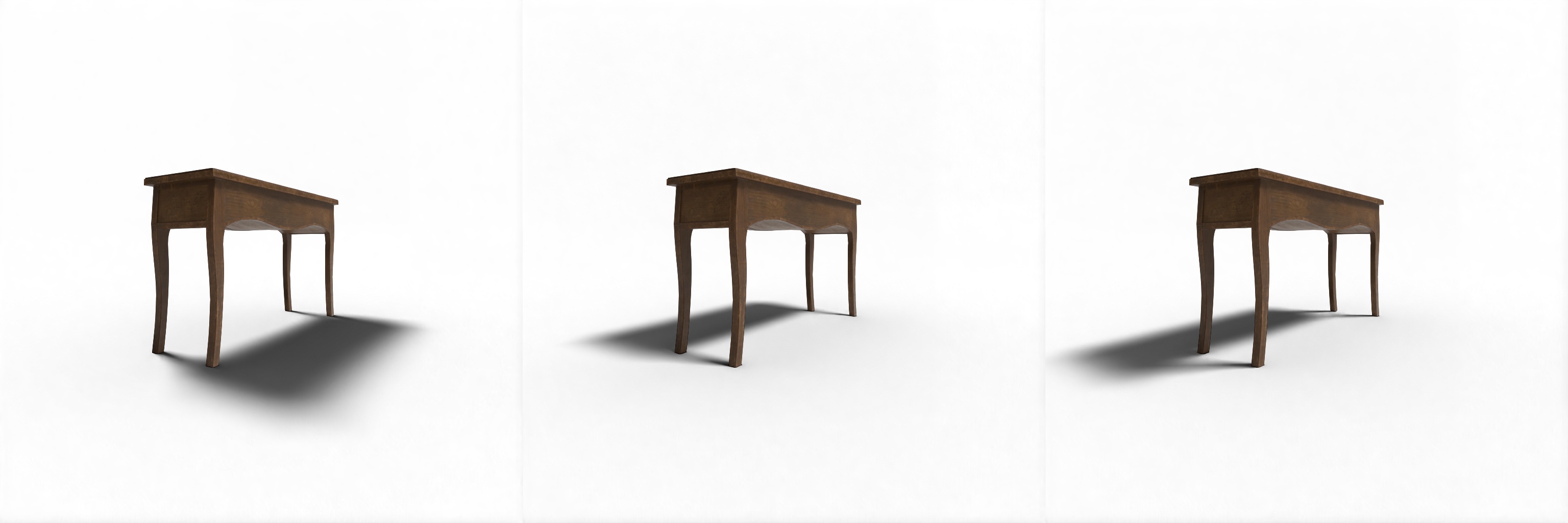}
    \caption{Direction control by moving the light source horizontally.}
    \end{subfigure}%
    \begin{subfigure}{0.42\textwidth}
     \centering
    \includegraphics[width=\textwidth]{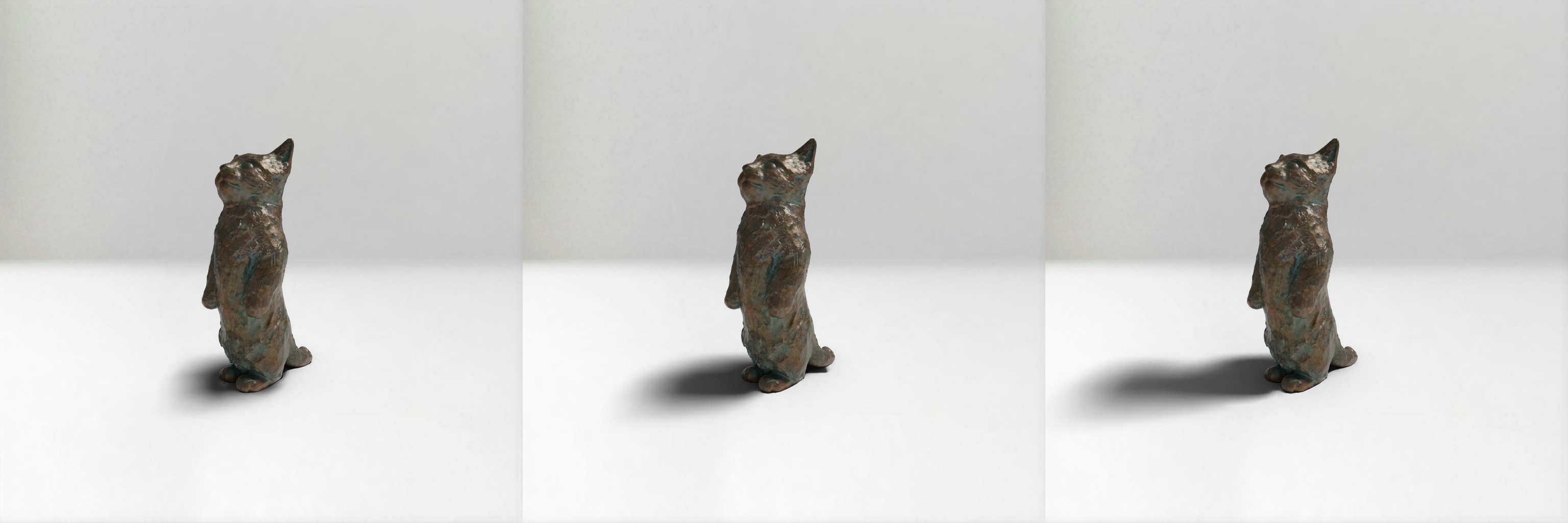}
    \caption{Direction control by moving the light source vertically.}
    \end{subfigure}

    \begin{subfigure}{0.42\textwidth}
     \centering
    \includegraphics[width=\textwidth]{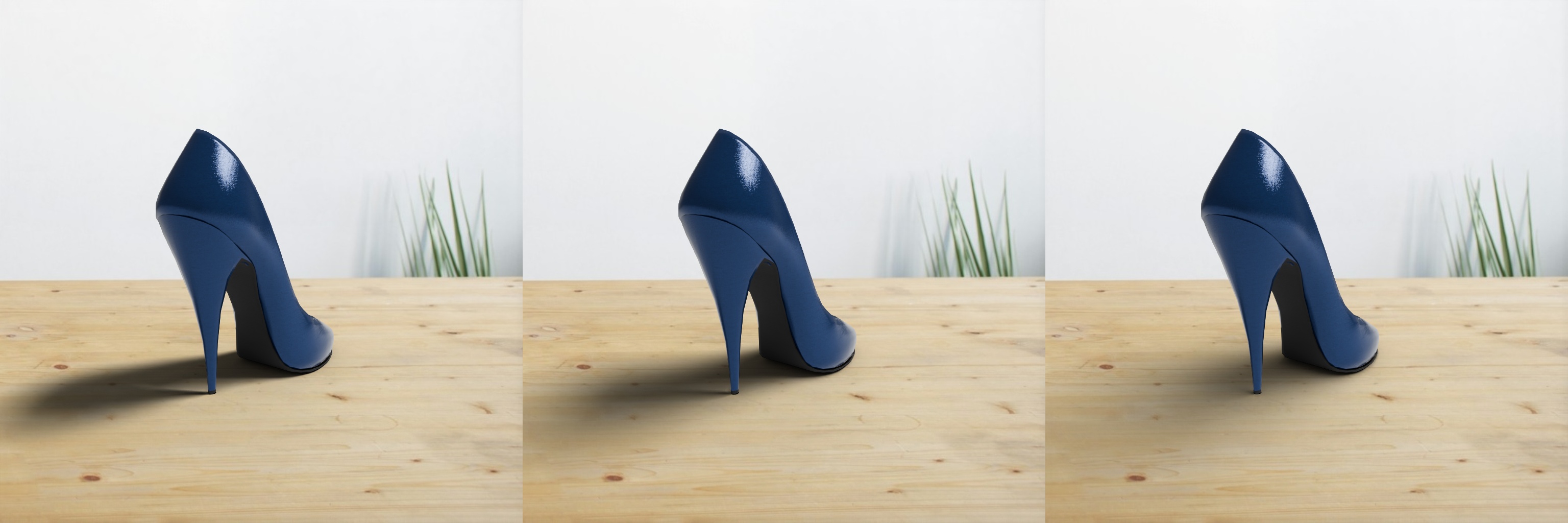}
    \caption{Softness control.}
    \end{subfigure}%
    \begin{subfigure}{0.42\textwidth}
     \centering
    \includegraphics[width=\textwidth]{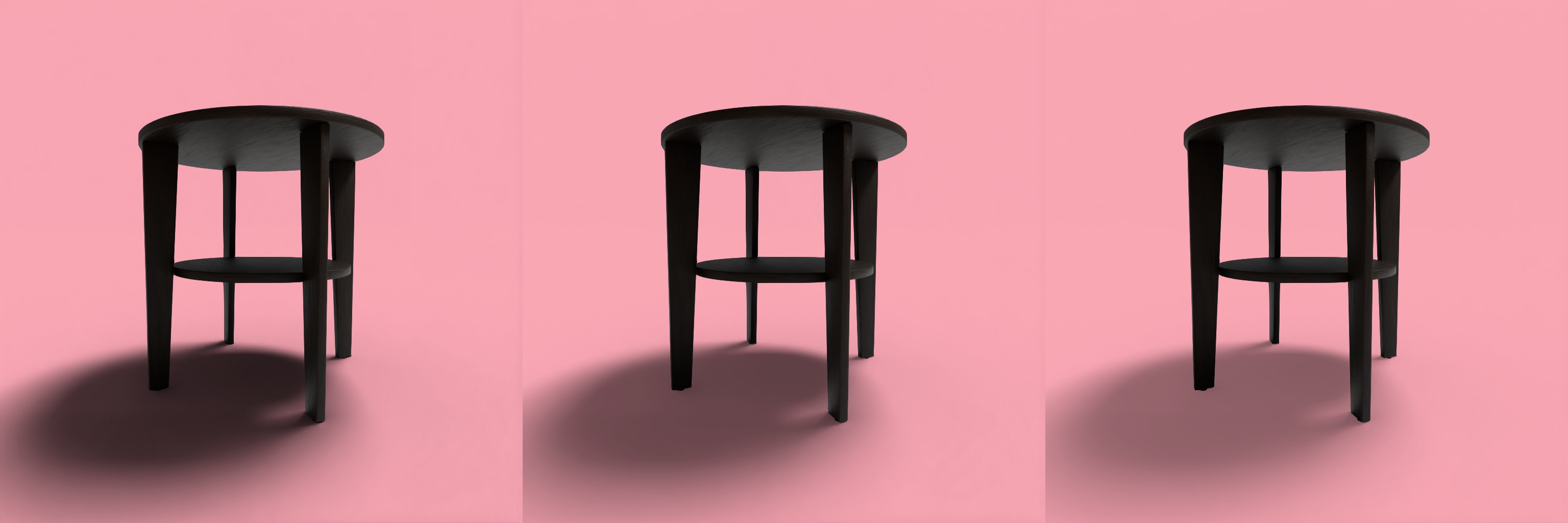}
    \caption{Intensity control.}
    \end{subfigure}
    \caption{Our single-step model enables the generation of realistic shadows with precise control over their direction, softness, and intensity.\vspace{2mm}}
    \label{fig:teaser}
\end{figure}
\end{minipage}
}]

\input{sections/abstract}

\input{sections/introduction}
\input{sections/related}

\input{sections/method}
\input{sections/experiments}
\input{sections/conclusion}

\bibliographystyle{ieeenat_fullname.bst}
\bibliography{main}

\input{sections/sm}

\end{document}

%% file: sections/abstract.tex
\begin{abstract}
Realistic shadow generation is a critical component for high-quality image compositing and visual effects, yet existing methods suffer from certain limitations: Physics-based approaches require a 3D scene geometry, which is often unavailable, while learning-based techniques struggle with control and visual artifacts. We introduce a novel method for fast, controllable, and background-free shadow generation for 2D object images. We create a large synthetic dataset using a 3D rendering engine to train a diffusion model for controllable shadow generation, generating shadow maps for diverse light source parameters. Through extensive ablation studies, we find that rectified flow objective achieves high-quality results with just a single sampling step enabling real-time applications. Furthermore, our experiments demonstrate that the model generalizes well to real-world images. To facilitate further research in evaluating quality and controllability in shadow generation, we release a new public benchmark containing a diverse set of object images and shadow maps in various settings. The project page is available at \href{https://gojasper.github.io/controllable-shadow-generation-project/}{this link}. 
\end{abstract}

%% file: sections/introduction.tex
\section{Introduction}
\label{sec:intro}

Generating shadows for object images is crucial for visually appealing content creation with a wide range of real-world applications such as product photography, packshots, and online marketing~\cite{product_photography}. A common approach for addressing this task entails constructing a 3D scene geometry based on an object image and configuring a light source to render the desired shadows by employing advanced rendering algorithms such as ray tracing~\cite{ray_tracing}. The cumbersome process of first creating a 3D scene and then executing a rendering algorithm is time-consuming, making this approach impractical for real-world applications. An alternative approach is to operate directly on images to predict shadows from the provided input. We focus on this strategy as it bypasses the lengthy 3D construction and rendering steps.

The main challenge in training a model that generates shadows from images is collecting a dataset suitable for this task. Requesting professional annotators to manually create shadows for object images is excessively labor-intensive and costly. It is also challenging for them to create geometrically and physically correct shadows, as these shadows are not generated by an actual light source. An even more complex challenge is to give a model control over specific attributes, such as direction, softness, and intensity of the generated shadows, as illustrated in Fig.~\ref{fig:teaser}, to accommodate a wide range of applications. There has been significant progress in realistic image generation, especially using diffusion models with various conditionings such as text~\cite{sdxl, latent_diffusion, playground_v25}, depth maps~\cite{t2i_adapter, controlnet, controlnet_plus_plus}, segmentation masks~\cite{controlnet, controlnet_plus_plus} etc. However, to the best of our knowledge, there is no prior work on conditioning diffusion models for controllable shadow generation based on the aforementioned specified light source attributes.

Over time, rendering engines have achieved a great level of realism that makes their generated images nearly indistinguishable from actual photos. In addition to rendering high-quality images, they are also able to create corresponding accurate pixel-wise annotations~\cite{blenderproc} such as semantic segmentation maps, depth maps, surface normals, and shadow maps with various light source settings, thereby automating the excessively time-consuming and costly data annotation process without the need for external annotators. Although rendering engines are excellent tools for generating synthetic annotated data, their use and generalization on real-world images have largely been under-explored in the research community, especially for shadow generation.

Given these challenges and limitations, we present a new fast and controllable shadow generation pipeline that is independent of the background image. To this end, we first generate a large synthetic dataset that includes high-quality images and shadows with various directions and softness by positioning a light source and moving across the surface of a sphere. We then train a single-step diffusion model conditioned on the spherical coordinates and other parameters of the light source, predicting controllable shadows as gray-scale images. We finally blend the object image and the predicted shadows into given target background images.

The main contributions of the paper are as follows:
\begin{itemize}
    \item Presenting a new shadow generation pipeline for object images that is robust to varying backgrounds.
    \item Creating a large synthetic dataset from a diverse collection of 3D meshes and showing that our model trained on fully synthetic data generalizes effectively to real images.
    \item Training a one-step diffusion model by proposing a simple yet effective conditioning mechanism to inject light source parameters, such as the spherical coordinates and size, allowing the control of shadow properties.
    \item Performing extensive ablation studies on several diffusion prediction types, number of sampling steps, number of training iterations, and various conditioning mechanisms.
    \item Publicly releasing three curated public test sets of a diverse distribution of objects to evaluate the performance of future works in the community for shadow shape consistency, shadow direction and softness control.
\end{itemize}

%% file: sections/related.tex
\section{Related Works}
Diffusion models~\cite{sohl2015deep,song2020score,ddpm}, have been demonstrated to represent the current state-of-the-art in generative modeling for image synthesis ~\cite{latent_diffusion, sdxl, playground_v25, sd3}. Using diffusion models, various dense prediction tasks like surface normal prediction~\cite{stablenormal,controlnet}, depth map estimation~\cite{depthgen, marigold,lotus,controlnet}, and image matting~\cite{matting_by_generation, diffmate} have also been explored. One major limitation of diffusion models for real-world applications is their slow inference time due to the iterative de-noising sampling process~\cite{ddpm}. There have been attempts to achieve one or a few steps of sampling by using latent consistency models~\cite{lcm},  distillation~\cite{hinton2015distilling,meng2023distillation,liu2023instaflow,xu2023ufogen,sdxl_turbo,flash_diffusion} and flow matching~\cite{lipmanflow}. The authors of lotus~\cite{lotus} proposed a single-step dense prediction approach based on a pre-trained diffusion model. However, fast and controllable shadow generation with diffusion models has remained unexplored.

Specifically for controllable shadow generation, non-diffusion-based methods have been proposed. For instance, SSN~\cite{ssn} presents a framework with two U-Nets \cite{ronneberger2015u}. The former predicts an ambient occlusion map and the latter outputs the final shadows. The main limitation of this work is that the performance of the first U-Net strongly depends on the view angle of the object. To mitigate this issue, this research has then been extended by replacing the first U-Net by pixel-high map estimation~\cite{ssn_extension}. Another extension of this work is PixHt-Lab~\cite{pixht_lab}, which also predicts reflections. However, this approach requires several intermediate steps—such as predicting surface normals and a depth map, followed by rendering—which makes the process lengthy.

There exists a line of research focusing on generating shadows for an object by analyzing the background scene, and the direction and intensity of the shadows of other objects within the image. This approach is specifically investigated using generative adversarial networks~\cite{gans} in Desoba~\cite{desoba} and ShadowGAN~\cite{shadowgan}, and using ControlNets~\cite{controlnet} in Desobav2~\cite{desobav2}. ObjectDrop~\cite{objectdrop} blends a foreground object into a background image by generating shadows and reflections consistent with the background scene using diffusion models. In addition to generating reflections and shadows, ObjectStitch~\cite{objectstitch} also changes the geometry and color of the foreground object by merging it with a background image. However, these approaches lack flexibility in controlling shadow direction, softness, and intensity.

Realistic and controllable shadow generation can also be achieved by performing an image-to-3D mesh generation followed by rendering shadows using the mesh~\cite{shadow_3d}. Although recent research has achieved remarkable performance on image-to-3D model tasks using neural radiance fields~\cite{nerf, latent_nerf, neurallift, nerdi}, diffusion models~\cite{zero1to3, one2345}, and Gaussian Splatting~\cite{gaussian_splatting}, they are not well suited towards the shadow generation for single object image task, since they either require multiple views of the object for 3D mesh generation or have long processing times.

%% file: sections/method.tex
\newcommand{\normal}{\mathcal{N}\left(0, \mathbf{I}\right)}

\section{Method}
\label{sec:method}

Creating a large, high-quality synthetic dataset and training a model on it for fast inference and precise control over specific shadow properties are two crucial components of our pipeline.
We start by providing a comprehensive overview of our synthetic dataset creation in Sec.~\ref{sec:method:subsec:dataset}, and we present our shadow generation training pipeline in Sec.~\ref{sec:method:subsec:shadow_generation_pipeline}.

\begin{figure}[t]
    \centering
    \begin{subfigure}[b]{0.244\linewidth}
        \centering
        \includegraphics[width=\linewidth]{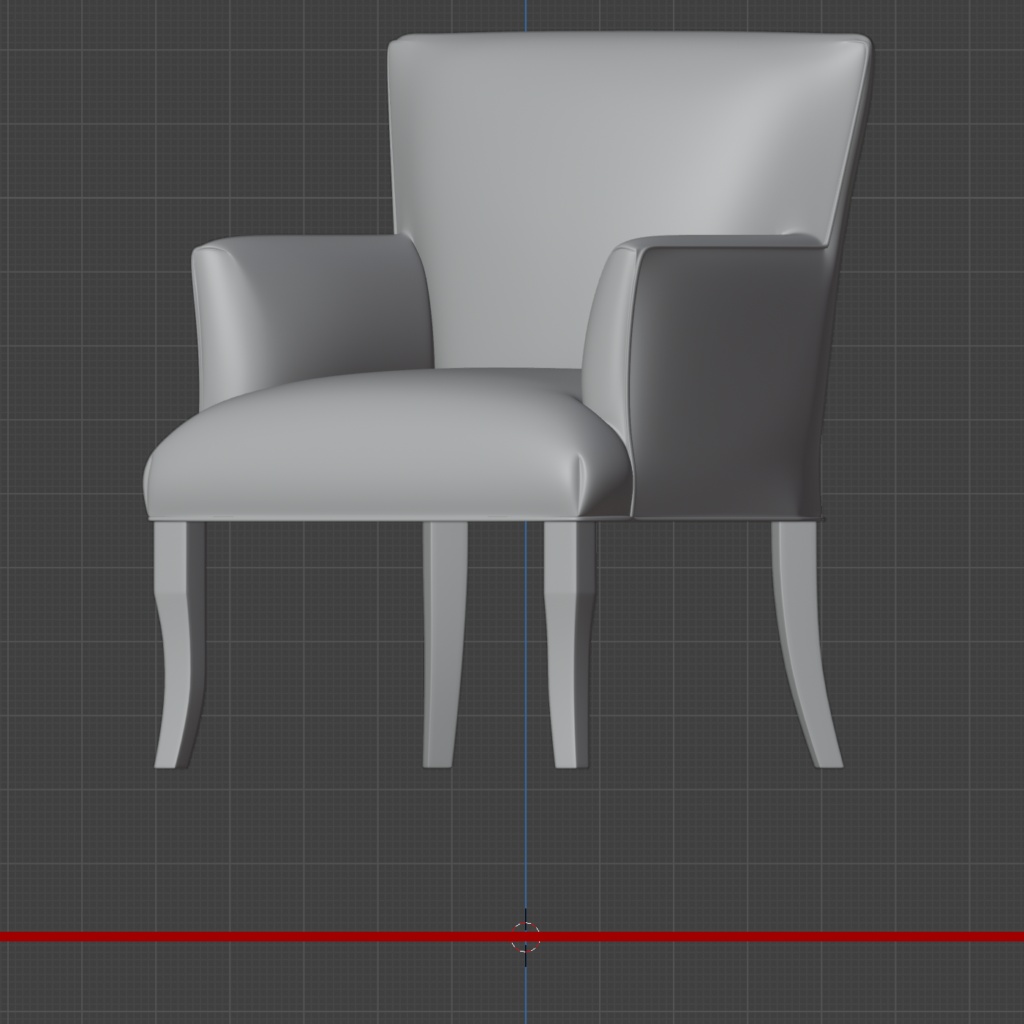}
    \end{subfigure}
    \hfill
    \centering
    \begin{subfigure}[b]{0.24\linewidth}
        \centering
        \includegraphics[width=\linewidth]{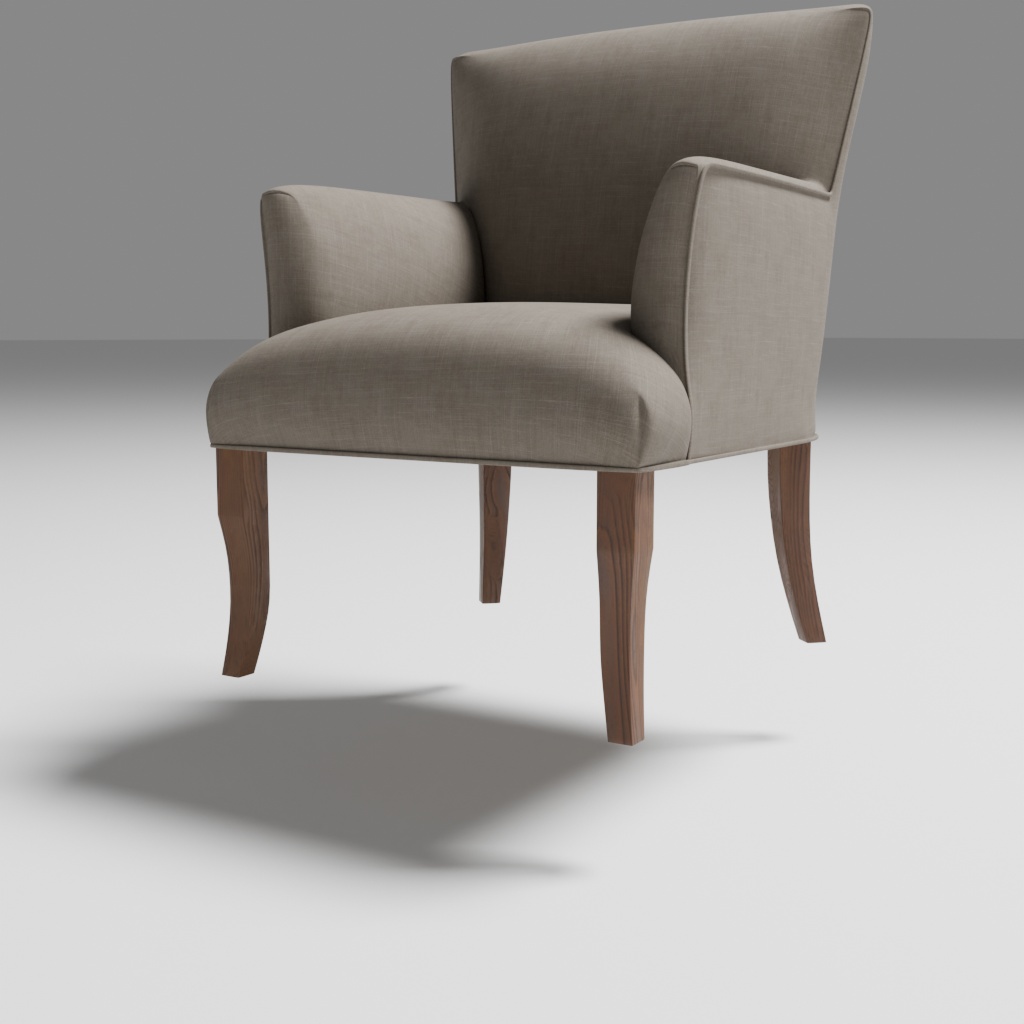}
    \end{subfigure}
    \hfill
    \begin{subfigure}[b]{0.24\linewidth}
        \centering
        \includegraphics[width=\linewidth]{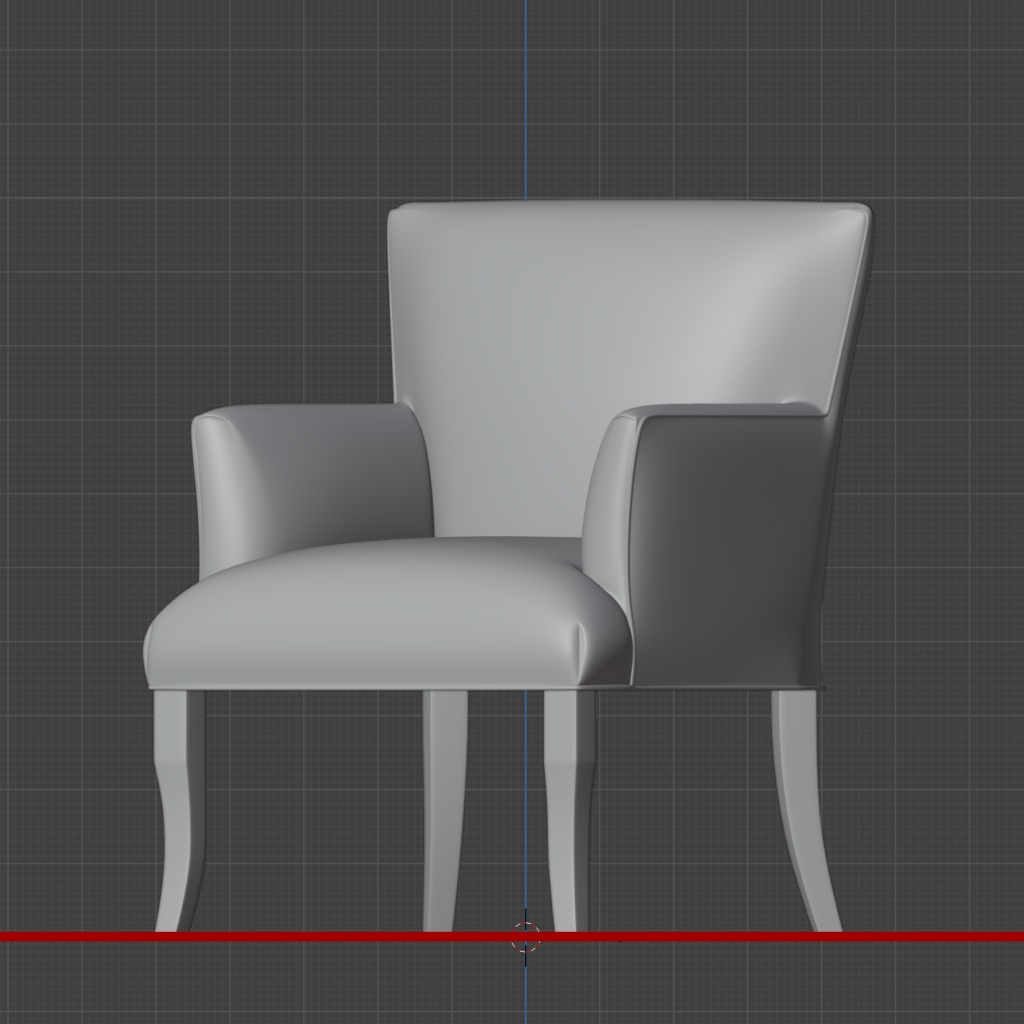}
    \end{subfigure}
    \hfill
    \begin{subfigure}[b]{0.24\linewidth}
        \centering
        \includegraphics[width=\linewidth]{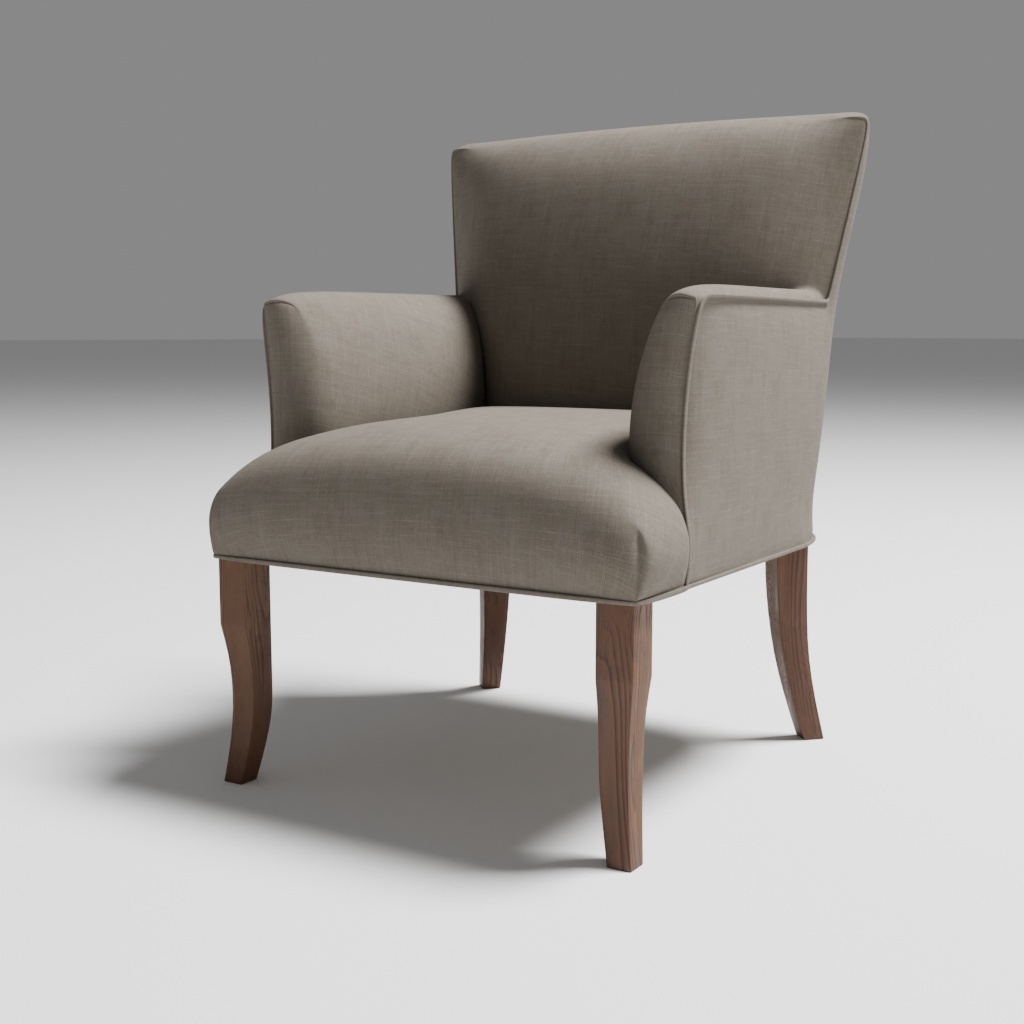}
    \end{subfigure}
    \caption{Example renders with unprocessed (first two) and processed (last two) meshes. The red line represents the ground.}
    \label{fig:flying_objects}
\end{figure}

\begin{figure}[t]
    \centering
    \begin{subfigure}[b]{\linewidth}
        \centering
        \includegraphics[width=\linewidth]{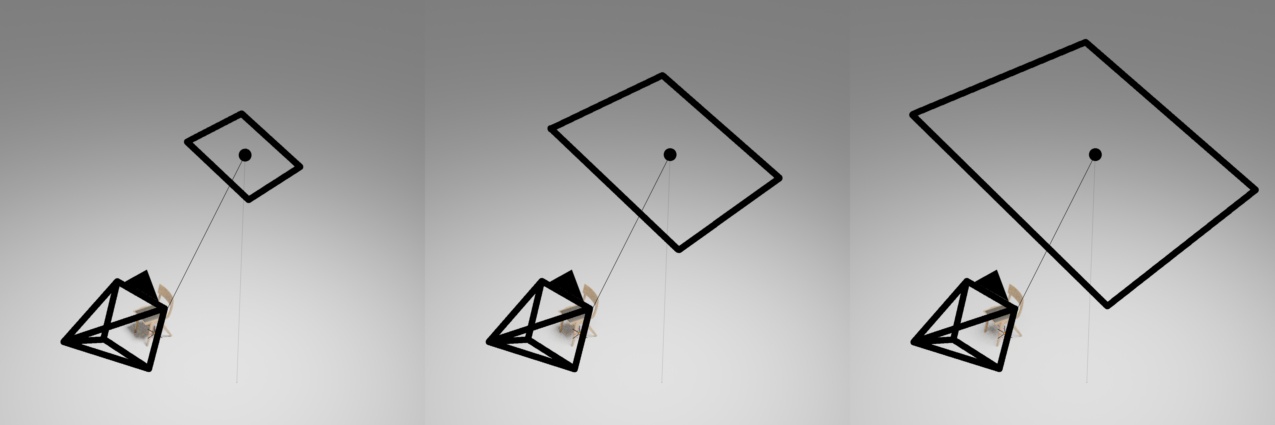}
        \caption{Scenes with varying area light sizes. The pyramid represents the camera. The area light is indicated by the square with a circle in the middle.}
    \end{subfigure}
    \hfill
    \centering
    \begin{subfigure}[b]{\linewidth}
        \centering
        \includegraphics[width=\linewidth]{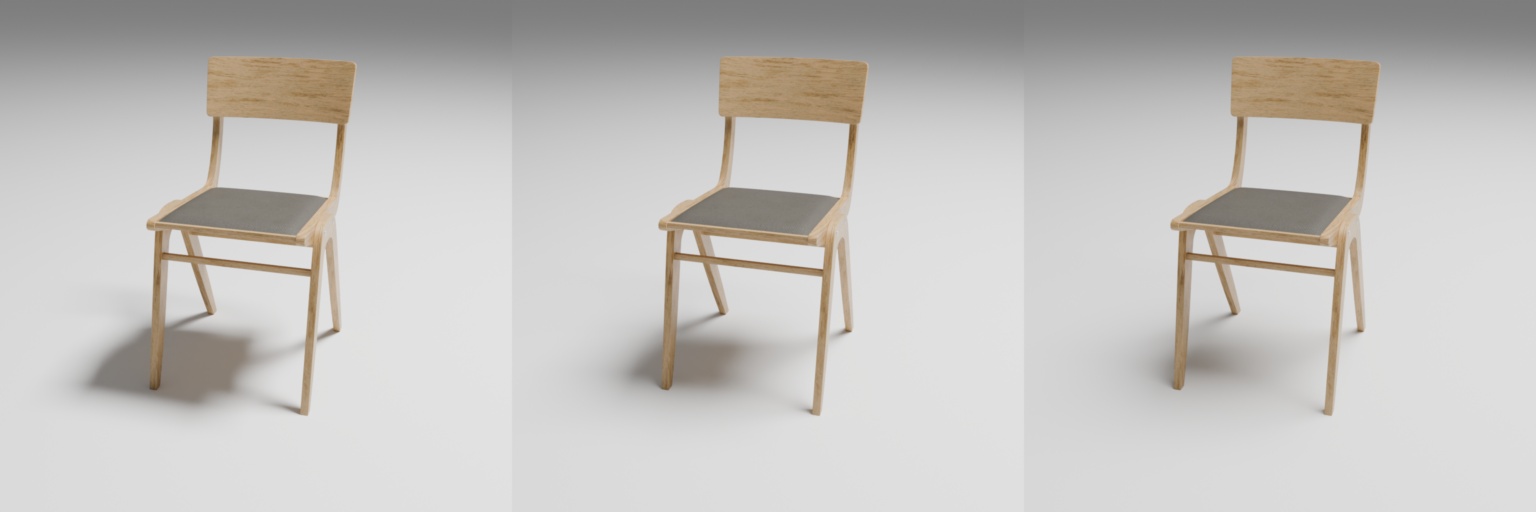}
        \caption{Renders with the area lights having sizes 1, 2, and 3 respectively.}
    \end{subfigure}    
    \caption{The effect of the area light size on the shadow's softness. The area light sizes are 1, 2, and 3 respectively from left to right.}
    \label{fig:area_light_softness}
\end{figure}

\begin{figure}[t]
    \centering
    \includegraphics[width=0.95\linewidth]{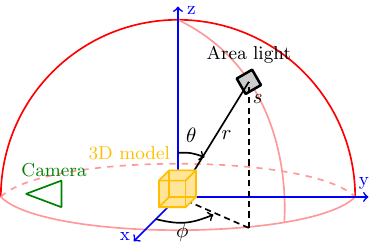}
    \caption{Spherical coordinate system. $\theta$, $\phi$, and $r$ represent the polar angle, azimuthal angle, and the radius. $s$ corresponds to the size of the area light. We place the camera at negative $y$-axis.}
    \label{fig:spherical_coordinate_system}
\end{figure}

\begin{figure}[t]
    \centering
    \begin{subfigure}[b]{0.32\linewidth}
        \centering
        \includegraphics[width=\linewidth]{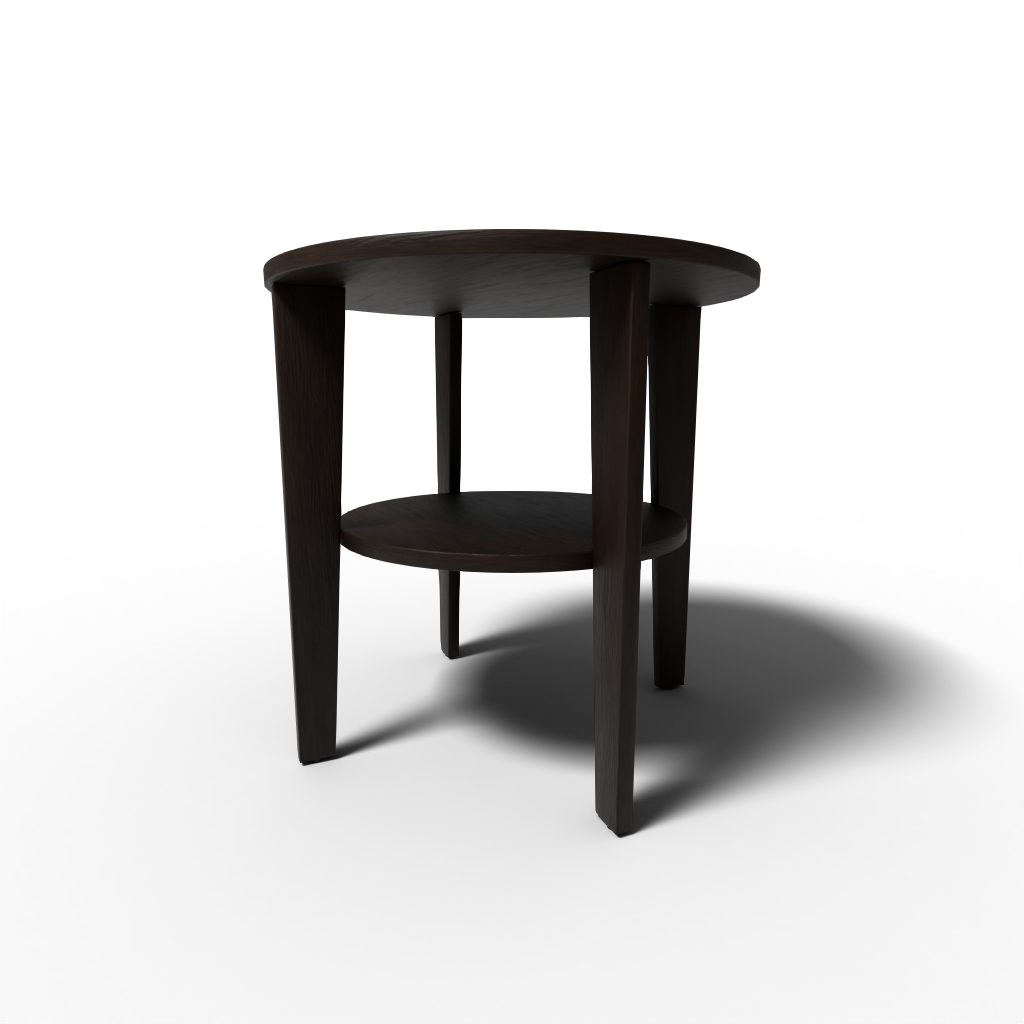}
        \caption{Image}
        \label{fig:render_example_image}
    \end{subfigure}
    \hfill
    \begin{subfigure}[b]{0.32\linewidth}
        \centering
        \includegraphics[width=\linewidth]{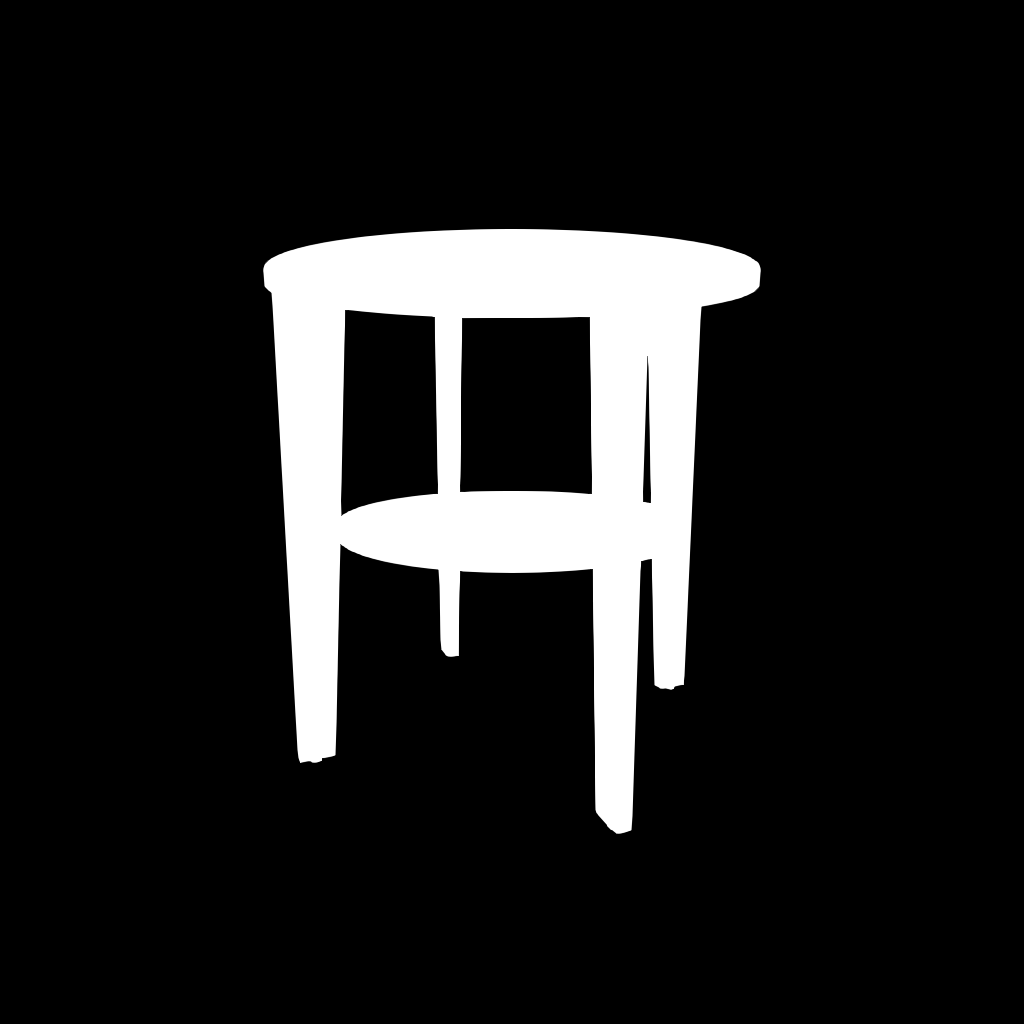}
        \caption{Mask}
        \label{fig:render_example_mask}
    \end{subfigure}
    \hfill
    \begin{subfigure}[b]{0.32\linewidth}
        \centering
        \includegraphics[width=\linewidth]{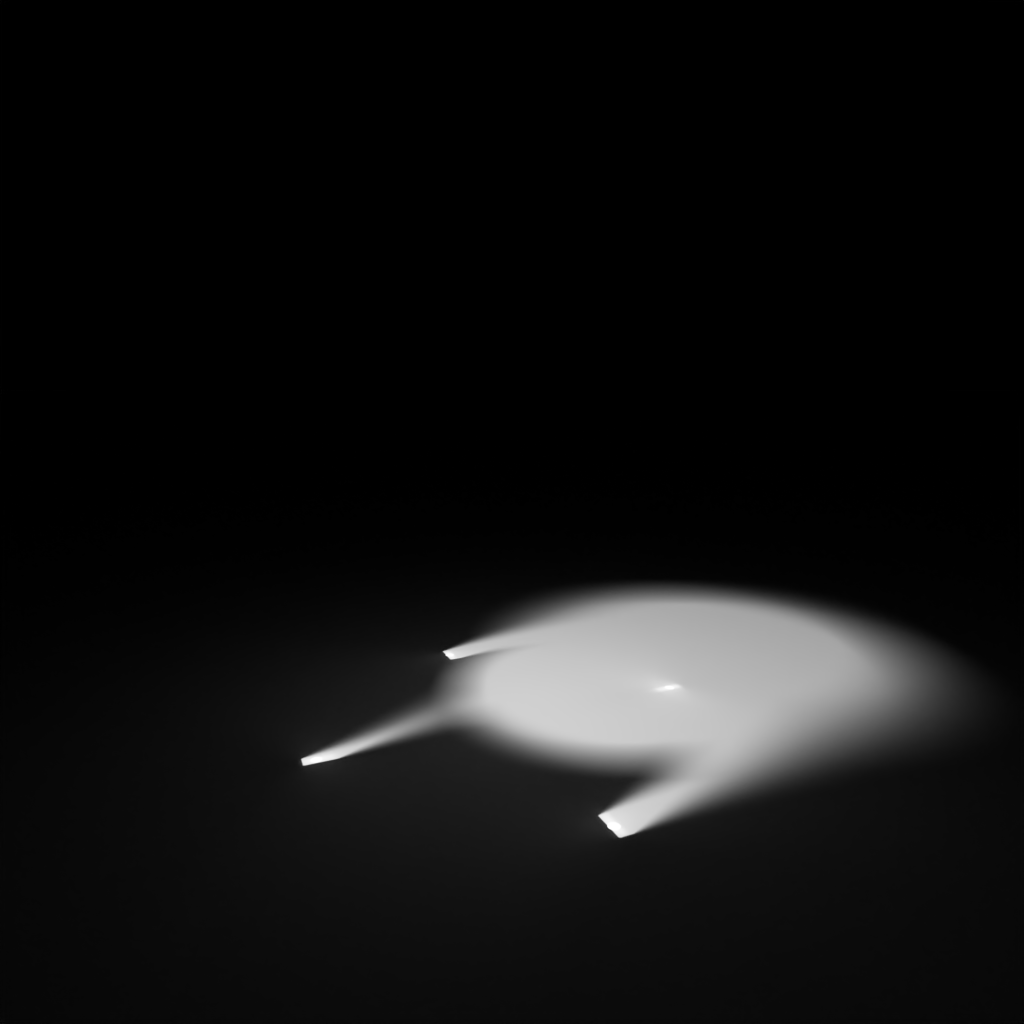}
        \caption{Shadow Map}
        \label{fig:render_example_shadow}
    \end{subfigure}
    \caption{An example image from our dataset and its annotations.}
    \label{fig:render_example}
\end{figure}

\subsection{Synthetic Dataset}
\label{sec:method:subsec:dataset}
Manually annotating images to build a dataset for the shadow generation task is inefficient, as it requires significant labor and poses challenges in generating geometrically accurate shadows without incorporating physics. We overcome these challenges by creating a synthetic dataset, which is therefore a crucial component in our pipeline.

We collected a large set of high-quality 3D meshes created by professional artists, licensed for free of use, covering a wide range of object categories (see SM.~\ref{sm:sec:synthetic_dataset} for more details). However, the 3D meshes created by the artists are usually placed at random locations in 3D space. As a result, when a 3D mesh is positioned above the ground, the shadow cast by the light source often does not connect with the object. This creates the illusion that the object is floating in the air, as illustrated in Fig. \ref{fig:flying_objects}. Training a model on such images would likely lead to the prediction of similar erroneous shadows positioned below the object. To overcome this limitation, we place a sufficiently large temporary horizontal plane on the ground, apply a rigid body physics~\cite{rigid_body} to each mesh, drop the 3D mesh from a certain height, and finally remove the plane. This pre-processing step is necessary to place the meshes on the ground to connect the shadow with the object. Fig.~\ref{fig:flying_objects}~shows example meshes before and after this processing step.

We use Blender~\footnote{https://www.blender.org} 3D engine to create our synthetic dataset. To render shadows for a given 3D mesh, we employ an \emph{area light}, a light source that emits light from a defined two-dimensional shape, such as a rectangle, square, or disc. We prefer to use a square area light, because it allows us to easily adjust the softness of the shadows by simply changing the size of the square area light. A smaller area light produces sharper shadows, while a larger one results in softer, more diffused shadows. Fig.~\ref{fig:area_light_softness} illustrates how the shadow's softness evolves when changing the light size $s$.

We propose to use the spherical coordinate system to position the light, where its location is determined by the polar angle $\theta$, the azimuthal angle $\phi$, and the radius of the sphere $r$. We place the 3D mesh at the center and the camera at the negative $y$-axis. Given a set of light source parameters $\mathcal{S}(\theta, \phi, s)$, we render an image of the object with its shadow (Fig.~\ref{fig:render_example_image}), a binary object mask (Fig.~\ref{fig:render_example_mask}), and a gray-scale shadow map (Fig.~\ref{fig:render_example_shadow}) during each rendering iteration.
Fig.~\ref{fig:spherical_coordinate_system} illustrates all light source parameters and the positioning of the camera and the 3D object within a spherical coordinate system. The intensity of the rendered shadow map can be easily adjusted by multiplying it with a scalar $I$.

\begin{figure*}[t]
    \centering
    \includegraphics[width=0.75\textwidth]{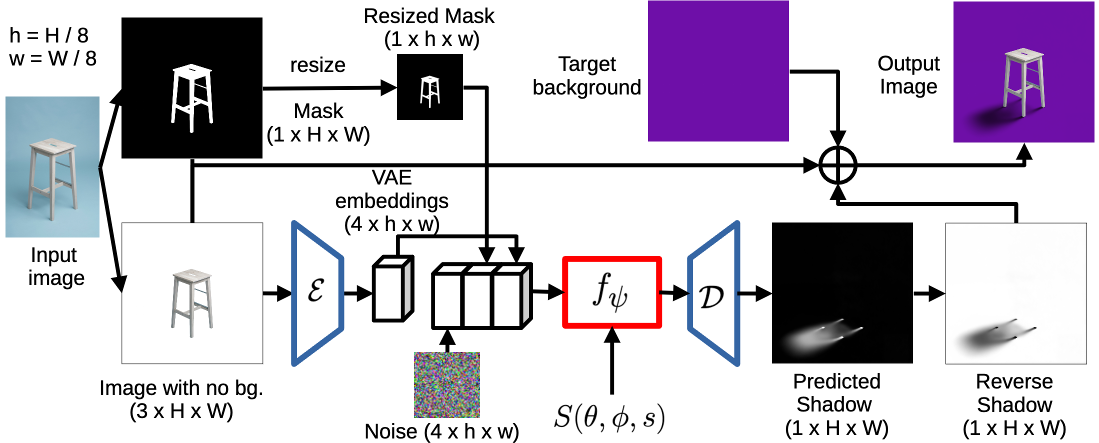}
    \caption{Our controllable shadow generation pipeline. We first remove the background of the input image, providing us with a binary mask. The VAE embeddings computed from the background-free input, and the resized mask are concatenated with the noise in the latent space. The denoiser $f_{\psi}$ is also conditioned on the light parameters $\mathcal{S}(\theta, \phi, s)$ through timestep embeddings to predict controllable shadows. We reverse the predicted shadow map, blend it with the object image and the target background to produce the final output.}
    \label{fig:shadow_gen_pipeline}
\end{figure*}

\subsection{The Shadow Generation Pipeline}\label{sec:method:subsec:shadow_generation_pipeline}

Our second main contribution focuses on training a model using our synthetic dataset, outlined in Sec.~\ref{sec:method:subsec:dataset}, to predict shadows. We propose developing a diffusion model conditioned on both the object image and the light parameters \(\mathcal{S}(\theta, \phi, s)\) to predict a controllable shadow map in a single step. In Sec.~\ref{sec:method:subsec:shadow_generation_pipeline:subsubsec:background_on_diffusion}, we provide an overview of diffusion models and the various prediction types for training. Additionally, in Sec.~\ref{sec:method:subsec:rectified_flows}, we revisit rectified flow~\cite{rectified_flow}, which has proven effective in generating shadows in just one step.

\subsubsection{Background on Diffusion Models}\label{sec:method:subsec:shadow_generation_pipeline:subsubsec:background_on_diffusion}

Diffusion models \cite{sohl2015deep,song2020score,ddpm} are probabilistic generative models that transform a simple noise distribution into a complex data distribution such as natural images, through a series of forward and reverse processes~\cite{diffusion_survey_pami, diffusion_survey_acm, variational_diffusion_models}. In the forward process, random noise is gradually added to the input data over a sequence of timesteps, transforming it into simple Gaussian noise. The reverse process aims to reverse this gradual corruption of the data by learning to denoise the noisy data. This allows for the generation of realistic samples from pure noise during the inference stage~\cite{ddpm}.

To reduce the computational complexity involved in training a diffusion model on high dimensional data, such as images, a common practice is to employ an auto-encoder~\cite{vae} with an encoder $\mathcal{E (\cdot)}$ and a decoder $\mathcal{D (\cdot)}$, to respectively map an input image into a smaller latent space and decode it back to the pixel space, i.e., $\mathcal{E}(x)=z^x$ and $\mathcal{D}(z^x)\approx x$, where $x \in \mathcal{X}$ is a set of input images deriving from an unknown distribution~\cite{latent_diffusion}. 

The forward process is controlled by two differentiable functions $\alpha(t)$, $\sigma(t)$ for any $t \in [0, 1]$ as follows:
\begin{equation}\label{eq:repametrization}
  z^x_t = \alpha(t) \cdot z^x_0 +  \sigma(t)\cdot \varepsilon\hspace{1em}\text{with}\hspace{1em}\varepsilon \sim \normal\,,
\end{equation}
where $z^x_0$ represents the embeddings computed by $\mathcal{E}$ from the input image $x_0$. As $t$ goes to 1, the noisy sample eventually resembles pure noise. In practice, a diffusion model consists in learning a function $f_{\psi}$ parametrized by a set of parameters $\psi$, conditioned on the timestep $t$ and taking as input the noisy sample $z^x_t$. Different choices of parametrization exist for $f_{\psi}$ leading to the following objectives:
\begin{itemize}
    \item \emph{$\varepsilon$-prediction}~\cite{ddpm}: predicting the amount of noise added to $z^x_0$ in Eq.~\eqref{eq:repametrization}, where the loss is
    \begin{equation}
        \mathcal{L} = \mathbb{E}_{z, t, \varepsilon}\left[\| \varepsilon - f_{\psi} (z^x_t) \|^2\right]\,,
    \end{equation}
    \item \emph{sample-prediction}: predicting the clean latents as
    \begin{equation}
        \mathcal{L} = \mathbb{E}_{z, t, \varepsilon}\left[\| z^x_0 - f_{\psi} (z^x_t) \|^2\right]\,,
    \end{equation}
    \item \emph{$v$-prediction}~\cite{prog_distillation}: which can be regarded as a combination of the two prediction types above
     \begin{equation}
        \mathcal{L} = \mathbb{E}_{z, t, \varepsilon}\left[\| \left(\alpha(t) \cdot \varepsilon - \sigma(t) \cdot z^x_0 \right) - f_{\psi} (z^x_t) \|^2\right]\,.
    \end{equation}
\end{itemize}

\subsubsection{Rectified Flows}
\label{sec:method:subsec:rectified_flows}
Let $p_0$ be an unknown target data distribution and $p_1$ a simple distribution easy to sample from, Flow Matching \cite{lipmanflow} is a type of generative model that aims to estimate a time-dependent vector field $v_t$ where the vector field defines the following Ordinary Differential Equation (ODE).
\begin{equation}\label{eq:ODE}
    \mathrm{d}x_t = v_t(x_t, t)\mathrm{d}t\hspace{1em}x_0 \sim p_0,~x_1 \sim p_1,~t \in [0, 1].
\end{equation}
The main idea behind flow matching is to estimate the vector field allowing to interpolate between $p_0$ and $p_1$  with a parametrized function $f_{\psi}$. In the specific case of Rectified Flows \cite{rectified_flow}, the vector field is trained to be constant and to follow the direction $x_1 - x_0$ using the following objective
\begin{equation}\label{eq:rectified flows}
    \min_{\psi} \mathbb{E}_{x_0,x_1,t}\left[\|(x_1 - x_0) - f_{\psi}(x_t, t) \|^2 \right]\,,
\end{equation}
where $x_t = x_1\cdot t + (1-t) \cdot x_0$,  $(x_0, x_1) \sim p_0 \times p_1$ and $t$ is sampled from a given timestep distribution $\pi$.

\subsubsection{Shadow Generation}
\label{sec:method:subsec:shadow_generation_pipeline:subsubsec:shadow_gen}
Our controllable shadow generation pipeline employs SDXL architecture~\cite{sdxl} as its backbone, but we remove all the cross-attention blocks originally used for text conditioning. Instead of using its original objective, we train it using a Rectified Flow objective  Eq.~\eqref{eq:rectified flows}. Since the VAE of SDXL has been trained to compress RGB images, we transform the target gray-scale shadow map into an RGB image by replicating it twice and concatenating them with the original. At inference, we use only the first channel as the predicted shadow. We reverse the predicted shadow map and finally blend it with the object image and the target background to produce the final output. Fig.~\ref{fig:shadow_gen_pipeline} depicts the full pipeline.

We condition the diffusion model $f_\psi$ on the \emph{object image} $o$ and its \emph{mask} $m$ to enforce the model to encode the geometry of the object. To achieve this, we map the object image $o \in \mathbb{R}^{3 \times H \times W}$ and its corresponding binary mask $m \in \mathbb{R}^{1 \times H \times W}$ to the latent space by encoding with the frozen encoder $\mathcal{E}$, $z^o = \mathcal{E}(o) \in \mathbb{R}^{c \times h \times w}$, $z^m = F(m) \in \mathbb{R}^{1 \times h \times w}$ with $W/w=H/h=8$, where $c=4$, and $F$ denotes the resizing operator. As the mask and the foreground are spatially aligned with the final output, we directly concatenate these conditionings to the noisy latent: $z^x_t \leftarrow [z^x_t,z^m, z^o] \in \mathbb{R}^{(2c+1) \times h \times w}$. Since the input latents to the denoising network have more channels, $2c+1$, than the original SDXL denoiser, $c$, we introduce new parameters in the first convolution block, initialized to zero. All other parameters are initialized using the SDXL weights.

To condition $f_\psi$ on the light parameters we propose to encode the scalar light source parameters $\mathcal{S}(\theta, \phi, s)$ with a sinusoidal embedding \cite{ddpm}, already used in SDXL~\cite{sdxl} to encode the timesteps $t$, as follows:
\begin{equation}
    \begin{split}
        e(t) &= \left[
            \left\{\cos\left(\omega_{i}^d \cdot t \right) \right\}_{i=0}^{d/2-1},
            \left\{\sin\left(\omega_{i}^d \cdot t \right) \right\}_{i=0}^{d/2-1}
        \right],
    \end{split}
\label{eq:timestep_emb}
\end{equation}
with the frequency $\omega_{i}^d = 2^{- \frac{i \cdot(i - 1)}{d/2 \cdot (d/2 - 1)} \log\left(10000\right)}$, $\forall i \in [0, d/2 - 1]$, where $d=256$ is the projection embedding dimension. The denoiser $f_\psi$ is therefore conditioned on the vector $[e(\theta), e(\phi), e(s)] \in \mathbb{R}^{768}$, which is added to the usual timestep embedding.
We use the above conditioning mechanism that allows to directly inject scalars into the denoiser, rather than requiring the creation of an external representation for the light parameters as proposed in SSN~\cite{ssn}. 

%% file: sections/experiments.tex
\begin{table}[t]
    \scriptsize
    \centering
    \resizebox{0.49\textwidth}{!}{
        \begin{tabular}{|c|c|c|c|c|}
        \hline
        \multicolumn{2}{|c|}{\textbf{Data}}  & \textbf{Param.} & \textbf{Interval} & \textbf{$\#$ Imgs.} \\
        
         \hline
        \multicolumn{2}{|c|}{\textbf{Training}}    & $\theta$ & $[ 0^\circ, 1^\circ, ..., 45^\circ ]$   &         \\
        \cline{3-4}
        \multicolumn{2}{|c|}{\textbf{Data}}      & $\phi$   & $[ 0^\circ, 1^\circ, ..., 360^\circ ]$  & 257,612 \\
        \cline{3-4} 
        \multicolumn{2}{|c|}{9,872 models}  & $s$      & $[ 2, 3, ..., 8]$                       &         \\
        \hline
        \hline
        \multirow{3}{*}{\rotatebox{90}{Track 1}}
        & \textbf{Softness}      & $\theta$ & $30^\circ$                              &         \\     
        \cline{3-4}
        & \textbf{Control}       & $\phi$   & $0^\circ$                               & 150     \\
        \cline{3-4}
        & 50 3D models     & $s$      & $[ 2, 4, 8]$                            &         \\
        \hline
        \hline
        \multirow{3}{*}{\rotatebox{90}{Track 2}}
        & \textbf{Horz. Direc.}  & $\theta$ & $35^\circ$                              &         \\
        \cline{3-4}
        & \textbf{Control}       & $\phi$   & $[ 0^\circ, 20^\circ, ..., 360^\circ ]$ & 270     \\    
        \cline{3-4}
        & 15 3D models    & $s$      & $2$                                     &         \\
        \hline
        \hline
        \multirow{3}{*}{\rotatebox{90}{Track 3}}
        & \textbf{Vert. Direc}.  & $\theta$ & $[ 5^\circ, 10^\circ, ..., 45^\circ  ]$ &         \\
        \cline{3-4}
        & \textbf{Control}       & $\phi$   & $0^\circ$                               & 135      \\    
        \cline{3-4}
        & 15 3D models     & $s$      & $2$                                     &         \\
        \hline 
        \end{tabular}
    }
    \caption{Light source parameters used to create the synthetic data.}
    \label{tab:light_param_intervals}
    \end{table}
    
    \section{Experiments}
    \label{sec:experiments}
    
    We start by outlining in Sec.~\ref{sec:experiments:subsec:dataset} the process of creating our synthetic dataset, a crucial component of our pipeline. Next, we include in Sec.~\ref{sec:experiments:subs:ablations}-\ref{sec:experiments:subs:intensity} detailed ablation studies and present in Sec.~\ref{sec:experiments:subs:real_images} qualitative results on real images.
    
    \begin{figure*}[t]
        \centering
        \begin{subfigure}[b]{0.33\linewidth}
            \centering
            \includegraphics[width=\linewidth]{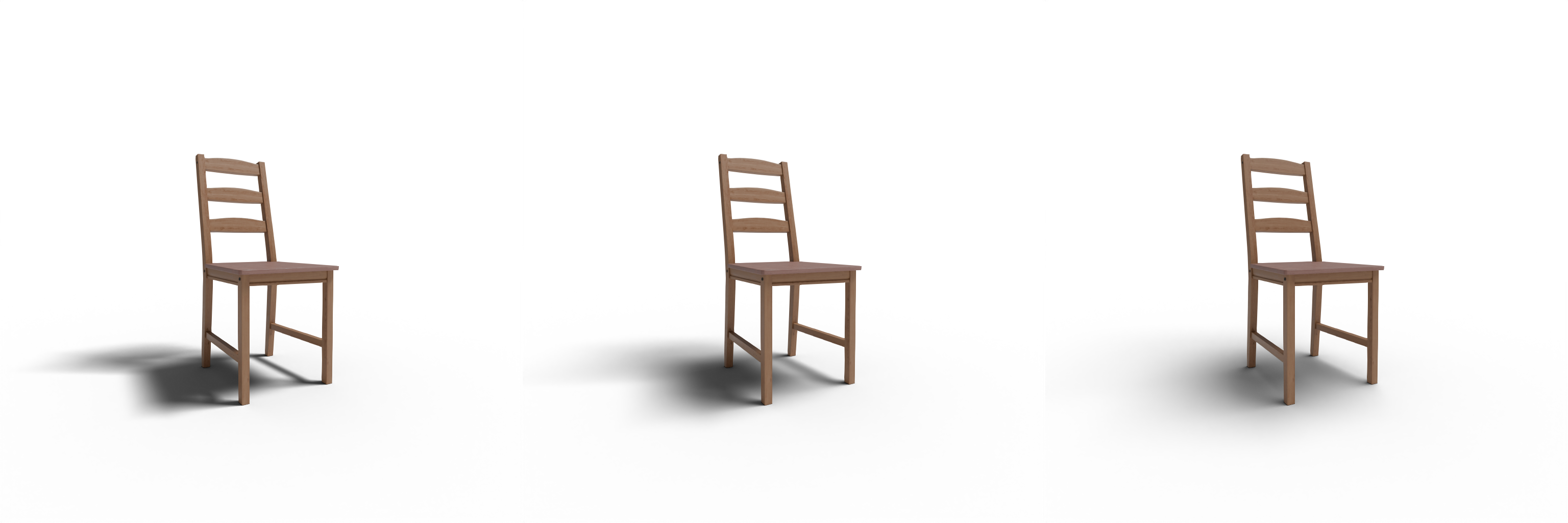}
            \caption{\centering Track 1: $\theta=30^\circ$, $\phi=0^\circ$, and $s=[2, 4, 8]$ from left to right.}
        \end{subfigure}
        \hfill
        \begin{subfigure}[b]{0.33\linewidth}
            \centering
            \includegraphics[width=\linewidth]{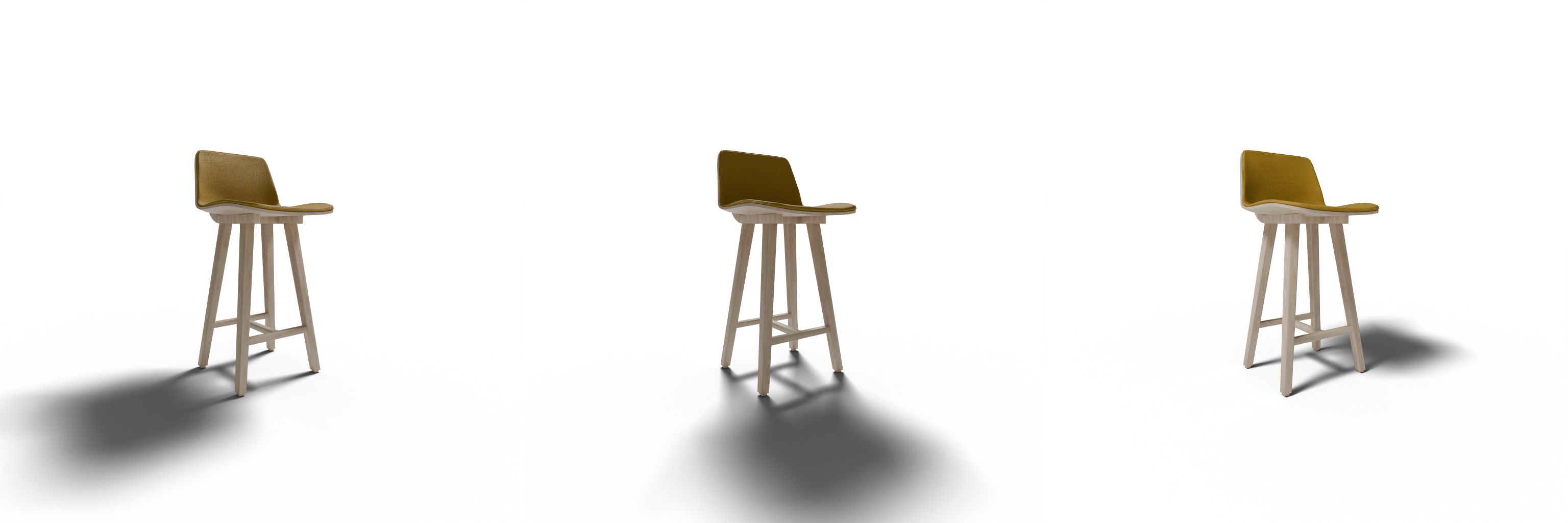}
            \caption{\centering Track 2: $s=2$, $\theta=35^\circ$, and $\phi = [ 40^\circ, 100^\circ, 220^\circ ]$ left to right.}
        \end{subfigure}
        \hfill
        \begin{subfigure}[b]{0.33\linewidth}
            \centering
            \includegraphics[width=\linewidth]{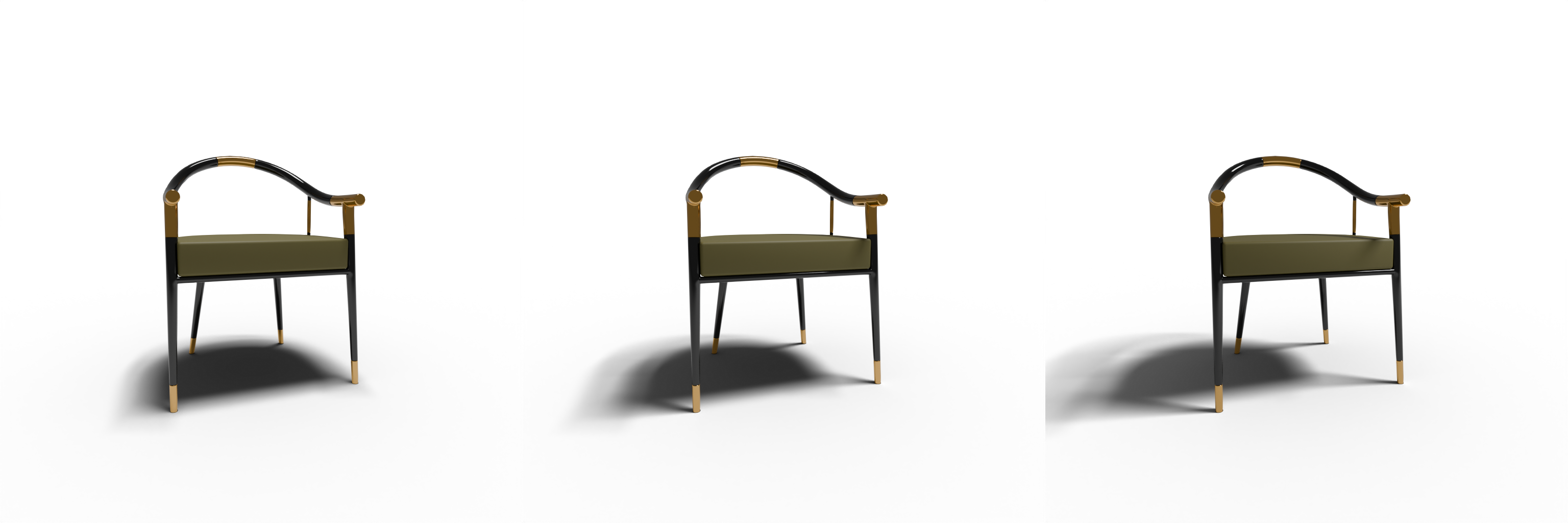}
            \caption{\centering Track 3: $s=2$, $\phi = 0^\circ$, and $\theta=[5^\circ, 20^\circ, 35^\circ]$ from left to right.}
        \end{subfigure}
        \hfill
        \caption{Example renders from each test track. Track 1: Softness control. Tracks 2-3: Horizontal and vertical shadow direction control.}
        \label{fig:example_renders_from_test_tracks}
    \end{figure*}
    
    \subsection{Dataset}
    \label{sec:experiments:subsec:dataset}
    
    To enable the model to generalize across a broad range of objects, it is crucial to gather a large dataset encompassing diverse high-quality 3D meshes. We collected 9,922 3D meshes manually designed by artists representing a wide variety of real-world objects (see more details in SM.~\ref{sm:sec:synthetic_dataset}). We use 50 models to create a test set and consider the remaining 9,872 models for the trainings. We rendered 257,612 training images with $(W, H)$=$(1024, 1024)$ resolution in Blender using \emph{cycles rendering engine}~\cite{blender_cycles}, a ray-trace-based production render engine~\cite{ray_tracing}. Although it is substantially slower, we prefer using \emph{cycles} over engines like \emph{eevee}~\cite{blender_eevee}, since it delivers the highest quality.
    
    To diversify the number of views for each object when creating the training set, in each rendering iteration, we instantiate a randomly selected 3D mesh at the center of the spherical coordinate system shown in Fig.~\ref{fig:spherical_coordinate_system} and rotate it around $z$-axis by a random degree between $0^\circ$ and $360^\circ$. We also position the camera at the negative $y$-axis and randomly move it along the $y$-axis to have images with varying object scales. We resize the 3D mesh proportionally by adjusting its largest dimension to a fixed value to maintain an appropriate scale, avoiding excessively large or small sizes. We set the sphere radius $r$ to 8. We randomize the light parameters $\mathcal{S}(\theta, \phi, s)$ (see Sec.~\ref{sec:method:subsec:dataset}) by selecting a value from the intervals indicated in Table~\ref{tab:light_param_intervals}. We render the shadow maps with a fixed intensity.
    
    With no existing dataset available to evaluate our pipeline's performance, we decide to create a new benchmark specifically for this task and make it publicly accessible. Our new test set includes three tracks, each carefully designed to assess the model's performance in controlling shadow softness, as well as horizontal and vertical direction. We create the samples for each track as:
    
    \begin{itemize}
        \item \textbf{Track 1}: Softness control. We fix $\theta$ and $\phi$ and use 3 different values for $s$.
        \item \textbf{Track 2}: Horizontal direction control. We fix $\theta$ and $s$ and use 18 different values for $\phi$.
        \item \textbf{Track 3}: Vertical direction control. We fix $\phi$ and $s$ and employ 9 distinct values for $\theta$. 
    \end{itemize}
    Table~\ref{tab:light_param_intervals} also reports the number of 3D models, values for the fixed and varying parameters, and the total number of rendered images with $1024 \times 1024$ resolution in each track. Fig.~\ref{fig:example_renders_from_test_tracks} depicts some example renders from each test track. More details can be found in the SM~Sec.~\ref{sm:sec:synthetic_dataset}.

    \begin{figure}[t]
        \begin{subfigure}[b]{\linewidth}
            \centering
            \includegraphics[width=\linewidth]{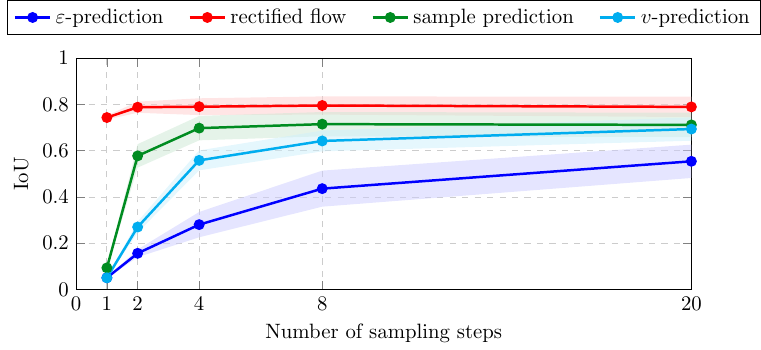}
        \end{subfigure}
        \begin{subfigure}[b]{0.945\linewidth}
            \centering
            \includegraphics[width=\linewidth]{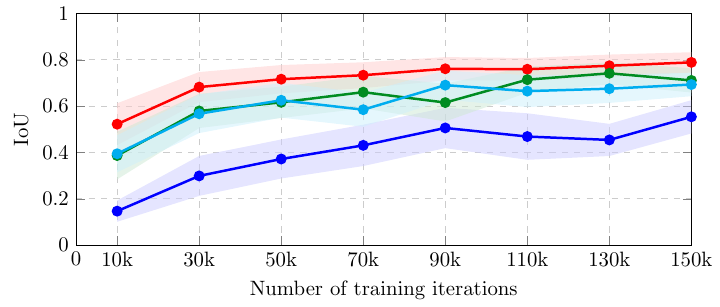}
        \end{subfigure}
        \caption{IoU vs number of sampling (first plot) and training iteration plots (second plot). In the first plot, the number of training iterations for each model is fixed to $150k$. In the second plot, the number of sampling step is set to $20$. Half transparent curve thickness represents the standard deviation.}
        \label{fig:iou_vs_steps_and_iters}
    \end{figure}

    \begin{table*}[t]
        \centering
        \scriptsize
        \begin{tabular}{|
        >{\centering\arraybackslash}m{0.067\textwidth}|
        >{\centering\arraybackslash}m{0.05\textwidth}|
        >{\centering\arraybackslash}m{0.05\textwidth}|
        >{\centering\arraybackslash}m{0.05\textwidth}|
        >{\centering\arraybackslash}m{0.05\textwidth}|
        >{\centering\arraybackslash}m{0.05\textwidth}|
        >{\centering\arraybackslash}m{0.05\textwidth}|
        >{\centering\arraybackslash}m{0.05\textwidth}|
        >{\centering\arraybackslash}m{0.05\textwidth}|
        >{\centering\arraybackslash}m{0.05\textwidth}|
        >{\centering\arraybackslash}m{0.05\textwidth}|
        >{\centering\arraybackslash}m{0.05\textwidth}|
        >{\centering\arraybackslash}m{0.05\textwidth}|
        }
            \hline
            \textbf{Pred Type} &
            \multicolumn{4}{|c|}{\textbf{Track 1 (Softness Control)}} &  
            \multicolumn{4}{|c|}{\textbf{Track 2 (Horz. Direc. Control)}} &  \multicolumn{4}{|c|}{\textbf{Track 3 (Vert. Direc. Control)}}\\
            \cline{2-13}
            & IoU $\uparrow$ & rmse $\downarrow$ & s-rmse $\downarrow$ & zncc $\uparrow$  & IoU $\uparrow$ & rmse $\downarrow$ & s-rmse $\downarrow$ & zncc $\uparrow$ & IoU $\uparrow$ & rmse $\downarrow$ & s-rmse $\downarrow$ & zncc $\uparrow$\\
            \hline
            $\varepsilon$ pred. & 0.048 & 0.421 & 0.106 & 0.021 & 0.056 & 0.425 & 0.118 & 0.018 & 0.045 & 0.426 & 0.109 & 0.016 \\
            sample & 0.098 & 0.141 & 0.106 & 0.049 & 0.094 & 0.151 & 0.118 & 0.041 & 0.087 & 0.144 & 0.108 & 0.040 \\
            $v$-pred. & 0.047 & 0.424 & 0.107 & 0.020 & 0.055 & 0.427 & 0.118 & 0.018 & 0.044 & 0.428 & 0.109 & 0.016 \\
            Rect. Flow & \textbf{0.768} & \textbf{0.020} & \textbf{0.019} & \textbf{0.982} & \textbf{0.732} & \textbf{0.027} & \textbf{0.026} & \textbf{0.968} & \textbf{0.736} & \textbf{0.026} & \textbf{0.025} & \textbf{0.971} \\
            \hline
            \hline
            $\varepsilon$ pred. & 0.598 & 0.029 & 0.027 & 0.970 & 0.540 & 0.041 & 0.037 & 0.942 & 0.532 & 0.036 & 0.034 & 0.946 \\
            sample & 0.745 & 0.030 & 0.028 & 0.956 & 0.705 & 0.039 & 0.037 & 0.929 & 0.687 & 0.039 & 0.036 & 0.928\\
            $v$-pred. & 0.722 & 0.022 & 0.021 & 0.978 & 0.685 & \textbf{0.029} & \textbf{0.027} & \textbf{0.958} & 0.678 & \textbf{0.027} & \textbf{0.026} & 0.964 \\
            Rect. Flow & \textbf{0.818} & \textbf{0.021} & \textbf{0.020} & \textbf{0.979} & \textbf{0.780} & 0.030 & 0.028 & \textbf{0.958} & \textbf{0.776} & 0.028 & 0.027 & \textbf{0.960} \\
            \hline
        \end{tabular}
        \caption{The performance of models trained for $150k$ iterations with different prediction types on each track in terms of \emph{IoU}, \emph{rmse}, \emph{s-rmse}, and \emph{zncc} metrics. The first and the last 4 rows report the quantitative results for 1 and 20 sampling steps respectively.}
        \label{tab:results_on_each_track}
    \end{table*}

    \subsection{Ablation Studies}
    \label{sec:experiments:subs:ablations}
    
    We propose an extensive analysis of each component of our pipeline by using soft intersection over union (\emph{IoU}), root mean squared error (\emph{rmse}), its scaled version (\emph{s-rmse})~\cite{portrait_relight} and zero-normalized cross-correlation (\emph{zncc})~\cite{template_matching} as the evaluation metrics.
    
    \subsubsection{Prediction Types}
    \label{sec:experiments:subs:ablations:subsubsec:predictiontype}
    We trained our diffusion model using 4 different prediction types presented in Sec.~\ref{sec:method:subsec:shadow_generation_pipeline:subsubsec:background_on_diffusion}, namely $\varepsilon$, $v$, sample predictions and rectified flow for $150k$ iterations with $\mathcal{S}(\theta, \phi, s)$ conditionings. We trained each model on 4 Nvidia H100-80GB GPUs with a batch size of 2 for about 2 days. We used AdamW optimizer~\cite{adamw} with a learning rate of $1e^{-5}$. Due to the stochastic nature of the diffusion models, the performance of the models varies depending on the initial noise. Hence, we compute mean and standard deviation values for each metric averaged over $10$ seeds for each image. 
    
    Our objective is to reduce the number of sampling steps as much as possible to satisfy the requirements of practical applications. Therefore, we first analyze how the models trained with different prediction types perform for different sampling steps. To this end, using all the images across the three tracks, we compare the models in terms of IoU values for $1, 2, 4, 8, 20$ sampling steps (more quantitative results can be found in the SM~Sec.~\ref{sm:sec:metrics}). The first plot in Fig.~\ref{fig:iou_vs_steps_and_iters} demonstrates that rectified flow with only 1 step outperforms the other prediction types with 20 inference steps. It also shows that the performance gap between 1 and 20 steps for rectified flow is significantly smaller than for the others. We report the quantitative results for 1 and 20 steps using all the metrics on each individual track in Table.~\ref{tab:results_on_each_track}.
    
    Secondly, we examine the behavior of each prediction type across varying numbers of training iterations. Our motivation is to determine whether certain types of predictions converge faster than others. For this purpose, we set the number of sampling steps to 20, and we compute IoU values for $10k$, $20k$, $\dots$, $150k$ training iterations. As expected, training longer improves the performance of the models, as shown in Fig.~\ref{fig:iou_vs_steps_and_iters}. Interestingly, rectified flow again outperforms others with fewer training iterations. Its performance with only $10k$ iterations is on par with the performance of $\varepsilon$ prediction with $150k$ iterations.
    
    \begin{figure}[t]
        \centering
        \includegraphics[width=\linewidth]{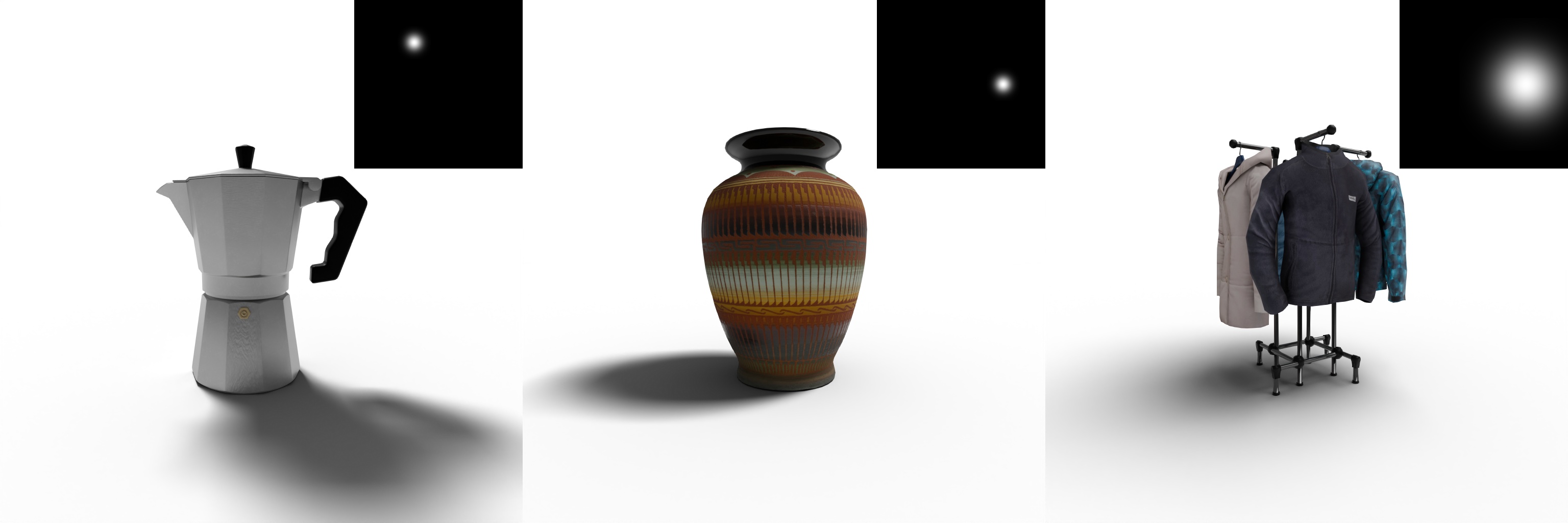}
        \caption{Example renders and the blob light maps (top-right). Blob location and size encode the light location ($x$, $y$) and size $s$.}
        \label{fig:blob_examples}
    \end{figure}
    
    \begin{figure}
        \centering
        \includegraphics[width=\linewidth]{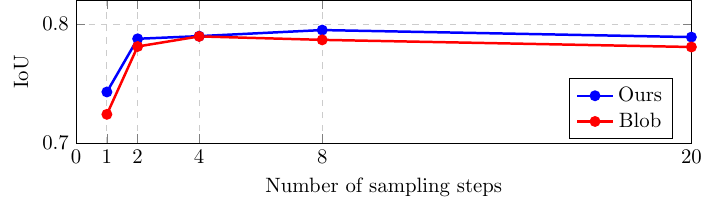}
        \caption{Comparison between our timestep and blob conditionings trained for $150k$ iterations and 1 sampling step.}
        \label{fig:model_comparison}
    \end{figure}

    \subsubsection{Other Conditioning Mechanisms}
    \label{sec:experiments:subs:ablations:subsubsec:conditioning}
    
    \begin{table*}[t]
        \centering
        \setlength{\tabcolsep}{1pt}
        \renewcommand{\arraystretch}{0.5} 
        \begin{tabular}{ccccccc} 
            & & Target Bg. Image & Object image & Result 1 & Result 2 & Result 3 \\
            \rotatebox{90}{~~~~Softness Control} & 
            \rotatebox{90}{~~~~$\theta=30^\circ$, $\phi=60^\circ$} &
            \includegraphics[width=0.175\textwidth]{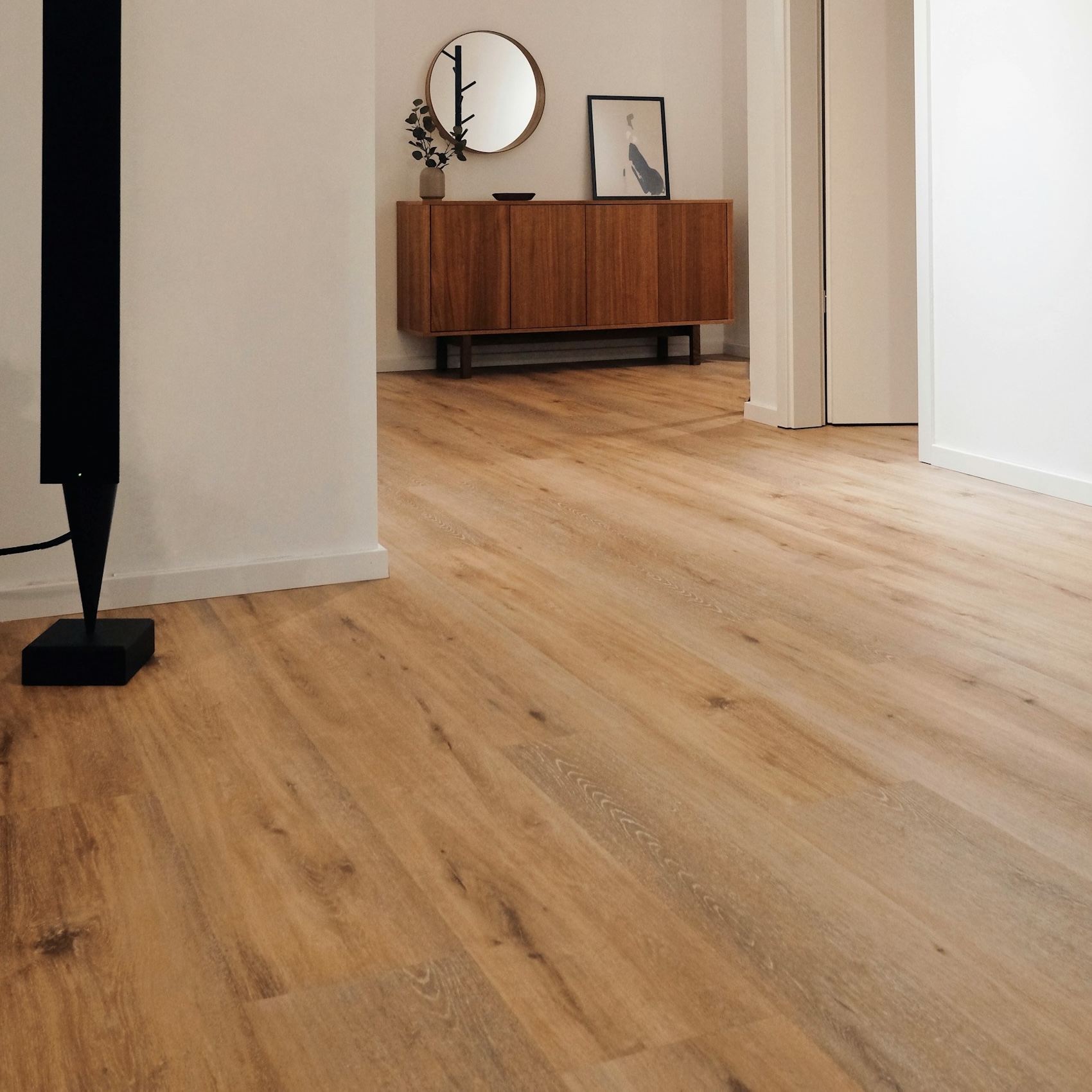} &
            \includegraphics[width=0.175\textwidth]{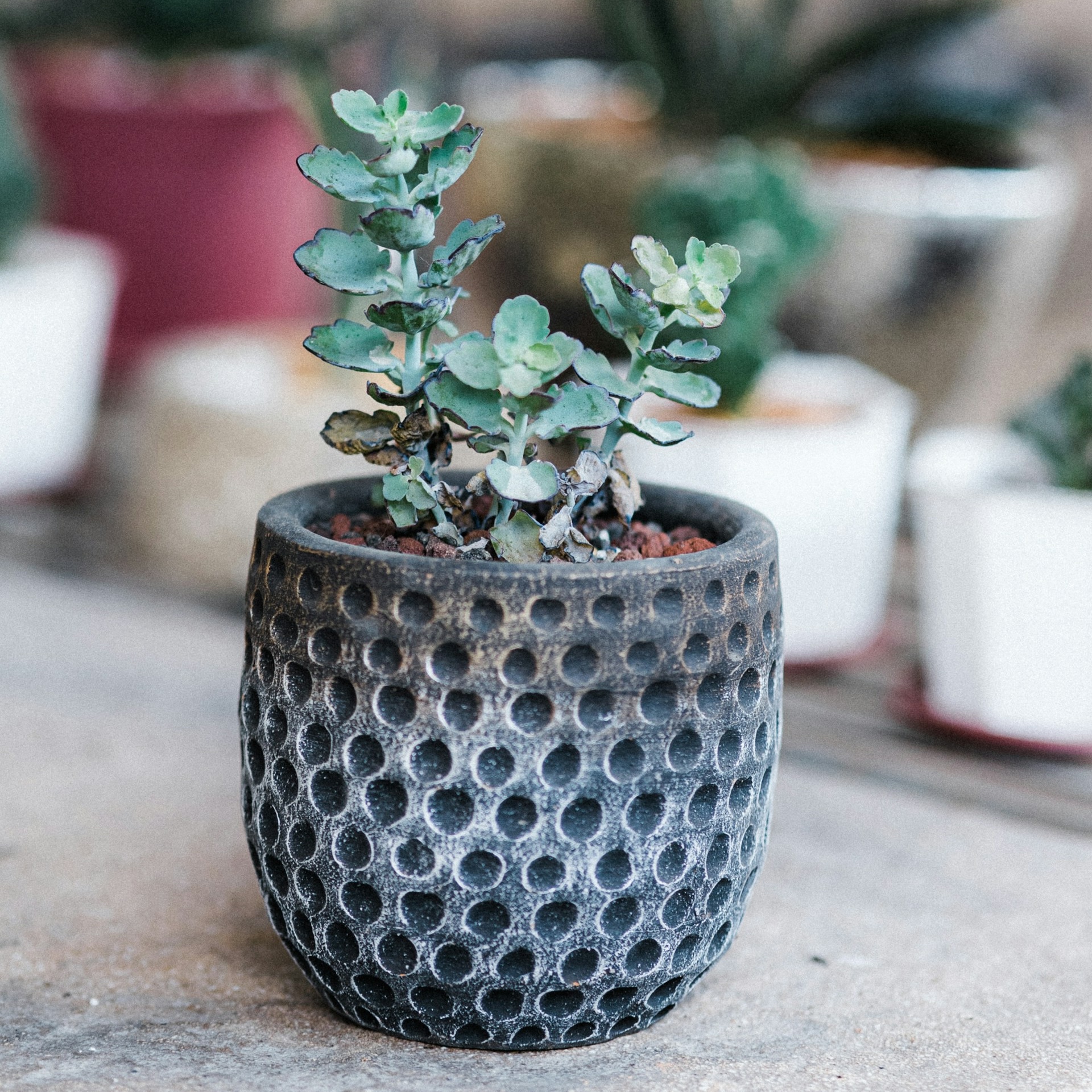} &
            \includegraphics[width=0.175\textwidth]{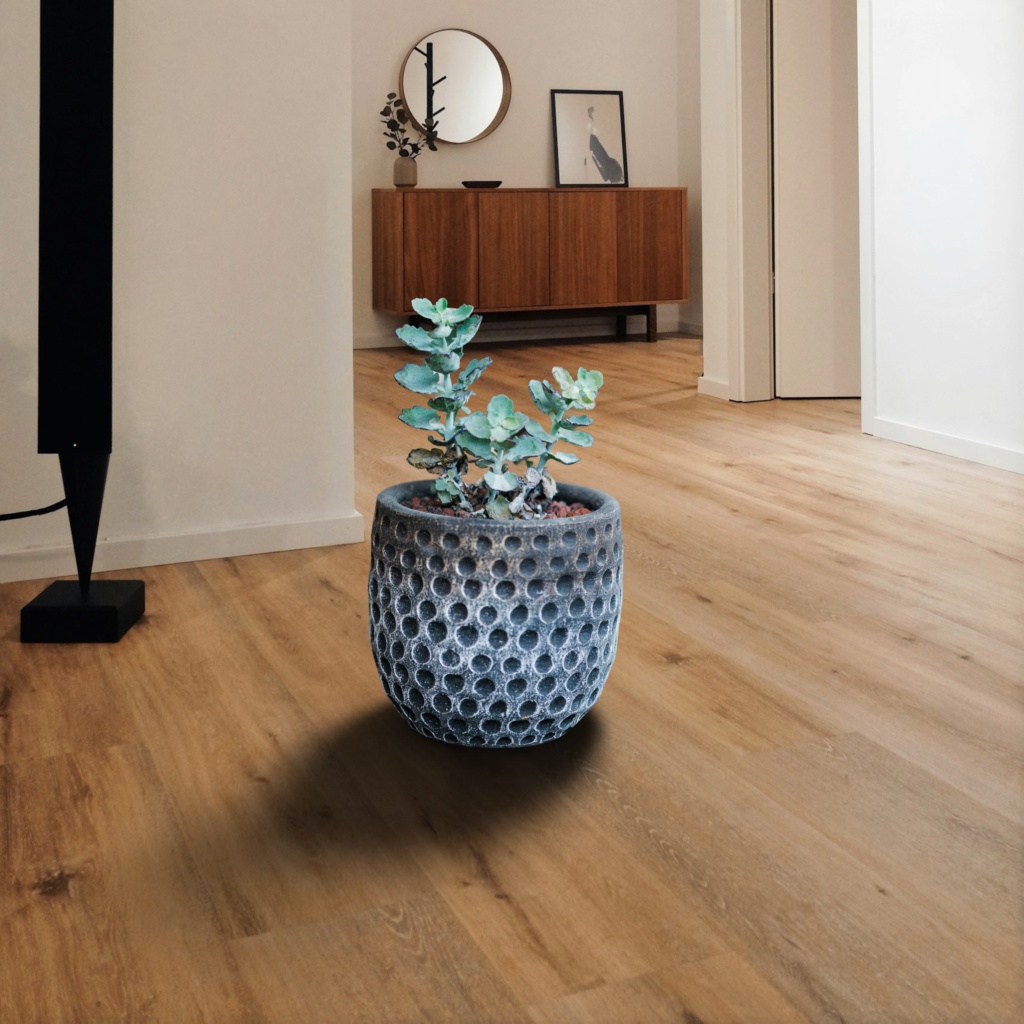} &
            \includegraphics[width=0.175\textwidth]{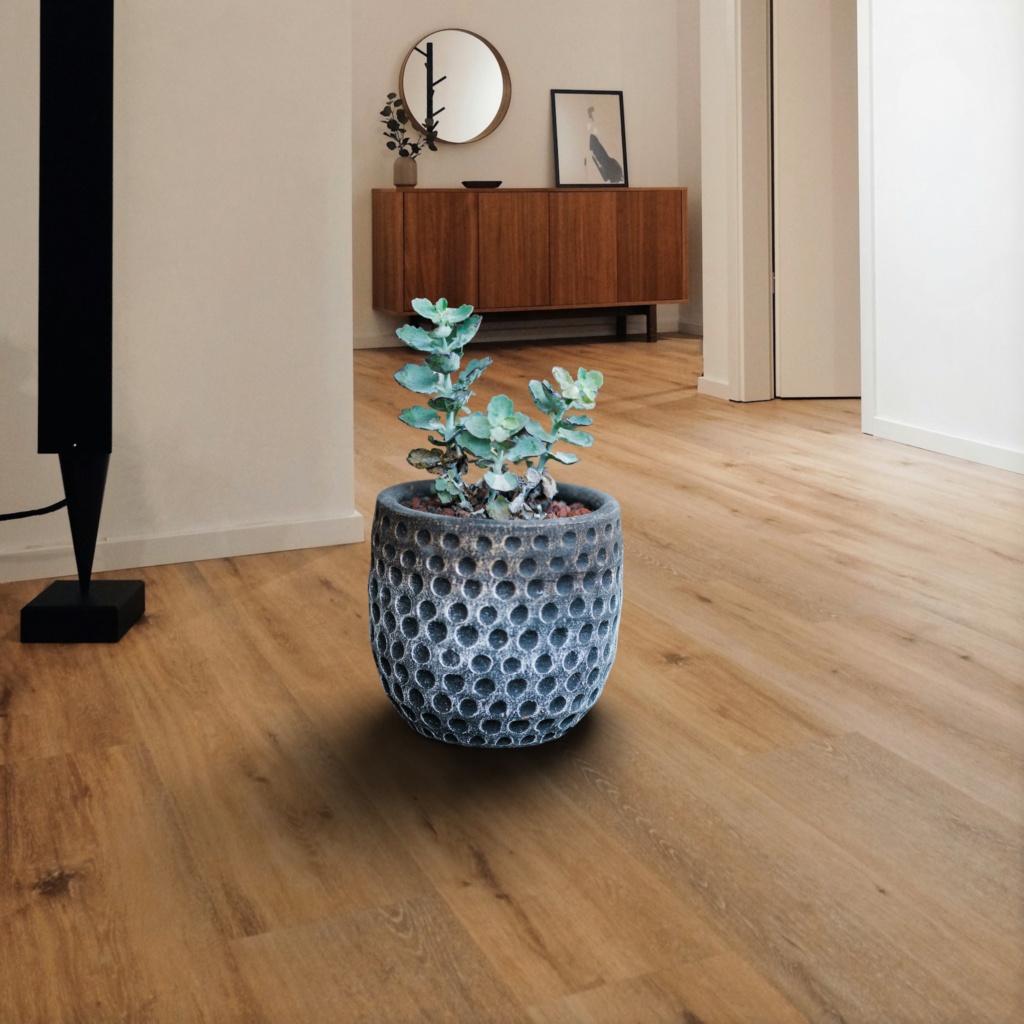} & 
            \includegraphics[width=0.175\textwidth]{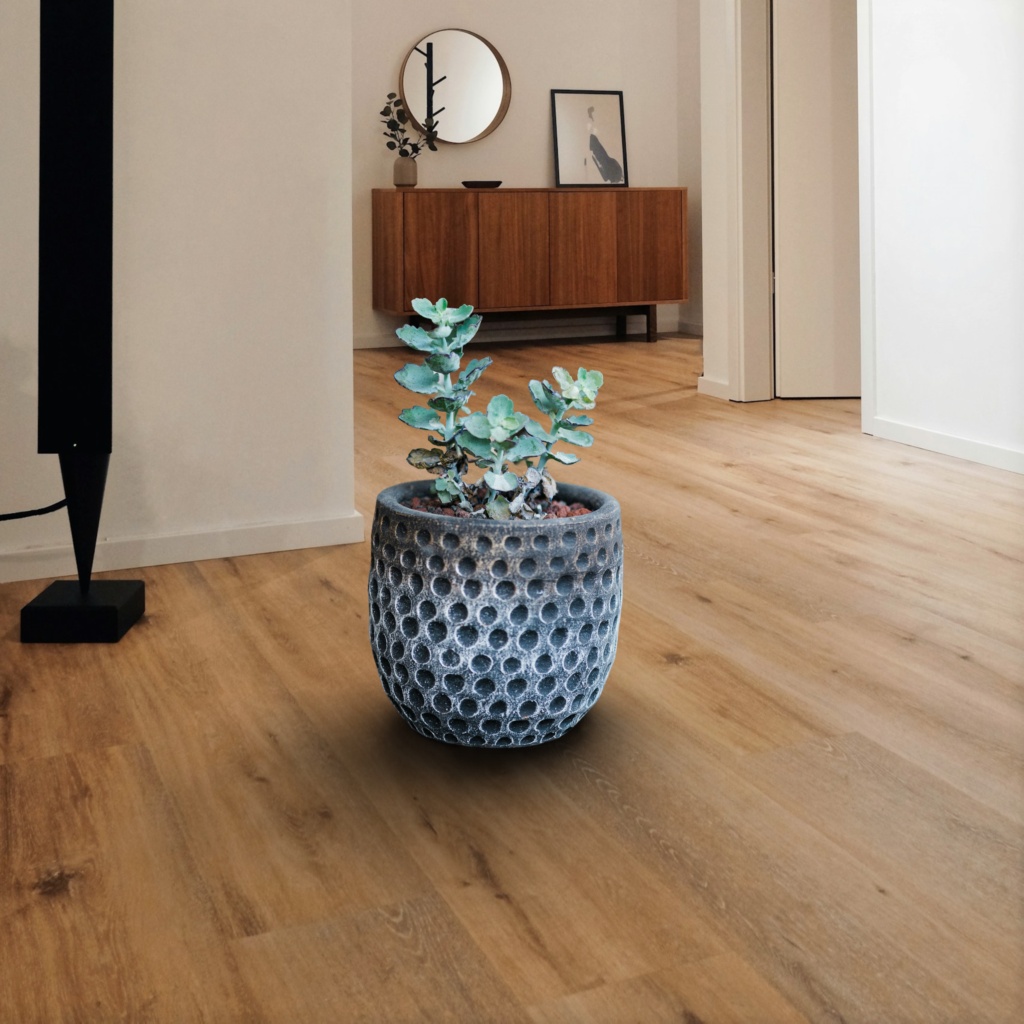} \\
    
            & & & & $s=2$ & $s=5$ & $s=8$ \\
    
            \rotatebox{90}{~~Horiz. Dir. Control} & 
            \rotatebox{90}{~~~~~~$\theta=30^\circ$, $s=2$} &
            \includegraphics[width=0.175\textwidth]{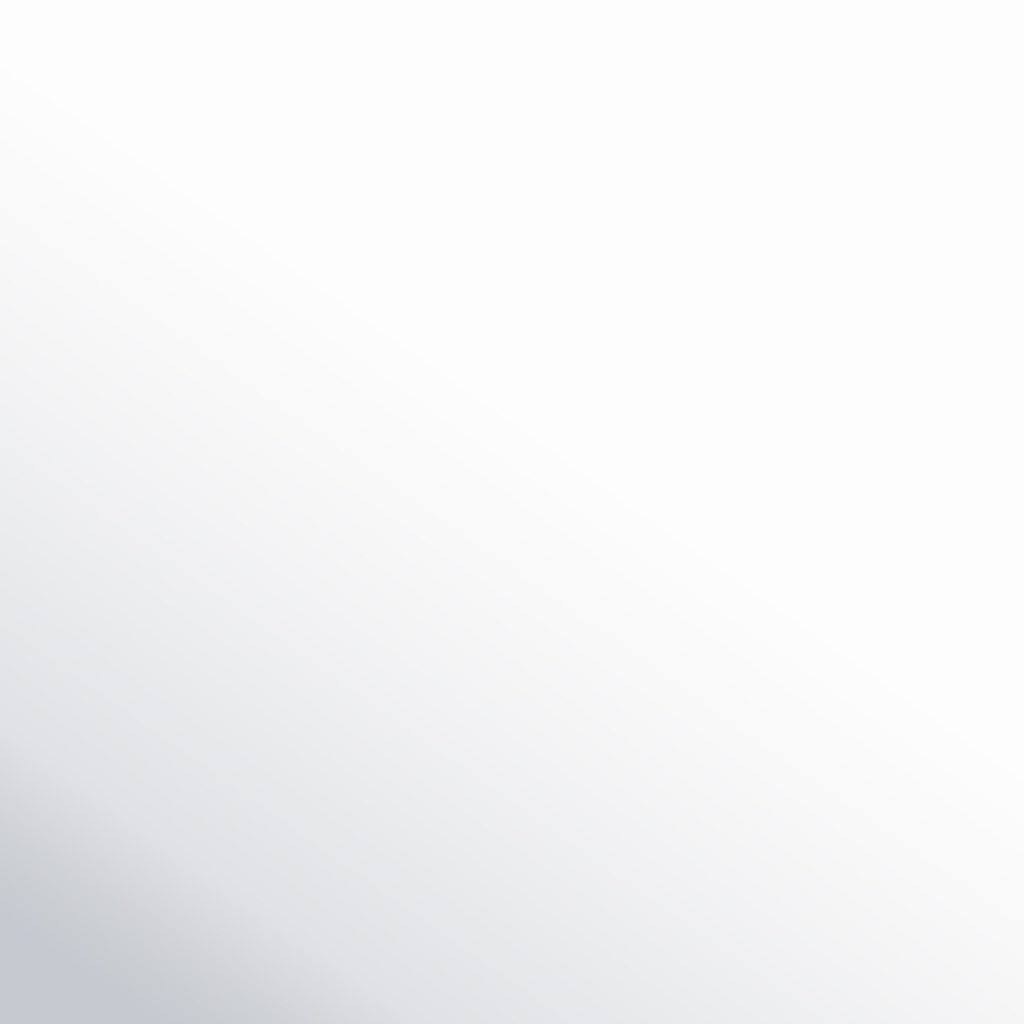} &
            \includegraphics[width=0.175\textwidth]{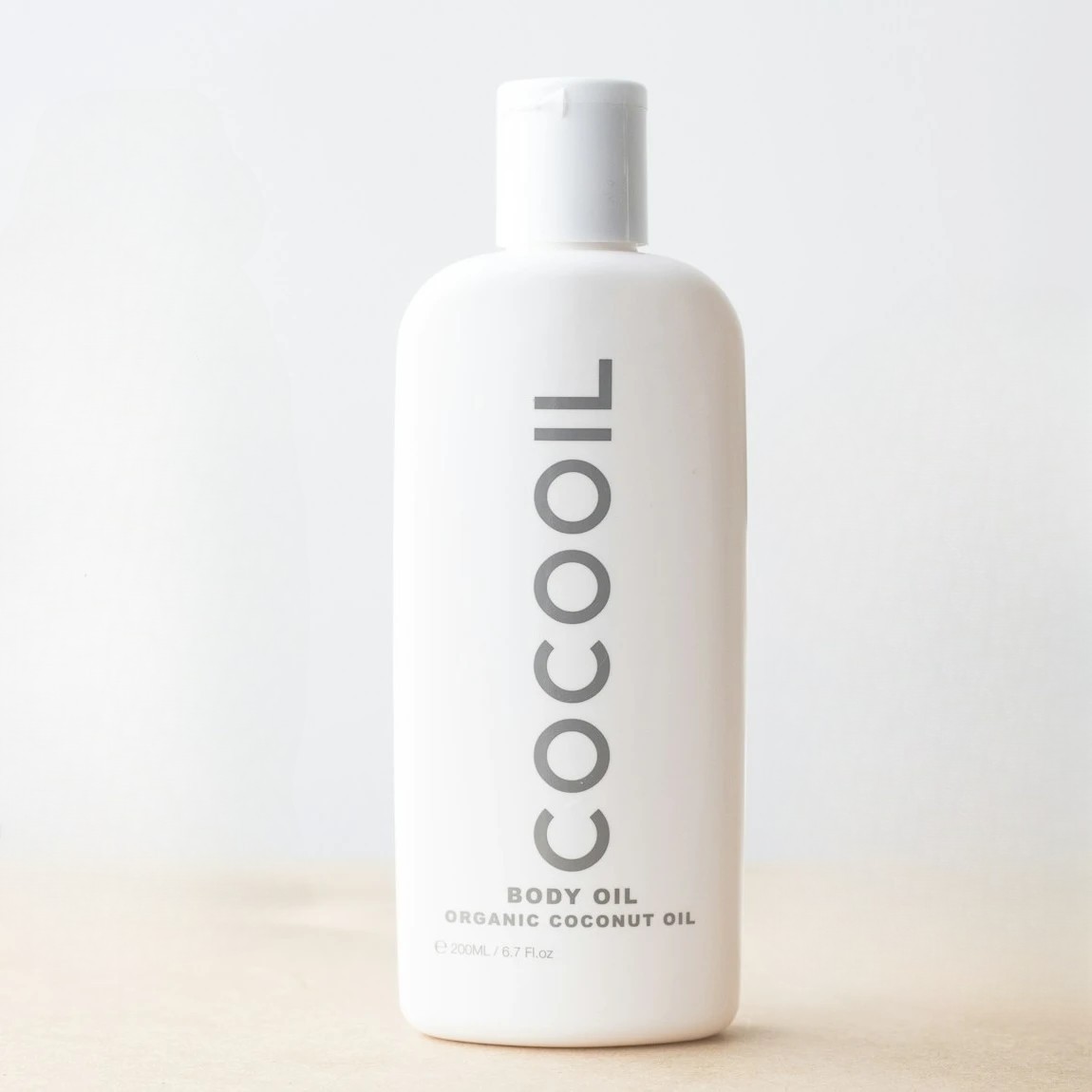} &
            \includegraphics[width=0.175\textwidth]{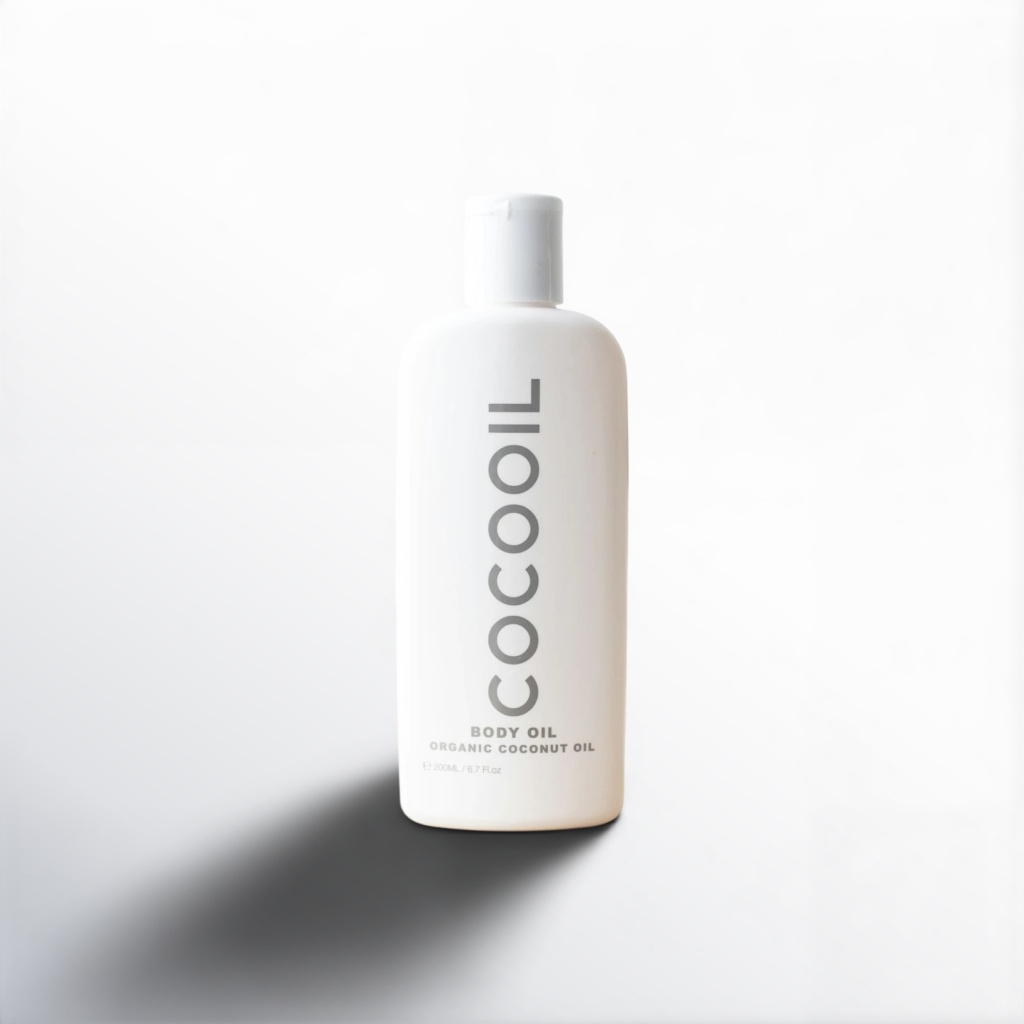} &
            \includegraphics[width=0.175\textwidth]{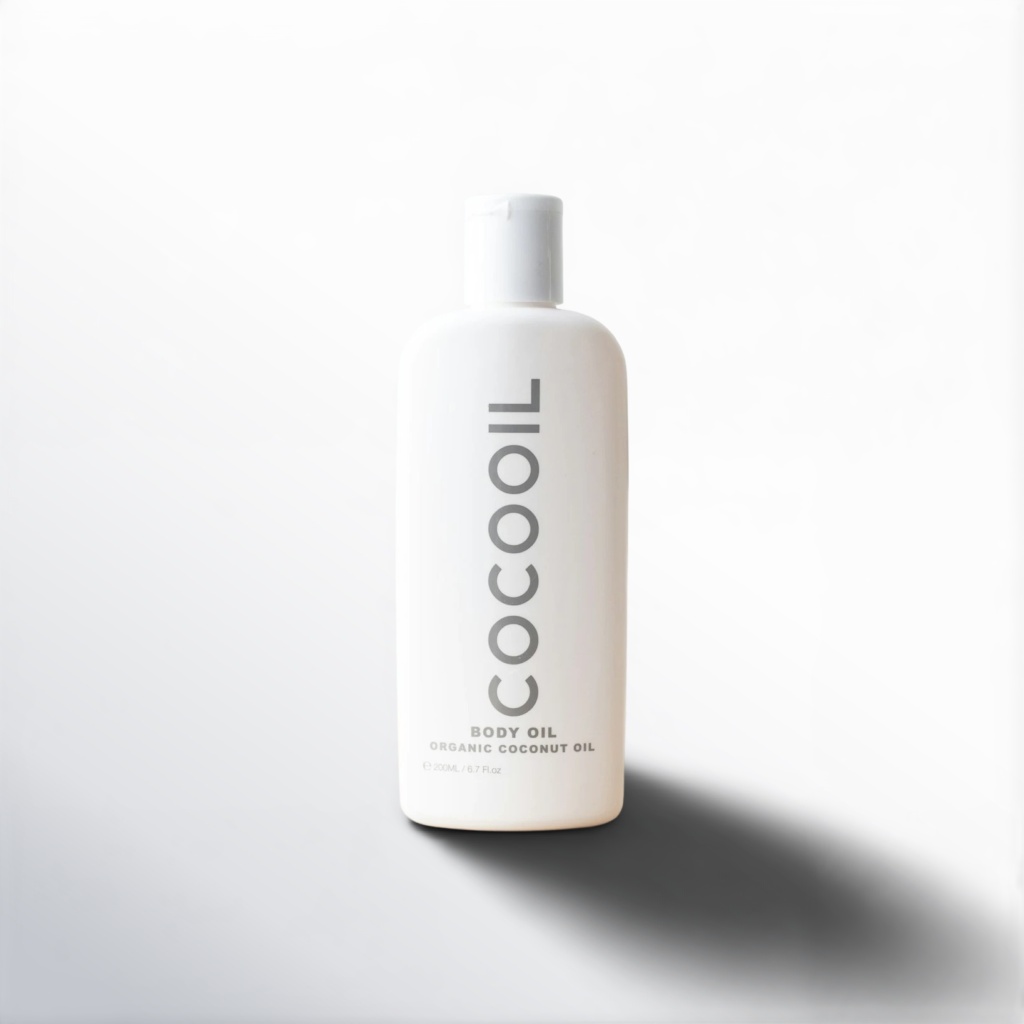} & 
            \includegraphics[width=0.175\textwidth]{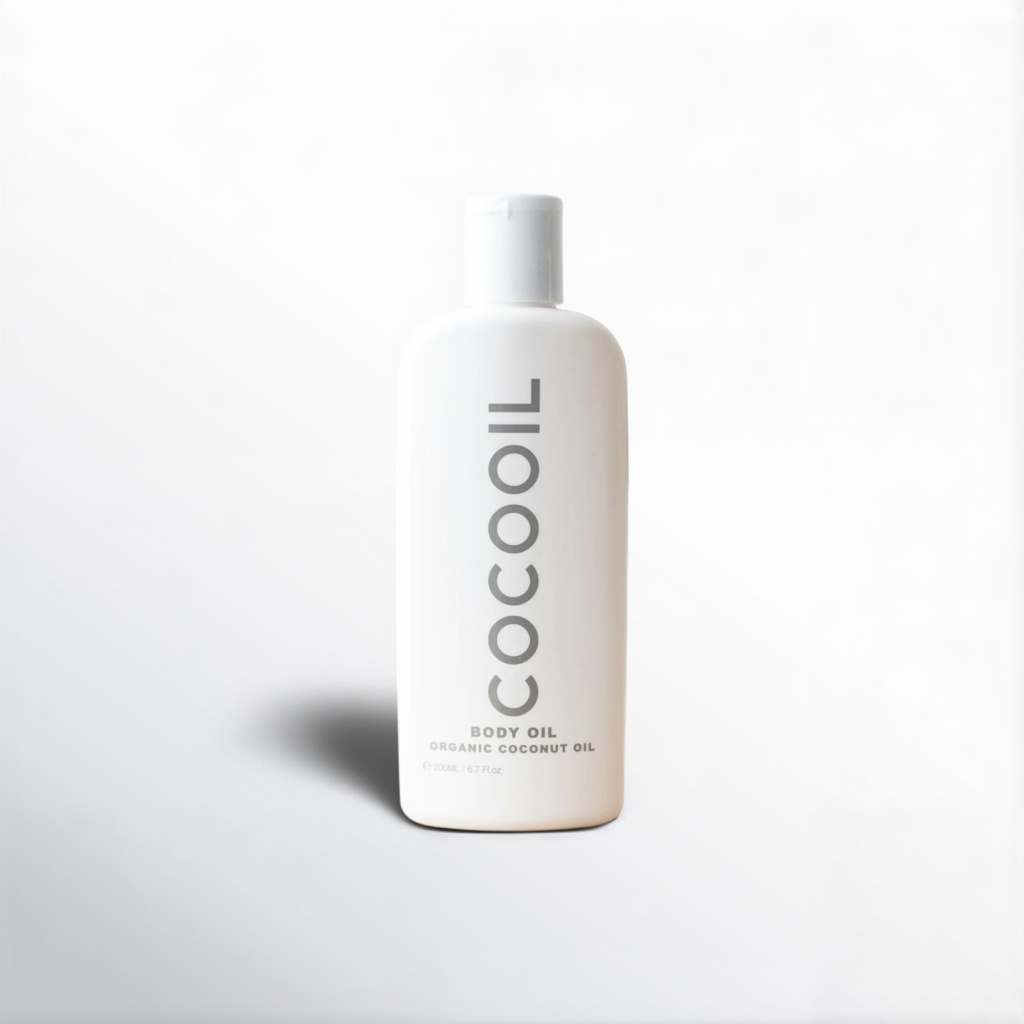}  \\
    
            & & & & $\phi=45^\circ$ & $\phi=135^\circ$ & $\phi=315^\circ$ \\
    
            \rotatebox{90}{~~~~Vert. Dir. Control} & 
            \rotatebox{90}{~~~~~~~$\phi=0^\circ$, $s=2$} &
            \includegraphics[width=0.175\textwidth]{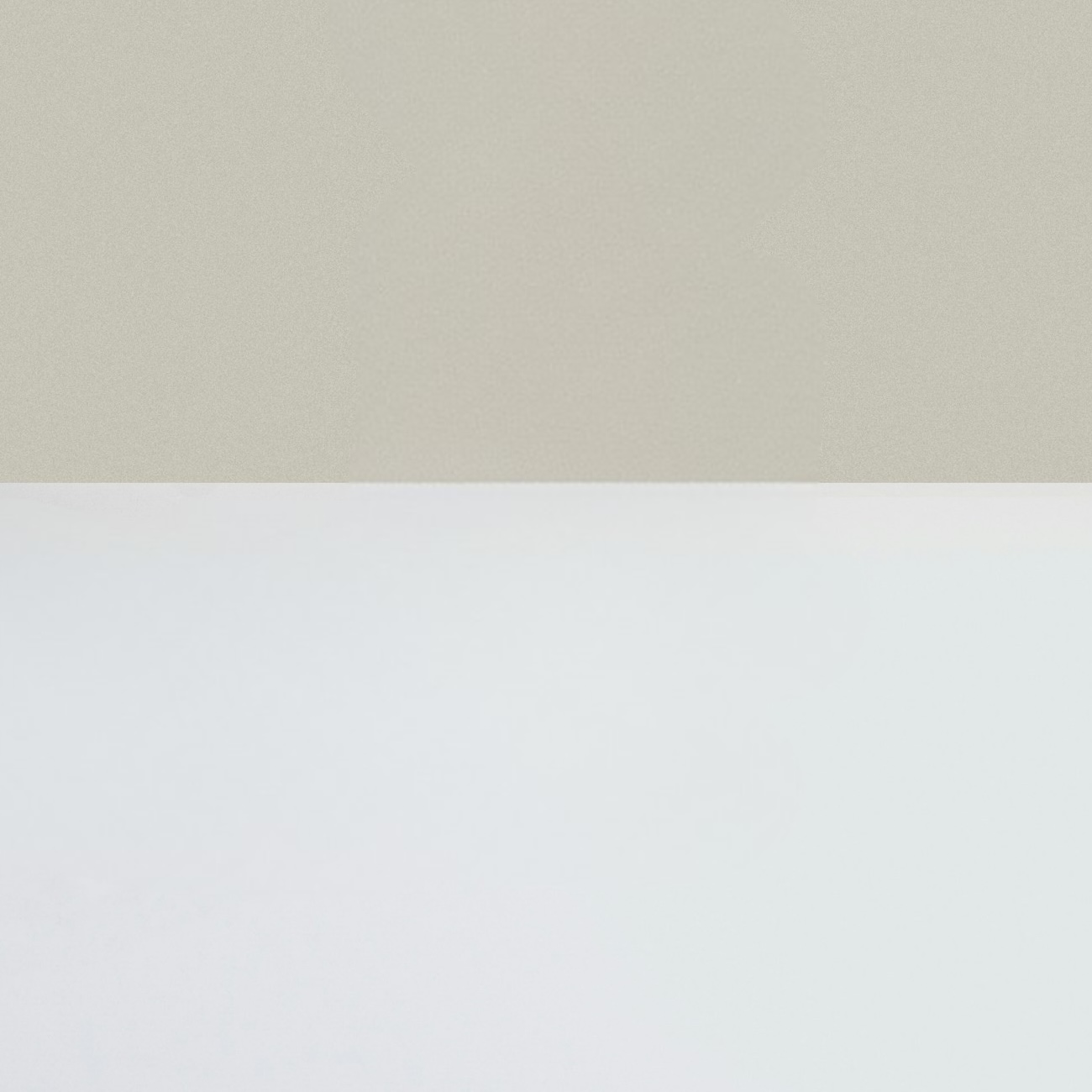} &
            \includegraphics[width=0.175\textwidth]{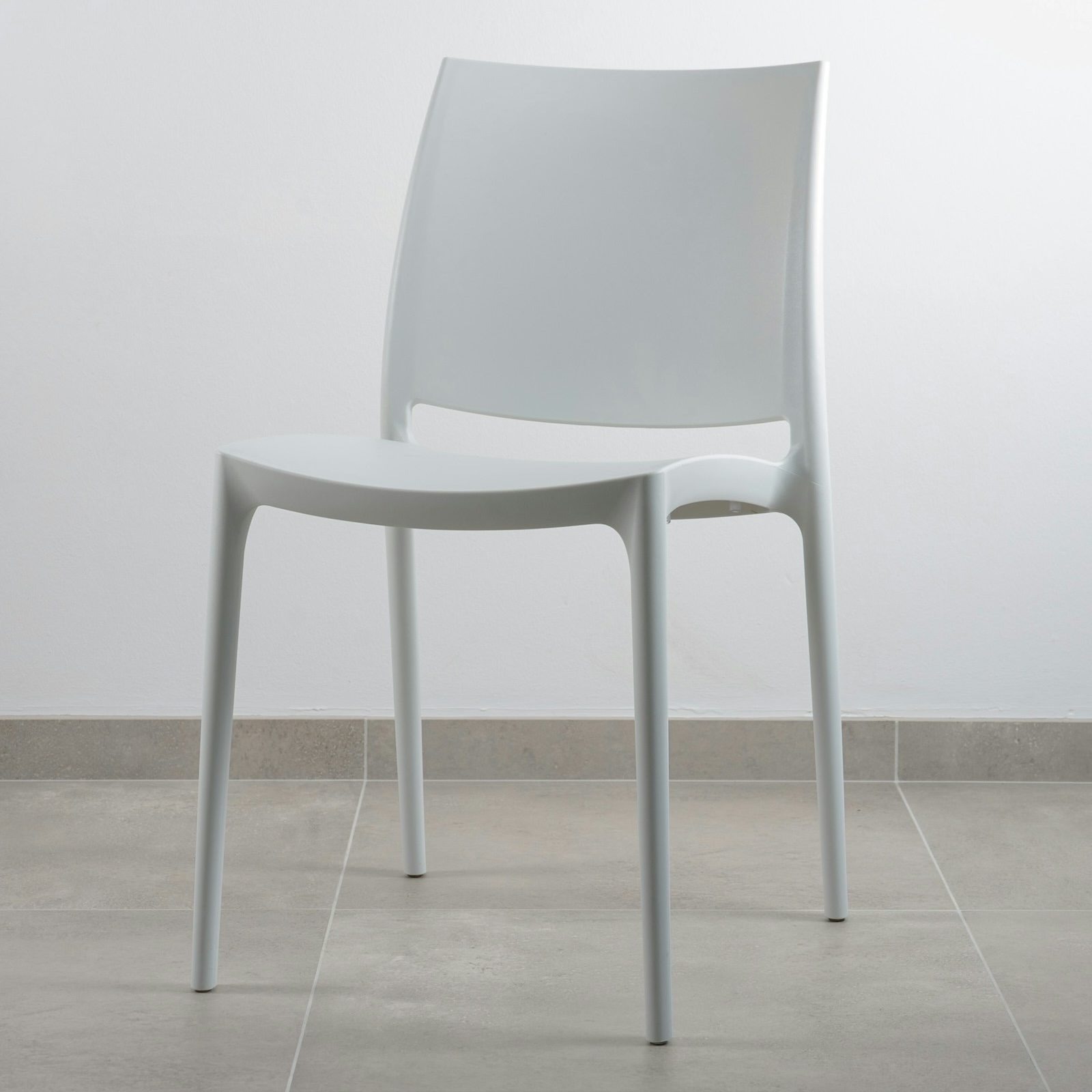} &
            \includegraphics[width=0.175\textwidth]{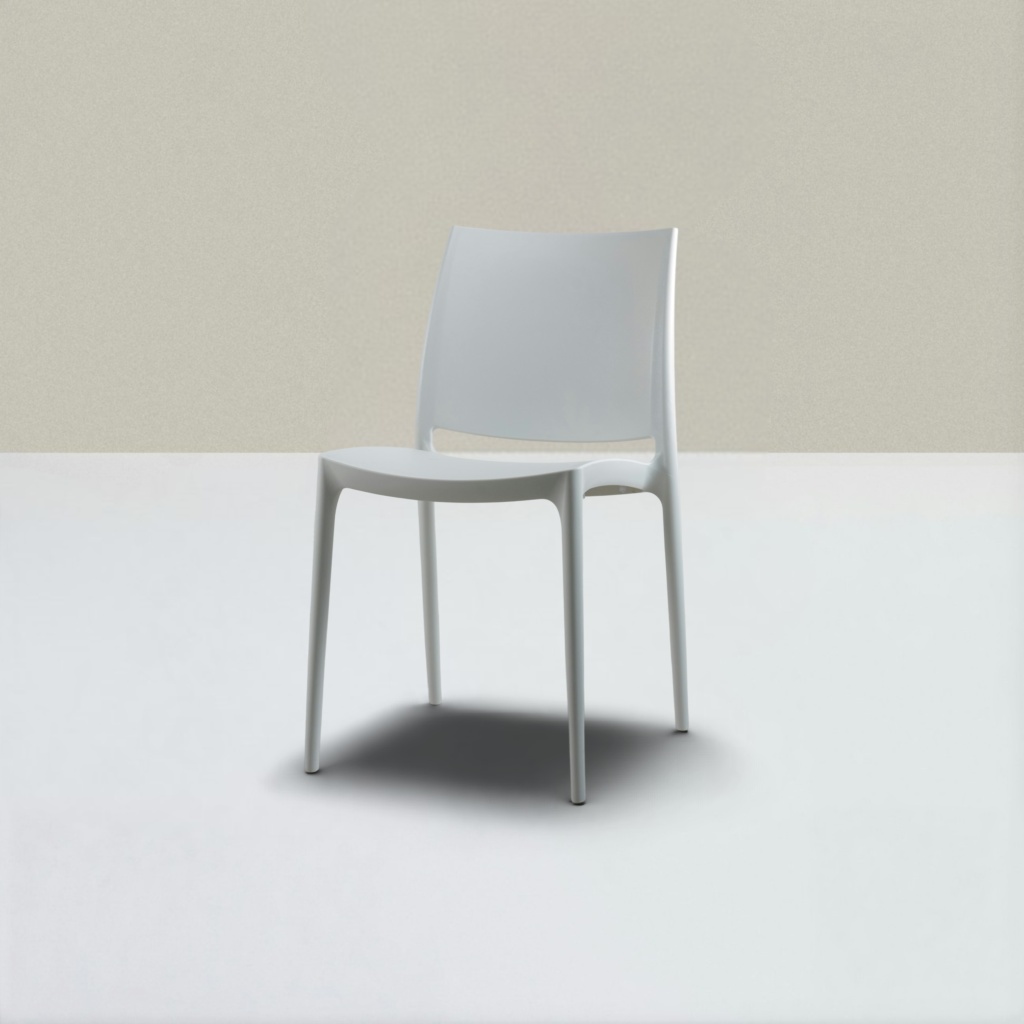} &
            \includegraphics[width=0.175\textwidth]{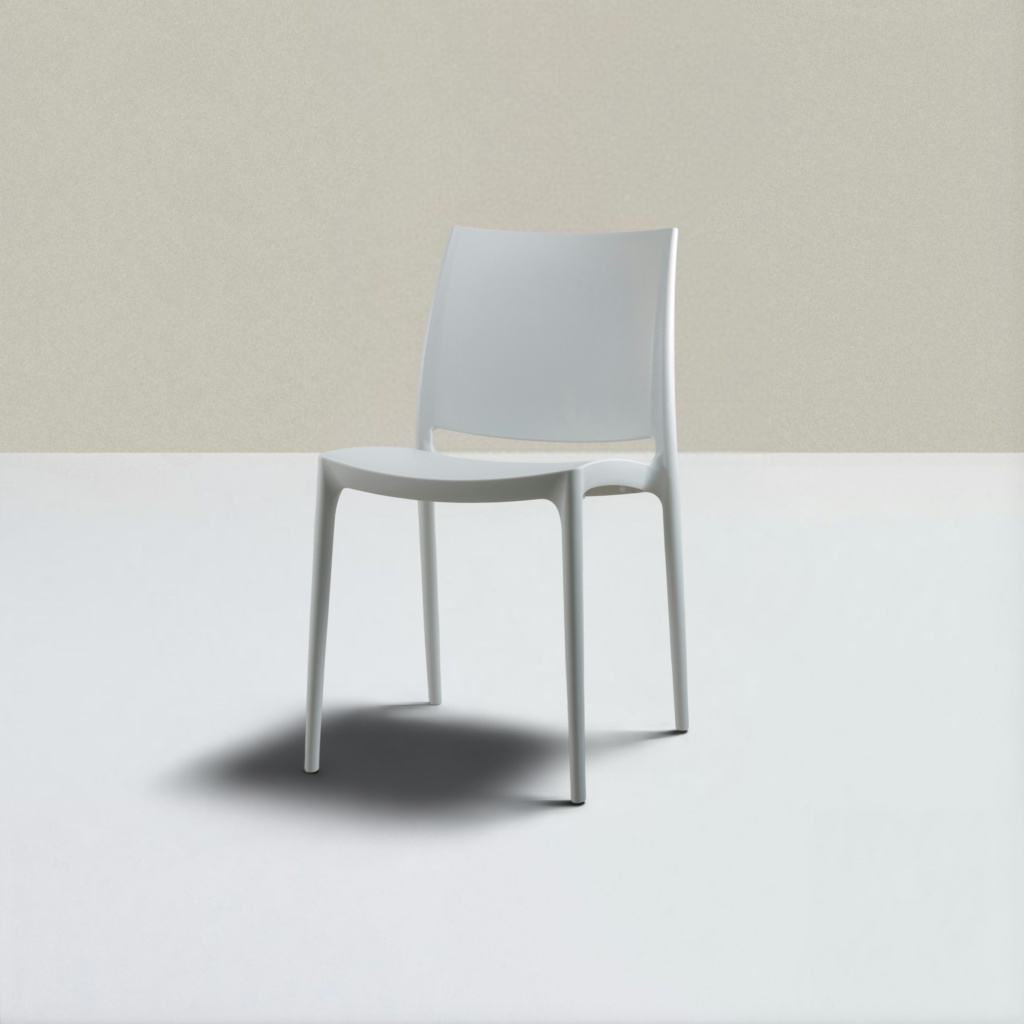} & 
            \includegraphics[width=0.175\textwidth]{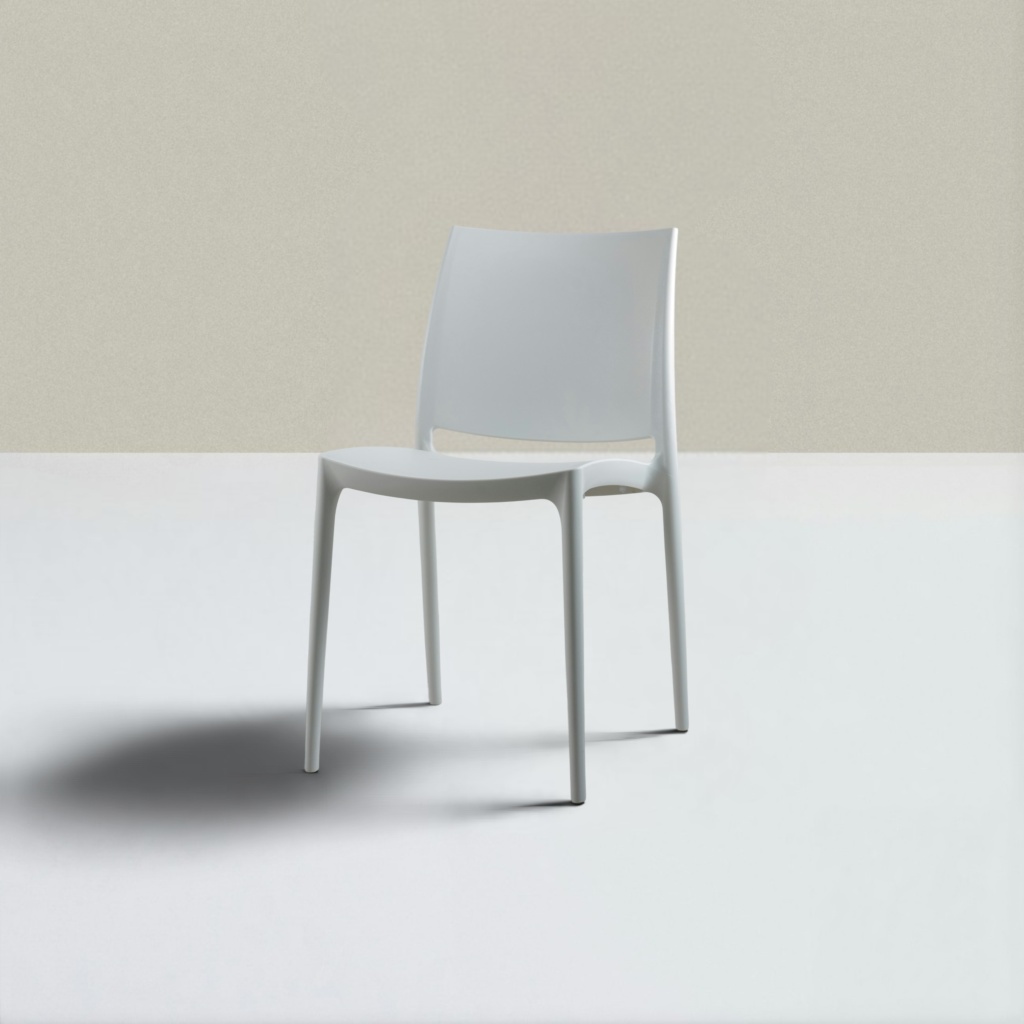} \\
    
            & & & & $\theta=0^\circ$ & $\theta=20^\circ$ & $\theta=40^\circ$
        \end{tabular}
        \caption{Shadow prediction on real images by using our method when controlling the softness $s$ and position $(\theta,\phi)$ of the light source.}
        \label{fig:shadow_gen_on_real_images}
    \end{table*}
    
    We compare our conditioning mechanism with other alternatives. SSN~\cite{ssn} proposes to represent the light sources as a mixture of Gaussian, whose amplitude and variance encode the softness $s$ and intensity $I$ of each light source. We adapt this idea to our setting to represent the light source as a Gaussian \emph{blob} in a $1024 \times 1024$ gray-scale image. 
    
    
    
    
    We generate a 2D Gaussian blob with a fixed size, resize it by a factor of $s$ to represent the softness through the Gaussian size, and position it on a black image using the cartesian coordinates of the light source ($x$, $y$) computed from its spherical coordinates as ($x = r \sin(\theta) \cos(\phi)$, $y = r \sin(\theta) \sin(\phi)$). Fig.~\ref{fig:blob_examples} illustrates example renders and the corresponding blob light maps. To condition the denoiser on the light map, we resize it to match the latent size and concatenate it with the noisy latent and the VAE embeddings of the object image and the resized mask, increasing the number of channels in the latent space to $2c + 2$.
    However, our conditioning approach presented in Sec.~\ref{sec:method:subsec:shadow_generation_pipeline:subsubsec:shadow_gen} is simpler as it directly injects scalars to the denoiser rather than using an external and complex light representation. It also does not require adding an extra channel in the latent space. However, interestingly, its performance is roughly on par with the model utilizing an external blob representation for the same number of training iterations with rectified flow and 1 sampling step, as demonstrated in Fig.~\ref{fig:model_comparison}.
    
    \begin{figure}[t]
        \centering
        \begin{subfigure}[b]{0.24\linewidth}
            \centering
            \includegraphics[width=\linewidth]{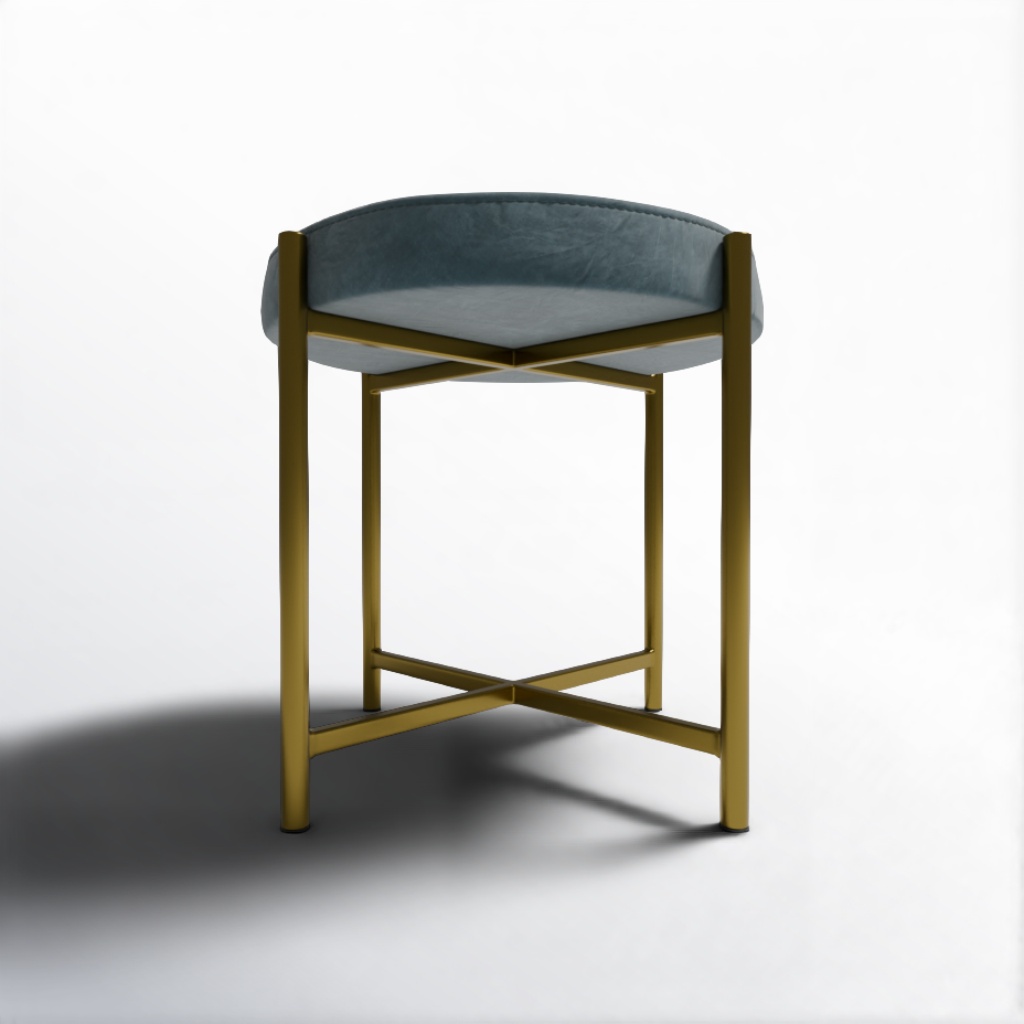}
        \end{subfigure}
        \hfill
        \begin{subfigure}[b]{0.24\linewidth}
            \centering
            \includegraphics[width=\linewidth]{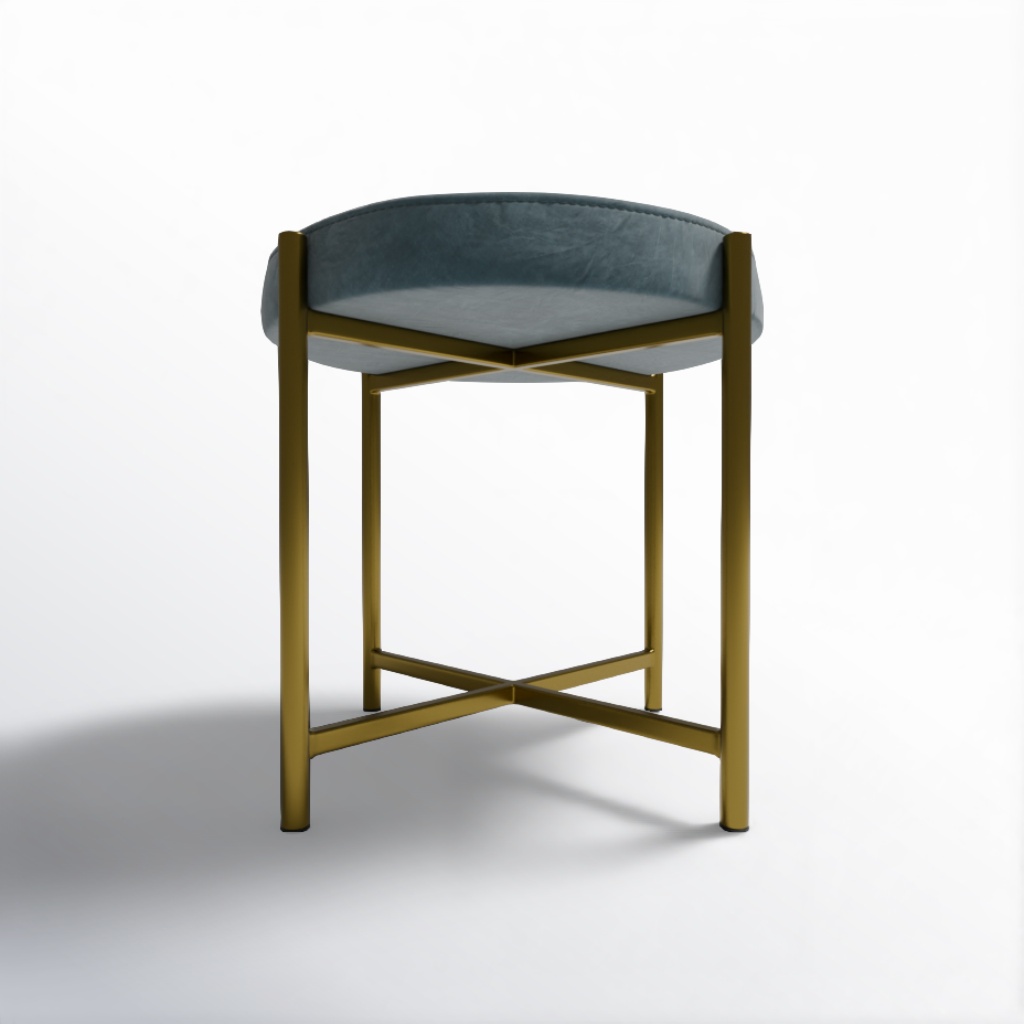}
        \end{subfigure}
        \hfill
        \begin{subfigure}[b]{0.24\linewidth}
            \centering
            \includegraphics[width=\linewidth]{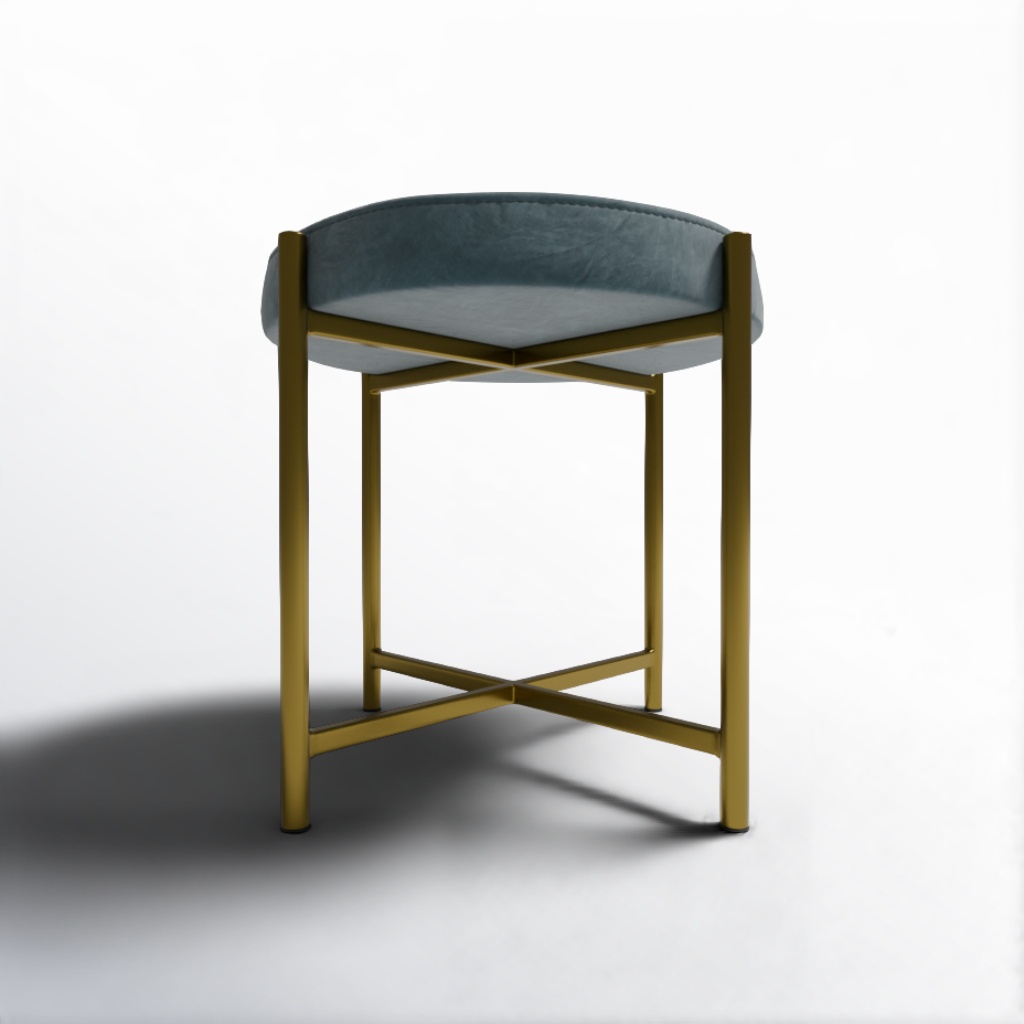}
        \end{subfigure}
        \hfill
        \begin{subfigure}[b]{0.24\linewidth}
            \centering
            \includegraphics[width=\linewidth]{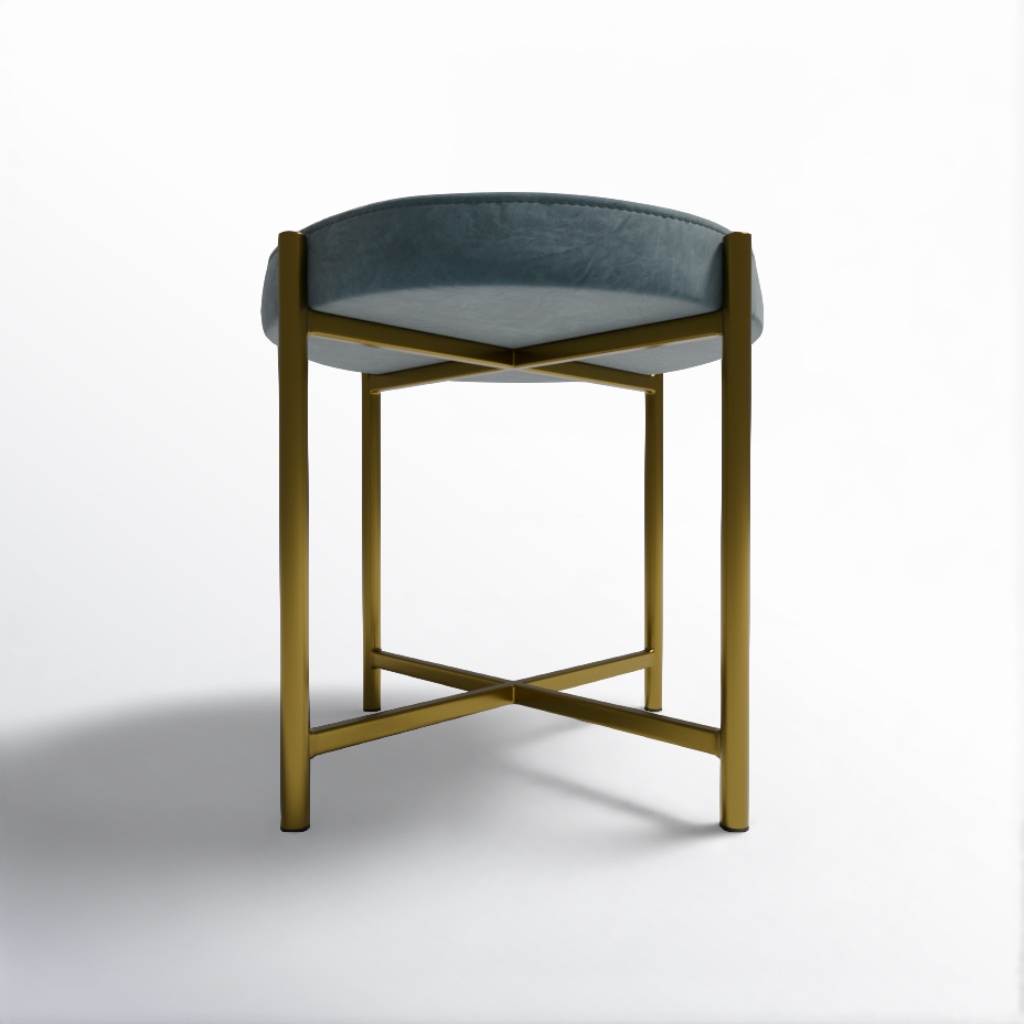}
        \end{subfigure}
        \caption{Intensity Control. The first two and the last two images show results for the models with $\mathcal{S}(\theta, \phi, s)$ and $\mathcal{S}(\theta, \phi, s, I)$ conditionings, respectively. We multiply the shadow map predicted by the first model by $I$ to control the shadow intensity. $I=1$ in the 1st and the 3rd images, $I=0.5$ in the 2nd and the 4th images.}
        \label{fig:intensity_control}
    \end{figure}
    
    \subsection{Intensity Conditioning}
    \label{sec:experiments:subs:intensity}
     While shadow intensity can be easily adjusted by multiplying the predicted shadow with a fixed intensity by a scalar $I$, we explore whether the intensity can be controlled in the same way as the other light parameters $\mathcal{S}(\theta, \phi, s)$ with our conditioning mechanism. When training the model with $\mathcal{S}(\theta, \phi, s, I)$ conditioning, we multiply the shadow maps from the training set by a random $I$ value, uniformly sampled between $0.1$ and $1.9$, in each iteration and use these modified maps as the ground truth. The output from the model conditioned on $\mathcal{S}(\theta, \phi, s)$ and scaled by $I$ is nearly identical to the output from the model conditioned on $\mathcal{S}(\theta, \phi, s, I)$ as shown in Fig.~\ref{fig:intensity_control}. This ablation confirms that our framework is flexible enough to easily incorporate additional shadow controls.

    \subsection{Qualitative Analysis on Real Images}
    \label{sec:experiments:subs:real_images}
    
    Given that our model is trained with fully synthetic data, the goal of this section is to assess whether it generalizes well to real images. To this end, we collected a set of foreground object images, as well as target backgrounds. We conduct a qualitative analysis for softness control, horizontal and vertical shadow direction control. We also seek to understand whether the predicted shadows are visually appealing, and their geometry aligns with the object as for synthetic data.
    
    We run our full pipeline, involving the model trained for $150k$ iterations, with rectified flow using only 1 inference step. For the softness control, we set the angles $\theta$ and $\phi$ to $30^\circ$ and $60^\circ$ (see Fig.~\ref{fig:spherical_coordinate_system}), respectively, while varying the parameter $s$ among the values $2$, $5$, and $8$. As illustrated in the top row of Table~\ref{fig:shadow_gen_on_real_images}, the predicted shadow softens progressively as the value of $s$ increases.
    
    For the horizontal shadow direction control, we set $\theta$ to $30^\circ$ and $s$ to $2$. We then move the light source horizontally along the surface of the sphere at angles of $\phi = 45^\circ$, $\phi = 135^\circ$, and $\phi = 315^\circ$. The middle row illustrates that the predicted shadow's direction corresponds with the varying degrees of $\phi$. For the analysis of vertical shadow direction control, we enforce the model to predict the shadow always on the left with a fixed softness by setting $\phi=0^\circ$ and $s=2$, and predict shadows by positioning the light source directly above the object at an angle of $\theta=0^\circ$, then move it along the sphere to the right at $\theta=20^\circ$ and $\theta=40^\circ$. As illustrated at the bottom row in Table~\ref{fig:shadow_gen_on_real_images}, the shadow initially falls directly beneath the object and extends to the left as we increase the value of $\theta$ and demonstrates as well that the predicted shadow aligns geometrically with the object.
    
    These observations indicate that our model, trained entirely on our synthetic datasets, demonstrates strong generalization capabilities when tested with real images. We provide more qualitative results in SM~Sec.~\ref{sm:sec:results_on_real_images}.

%% file: sections/conclusion.tex
\section{Conclusion}
\label{sec:conclusion}

We presented a novel method for fast, controllable, and background-free shadow generation for object images. To achieve this, we created a large synthetic dataset using a 3D rendering engine, generating diverse shadow maps with varying light parameters. We then trained a diffusion model with a rectified flow objective enabling fast and high-quality shadow generation while providing control over shadow direction, softness, and intensity. Furthermore, we demonstrated that our model generalizes well to real-world images, producing realistic and controllable shadows without introducing any artifacts or color shifts on the background. To enable further research in this area, we also released a public benchmark dataset containing a diverse set of object images and their corresponding shadow maps under various settings. By harnessing the power of synthetic data and diffusion models, we opened up new possibilities for creative content creation.

%% file: sections/sm.tex
\clearpage

\setcounter{page}{1}

\twocolumn[{%
\renewcommand\twocolumn[1][]{#1}%
\maketitlesupplementary
\begin{center}
    \centering
    \captionsetup{type=figure}
    \includegraphics[width=\textwidth]{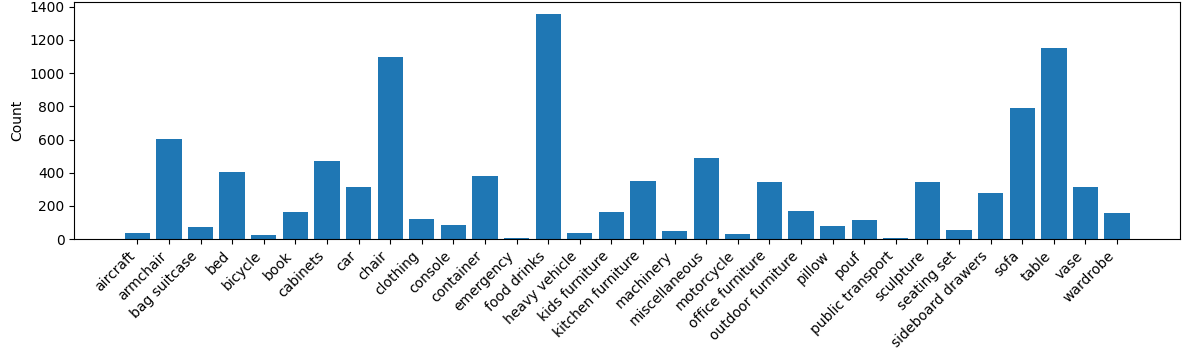}
    \captionof{figure}{Number of 3D meshes for each category in our synthetic dataset.}
    \label{fig:sm_num_meshes_per_cat}
\end{center}
}]



\begin{figure*}[t]
    \centering

    \begin{subfigure}{0.16\textwidth} 
        \centering
        \includegraphics[width=\textwidth]{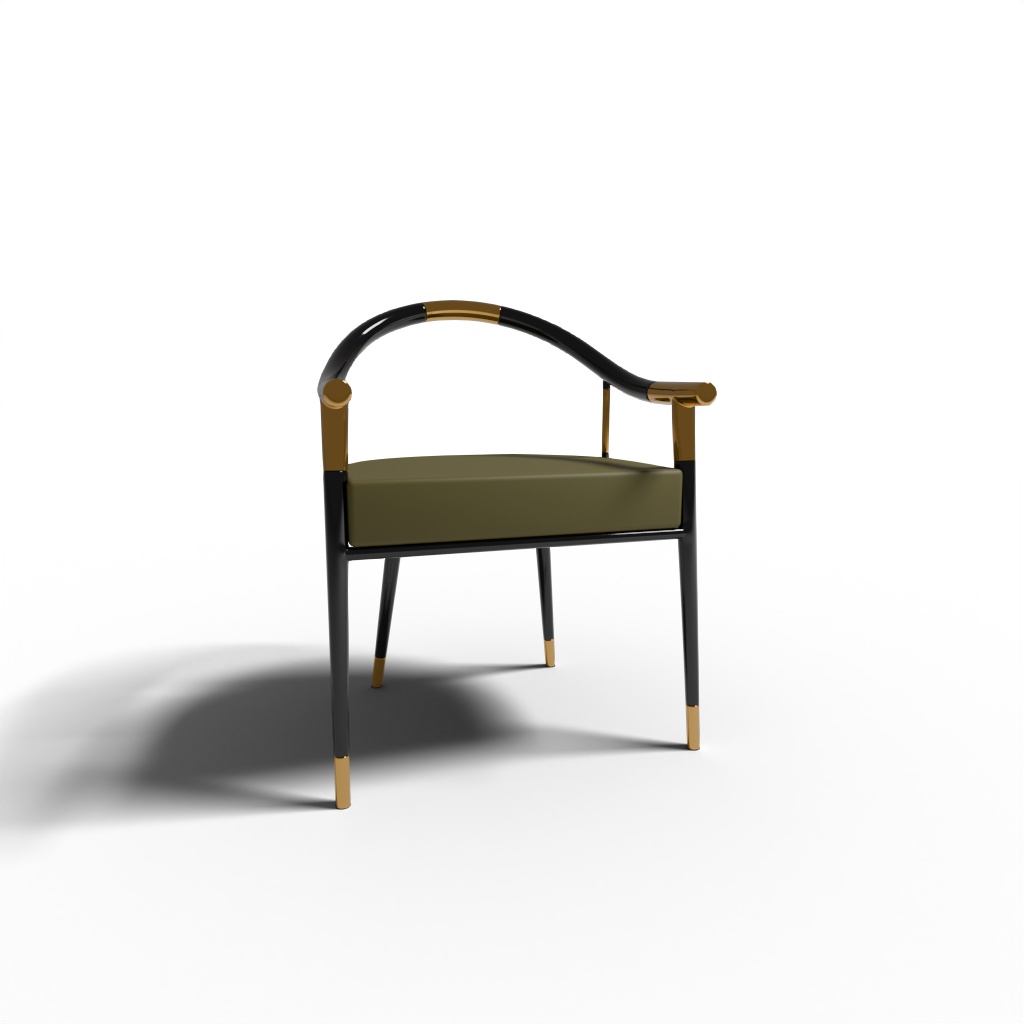} 
        \caption{$\phi=0^\circ$}
    \end{subfigure}
    \hfill
    \begin{subfigure}{0.16\textwidth} 
        \centering
        \includegraphics[width=\textwidth]{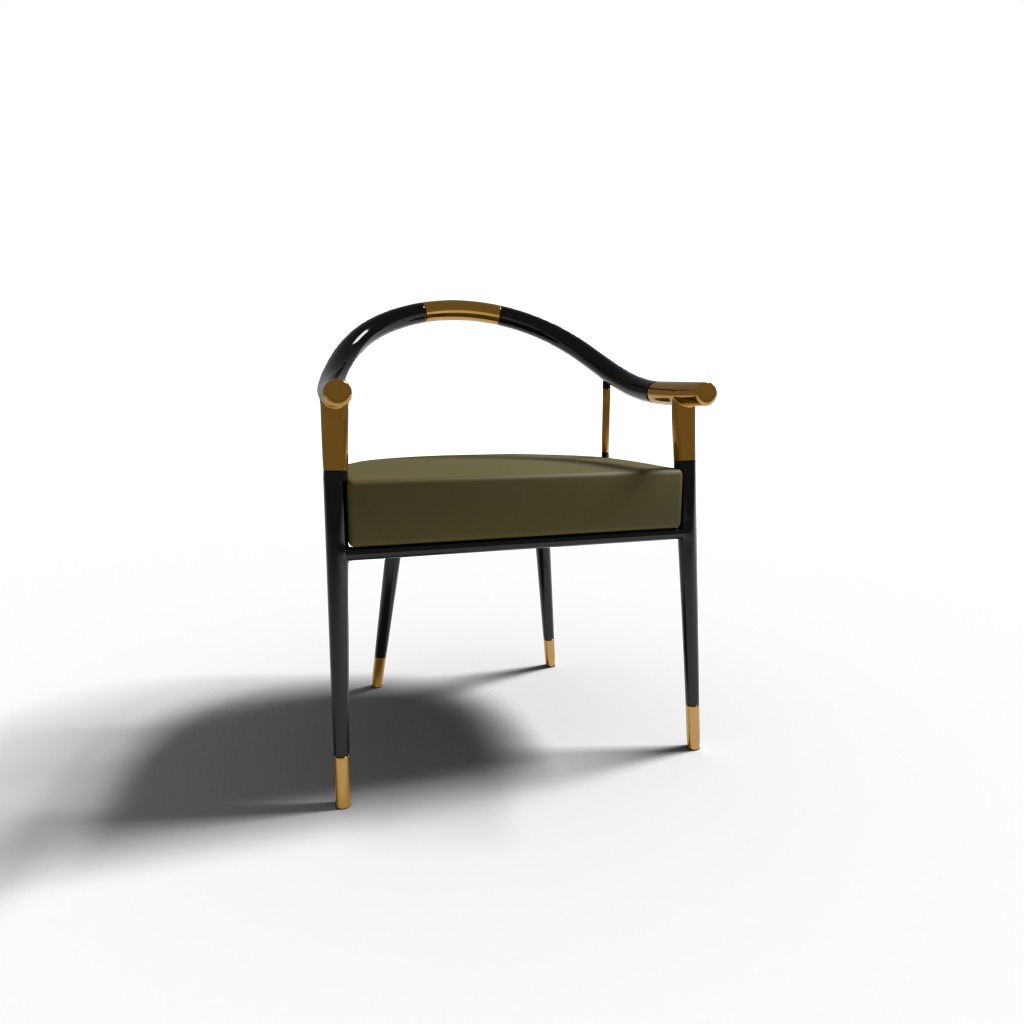}
        \caption{$\phi=20^\circ$}
    \end{subfigure}
    \hfill
    \begin{subfigure}{0.16\textwidth} 
        \centering
        \includegraphics[width=\textwidth]{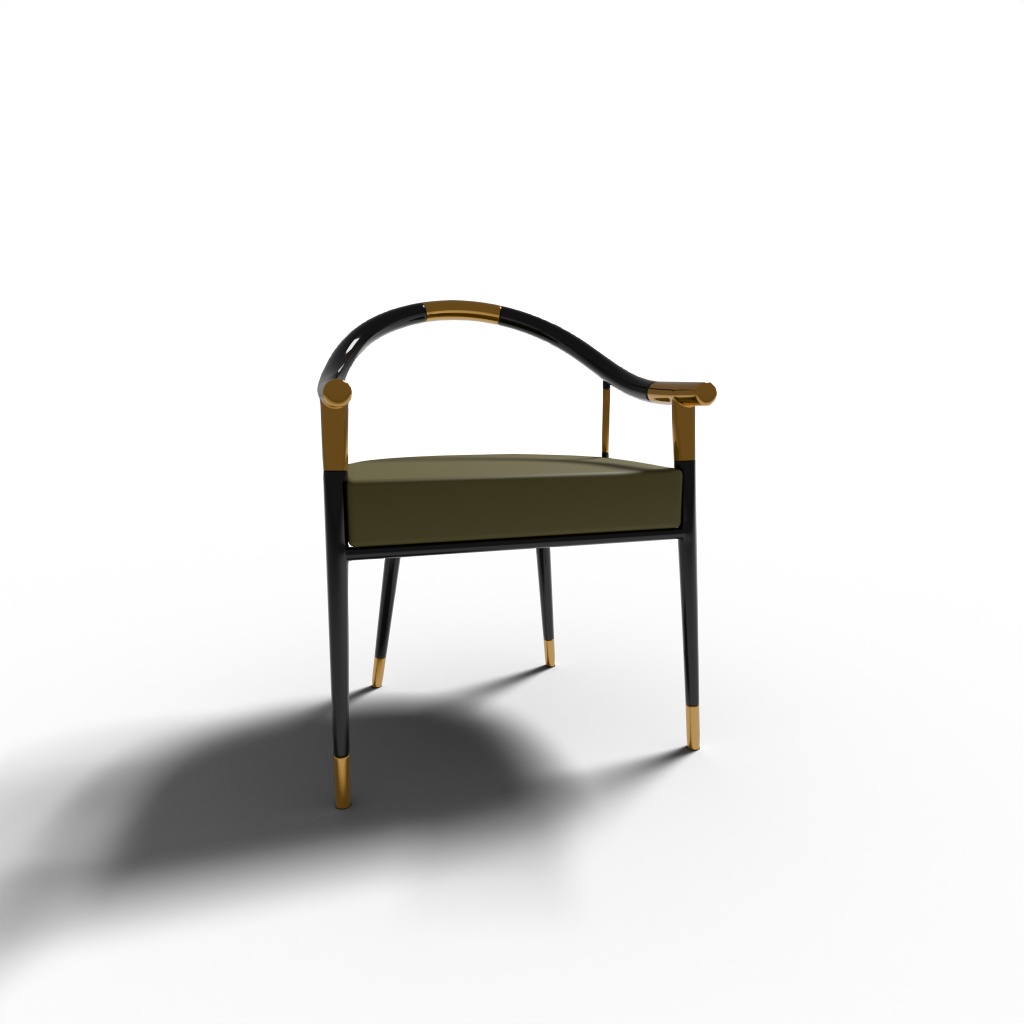}
        \caption{$\phi=40^\circ$}
    \end{subfigure}
    \hfill
    \begin{subfigure}{0.16\textwidth} 
        \centering
        \includegraphics[width=\textwidth]{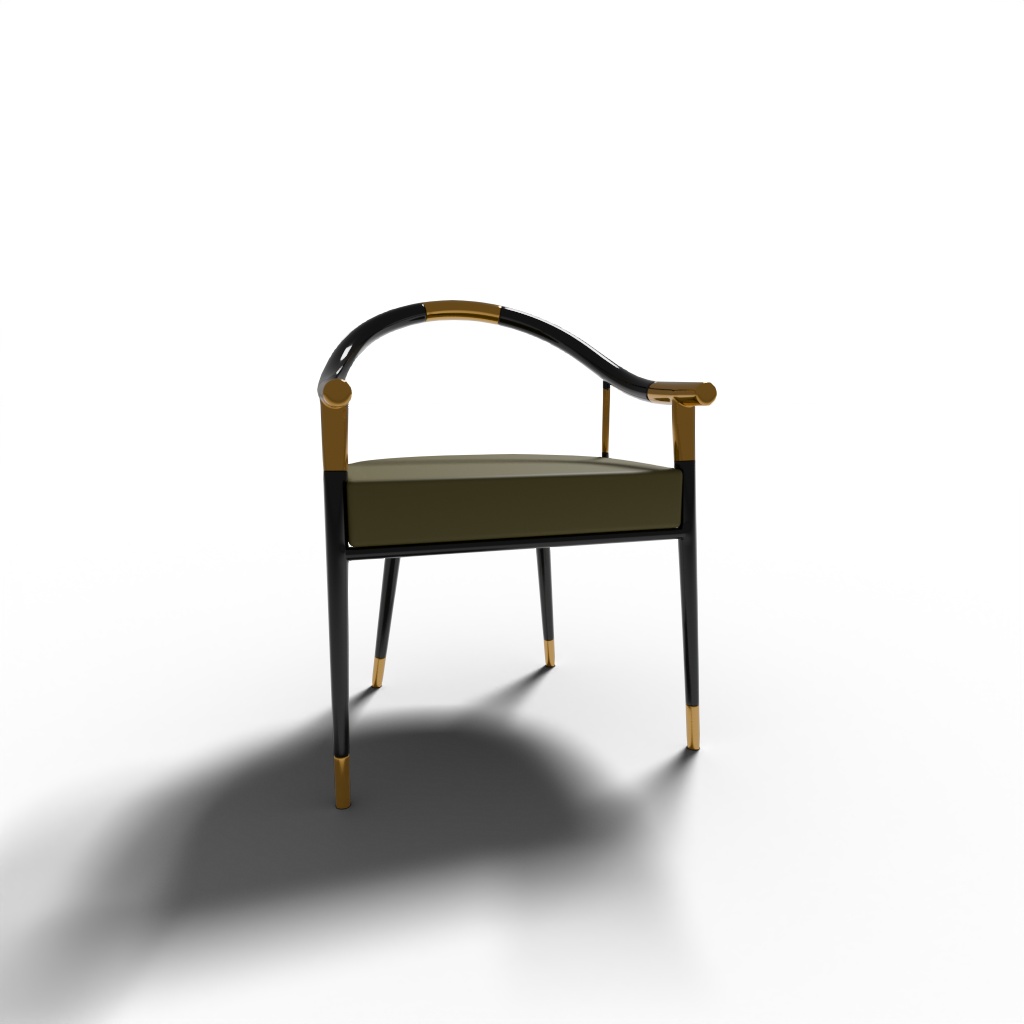}
        \caption{$\phi=60^\circ$}
    \end{subfigure}
    \hfill
    \begin{subfigure}{0.16\textwidth} 
        \centering
        \includegraphics[width=\textwidth]{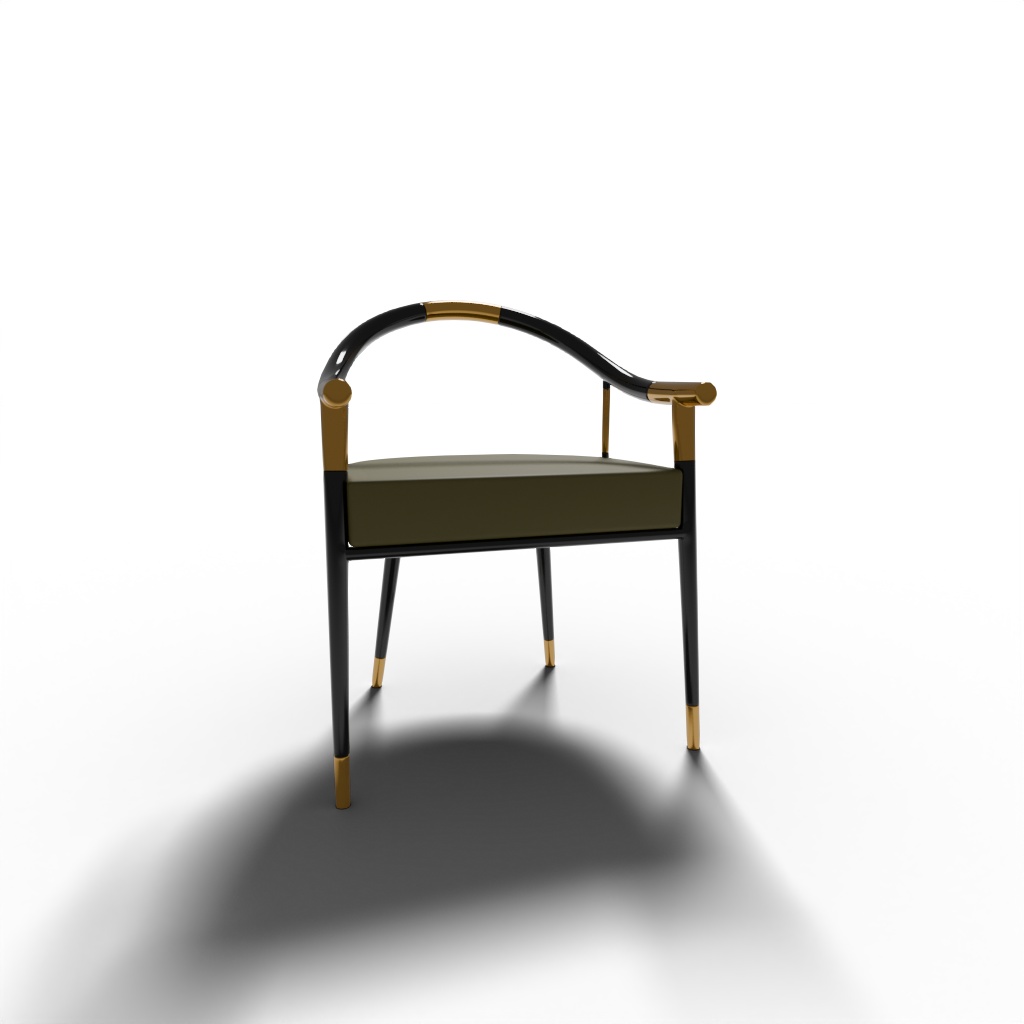}
        \caption{$\phi=80^\circ$}
    \end{subfigure}
    \hfill
    \begin{subfigure}{0.16\textwidth} 
        \centering
        \includegraphics[width=\textwidth]{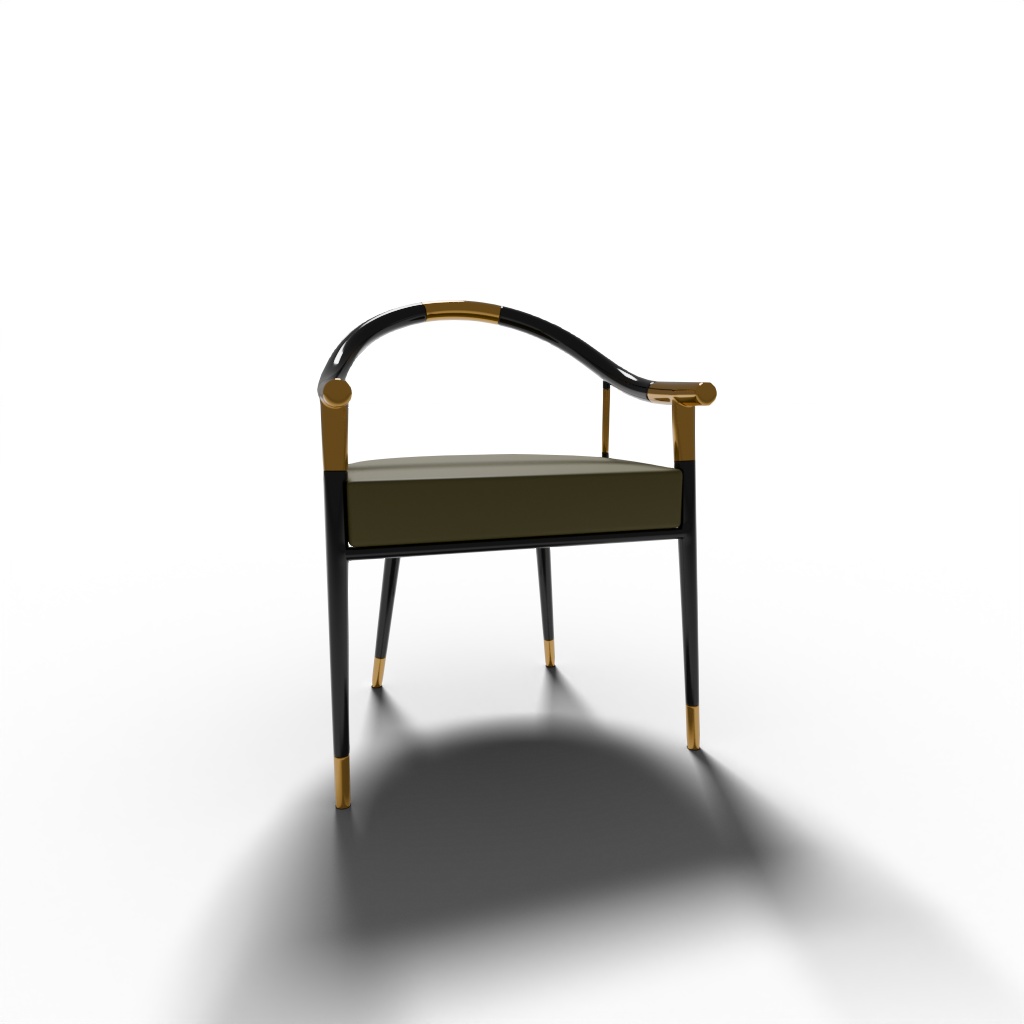}
        \caption{$\phi=100^\circ$}
    \end{subfigure}
    
    \begin{subfigure}{0.16\textwidth} 
        \centering
        \includegraphics[width=\textwidth]{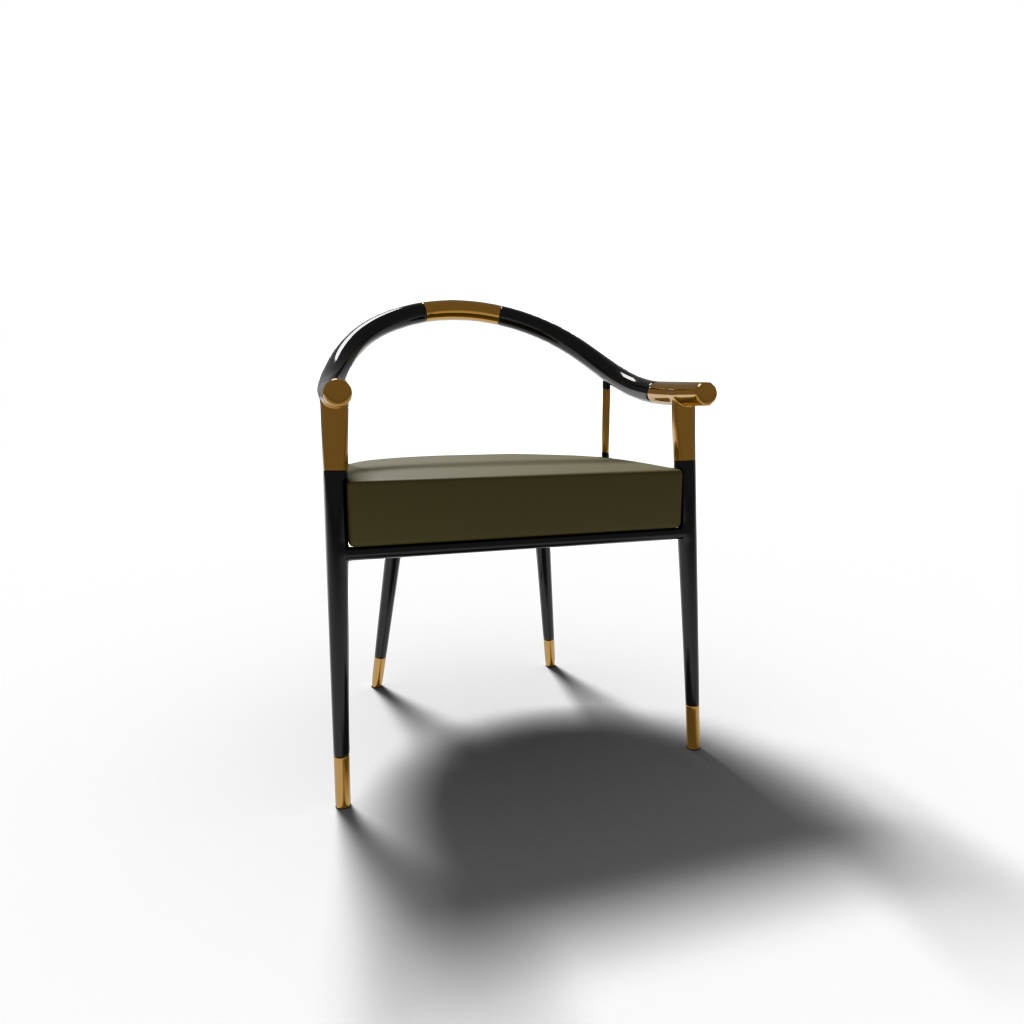}
        \caption{$\phi=120^\circ$}
    \end{subfigure}
    \hfill
    \begin{subfigure}{0.16\textwidth} 
        \centering
        \includegraphics[width=\textwidth]{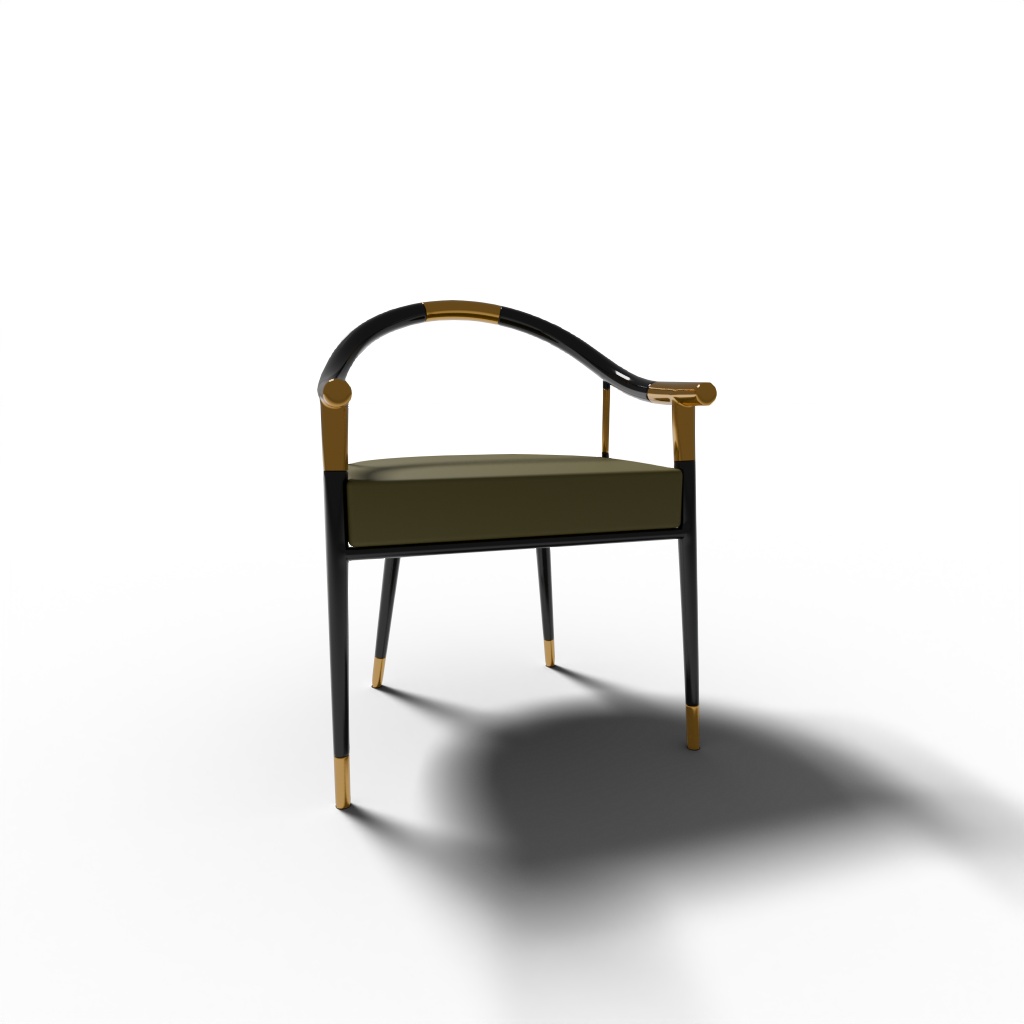}
        \caption{$\phi=140^\circ$}
    \end{subfigure}
    \hfill
    \begin{subfigure}{0.16\textwidth} 
        \centering
        \includegraphics[width=\textwidth]{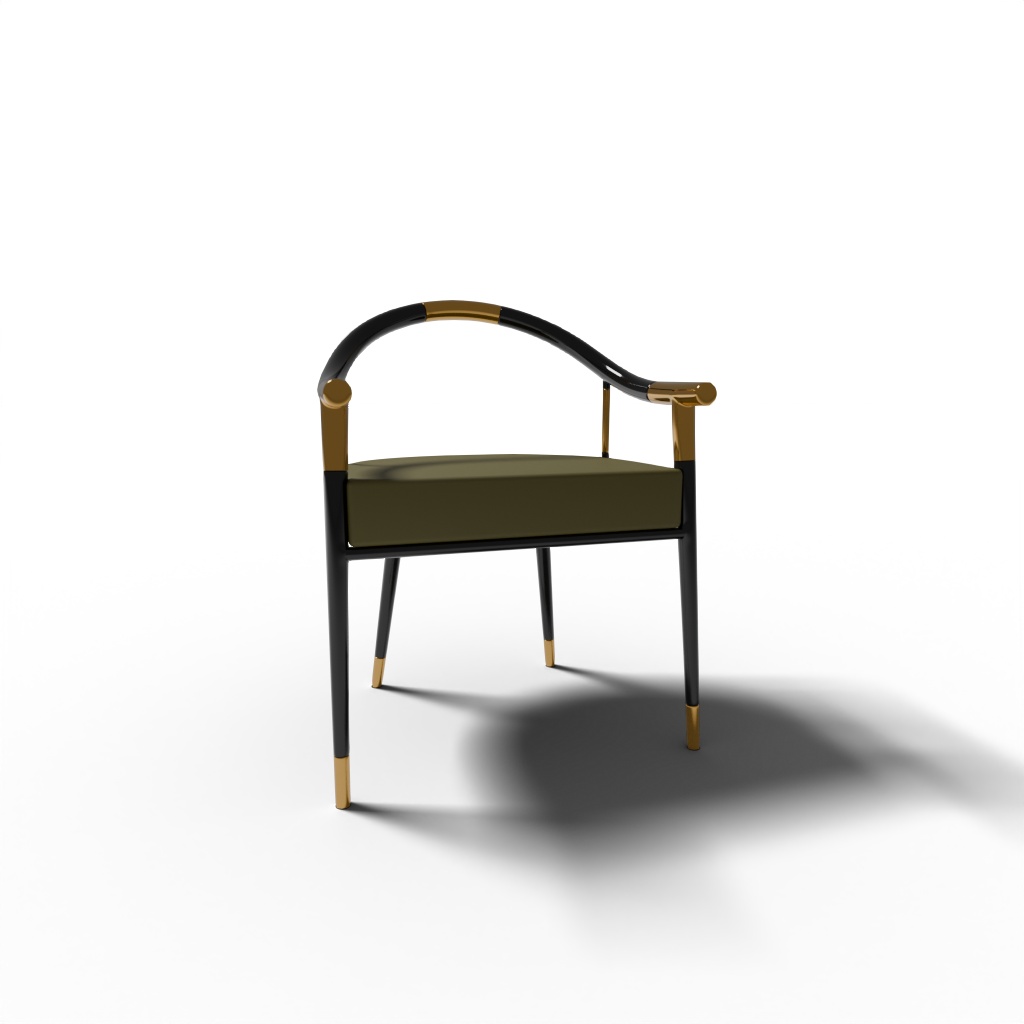}
        \caption{$\phi=160^\circ$}
    \end{subfigure}
    \hfill
    \begin{subfigure}{0.16\textwidth} 
        \centering
        \includegraphics[width=\textwidth]{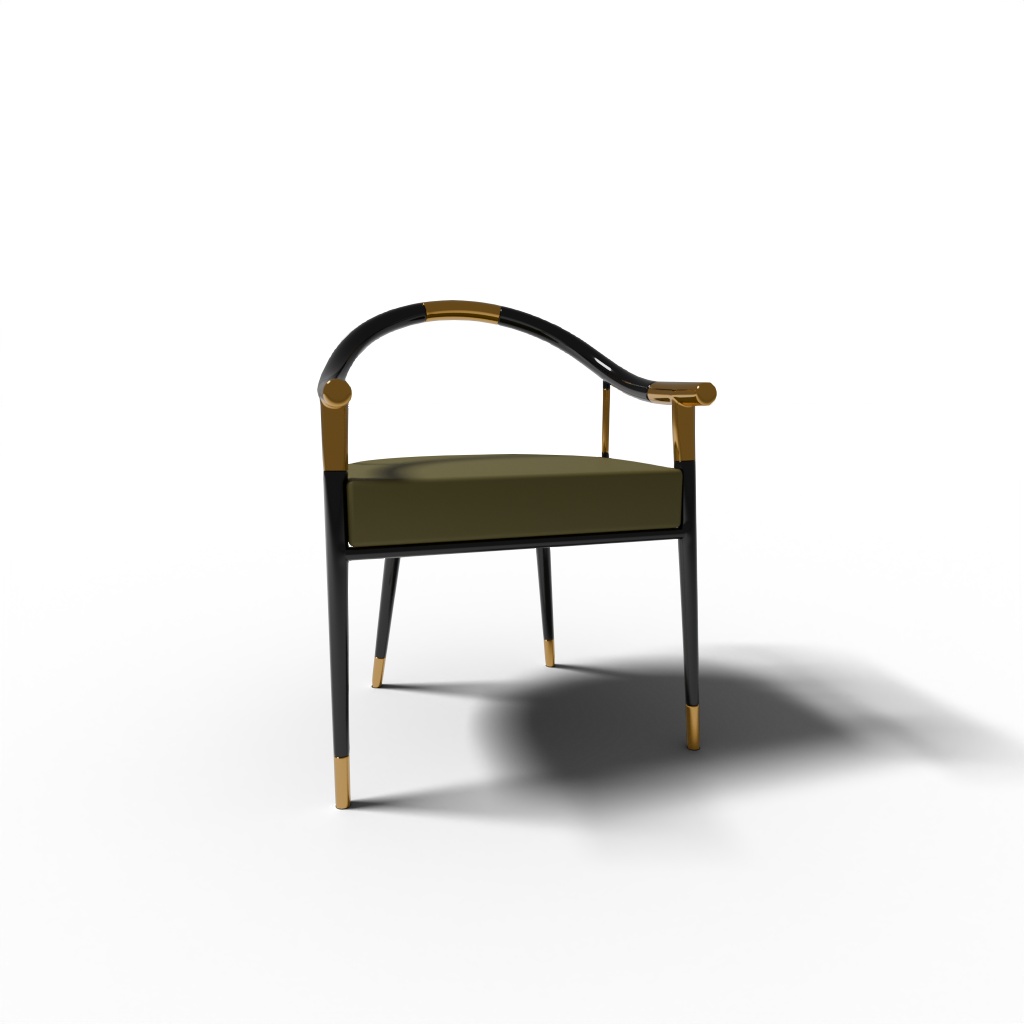}
        \caption{$\phi=180^\circ$}
    \end{subfigure}
    \hfill
    \begin{subfigure}{0.16\textwidth} 
        \centering
        \includegraphics[width=\textwidth]{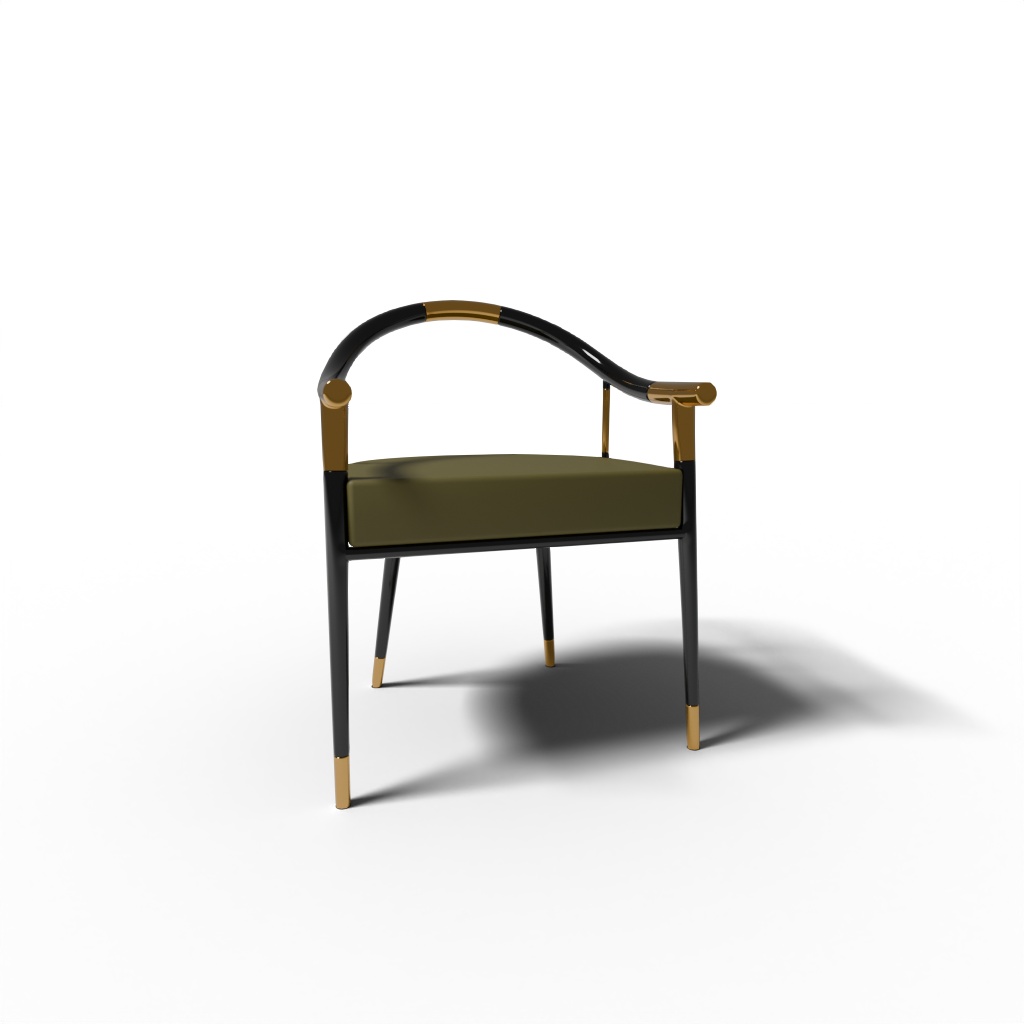}
        \caption{$\phi=200^\circ$}
    \end{subfigure}
    \hfill
    \begin{subfigure}{0.16\textwidth} 
        \centering
        \includegraphics[width=\textwidth]{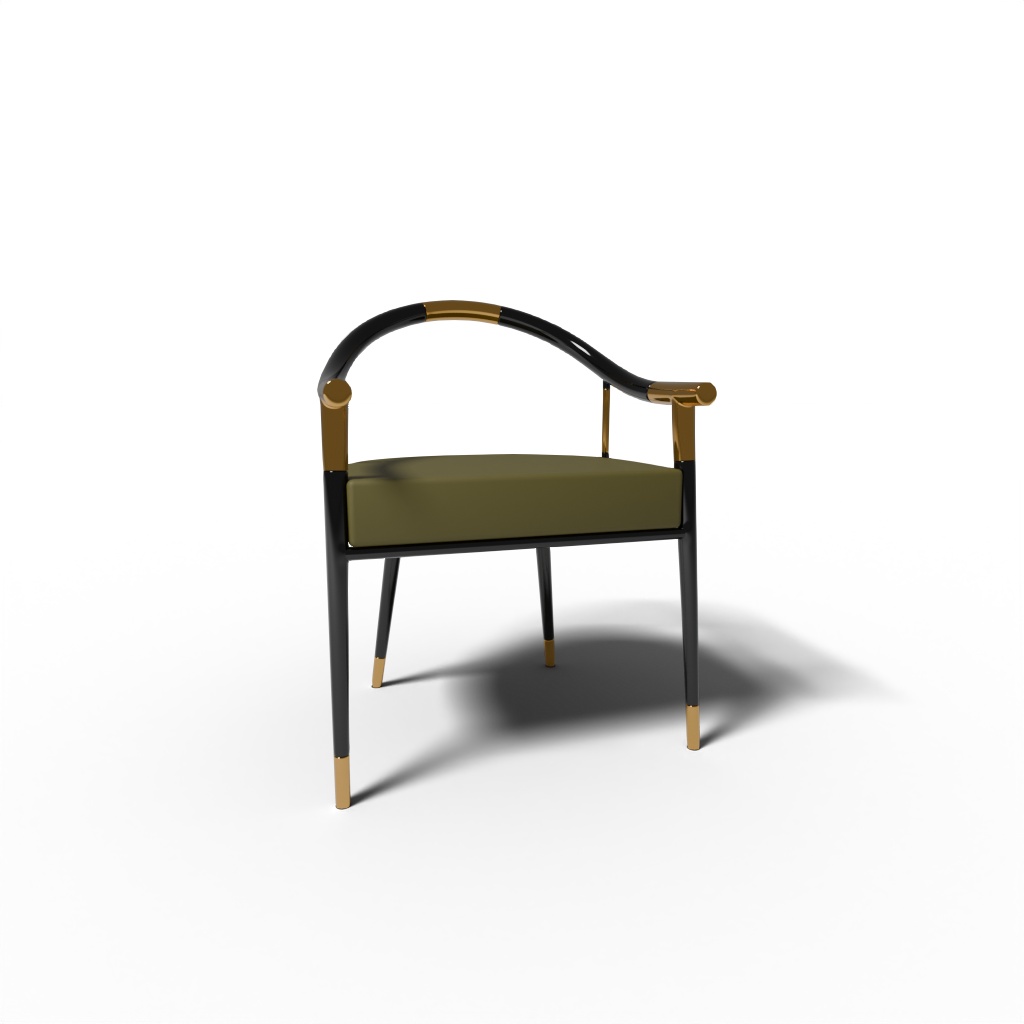}
        \caption{$\phi=220^\circ$}
    \end{subfigure}

    \begin{subfigure}{0.16\textwidth} 
        \centering
        \includegraphics[width=\textwidth]{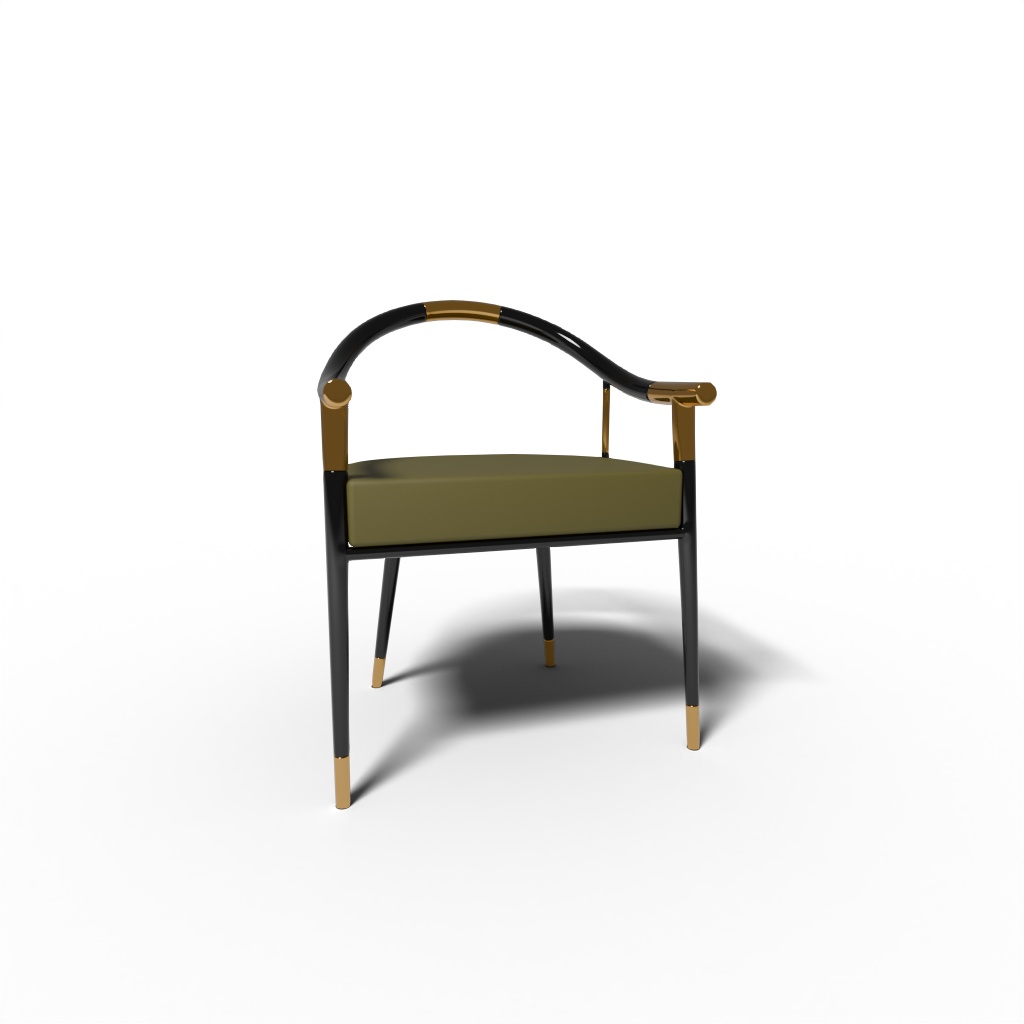}
        \caption{$\phi=240^\circ$}
    \end{subfigure}
    \hfill
    \begin{subfigure}{0.16\textwidth} 
        \centering
        \includegraphics[width=\textwidth]{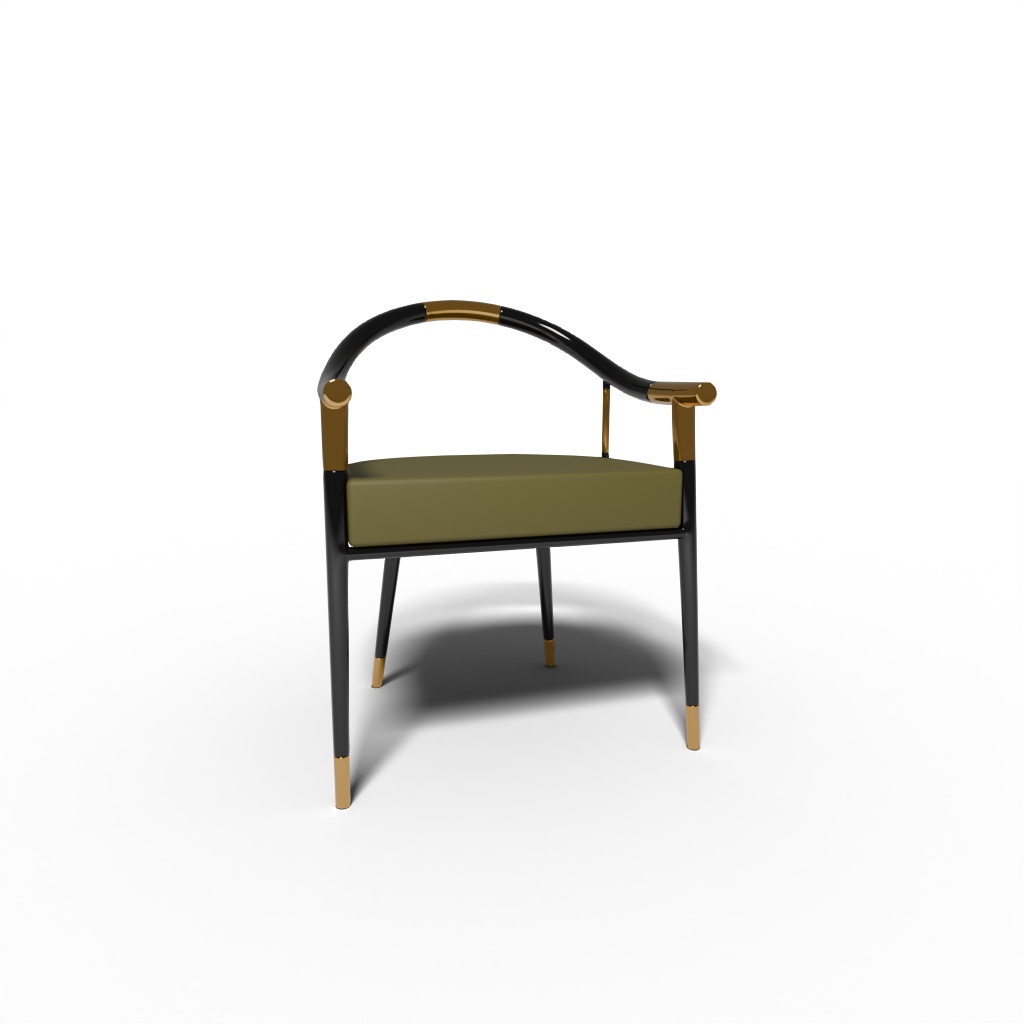}
        \caption{$\phi=260^\circ$}
    \end{subfigure}
    \hfill
    \begin{subfigure}{0.16\textwidth} 
        \centering
        \includegraphics[width=\textwidth]{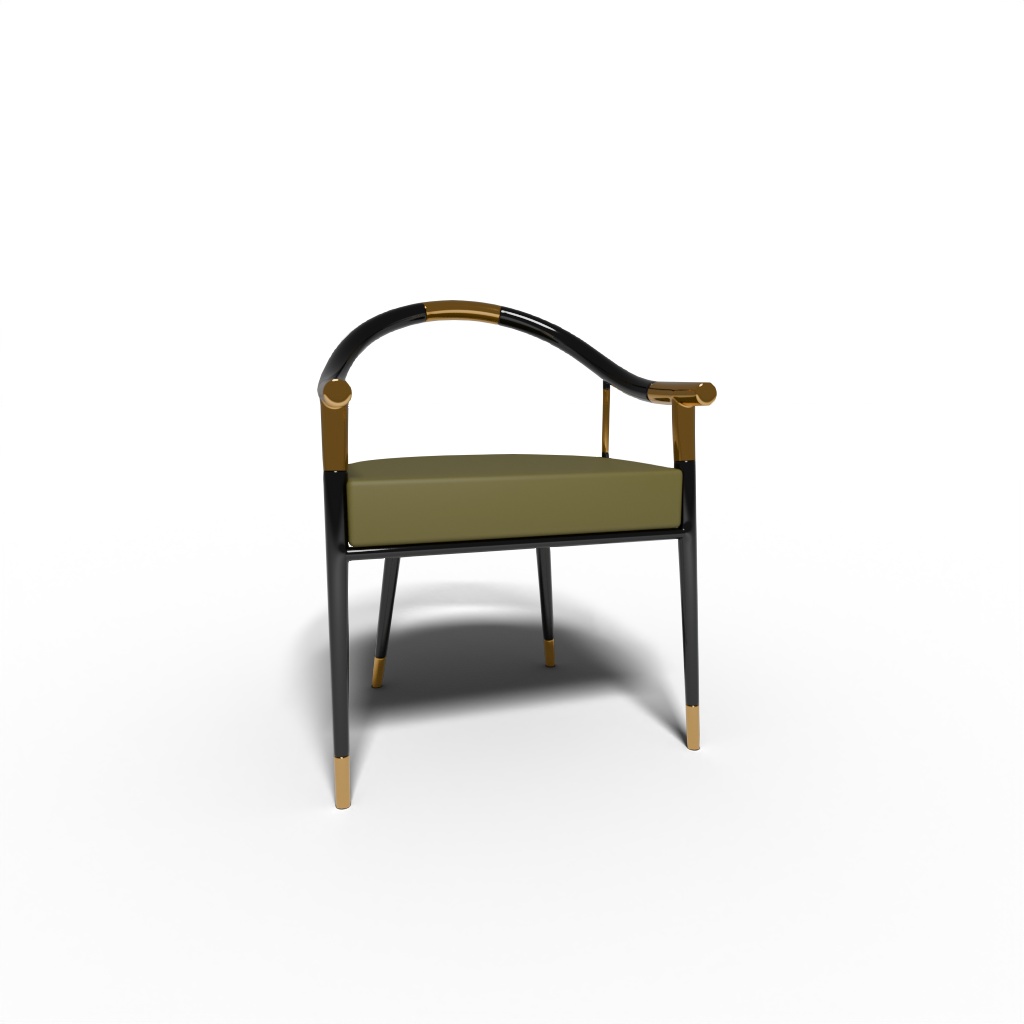}
        \caption{$\phi=280^\circ$}
    \end{subfigure}
    \hfill
    \begin{subfigure}{0.16\textwidth} 
        \centering
        \includegraphics[width=\textwidth]{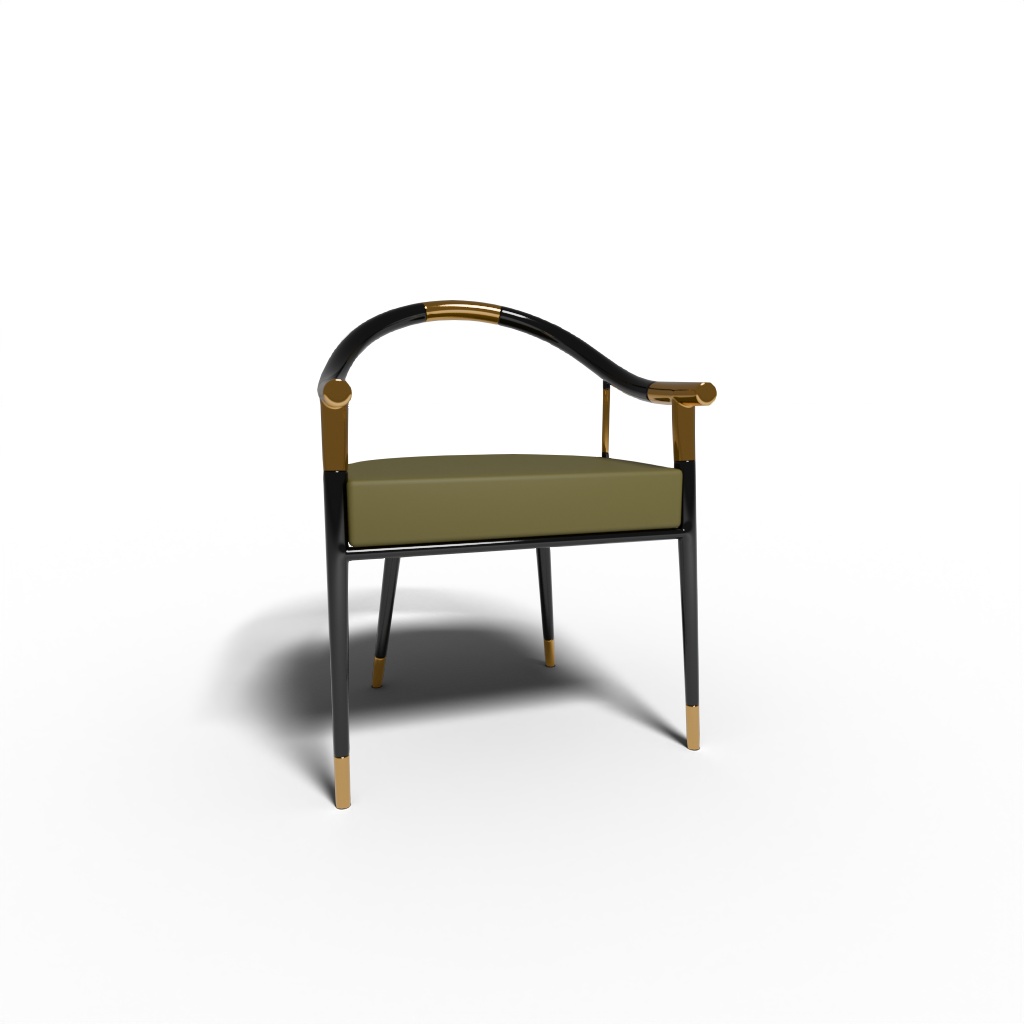}
        \caption{$\phi=300^\circ$}
    \end{subfigure}
    \hfill
    \begin{subfigure}{0.16\textwidth} 
        \centering
        \includegraphics[width=\textwidth]{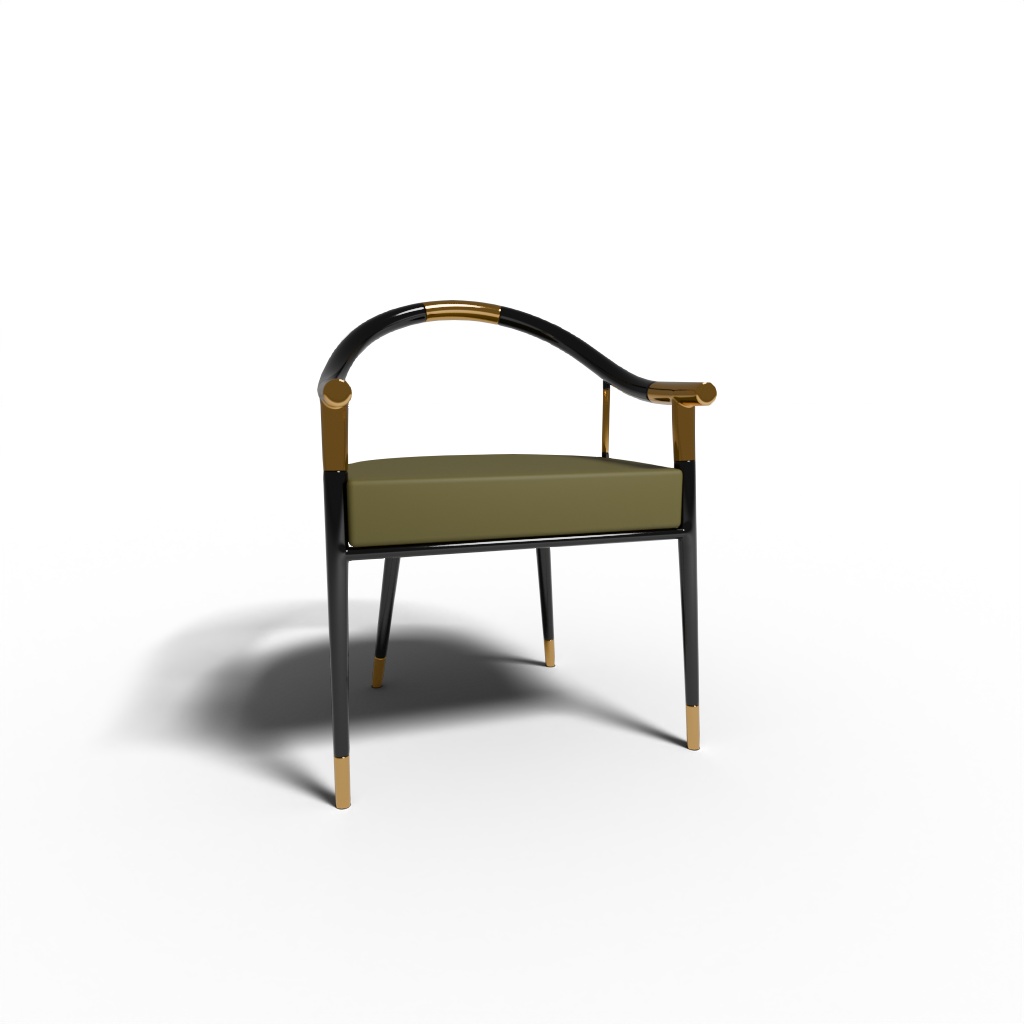}
        \caption{$\phi=320^\circ$}
    \end{subfigure}
    \hfill
    \begin{subfigure}{0.16\textwidth} 
        \centering
        \includegraphics[width=\textwidth]{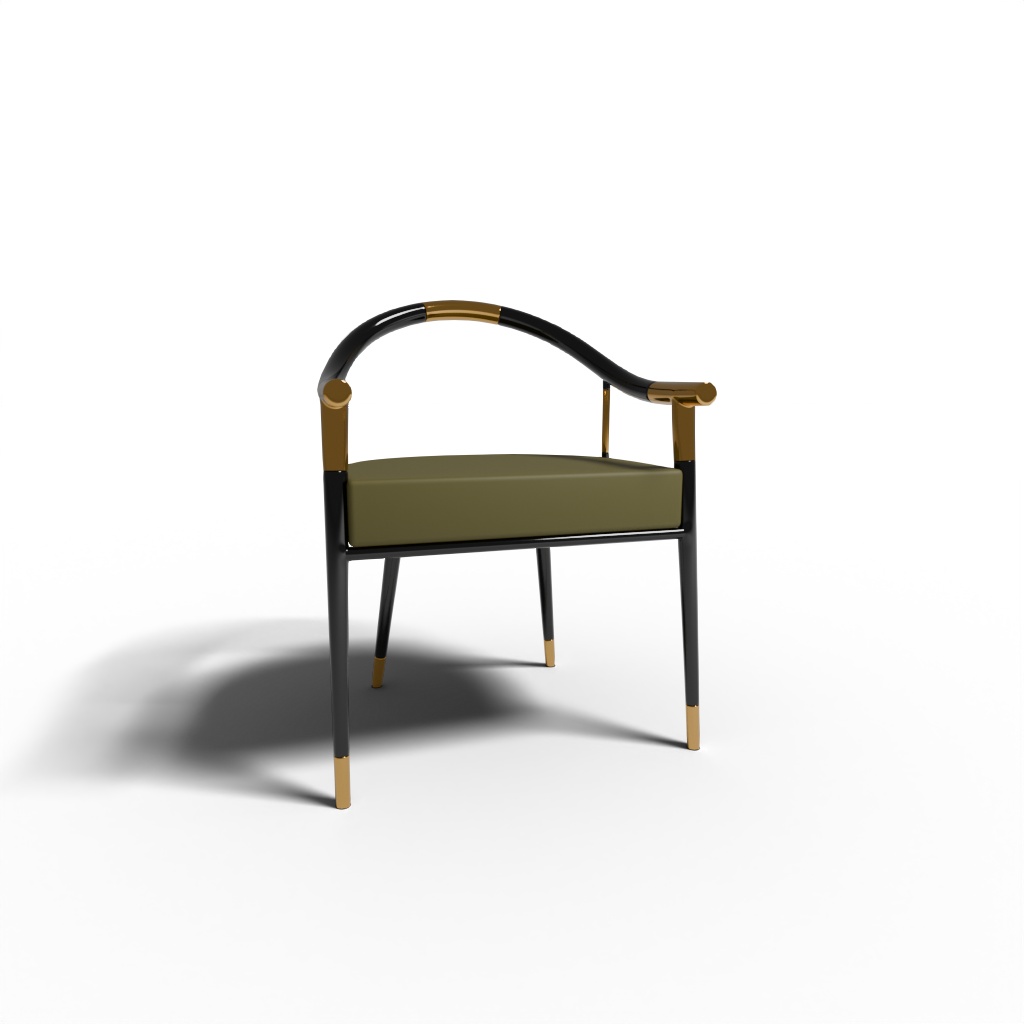}
        \caption{$\phi=340^\circ$}
    \end{subfigure}
    
    \caption{Renders for one 3D mesh from the horizontal shadow direction control track. $\theta=30^\circ$ and $s=2$.}
    \label{fig:sm_track_2}
\end{figure*}

\begin{figure*}[t]
    \centering

    \begin{subfigure}{0.16\textwidth} 
        \centering
        \includegraphics[width=\textwidth]{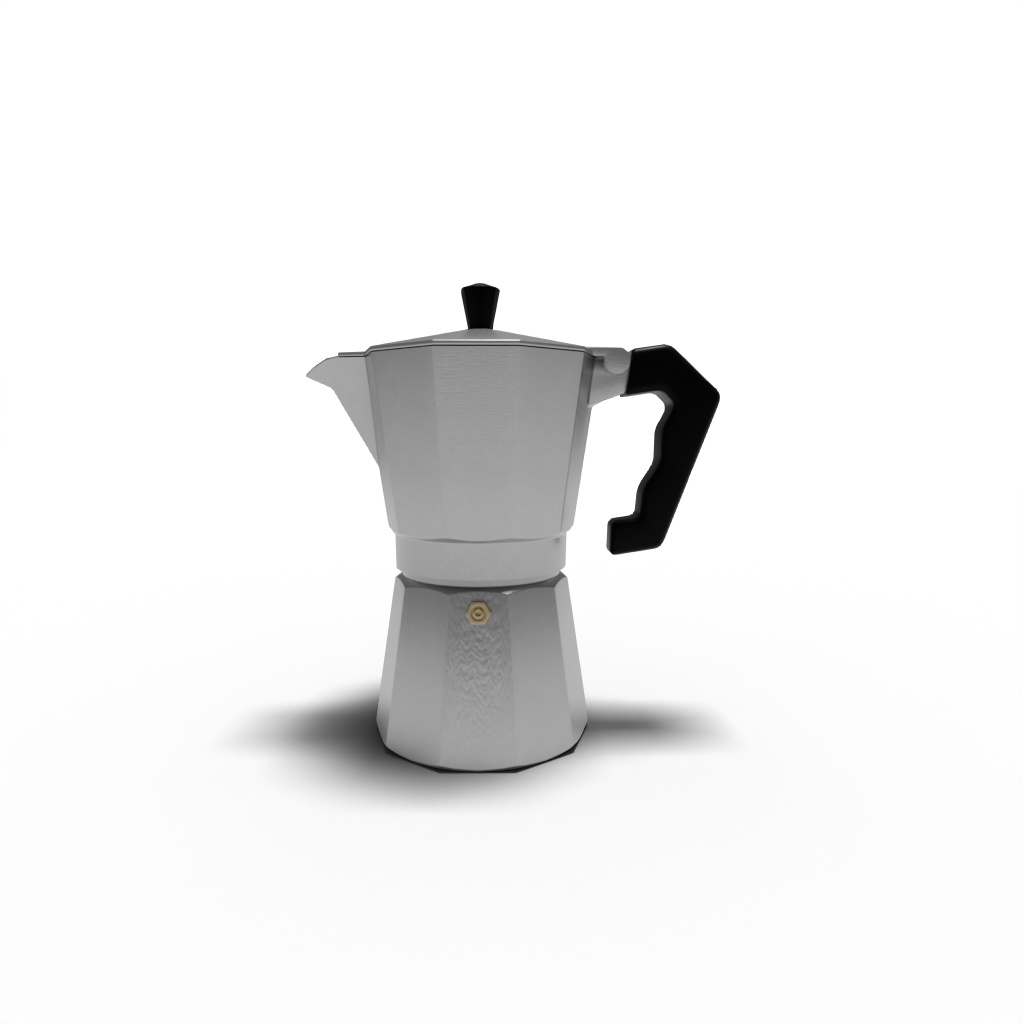} 
        \caption{$\theta=10^\circ$}
    \end{subfigure}
    \hfill
    \begin{subfigure}{0.16\textwidth} 
        \centering
        \includegraphics[width=\textwidth]{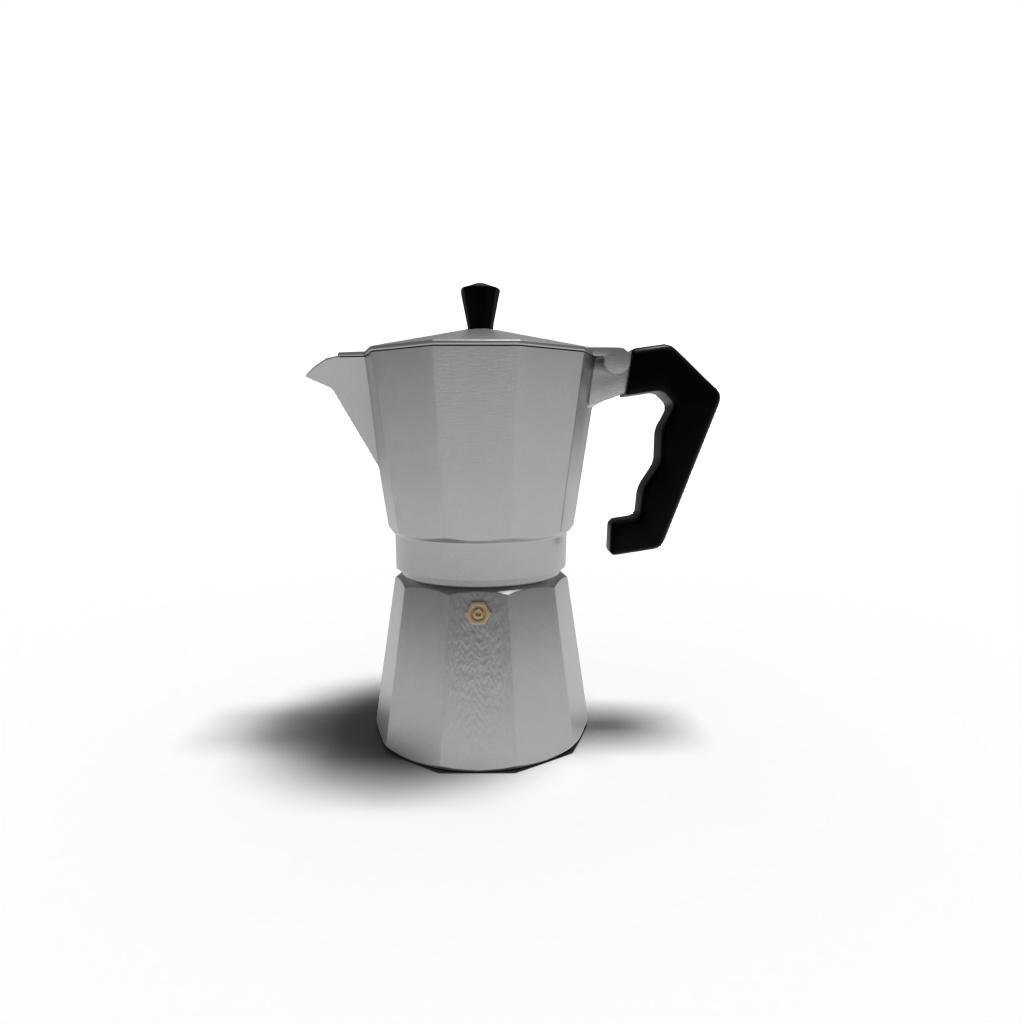}
        \caption{$\theta=15^\circ$}
    \end{subfigure}
    \hfill
    \begin{subfigure}{0.16\textwidth} 
        \centering
        \includegraphics[width=\textwidth]{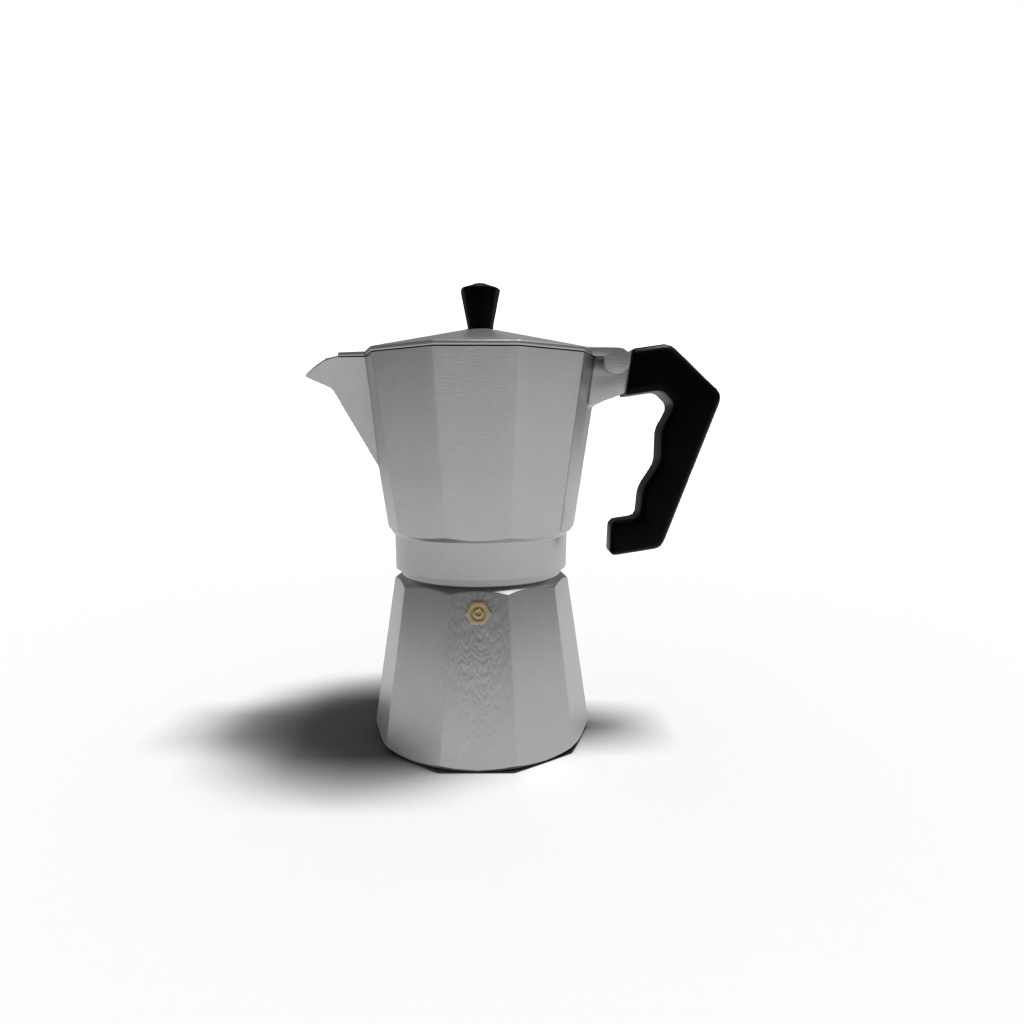}
        \caption{$\theta=20^\circ$}
    \end{subfigure}
    \hfill
    \begin{subfigure}{0.16\textwidth} 
        \centering
        \includegraphics[width=\textwidth]{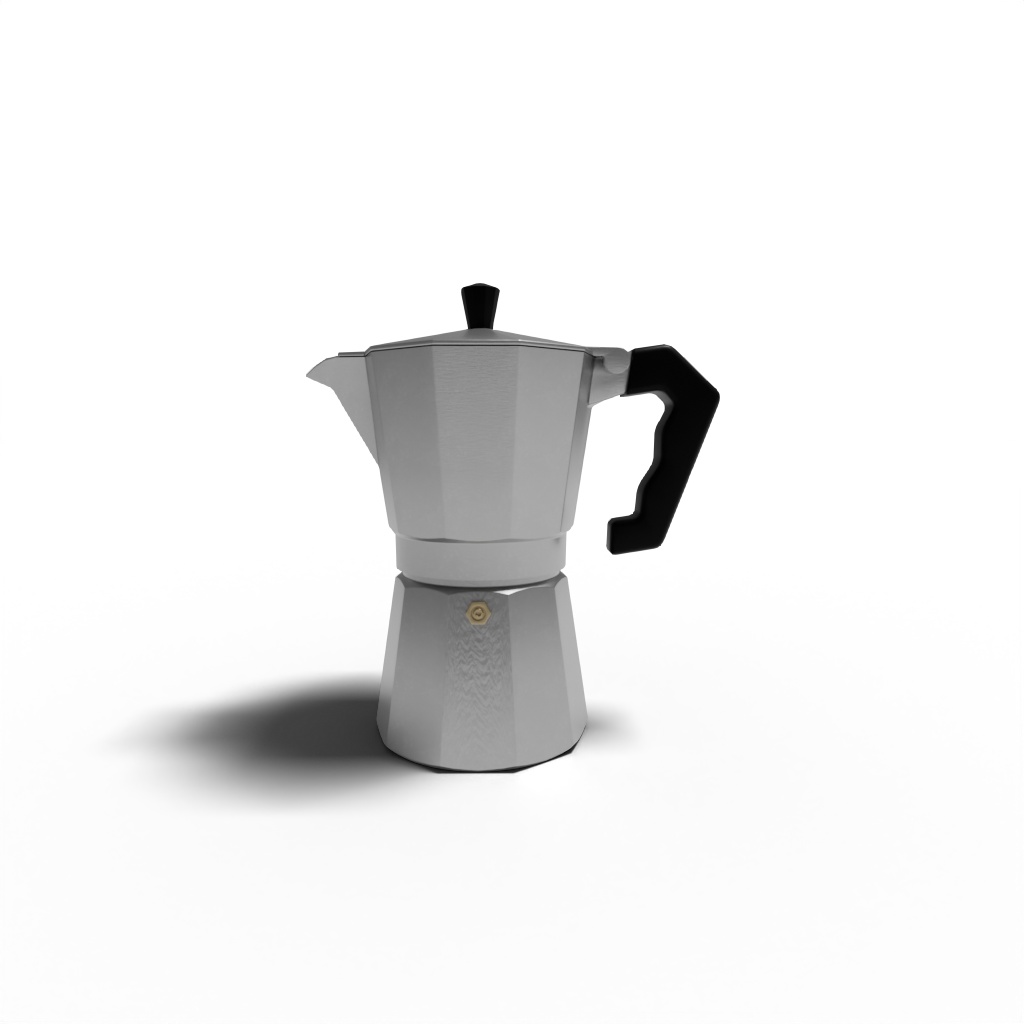}
        \caption{$\theta=25^\circ$}
    \end{subfigure}
    \hfill
    \begin{subfigure}{0.16\textwidth} 
        \centering
        \includegraphics[width=\textwidth]{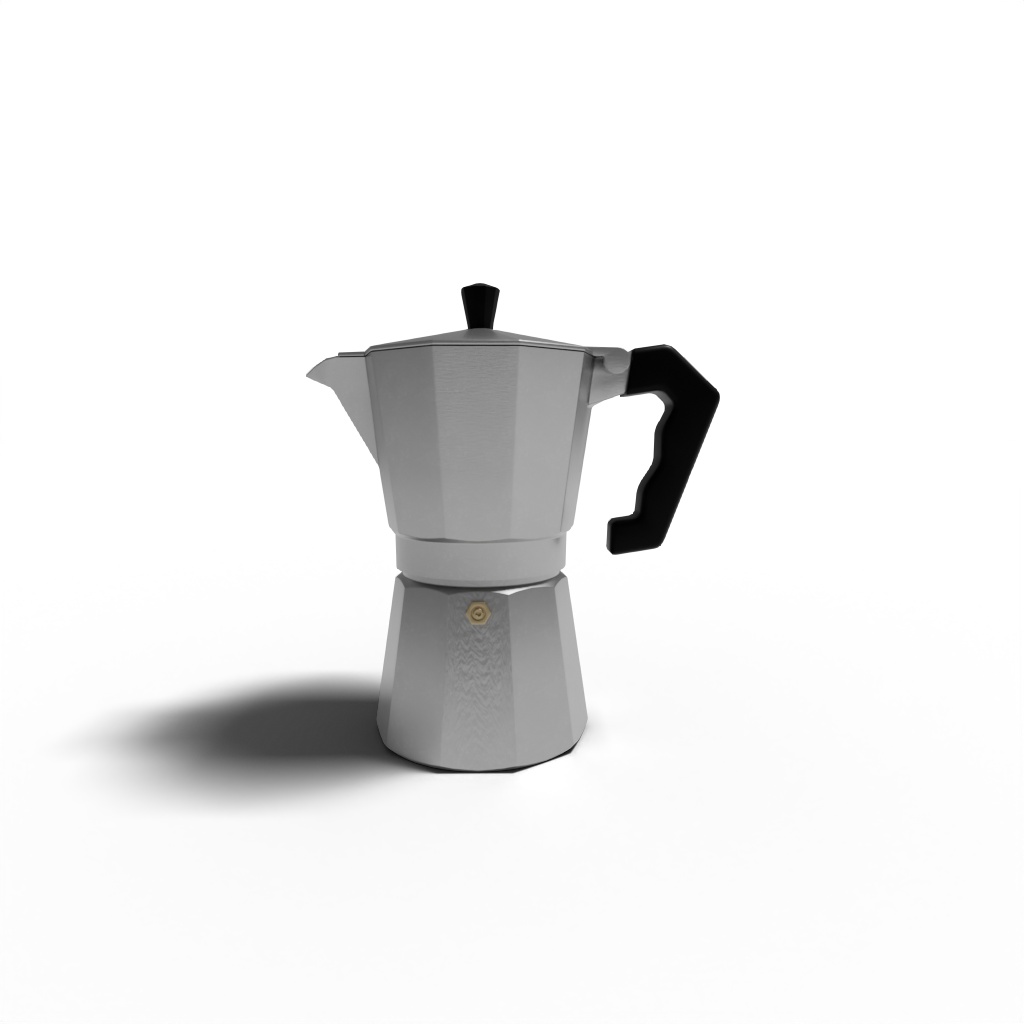}
        \caption{$\theta=30^\circ$}
    \end{subfigure}
    \hfill
    \begin{subfigure}{0.16\textwidth} 
        \centering
        \includegraphics[width=\textwidth]{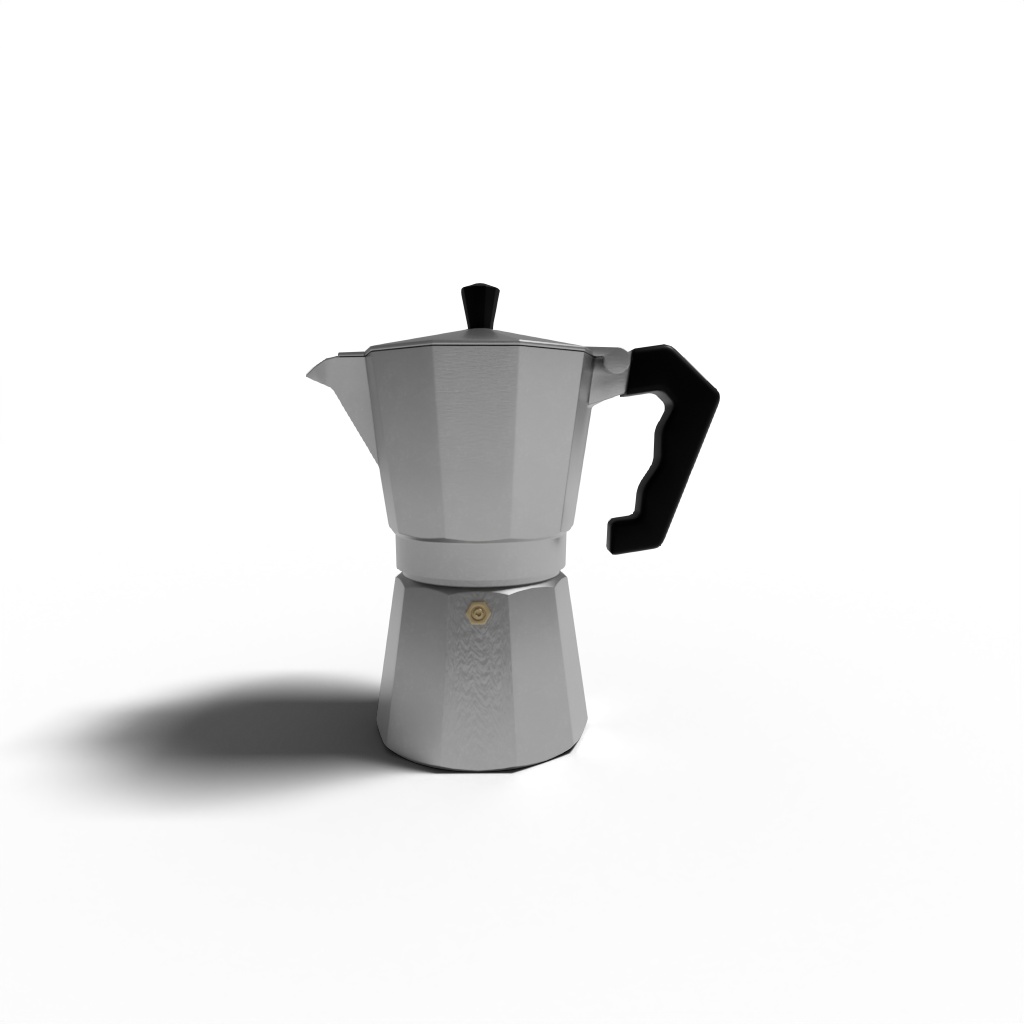}
        \caption{$\theta=35^\circ$}
    \end{subfigure}

    \begin{subfigure}{0.16\textwidth} 
        \centering
        \includegraphics[width=\textwidth]{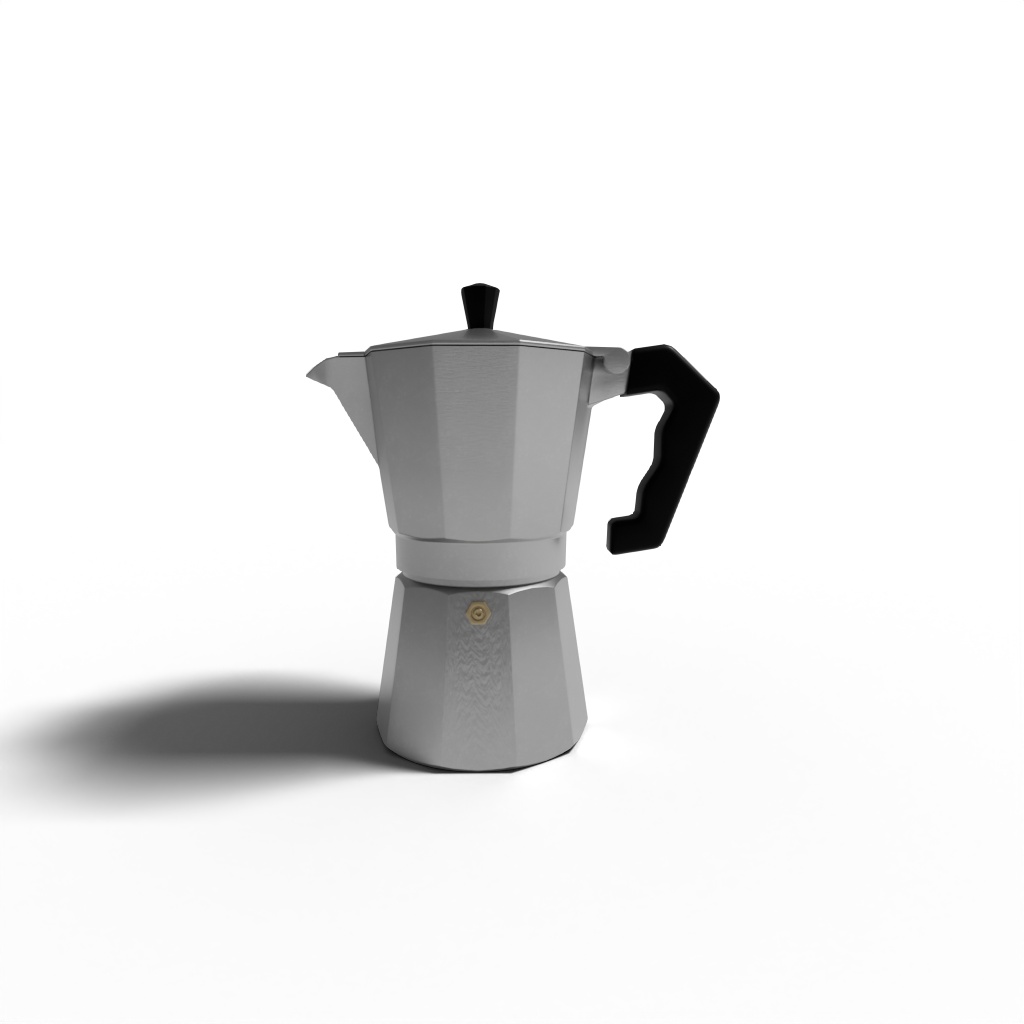}
        \caption{$\theta=40^\circ$}
    \end{subfigure}
    \hfill
    \begin{subfigure}{0.16\textwidth} 
        \centering
        \includegraphics[width=\textwidth]{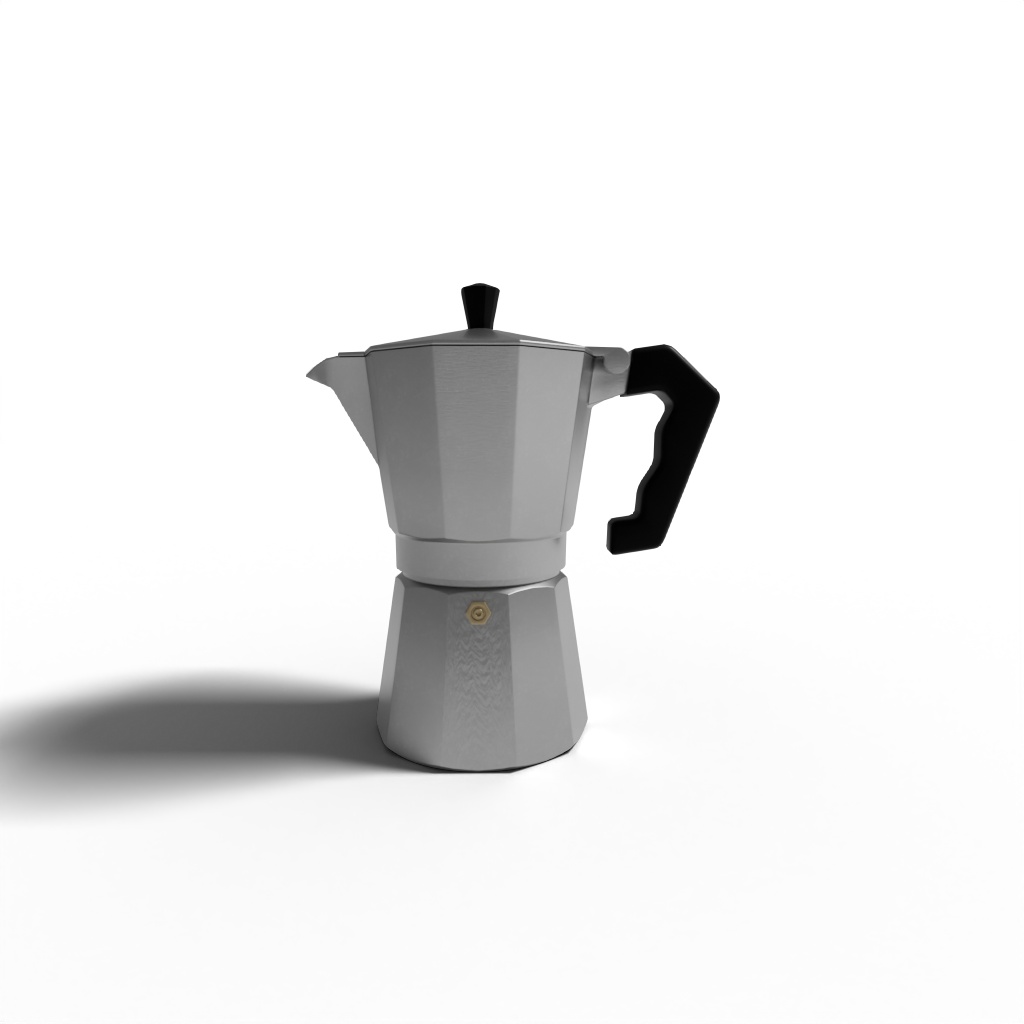}
        \caption{$\theta=45^\circ$}
    \end{subfigure}
    \hfill
    \begin{subfigure}{0.16\textwidth} 
        \centering
        \includegraphics[width=\textwidth]{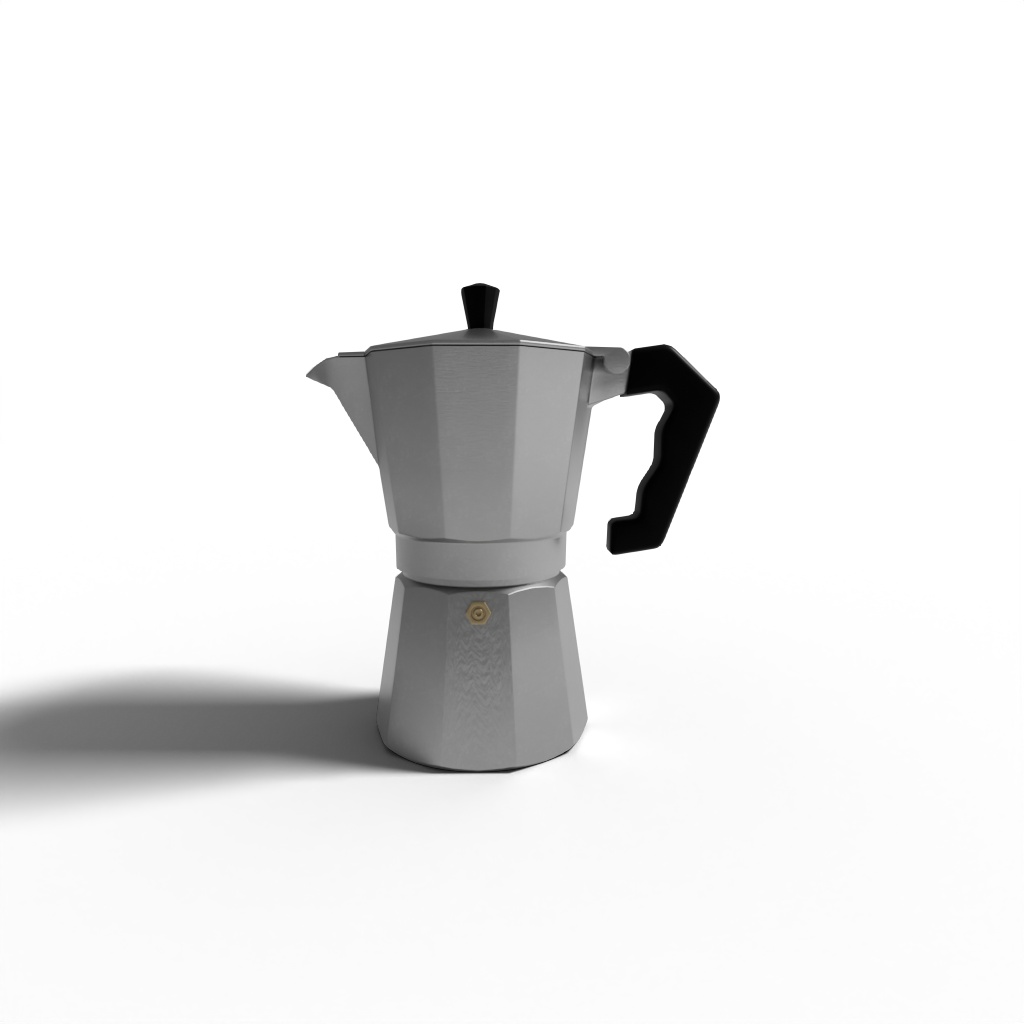}
        \caption{$\theta=45^\circ$}
    \end{subfigure}
    \hfill
    \begin{subfigure}{0.16\textwidth} 
        \centering
        \includegraphics[width=\textwidth]{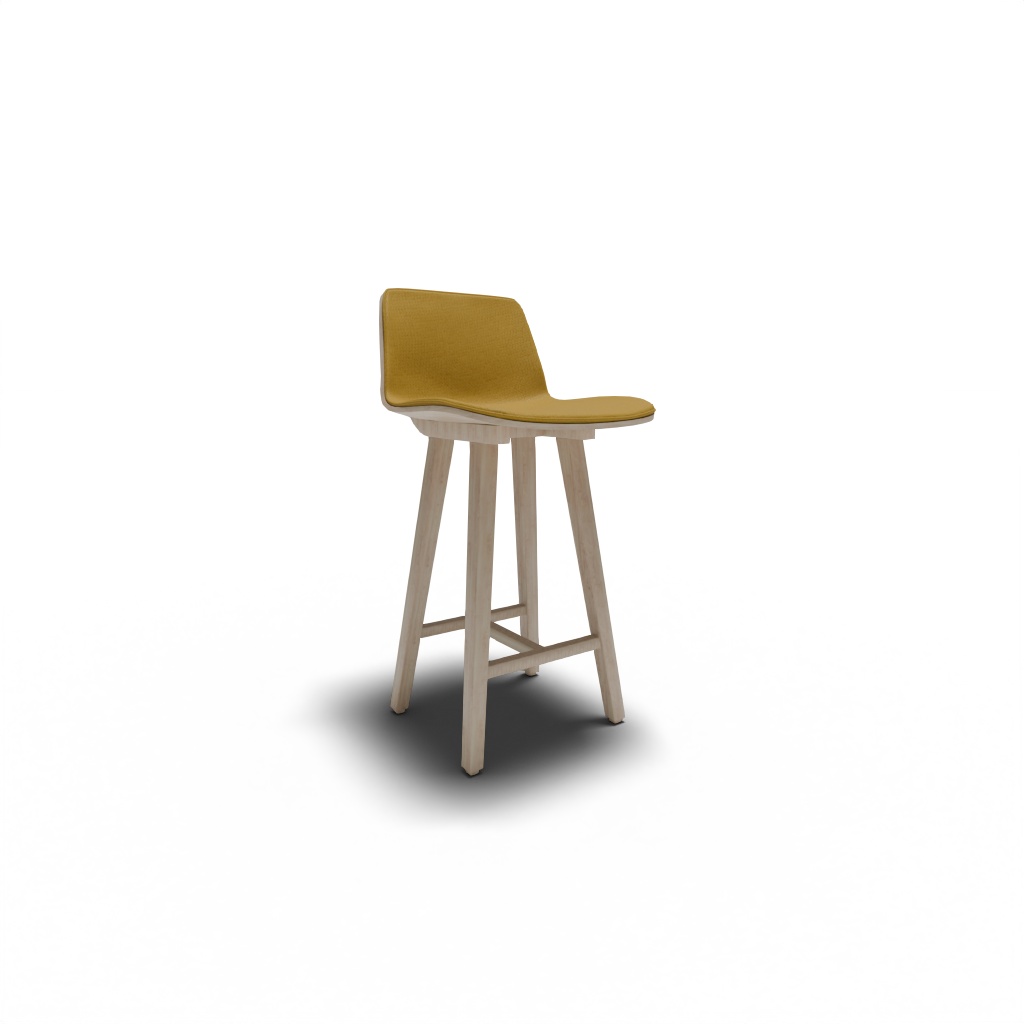}
        \caption{$\theta=5^\circ$}
    \end{subfigure}
    \hfill
    \begin{subfigure}{0.16\textwidth} 
        \centering
        \includegraphics[width=\textwidth]{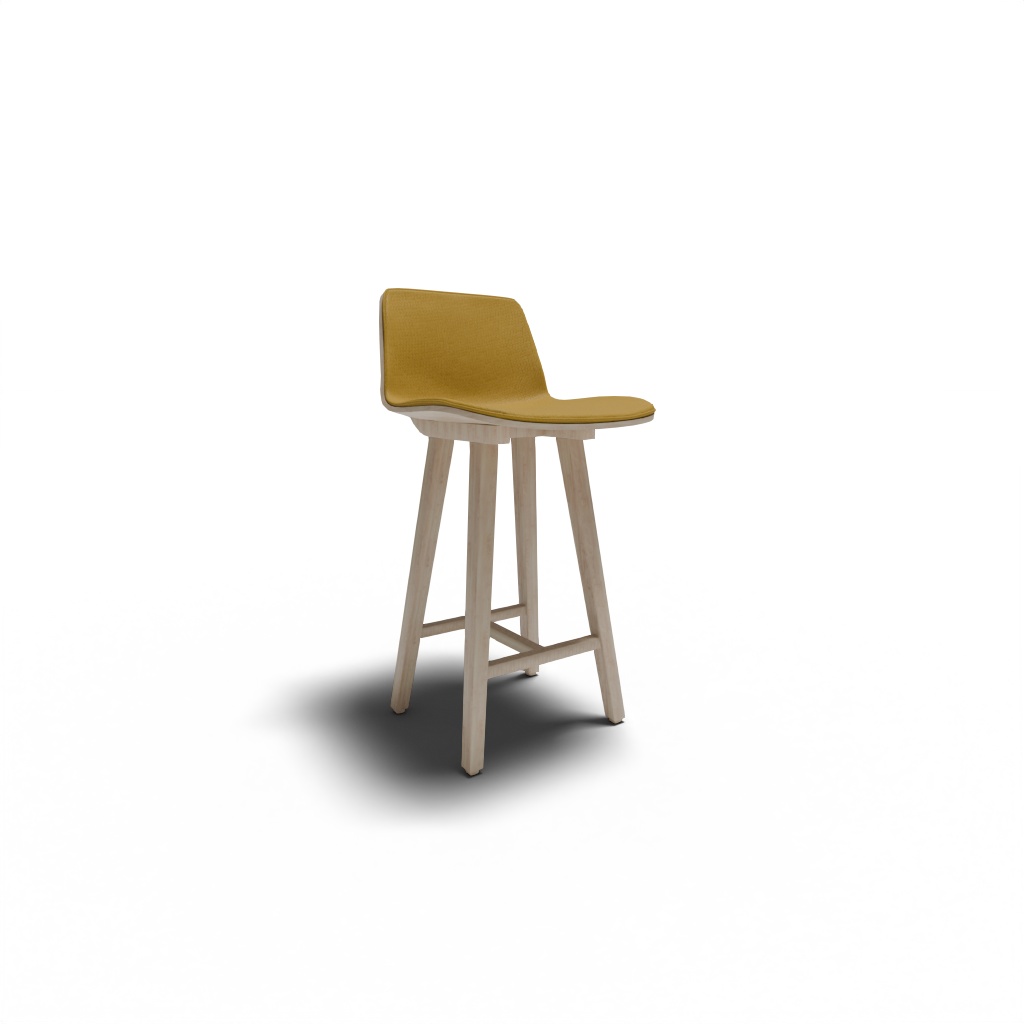}
        \caption{$\theta=10^\circ$}
    \end{subfigure}
    \hfill
    \begin{subfigure}{0.16\textwidth} 
        \centering
        \includegraphics[width=\textwidth]{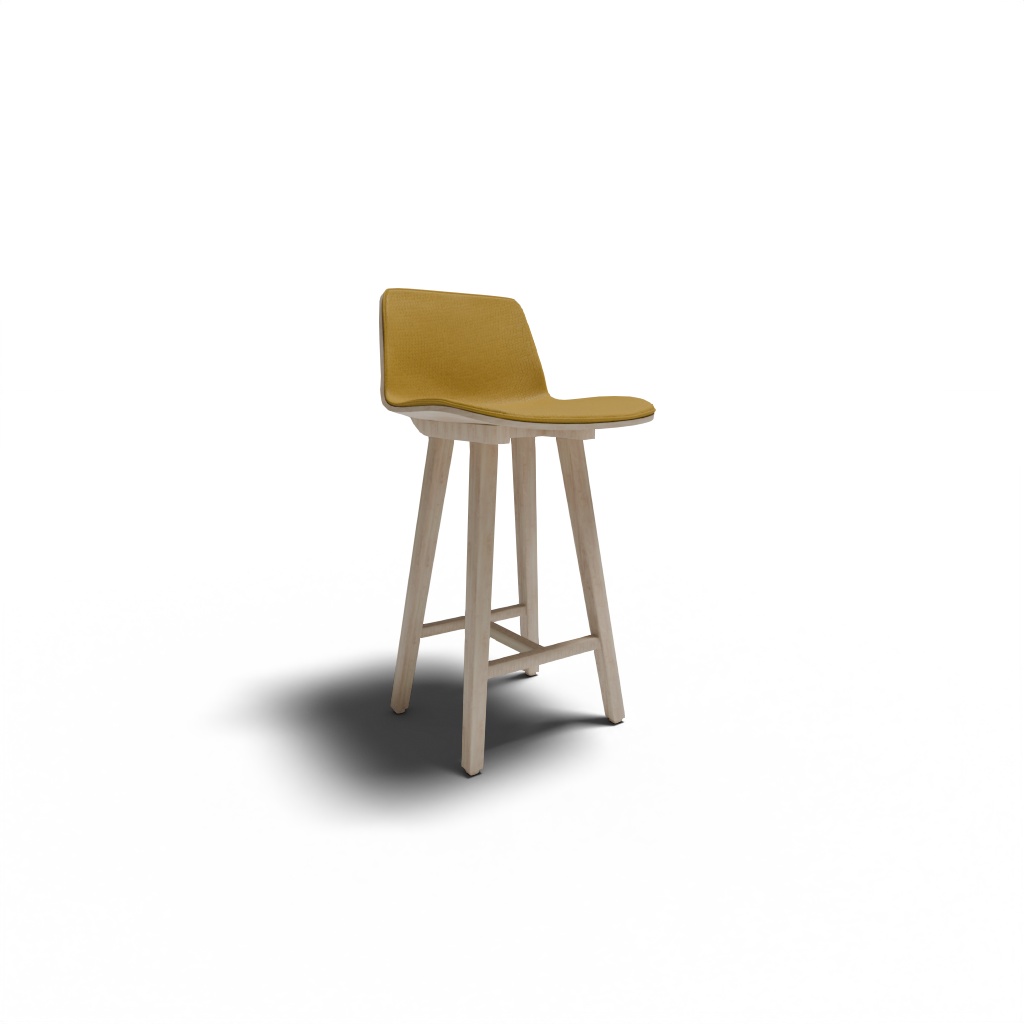}
        \caption{$\theta=15^\circ$}
    \end{subfigure}
    
    \begin{subfigure}{0.16\textwidth} 
        \centering
        \includegraphics[width=\textwidth]{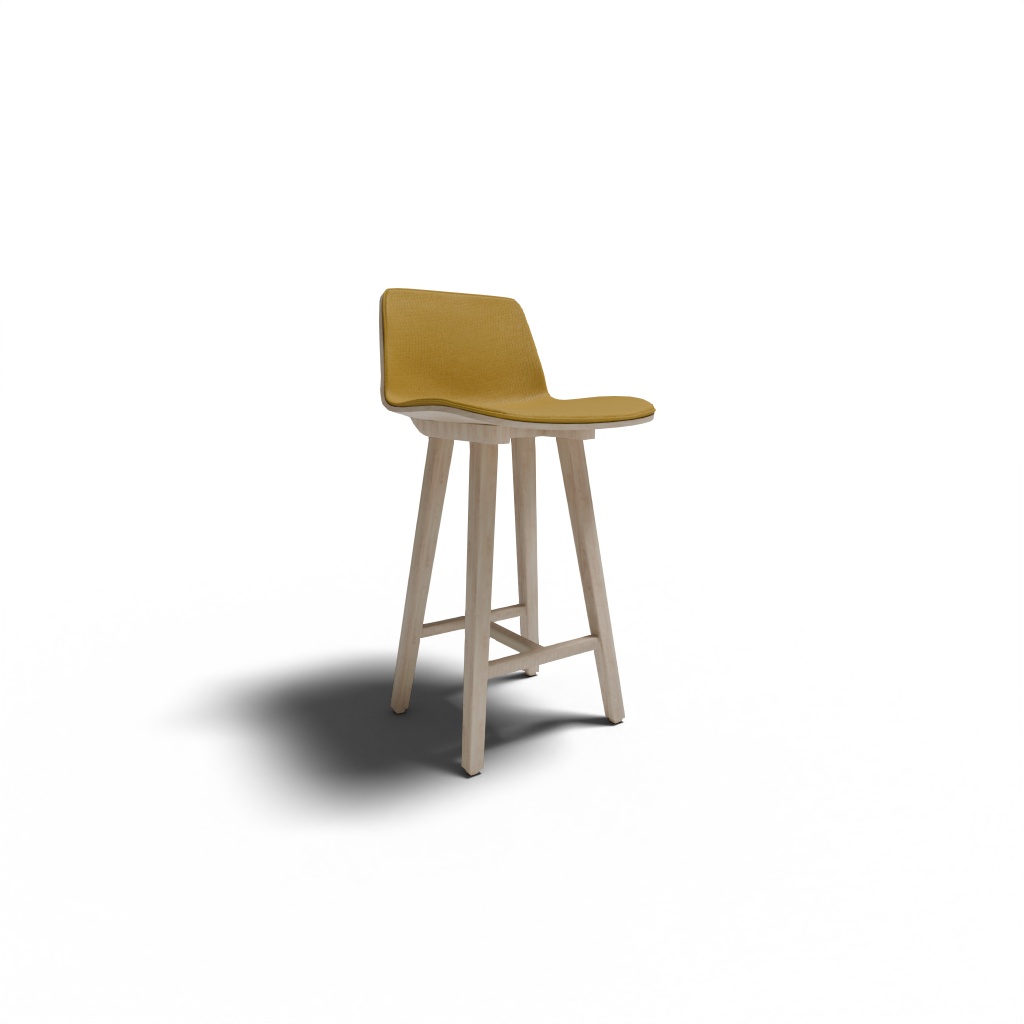}
        \caption{$\theta=20^\circ$}
    \end{subfigure}
    \hfill
    \begin{subfigure}{0.16\textwidth} 
        \centering
        \includegraphics[width=\textwidth]{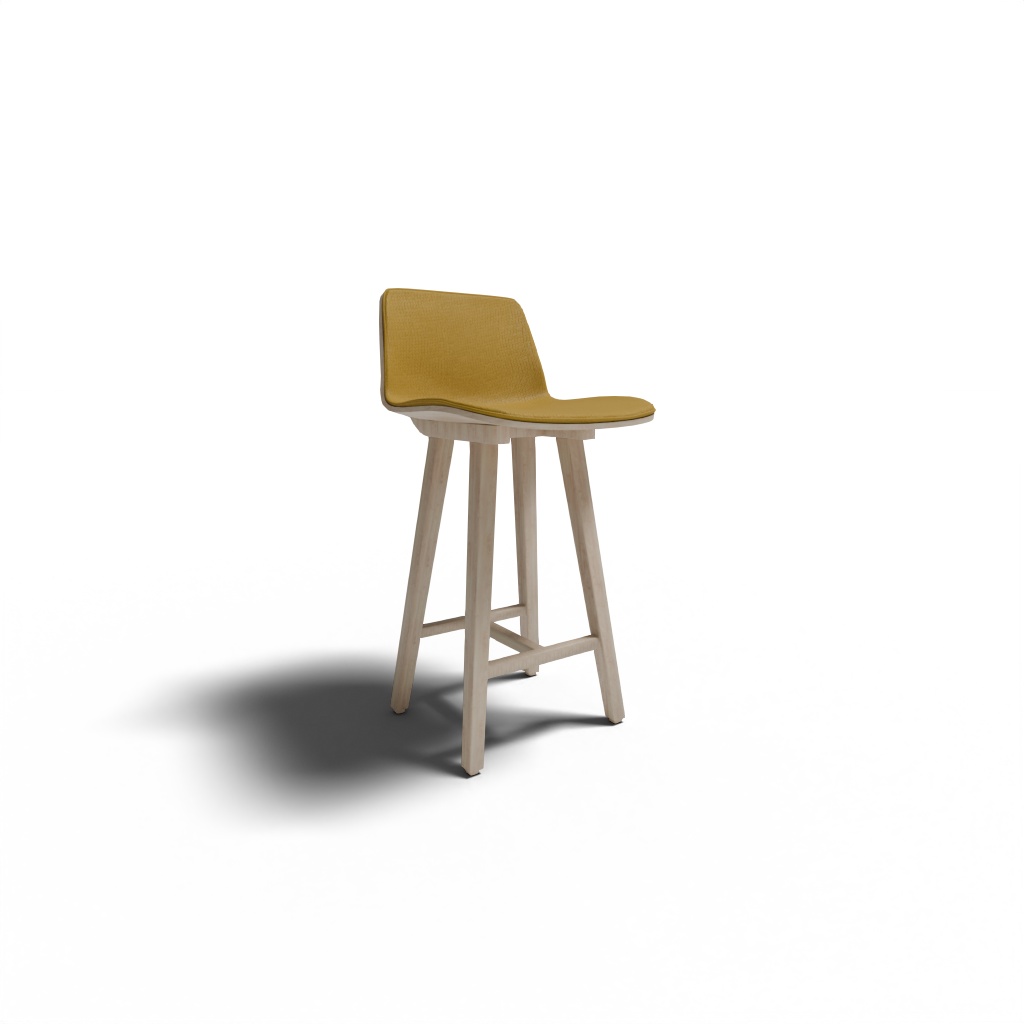}
        \caption{$\theta=25^\circ$}
    \end{subfigure}
    \hfill
    \begin{subfigure}{0.16\textwidth} 
        \centering
        \includegraphics[width=\textwidth]{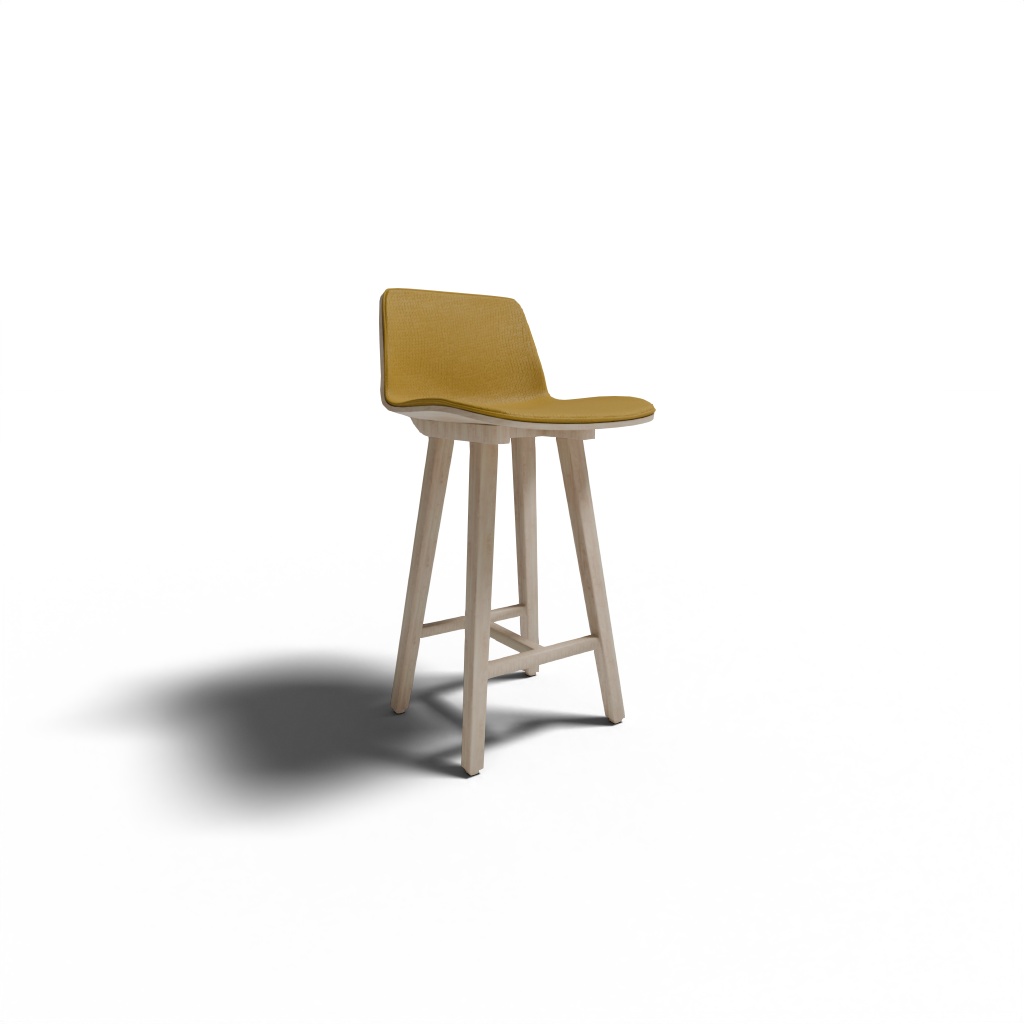}
        \caption{$\theta=30^\circ$}
    \end{subfigure}
    \hfill
    \begin{subfigure}{0.16\textwidth} 
        \centering
        \includegraphics[width=\textwidth]{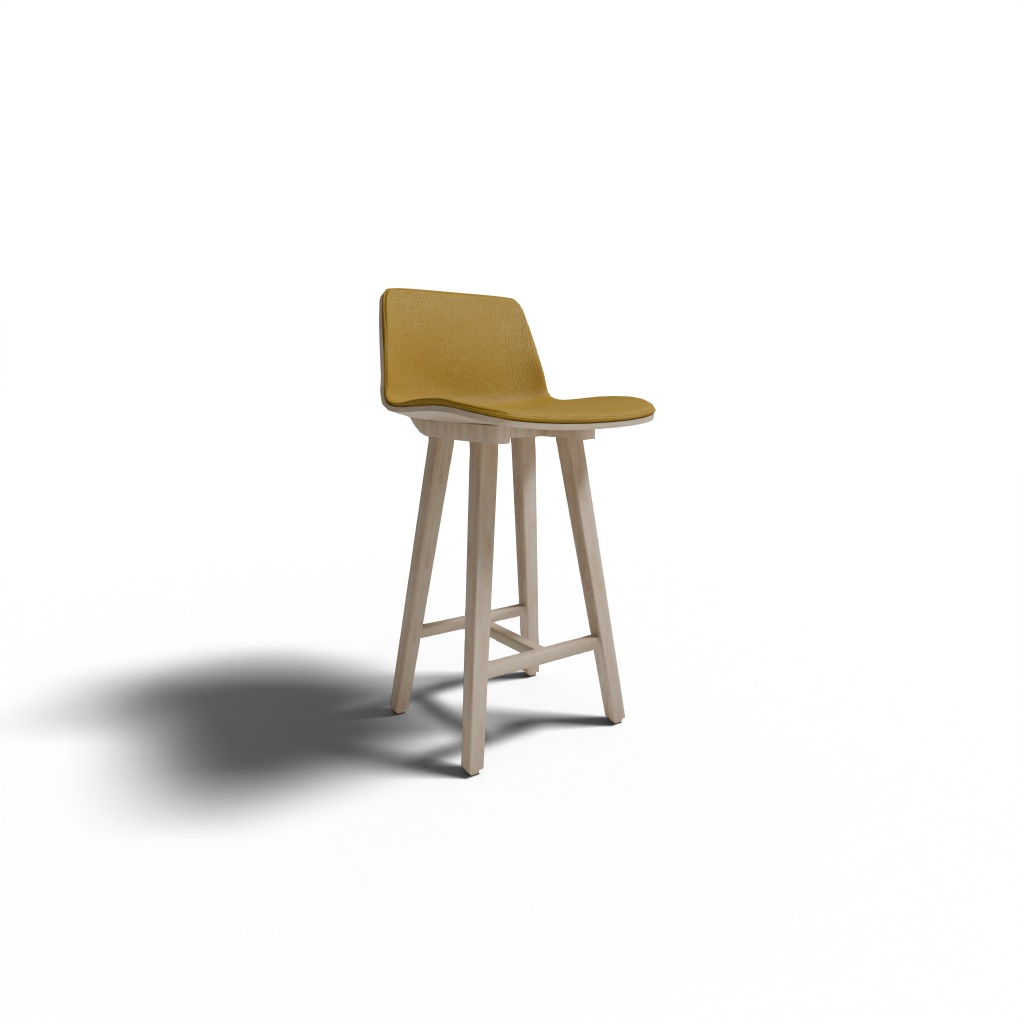}
        \caption{$\theta=35^\circ$}
    \end{subfigure}
    \hfill
    \begin{subfigure}{0.16\textwidth} 
        \centering
        \includegraphics[width=\textwidth]{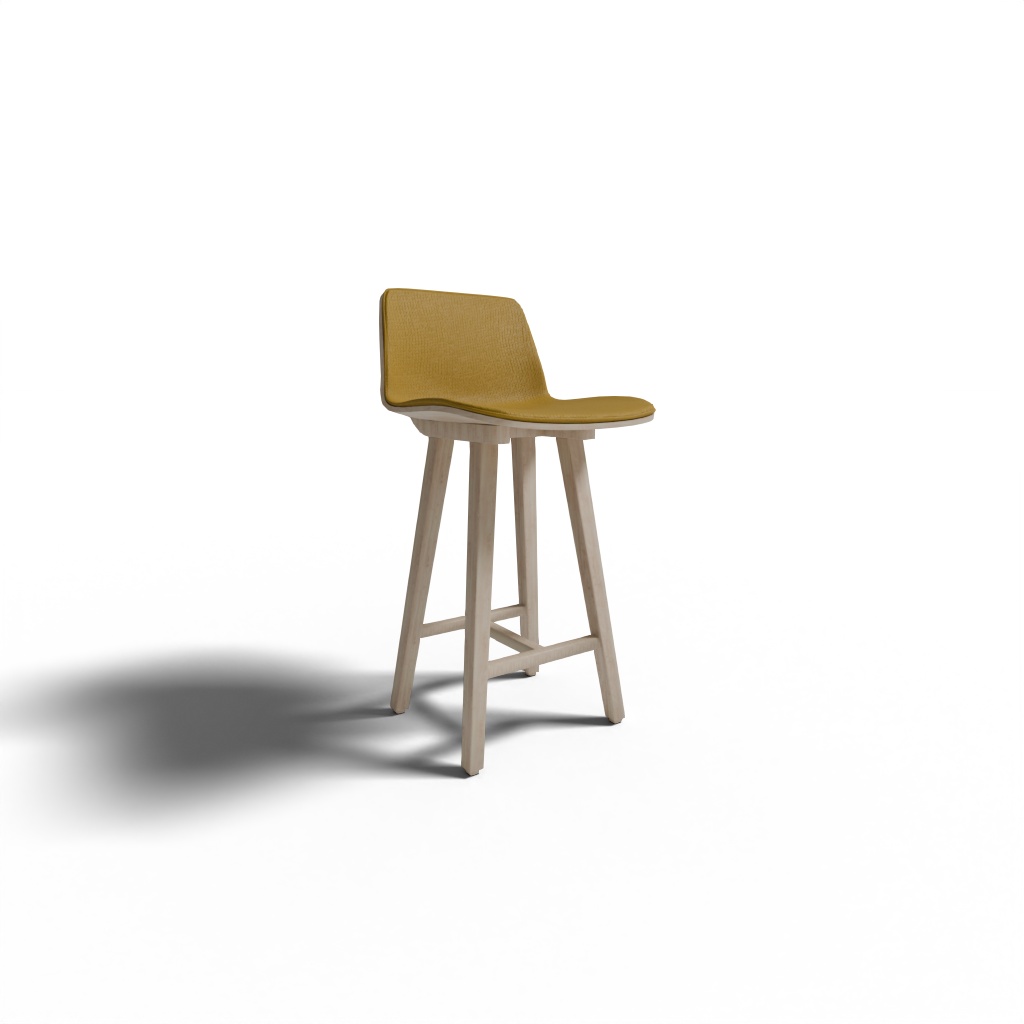}
        \caption{$\theta=40^\circ$}
    \end{subfigure}
    \hfill
    \begin{subfigure}{0.16\textwidth} 
        \centering
        \includegraphics[width=\textwidth]{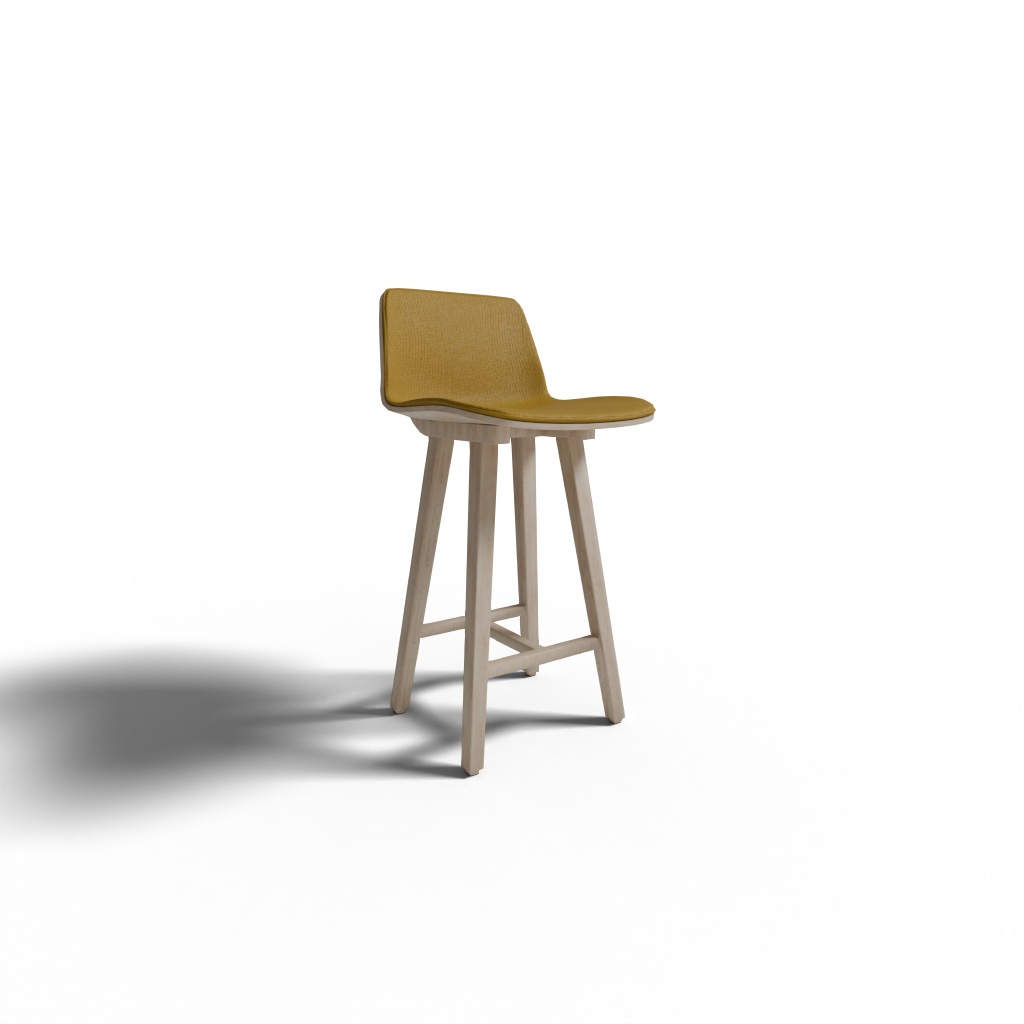}
        \caption{$\theta=45^\circ$}
    \end{subfigure}
    
    \caption{Renders for two 3D meshes from the vertical shadow direction control track. $\phi=0^\circ$ and $s=2$.}
    \label{fig:sm_track_3}
\end{figure*}

\begin{figure*}[t]
    \centering
    \begin{subfigure}{0.16\textwidth} 
        \centering
        \includegraphics[width=\textwidth]{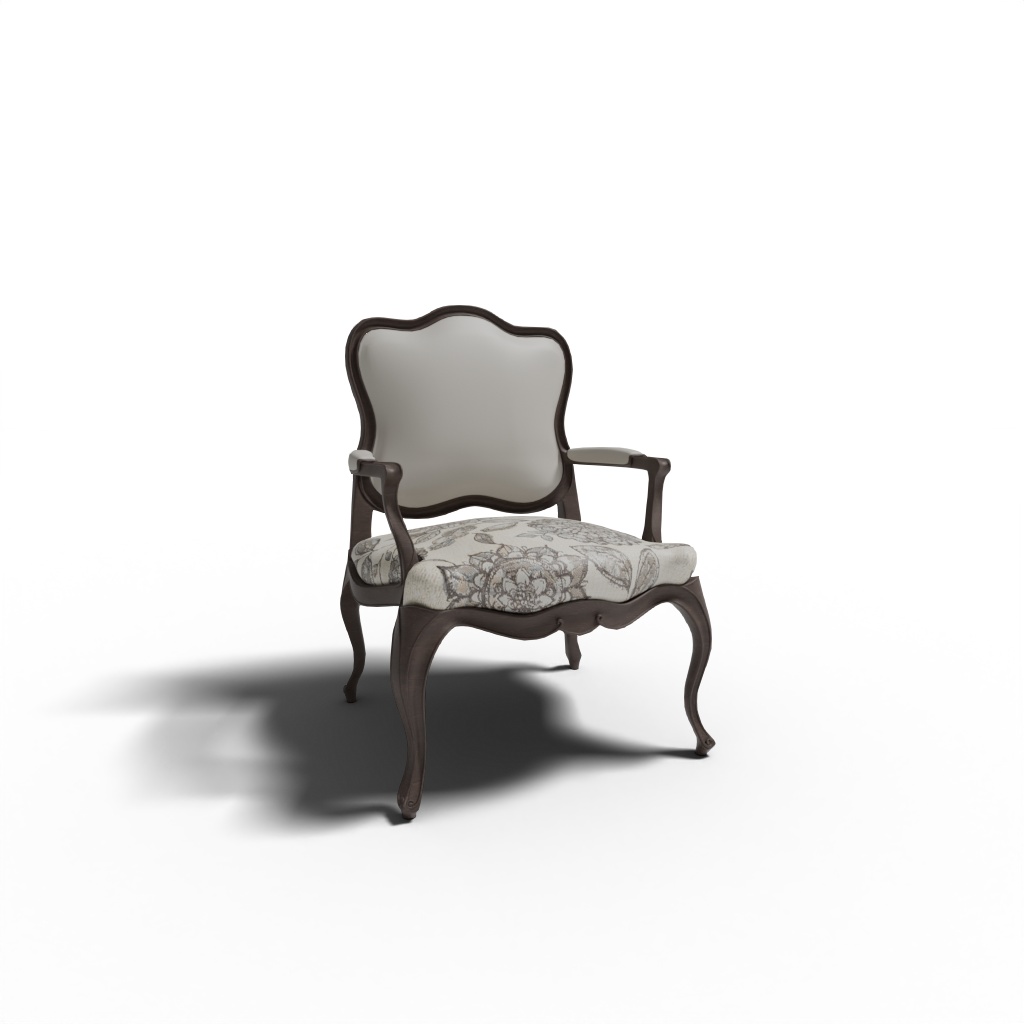} 
        \caption{$s=2$}
    \end{subfigure}
    \begin{subfigure}{0.16\textwidth} 
        \centering
        \includegraphics[width=\textwidth]{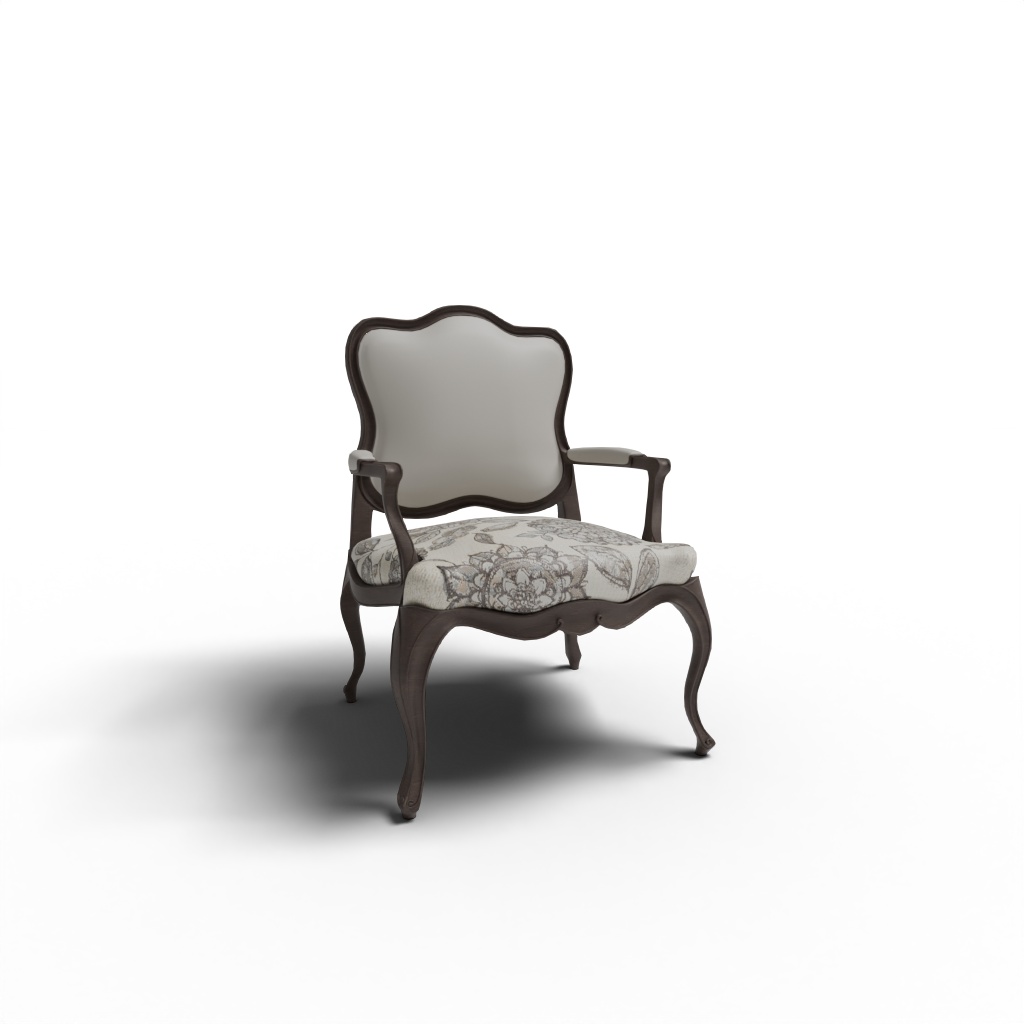} 
        \caption{$s=4$}
    \end{subfigure}
    \begin{subfigure}{0.16\textwidth} 
        \centering
        \includegraphics[width=\textwidth]{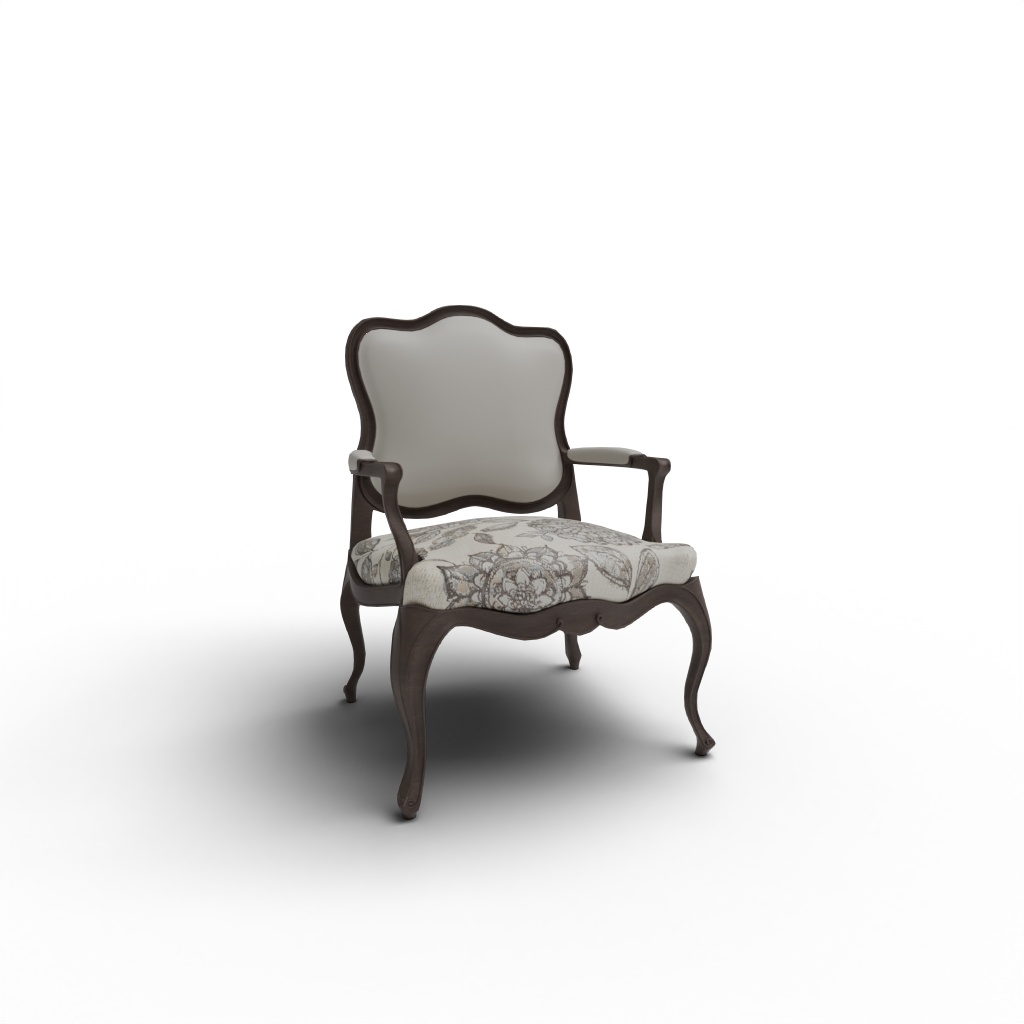} 
        \caption{$s=8$}
    \end{subfigure}
    \begin{subfigure}{0.16\textwidth} 
        \centering
        \includegraphics[width=\textwidth]{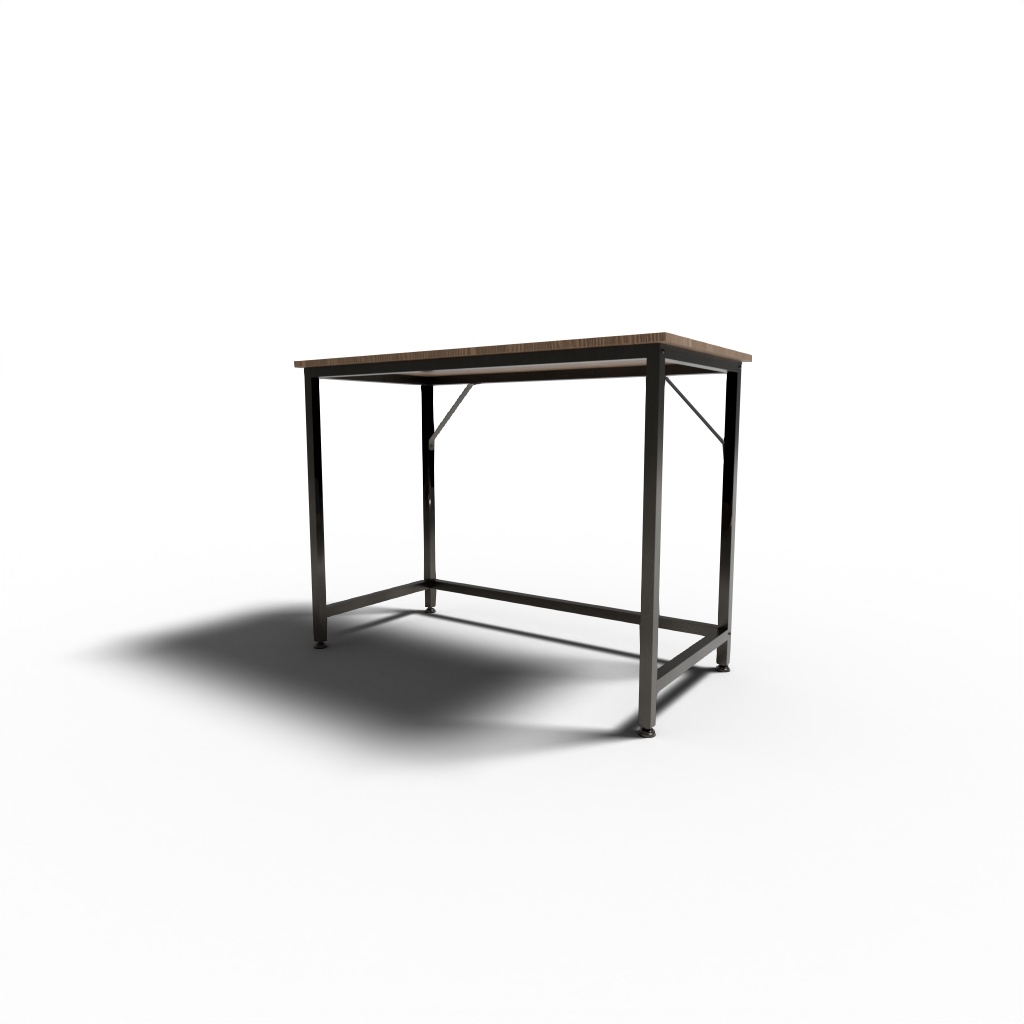} 
        \caption{$s=2$}
    \end{subfigure}
    \begin{subfigure}{0.16\textwidth} 
        \centering
        \includegraphics[width=\textwidth]{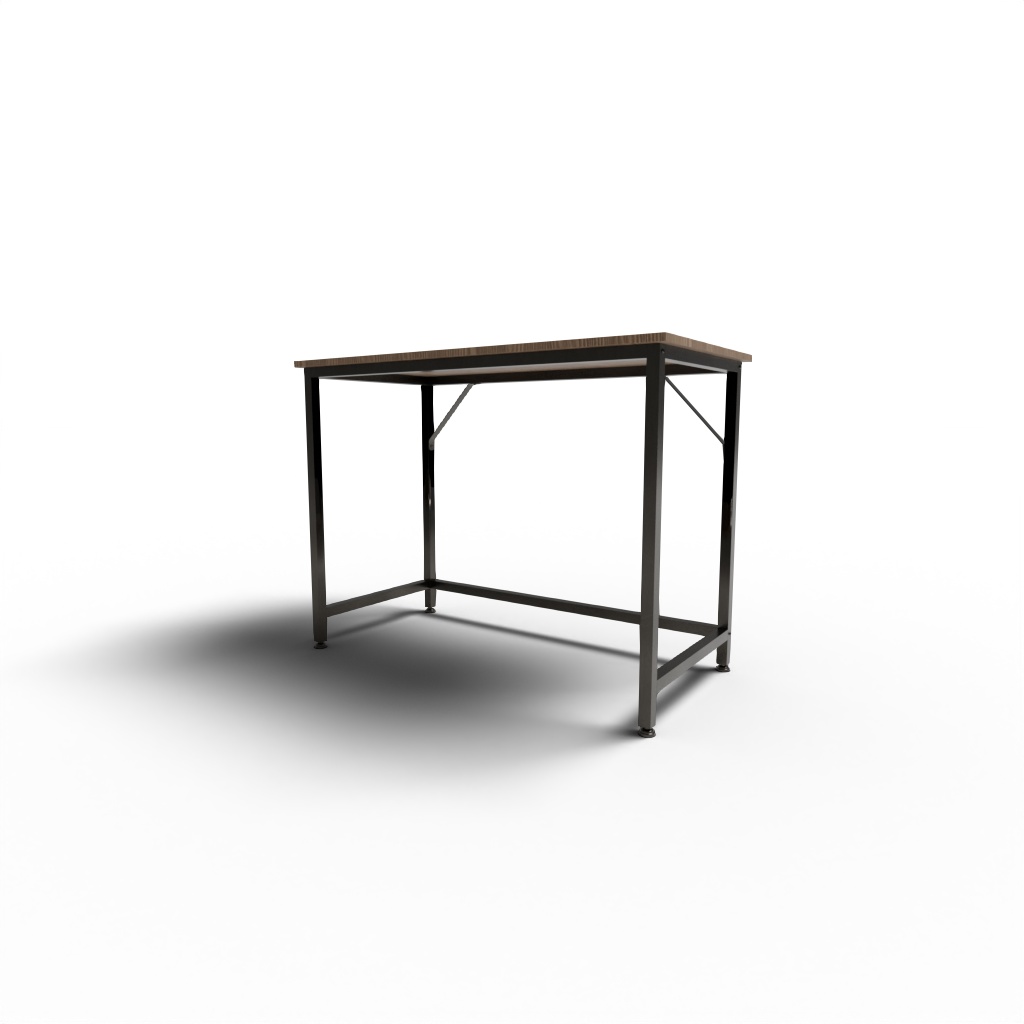} 
        \caption{$s=4$}
    \end{subfigure}
    \begin{subfigure}{0.16\textwidth} 
        \centering
        \includegraphics[width=\textwidth]{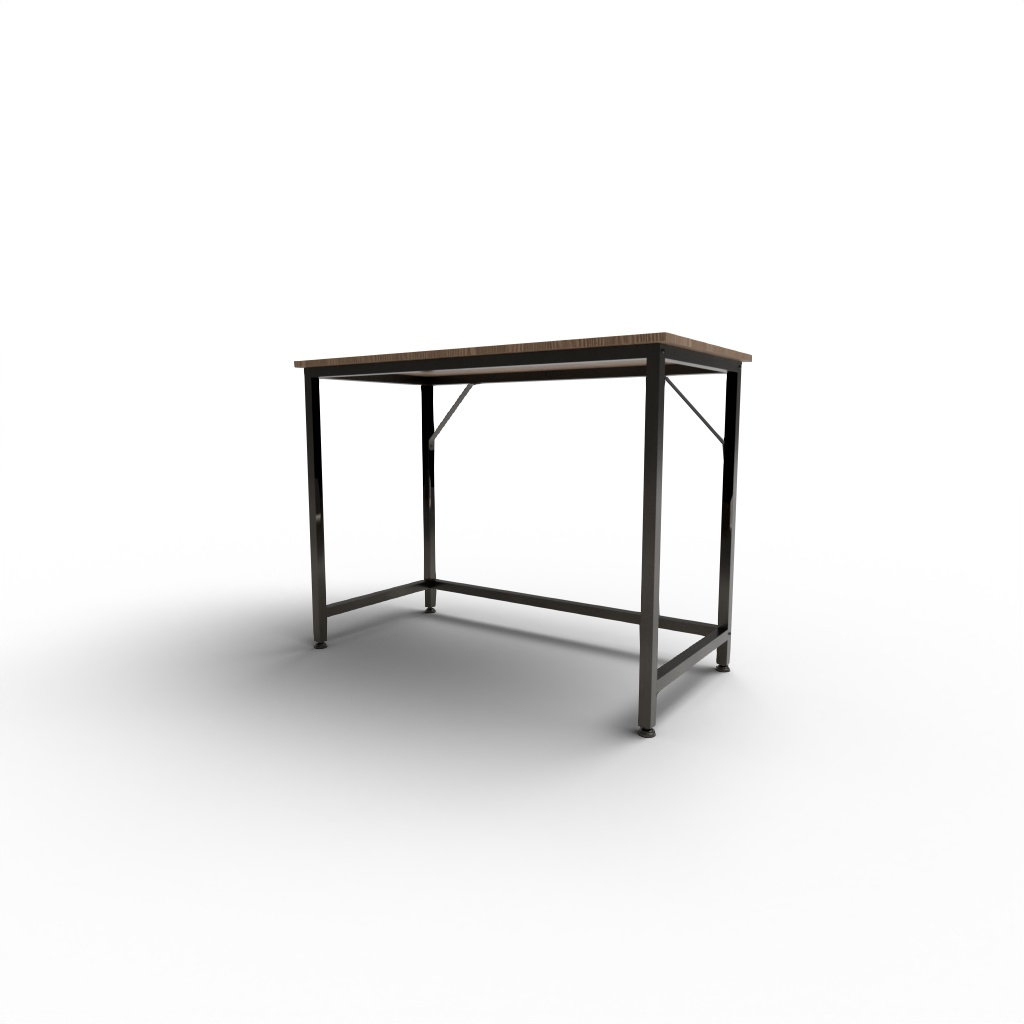} 
        \caption{$s=8$}
    \end{subfigure}
    \caption{Renders for two 3D meshes from the softness control track. $\theta=30^\circ$ and $\phi=0^\circ$.}
    \label{fig:sm_track_1}
\end{figure*}

\begin{figure}[t]
    \centering
    \begin{subfigure}{0.48\textwidth} 
        \centering
        \includegraphics[width=\textwidth]{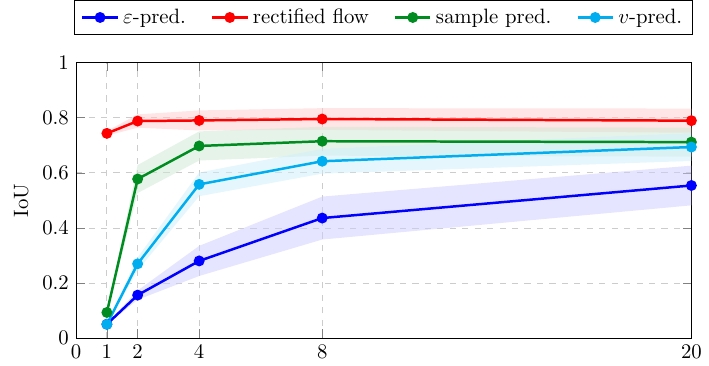} 
    \end{subfigure}
    \begin{subfigure}{0.48\textwidth} 
        \centering
        \includegraphics[width=\textwidth]{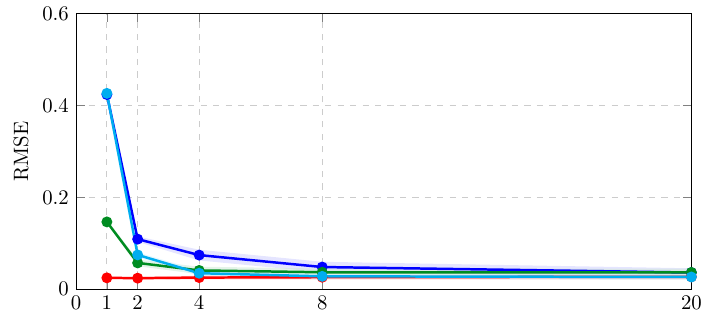} 
    \end{subfigure}
    \begin{subfigure}{0.48\textwidth} 
        \centering
        \includegraphics[width=\textwidth]{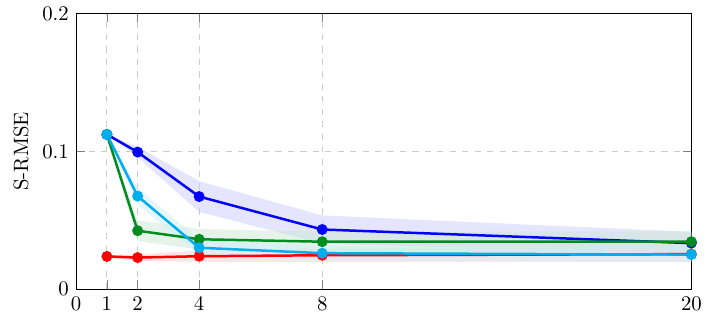} 
    \end{subfigure}
    \begin{subfigure}{0.48\textwidth} 
        \centering
        \includegraphics[width=\textwidth]{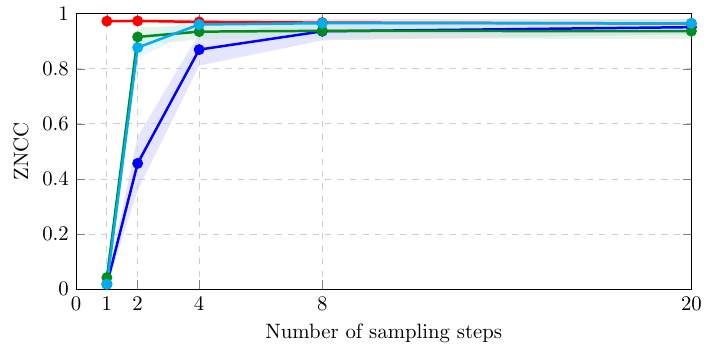} 
    \end{subfigure}
    \caption{Plots comparing methods trained for $150k$ iterations, across 4 prediction types and 4 metrics, for various sampling steps. Half transparent margin represents the standard deviation.}
    \label{fig:sm_num_steps_plots}
\end{figure}

\begin{figure}[t]
    \centering
    \begin{subfigure}{0.48\textwidth} 
        \centering
        \includegraphics[width=\textwidth]{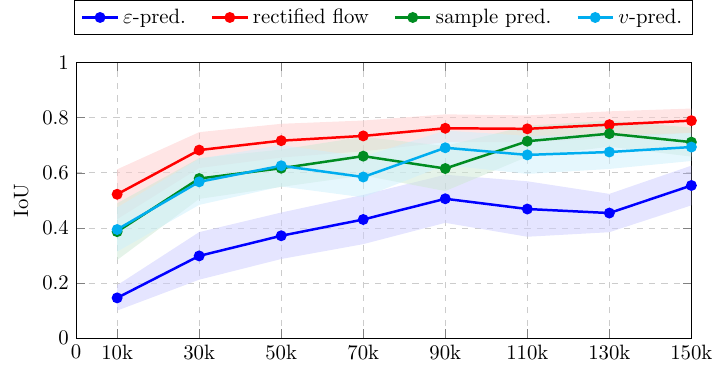} 
    \end{subfigure}
    \begin{subfigure}{0.48\textwidth} 
        \centering
        \includegraphics[width=\textwidth]{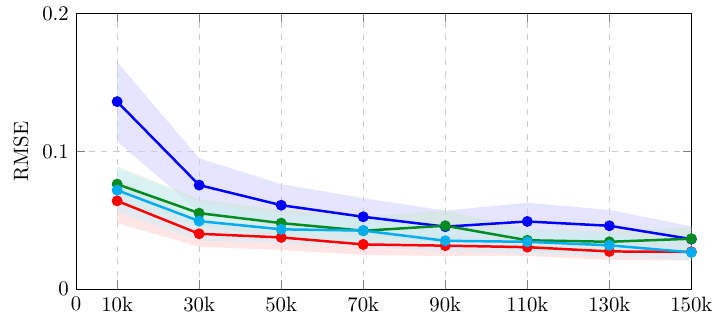} 
    \end{subfigure}
    \begin{subfigure}{0.48\textwidth} 
        \centering
        \includegraphics[width=\textwidth]{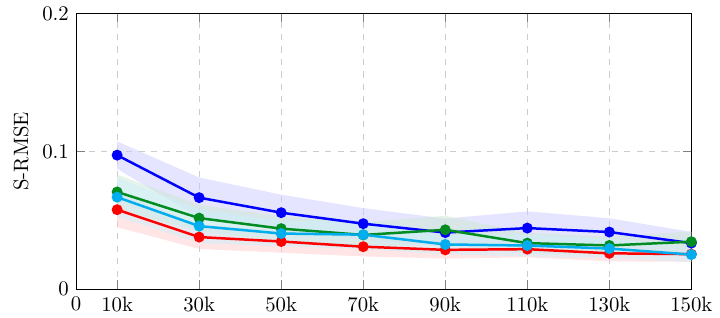} 
    \end{subfigure}
    \begin{subfigure}{0.48\textwidth} 
        \centering
        \includegraphics[width=\textwidth]{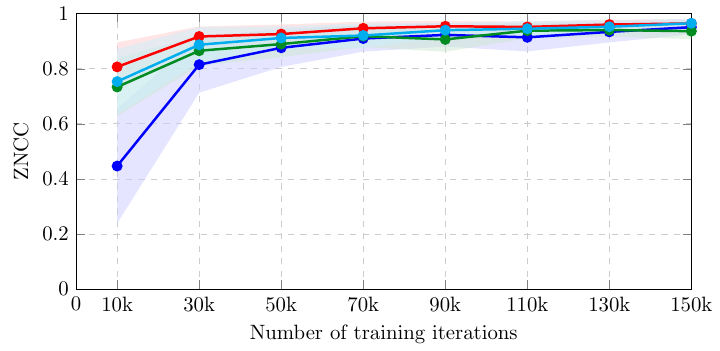} 
    \end{subfigure}
    \caption{Plots comparing methods trained for various iterations, across 4 prediction types and 4 metrics, for $20$ sampling steps. Half transparent margin represents the standard deviation.}
    \label{fig:sm_num_iters_plots}
\end{figure}

\begin{figure*}
    \centering
    \begin{subfigure}{0.24\textwidth} 
        \centering
        \includegraphics[width=\textwidth]{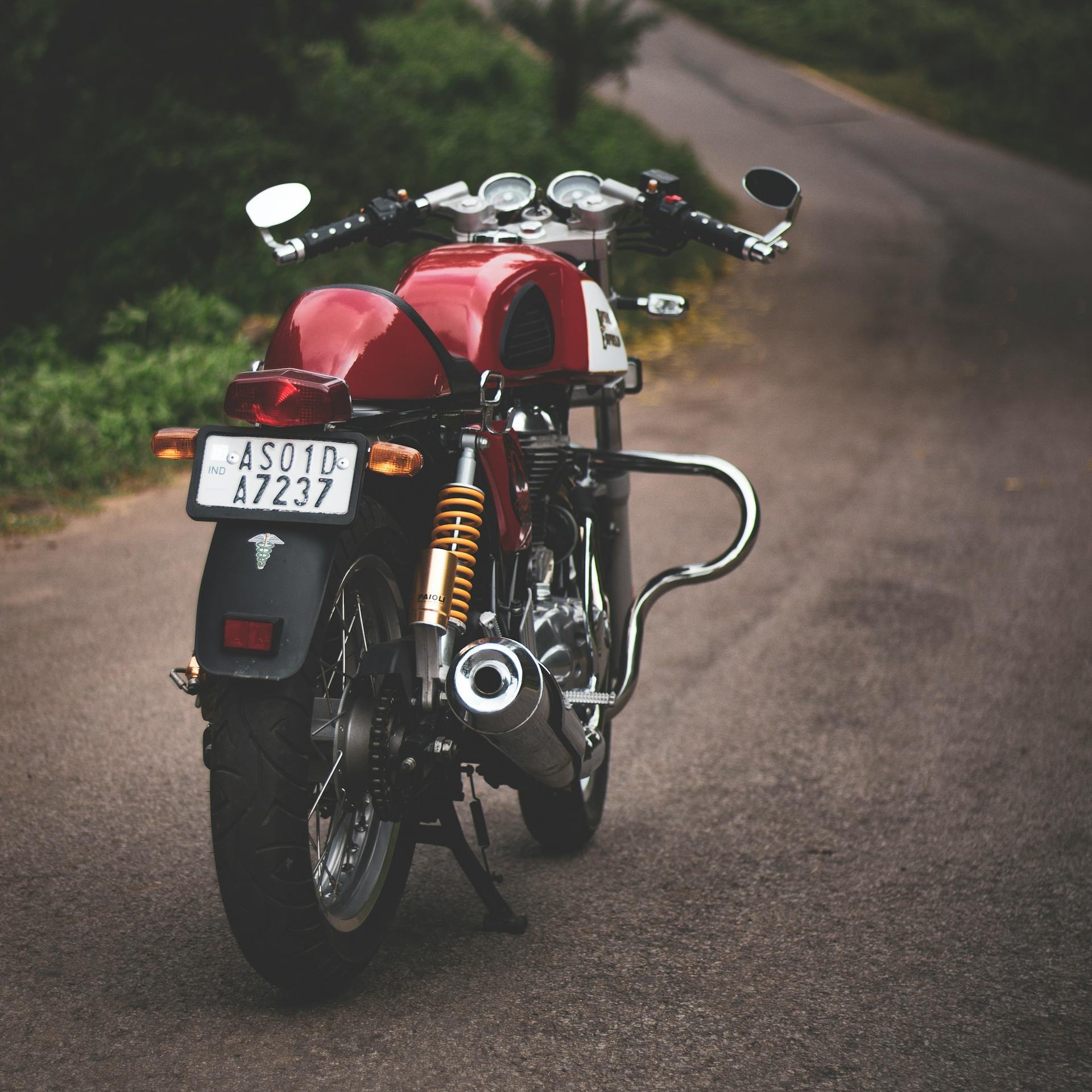}
        \caption{Input Image}
    \end{subfigure}
    \hfill
    \begin{subfigure}{0.24\textwidth} 
        \centering
        \includegraphics[width=\textwidth]{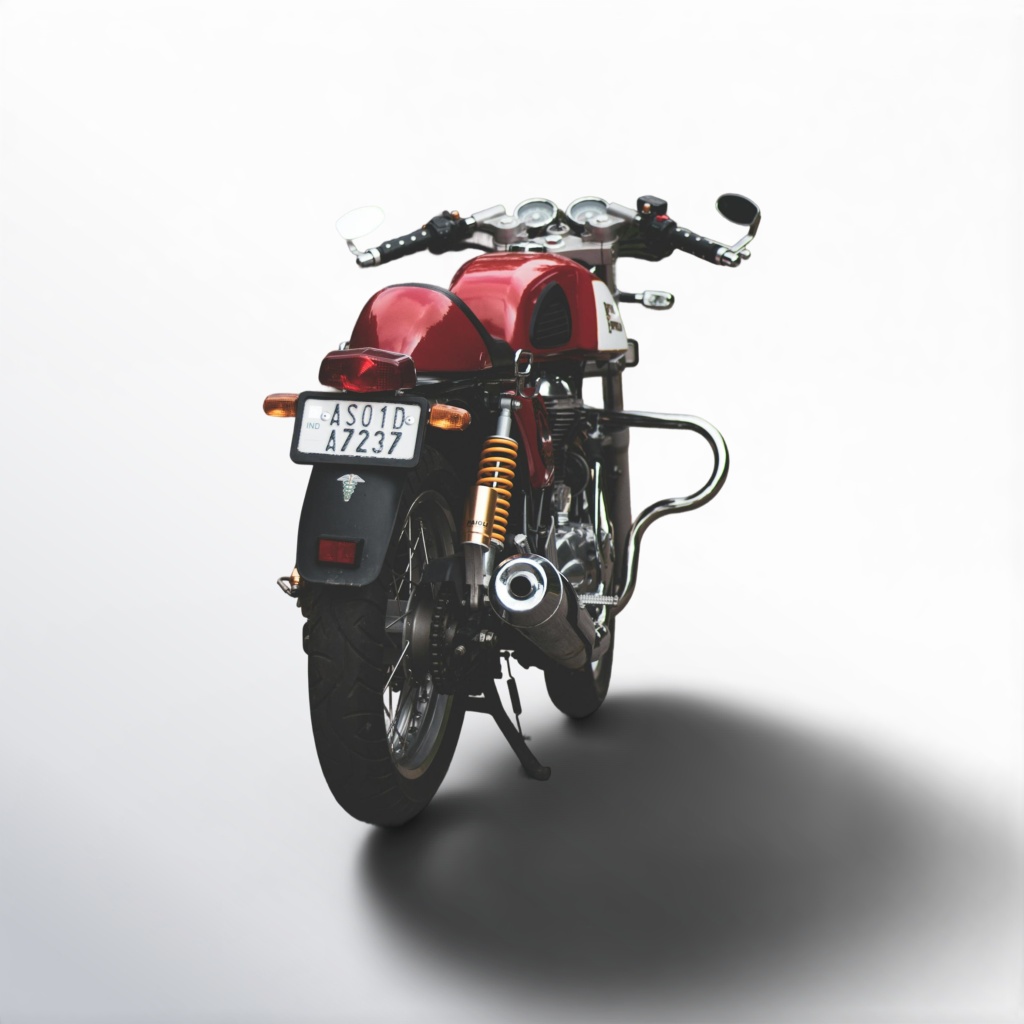}
        \caption{$s=2$}
    \end{subfigure}
    \hfill
    \begin{subfigure}{0.24\textwidth} 
        \centering
        \includegraphics[width=\textwidth]{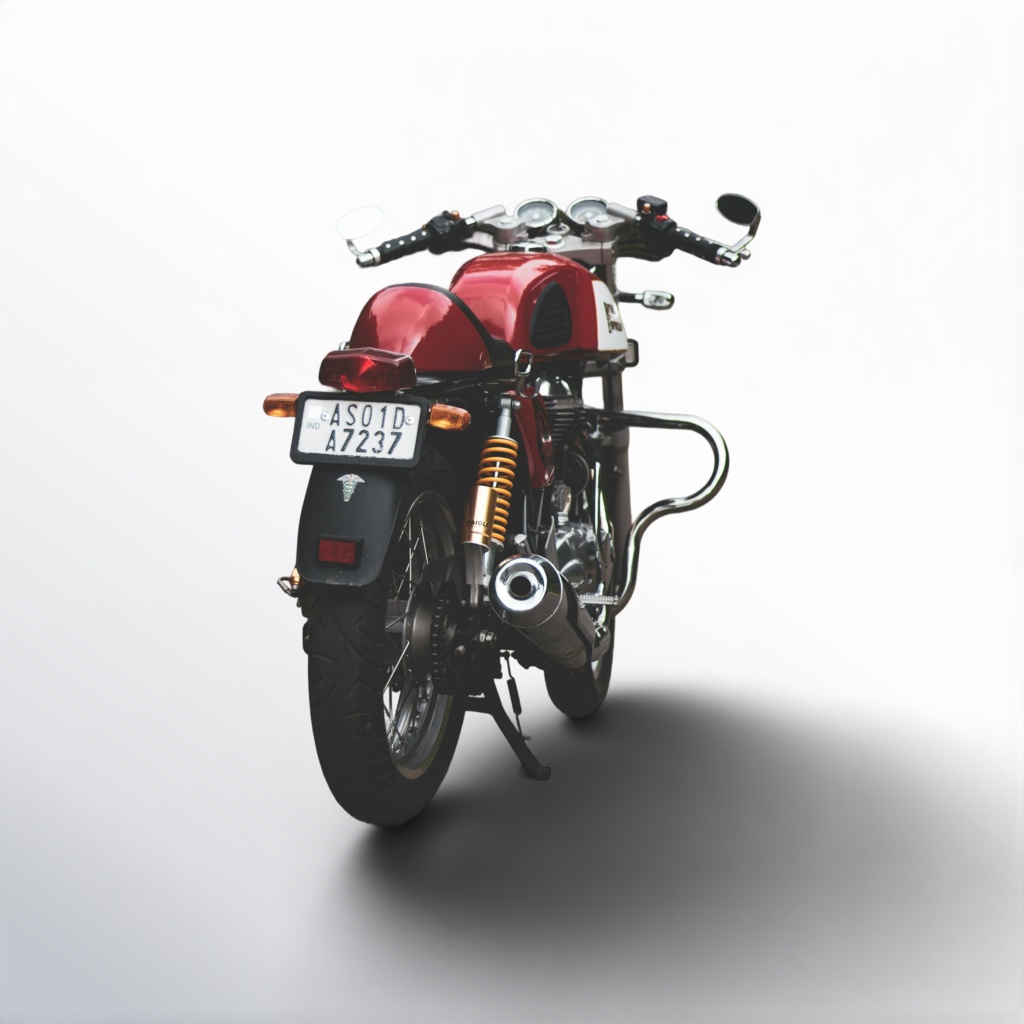}
        \caption{$s=5$}
    \end{subfigure}
    \hfill
    \begin{subfigure}{0.24\textwidth} 
        \centering
        \includegraphics[width=\textwidth]{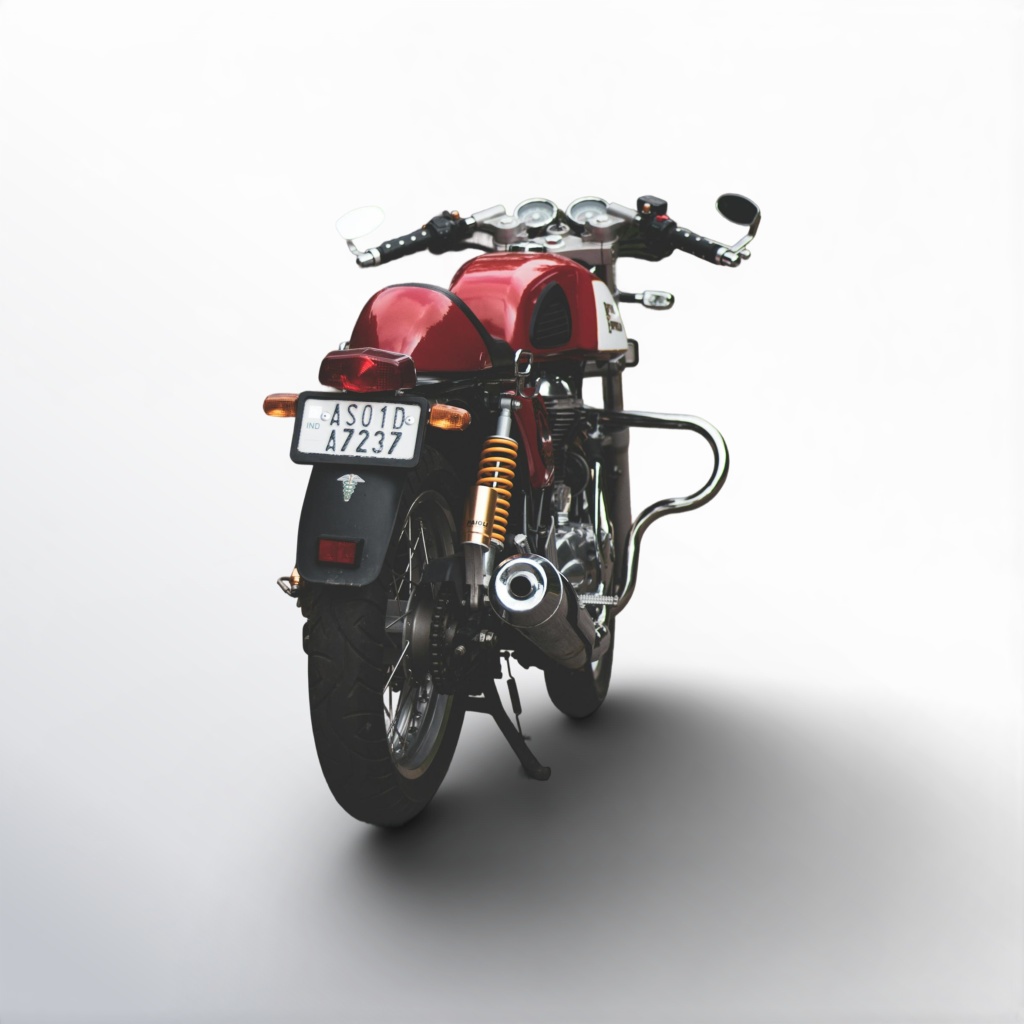}
        \caption{$s=8$}
    \end{subfigure}

    \begin{subfigure}{0.24\textwidth} 
        \centering
        \includegraphics[width=\textwidth]{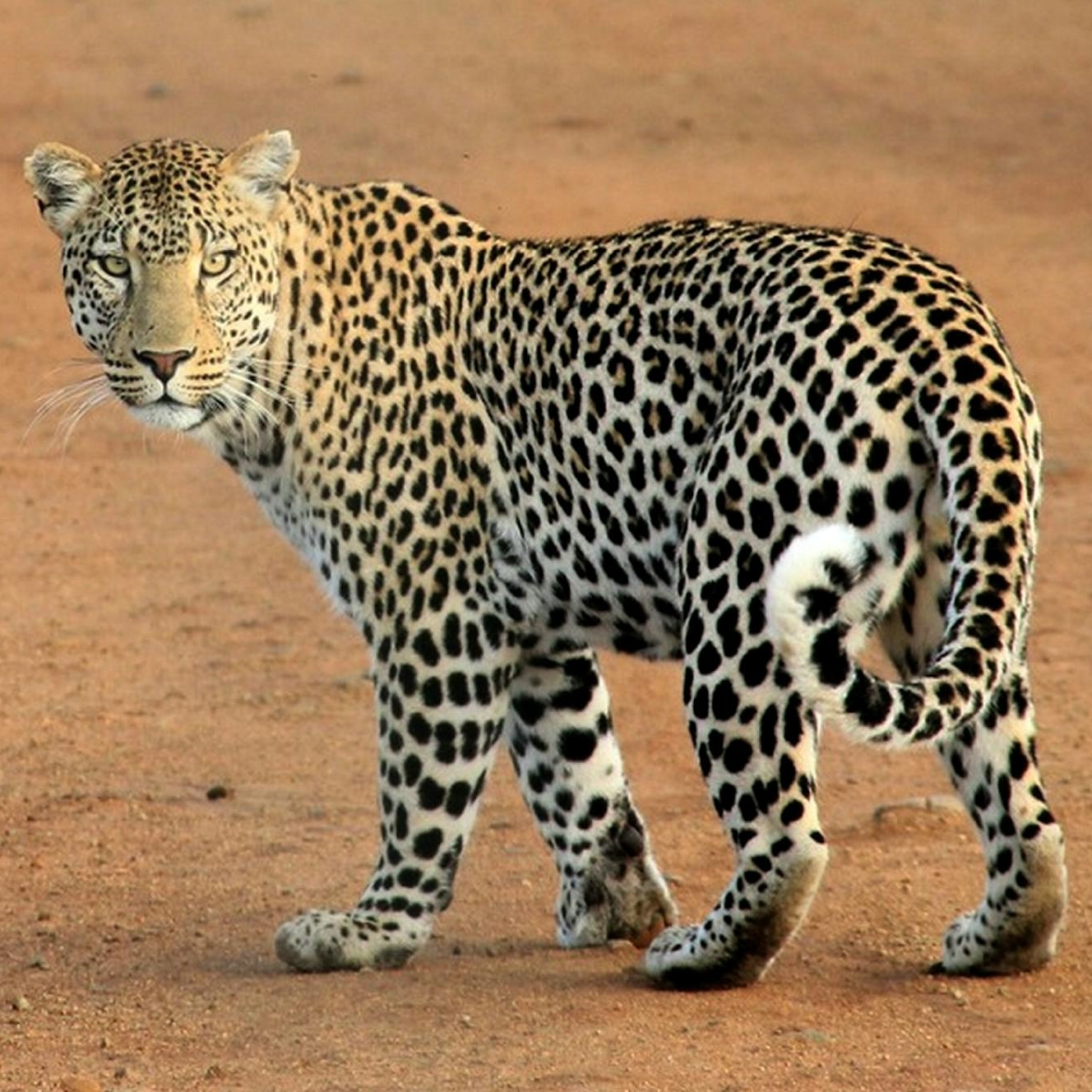}
        \caption{Input Image}
    \end{subfigure}
    \hfill
    \begin{subfigure}{0.24\textwidth} 
        \centering
        \includegraphics[width=\textwidth]{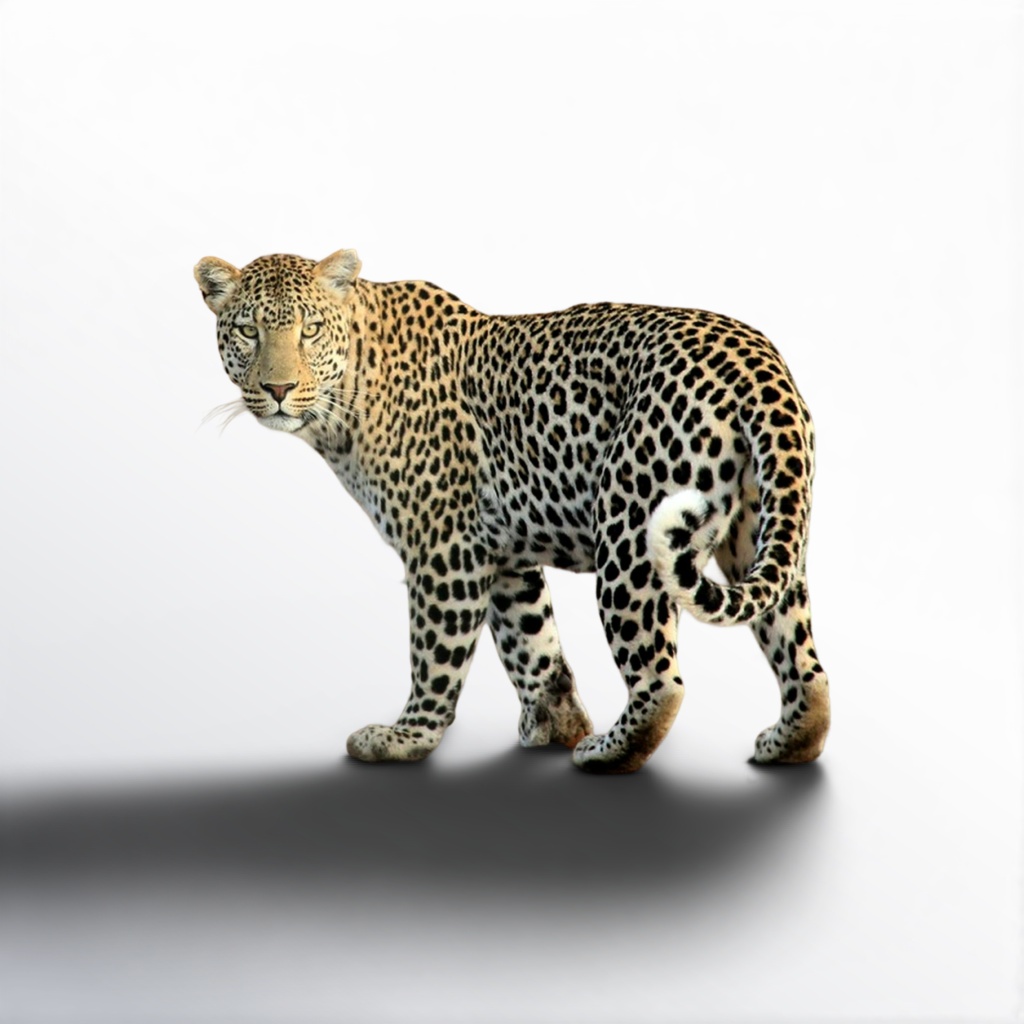}
        \caption{$s=2$}
    \end{subfigure}
    \hfill
    \begin{subfigure}{0.24\textwidth} 
        \centering
        \includegraphics[width=\textwidth]{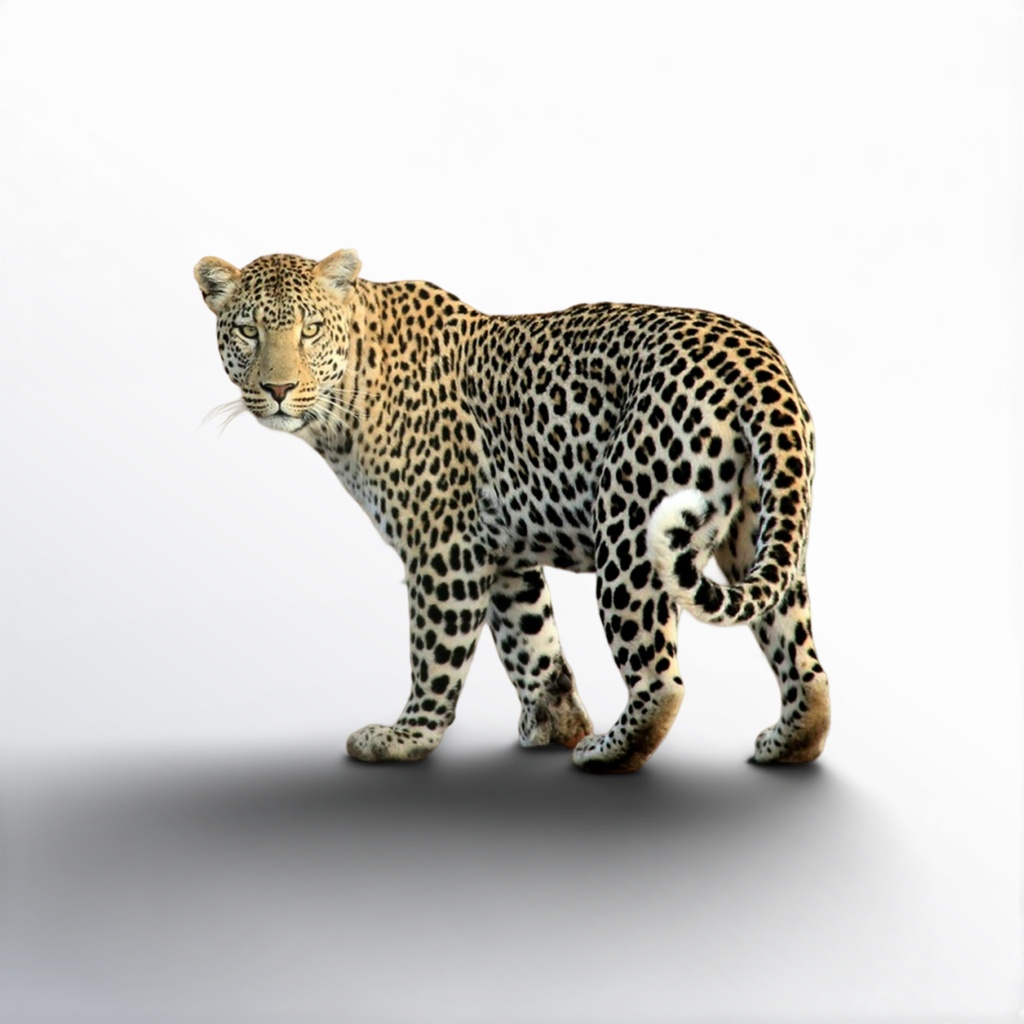}
        \caption{$s=5$}
    \end{subfigure}
    \hfill
    \begin{subfigure}{0.24\textwidth} 
        \centering
        \includegraphics[width=\textwidth]{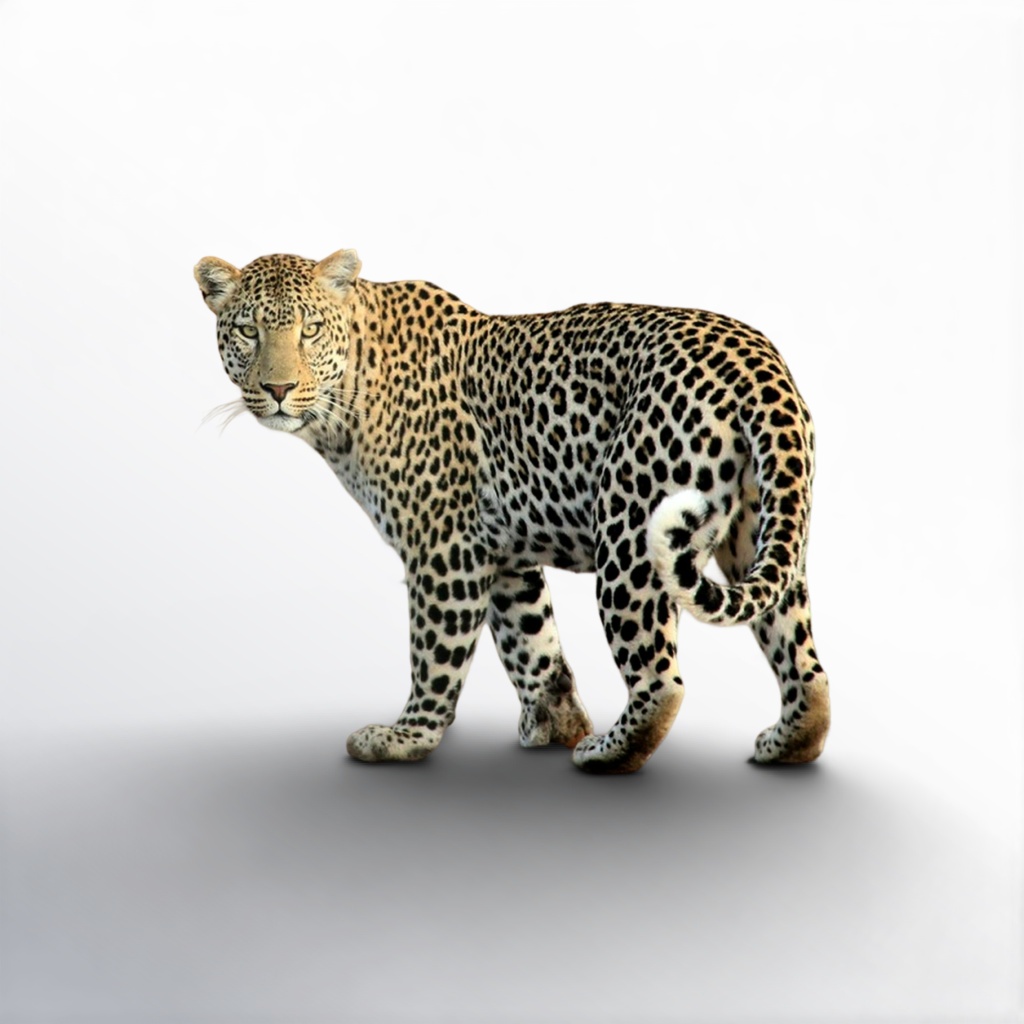}
        \caption{$s=8$}
    \end{subfigure}
    \caption{Softness control. The fixed light parameters are $\theta=30^\circ$ and $\phi=150^\circ$ for the top row, $\theta=30^\circ$ and $\phi=50^\circ$ for the bottom row. The varying parameter in both cases is $s$ to change the softness.}
    \label{fig:sm_real_images_softness_control_1}
\end{figure*}

\begin{figure*}[t]
    \centering
    \begin{subfigure}{0.24\textwidth} 
        \centering
        \includegraphics[width=\textwidth]{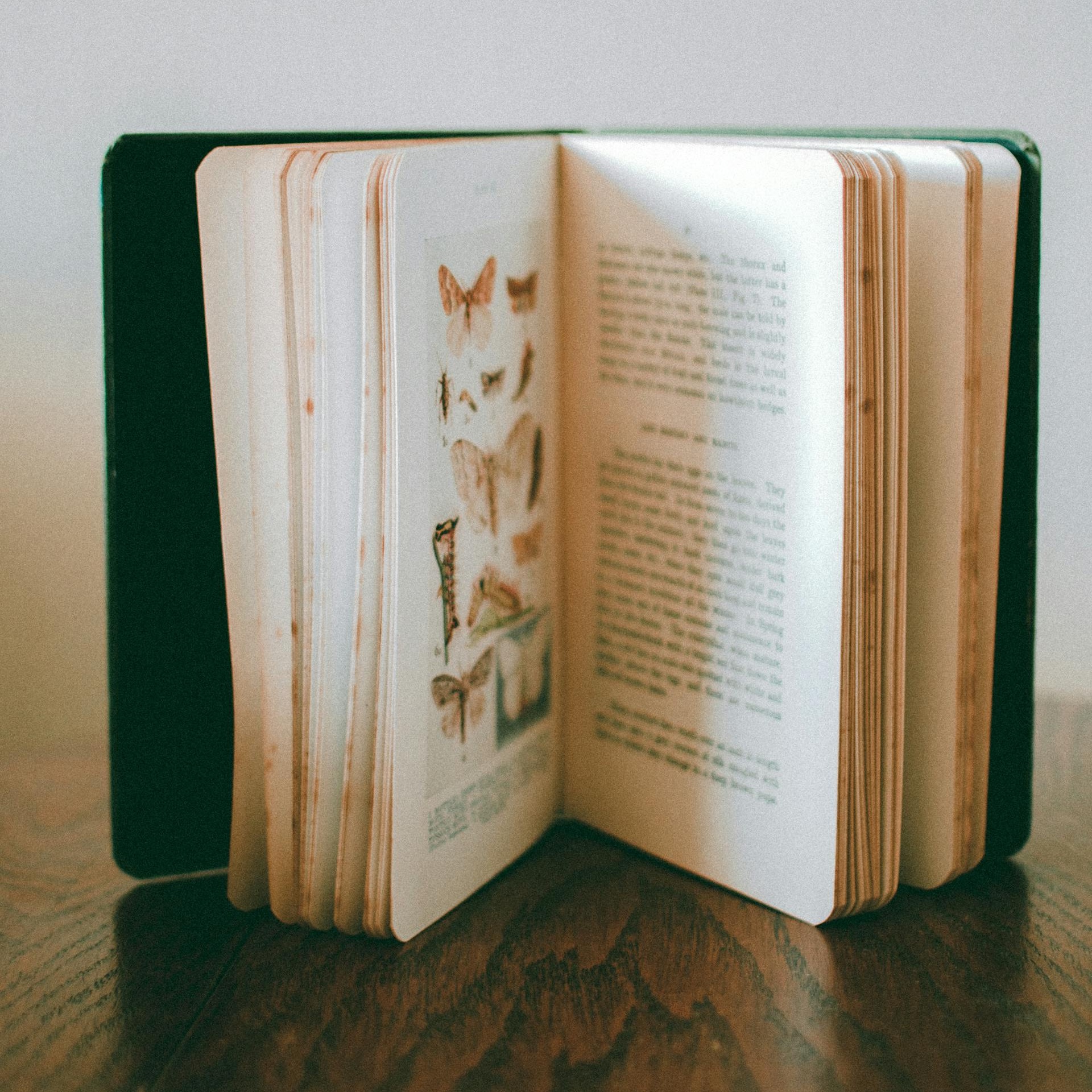}
        \caption{Input image}
    \end{subfigure}
    \hfill
    \begin{subfigure}{0.24\textwidth} 
        \centering
        \includegraphics[width=\textwidth]{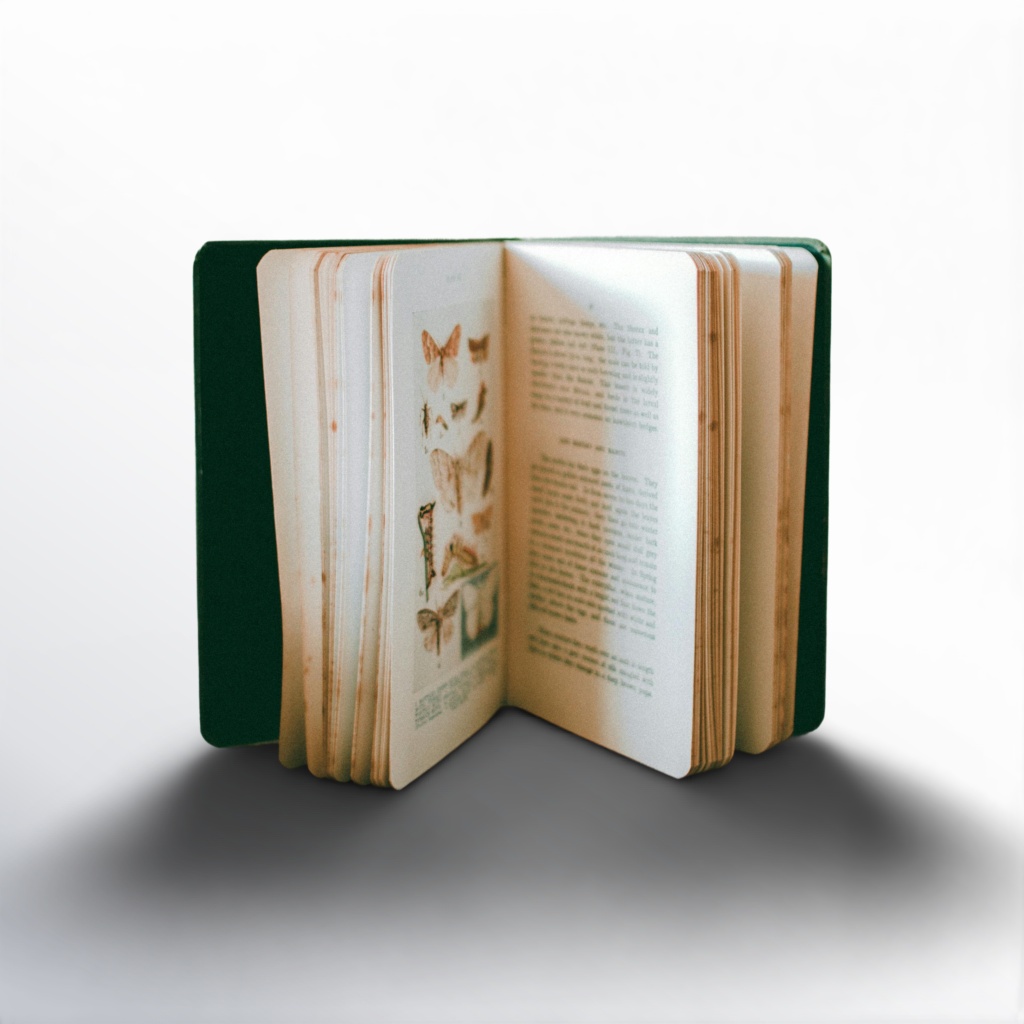}
        \caption{$s=2$}
    \end{subfigure}
    \hfill
    \begin{subfigure}{0.24\textwidth} 
        \centering
        \includegraphics[width=\textwidth]{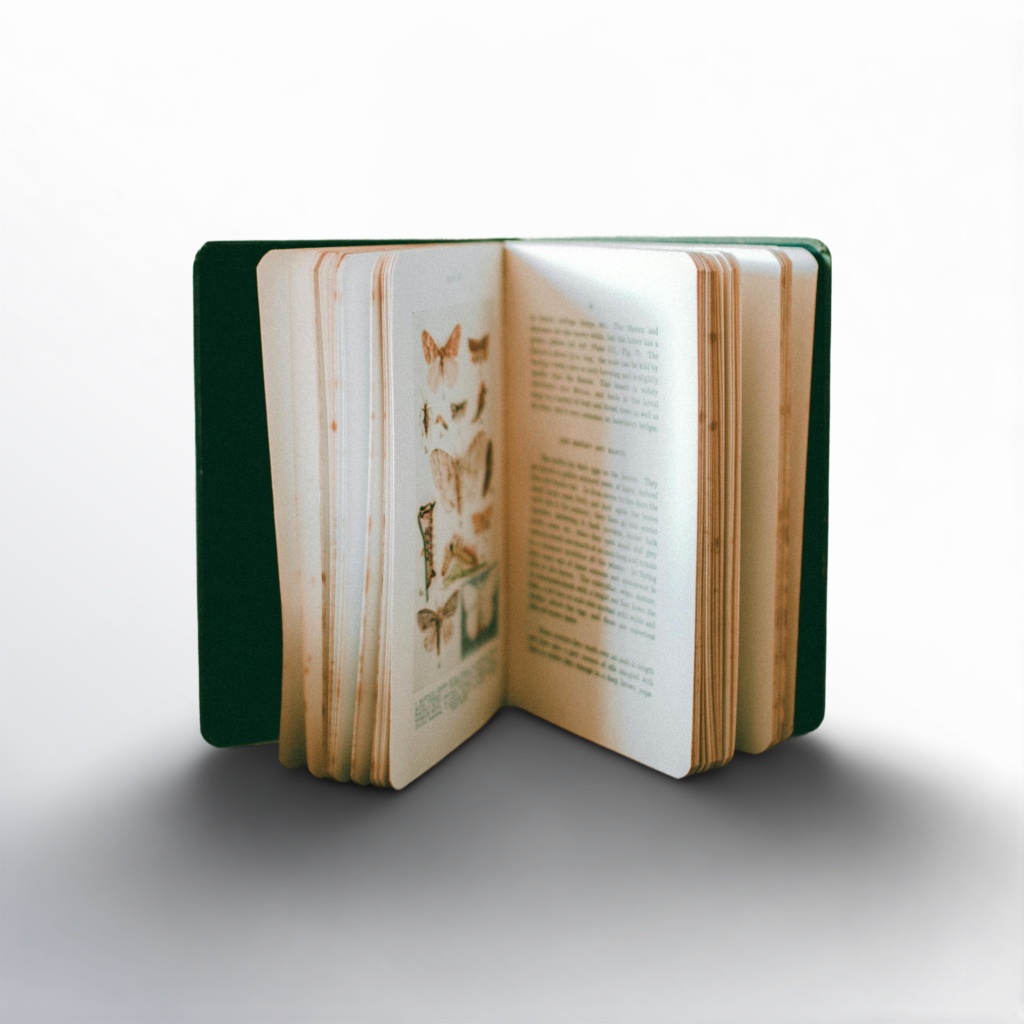}
        \caption{$s=4$}
    \end{subfigure}
    \hfill
    \begin{subfigure}{0.24\textwidth} 
        \centering
        \includegraphics[width=\textwidth]{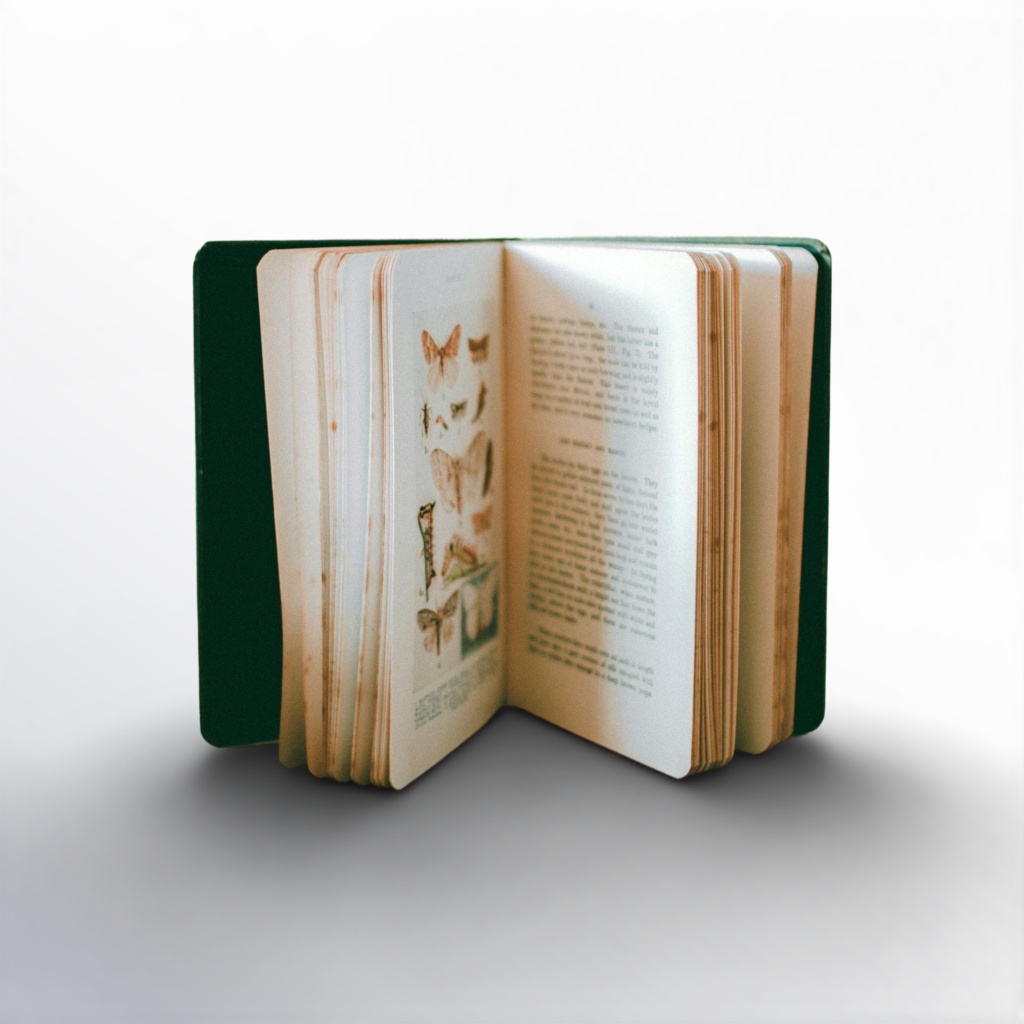}
        \caption{$s=6$}
    \end{subfigure}
    \hfill

    \begin{subfigure}{0.24\textwidth} 
        \centering
        \includegraphics[width=\textwidth]{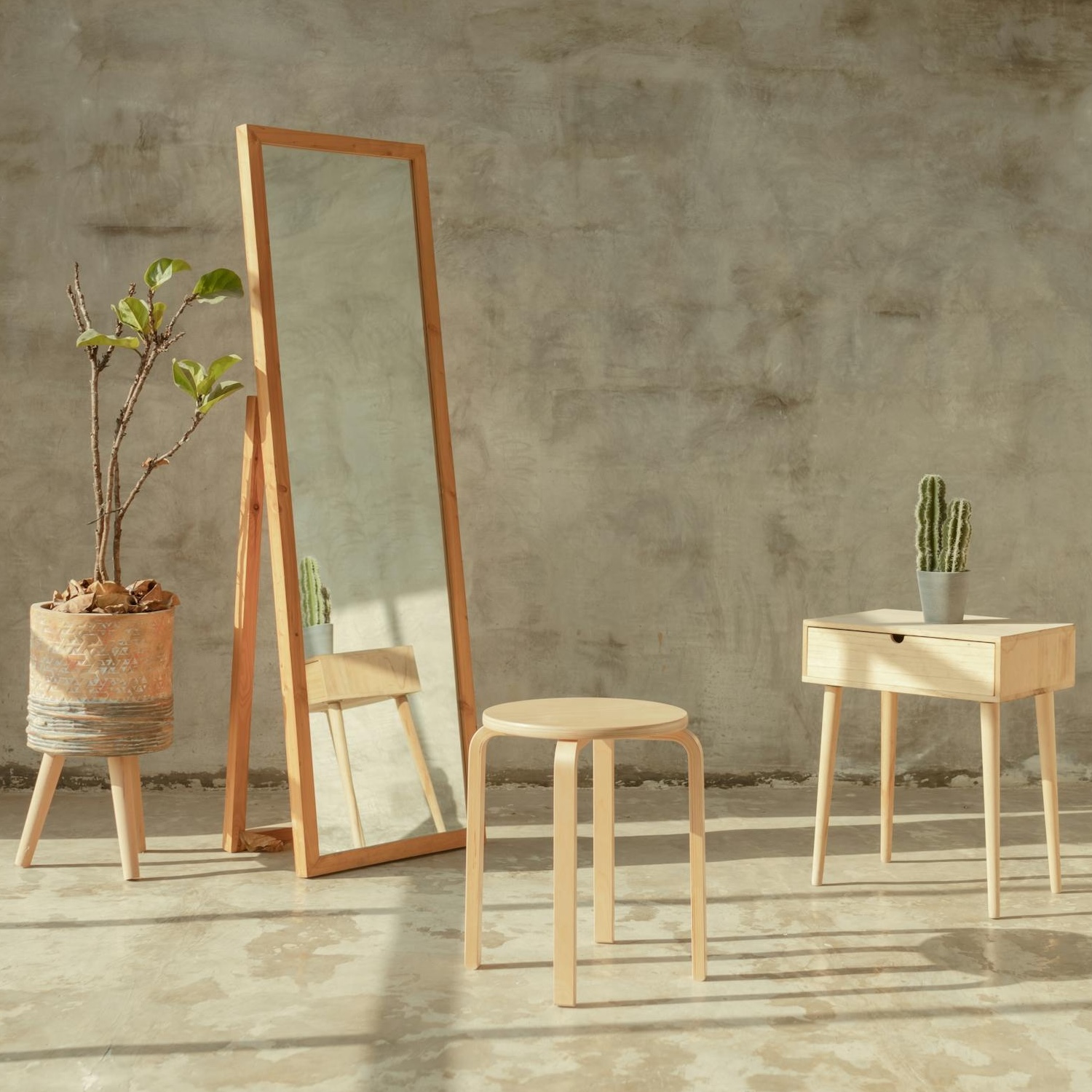}
        \caption{Input image}
    \end{subfigure}
    \hfill
    \begin{subfigure}{0.24\textwidth} 
        \centering
        \includegraphics[width=\textwidth]{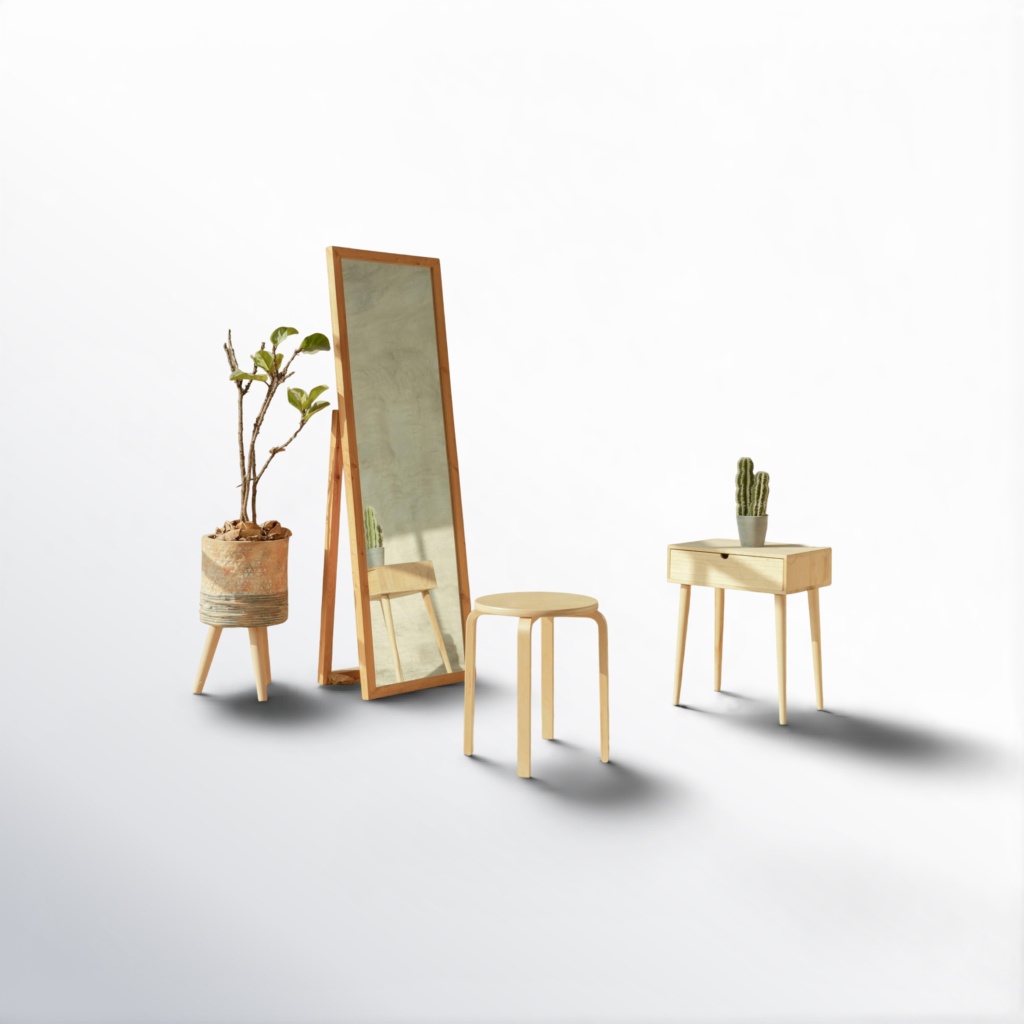}
        \caption{$s=3$}
    \end{subfigure}
    \hfill
    \begin{subfigure}{0.24\textwidth} 
        \centering
        \includegraphics[width=\textwidth]{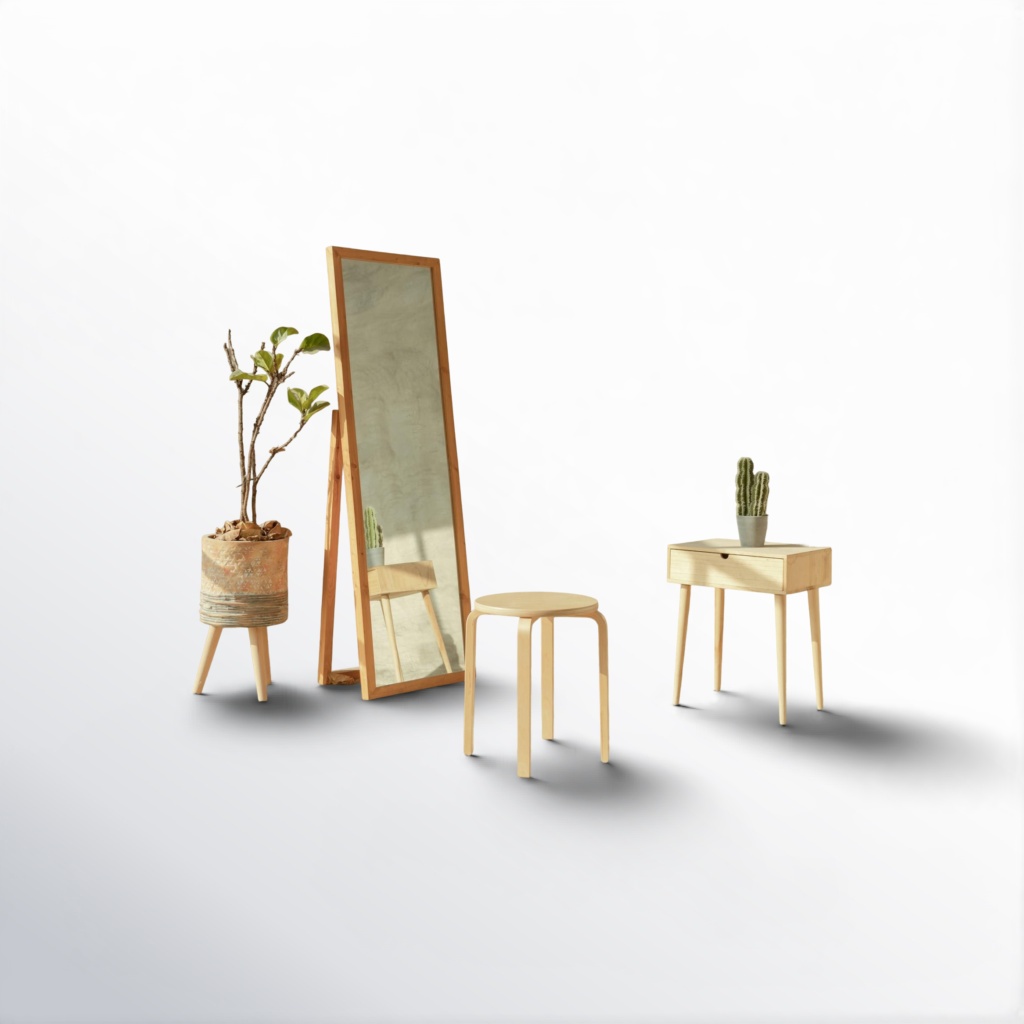}
        \caption{$s=5$}
    \end{subfigure}
    \hfill
    \begin{subfigure}{0.24\textwidth} 
        \centering
        \includegraphics[width=\textwidth]{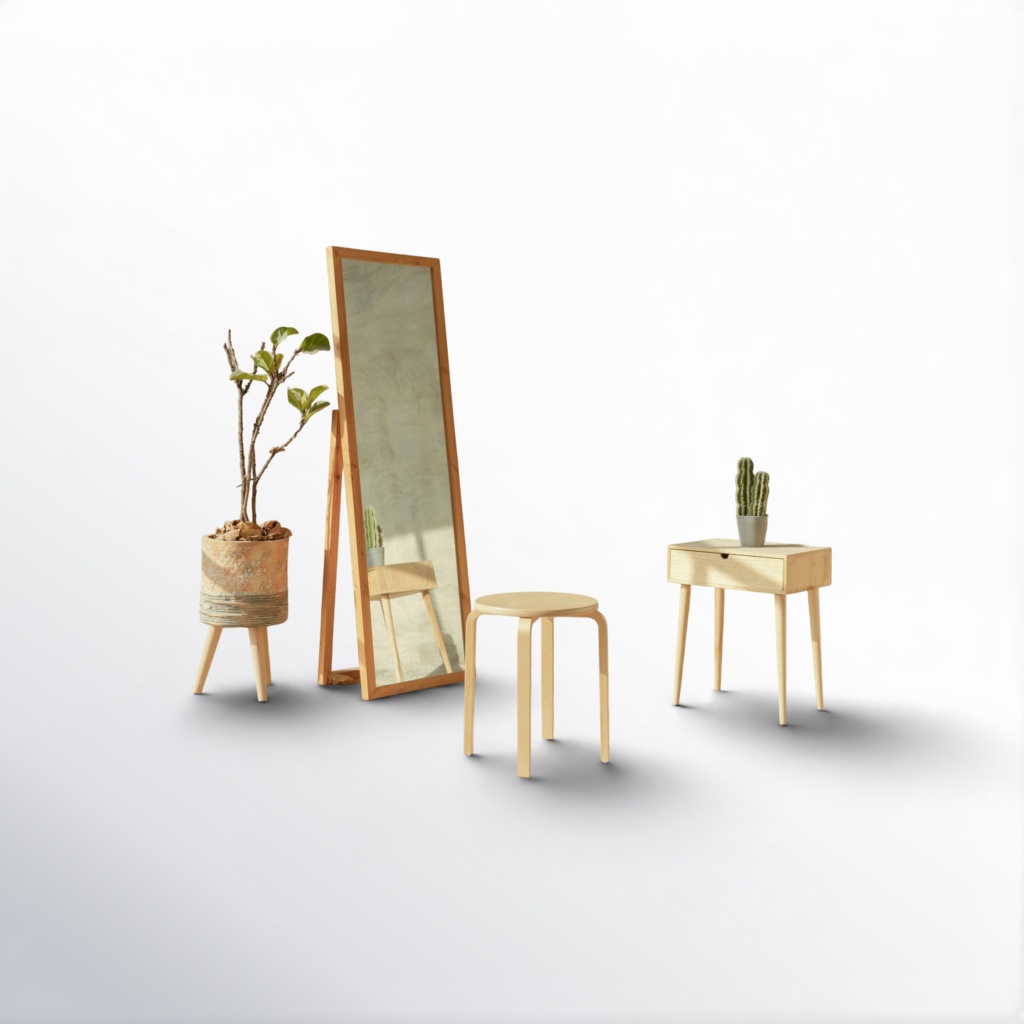}
        \caption{$s=7$}
    \end{subfigure}
    \caption{Softness control. The fixed light parameters are $\theta=20^\circ$ and $\phi=90^\circ$ for the top row, $\theta=35^\circ$ and $\phi=135^\circ$ for the bottom row. The varying parameter in both cases is $s$ to change the softness.}
    \label{fig:sm_real_images_softness_control_2}
\end{figure*}

\begin{figure*}[t]
    \centering

    \begin{subfigure}{0.24\textwidth} 
        \centering
        \includegraphics[width=\textwidth]{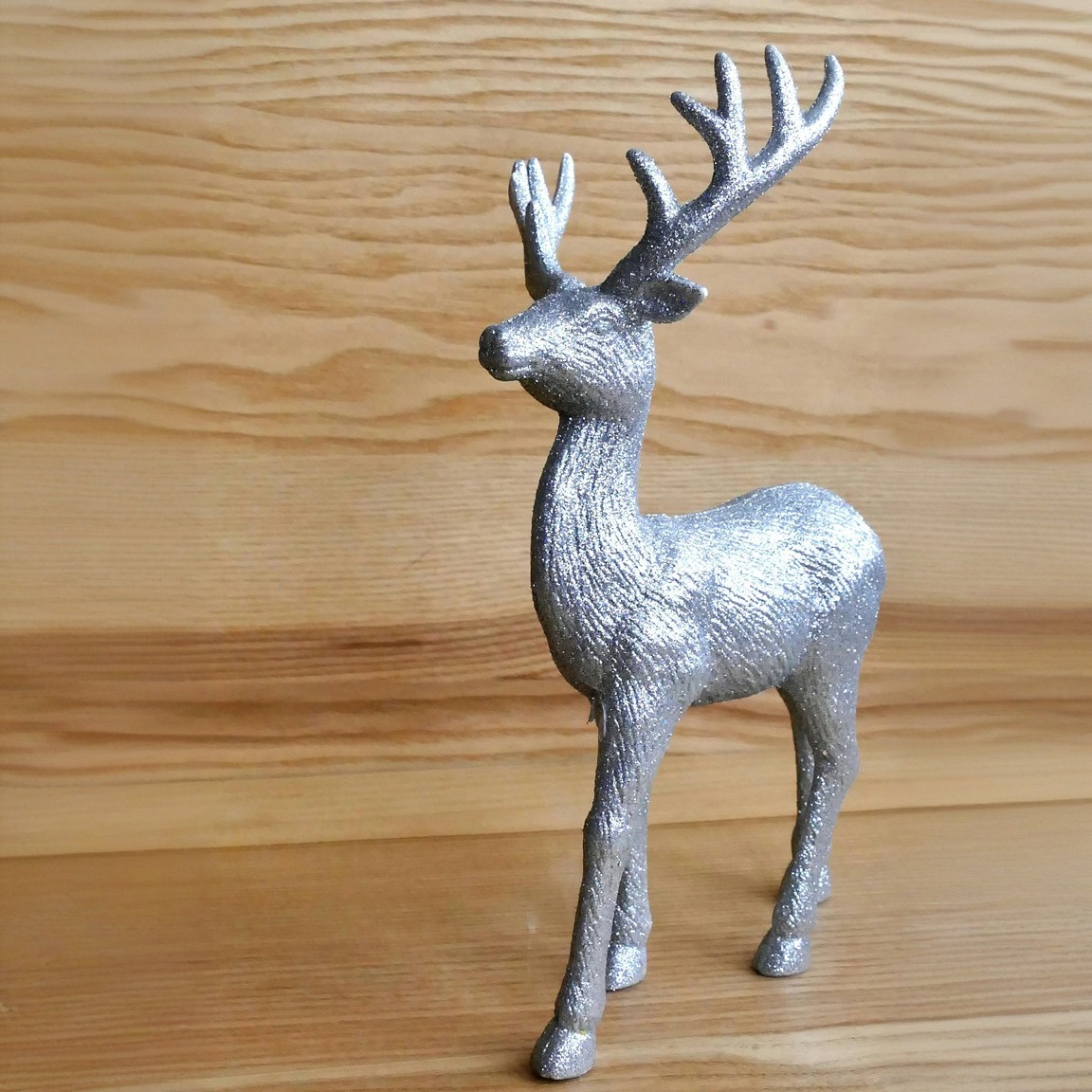}
        \caption{Input image}
    \end{subfigure}
    \hfill
    \begin{subfigure}{0.24\textwidth} 
        \centering
        \includegraphics[width=\textwidth]{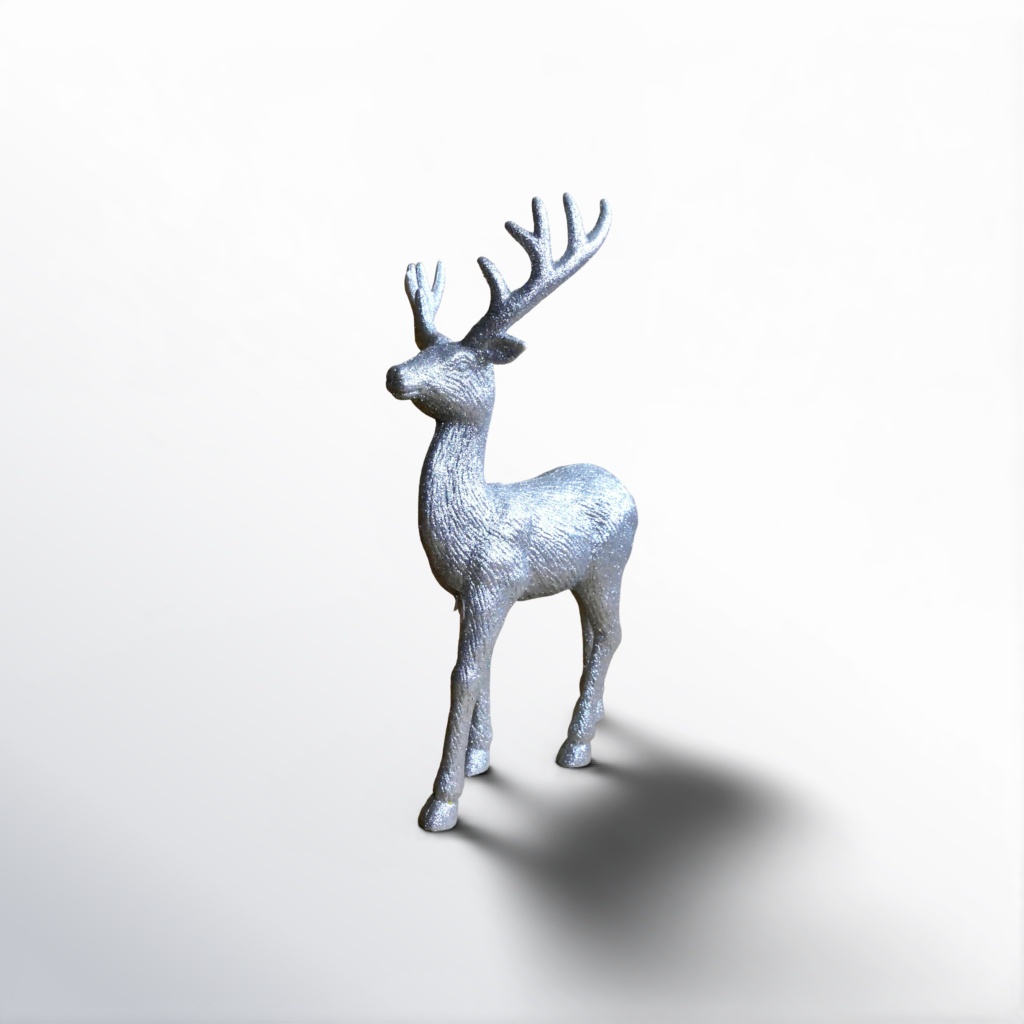}
        \caption{$\phi=135^\circ$}
    \end{subfigure}
    \hfill
    \begin{subfigure}{0.24\textwidth} 
        \centering
        \includegraphics[width=\textwidth]{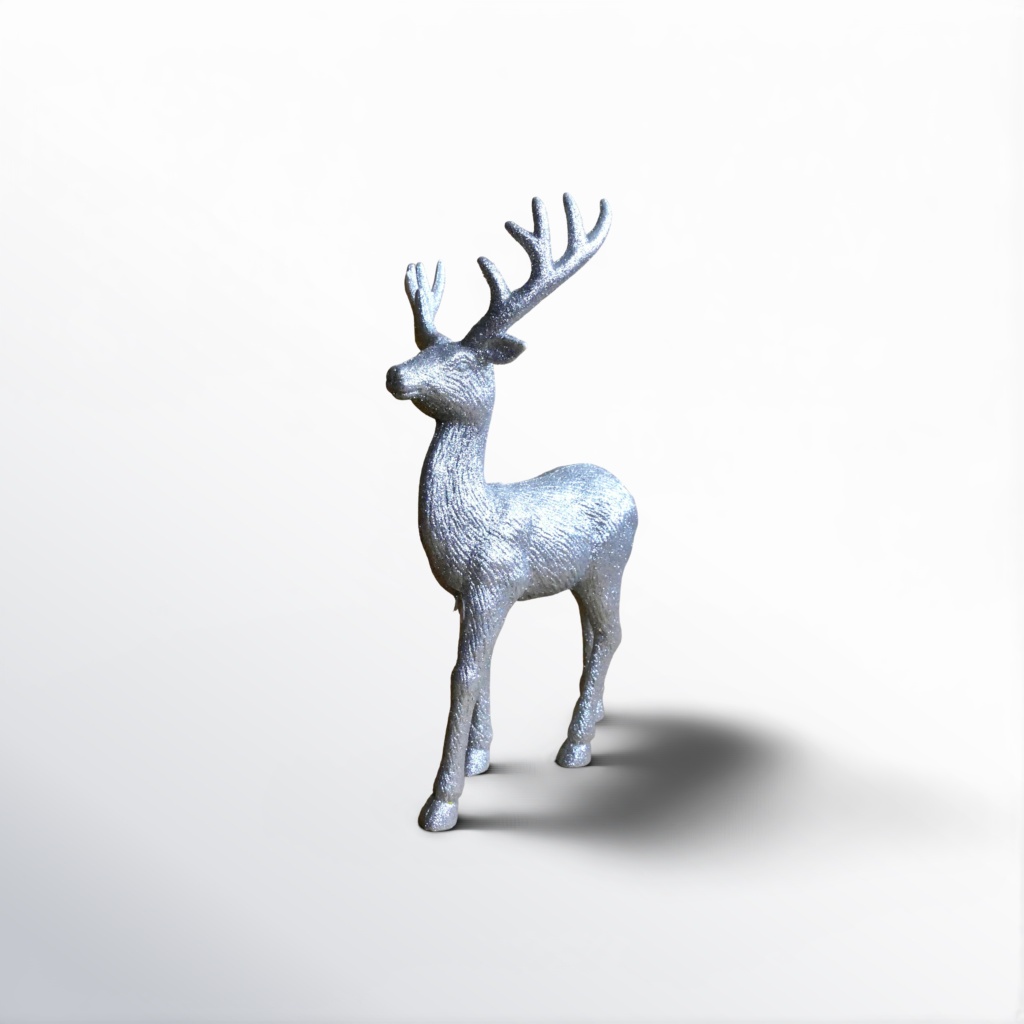}
        \caption{$\phi=180^\circ$}
    \end{subfigure}
    \hfill
    \begin{subfigure}{0.24\textwidth} 
        \centering
        \includegraphics[width=\textwidth]{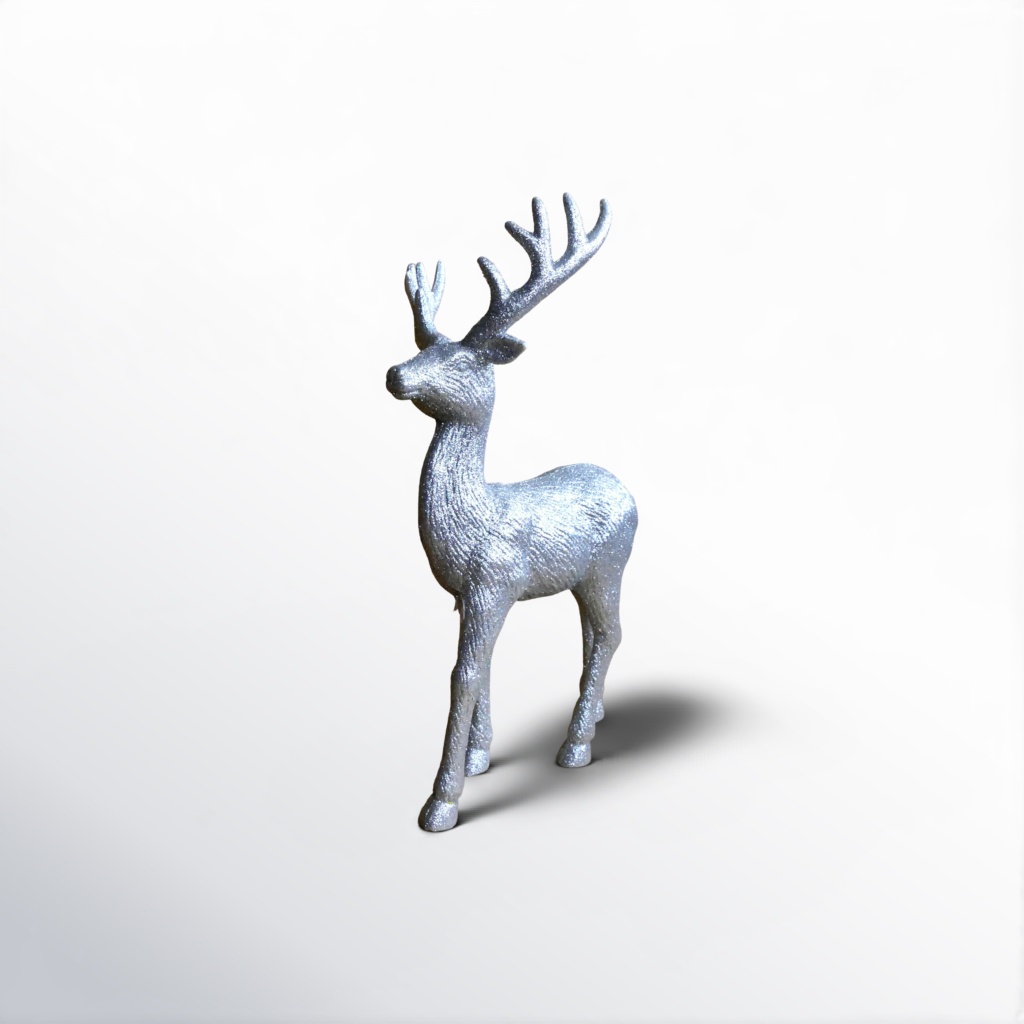}
        \caption{$\phi=225^\circ$}
    \end{subfigure}

    \begin{subfigure}{0.24\textwidth} 
        \centering
        \includegraphics[width=\textwidth]{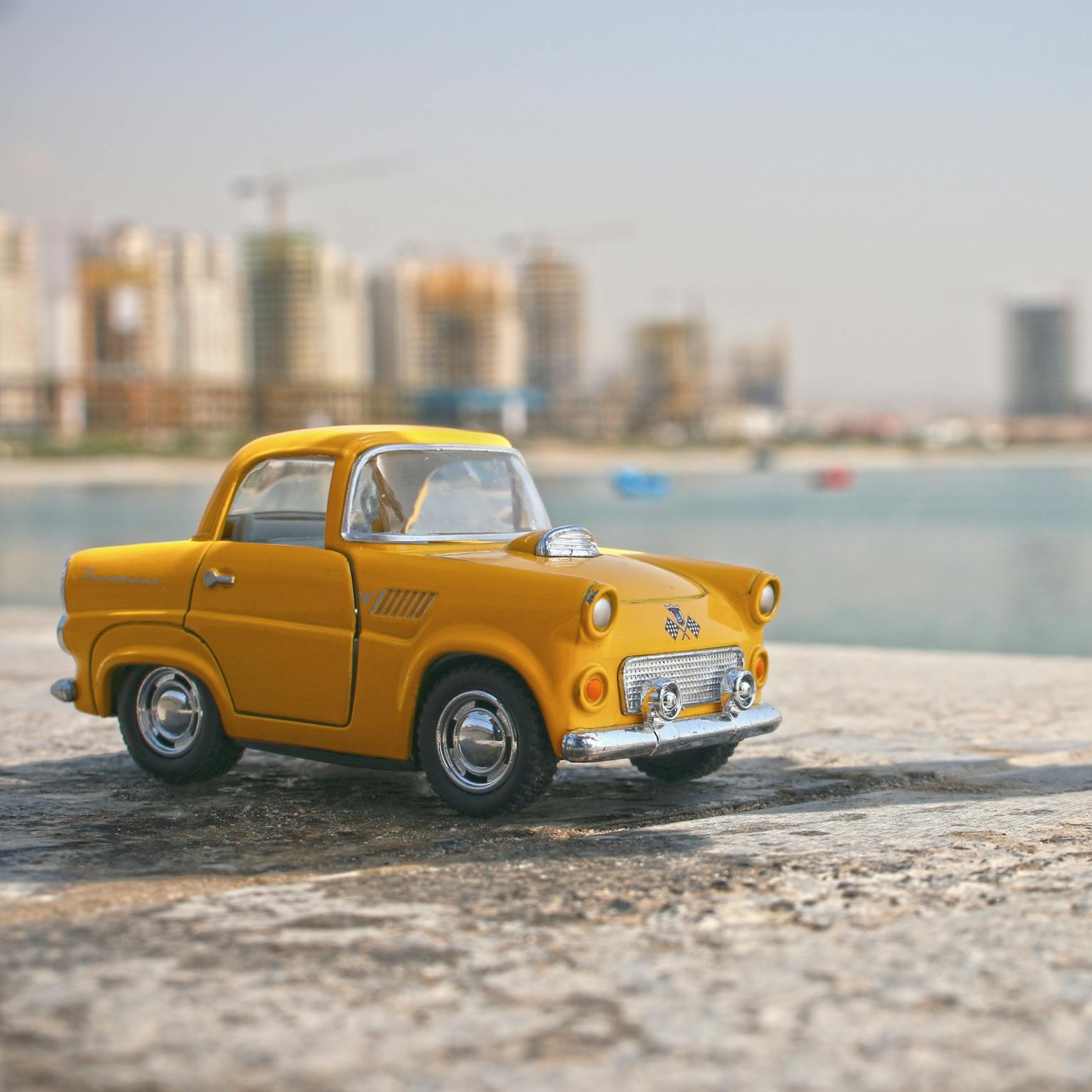}
        \caption{Input image}
    \end{subfigure}
    \hfill
    \begin{subfigure}{0.24\textwidth} 
        \centering
        \includegraphics[width=\textwidth]{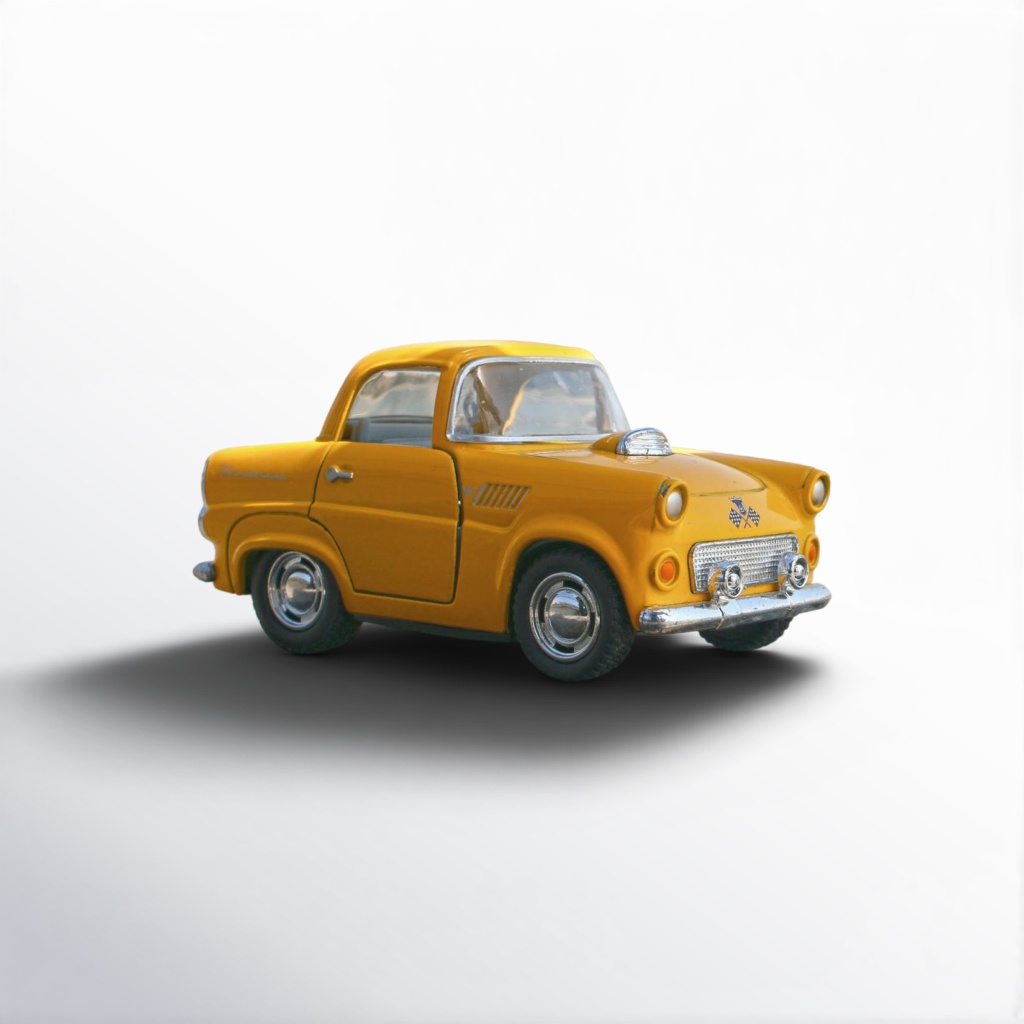}
        \caption{$\phi=50^\circ$}
    \end{subfigure}
    \hfill
    \begin{subfigure}{0.24\textwidth} 
        \centering
        \includegraphics[width=\textwidth]{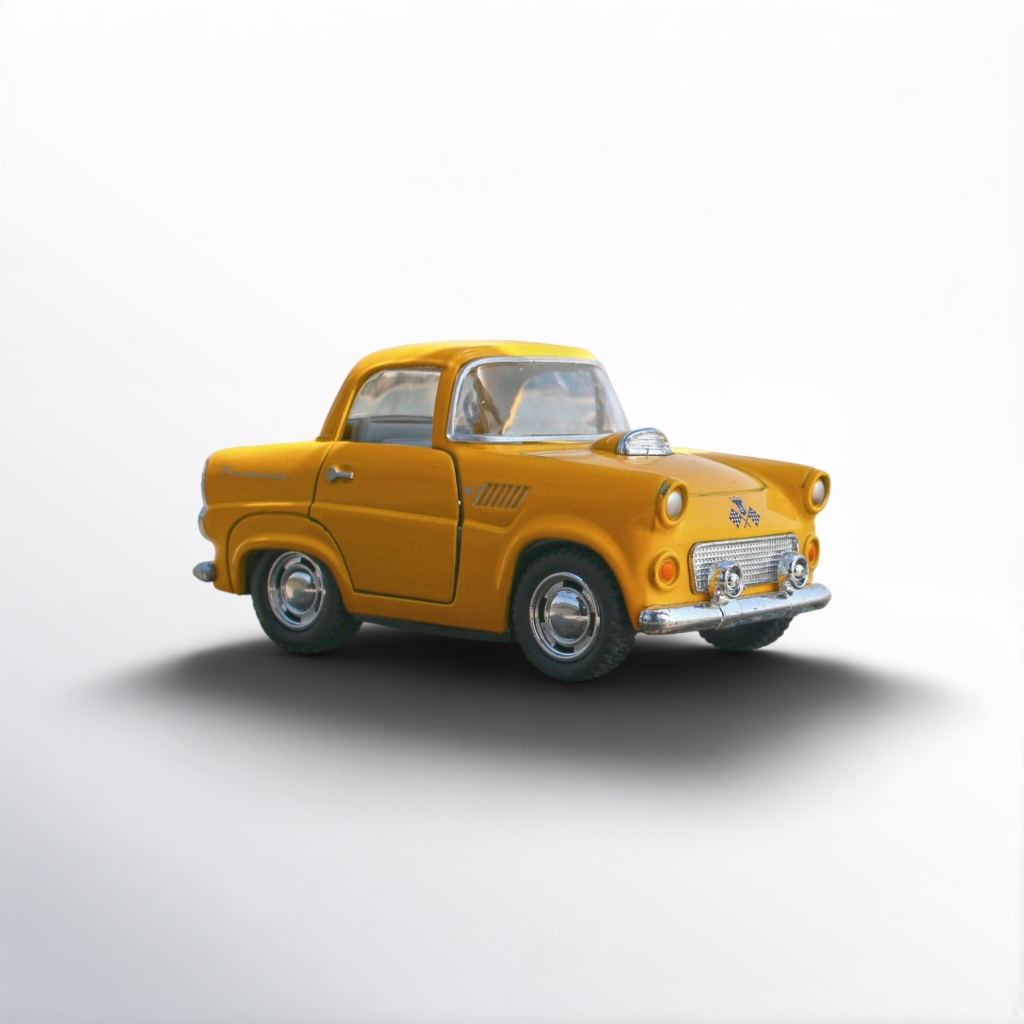}
        \caption{$\phi=90^\circ$}
    \end{subfigure}
    \hfill
    \begin{subfigure}{0.24\textwidth} 
        \centering
        \includegraphics[width=\textwidth]{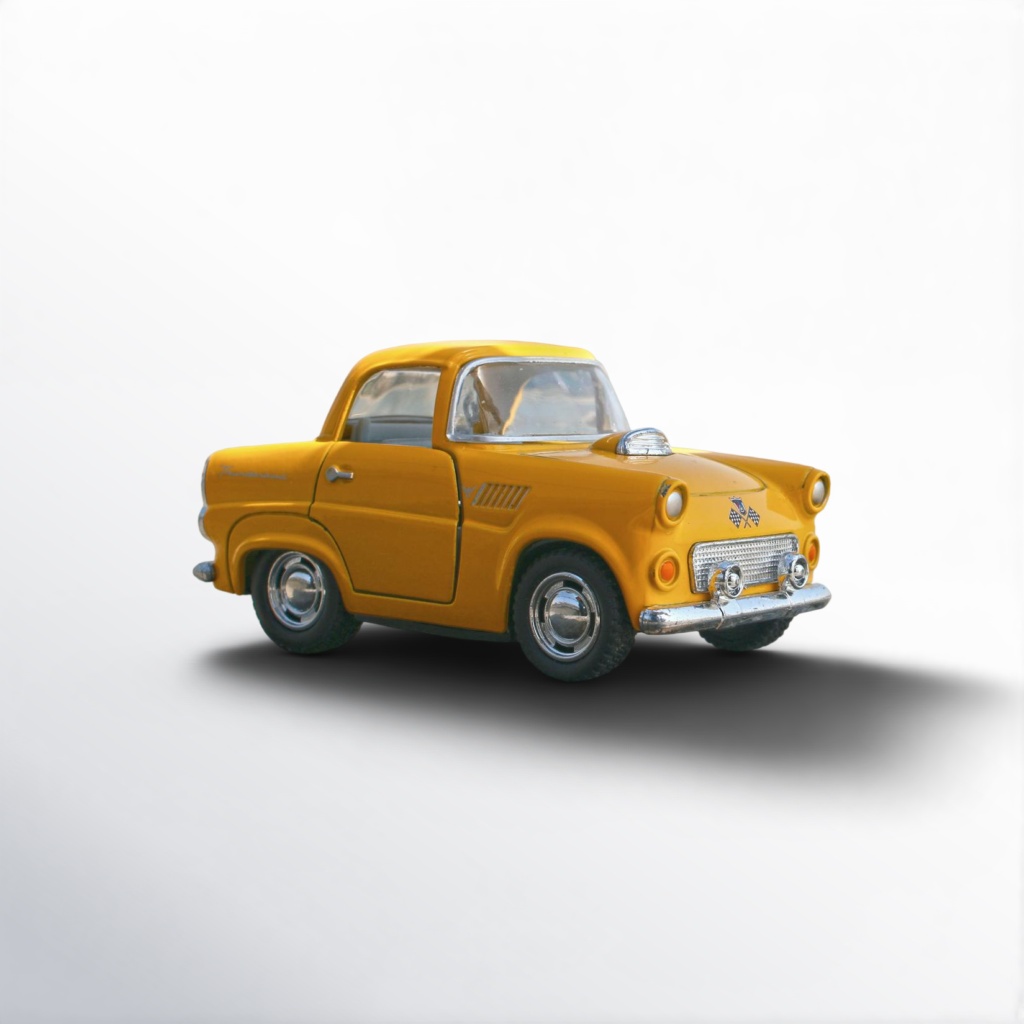}
        \caption{$\phi=120^\circ$}
    \end{subfigure}
    
    \caption{Horizontal shadow direction control. The fixed light parameters are $\theta=30^\circ$ and $s=2$ for the top row, $\theta=40^\circ$ and $s=3$ for the bottom row. The varying parameter in both cases is $\phi$ to move the shadow horizontally.}
    \label{fig:sm_real_images_horz_control}
\end{figure*}

\begin{figure*}[t]
    \centering
    \begin{subfigure}{0.24\textwidth} 
        \centering
        \includegraphics[width=\textwidth]{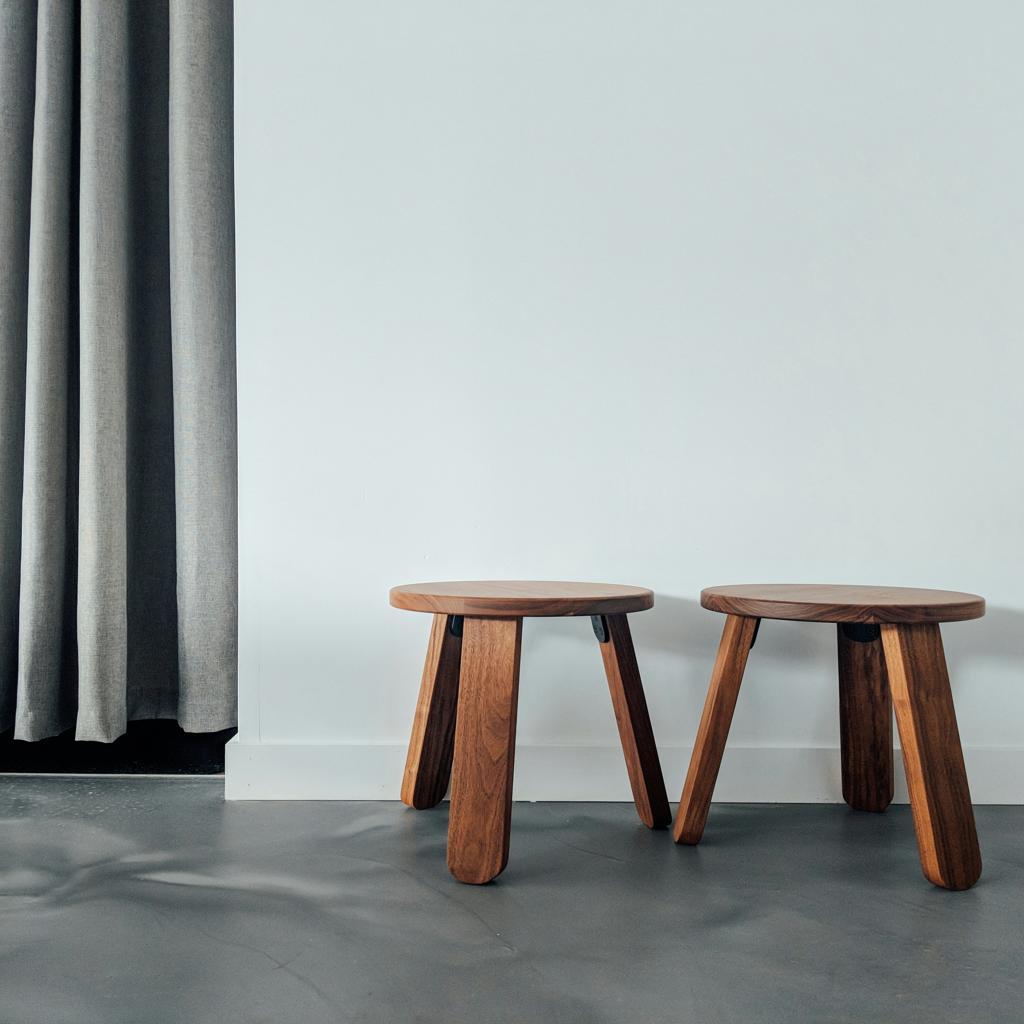}
        \caption{Input image}
    \end{subfigure}
    \hfill
    \begin{subfigure}{0.24\textwidth} 
        \centering
        \includegraphics[width=\textwidth]{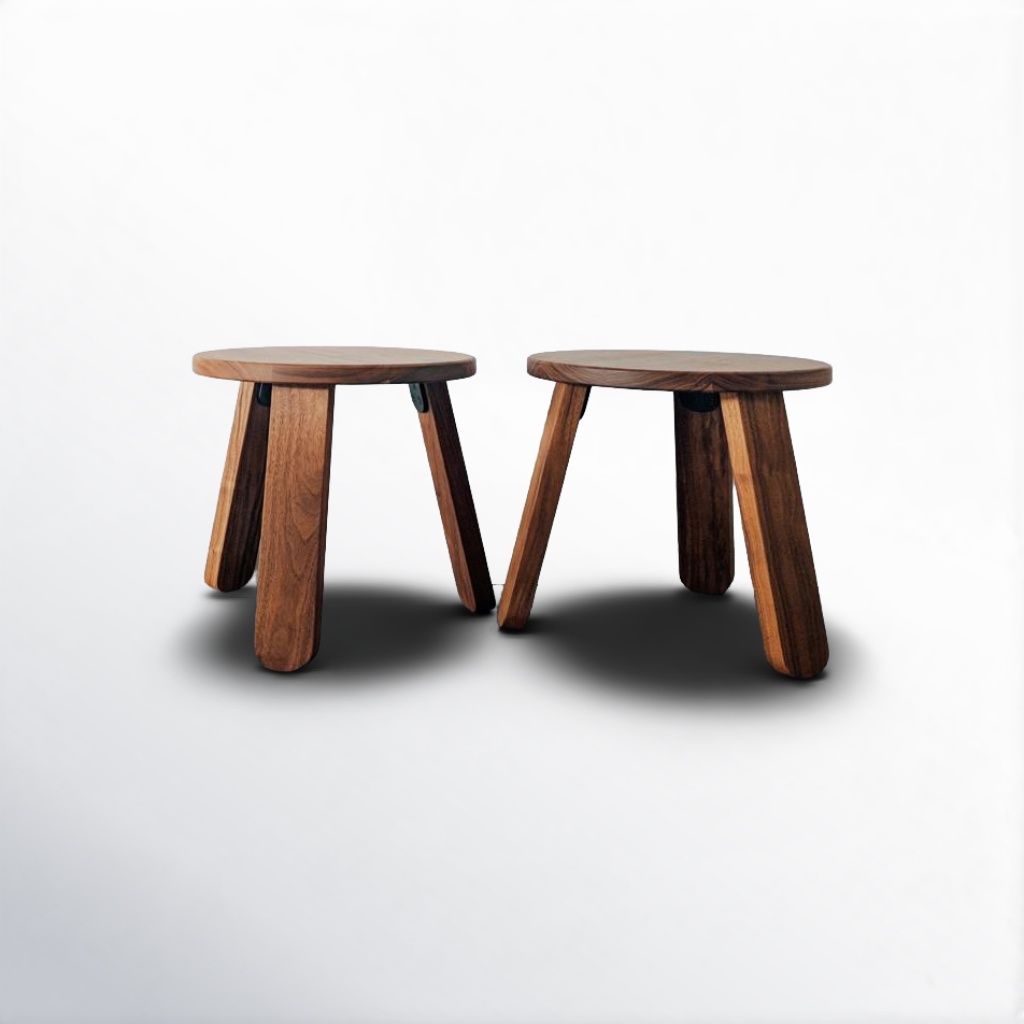}
        \caption{$\theta=0^\circ$}
    \end{subfigure}
    \hfill
    \begin{subfigure}{0.24\textwidth} 
        \centering
        \includegraphics[width=\textwidth]{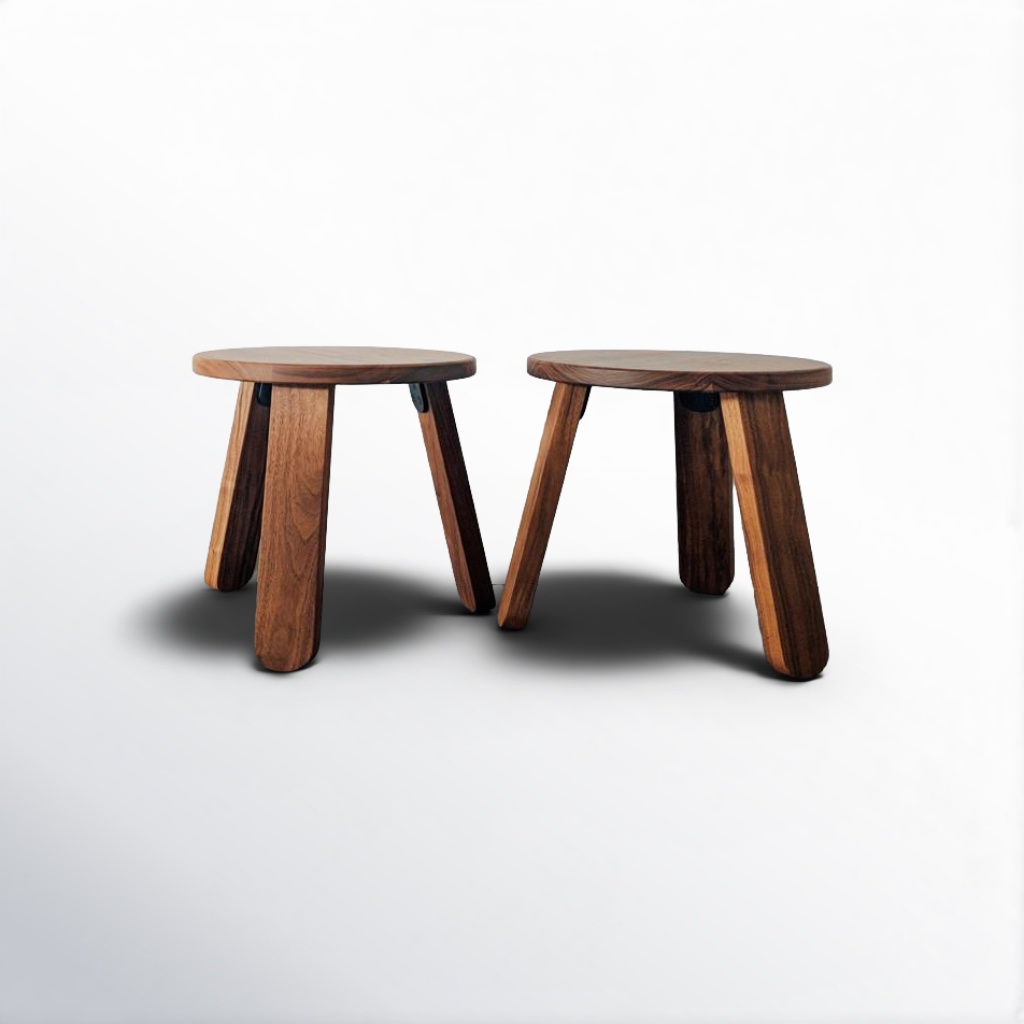}
        \caption{$\theta=25^\circ$}
    \end{subfigure}
    \hfill
    \begin{subfigure}{0.24\textwidth} 
        \centering
        \includegraphics[width=\textwidth]{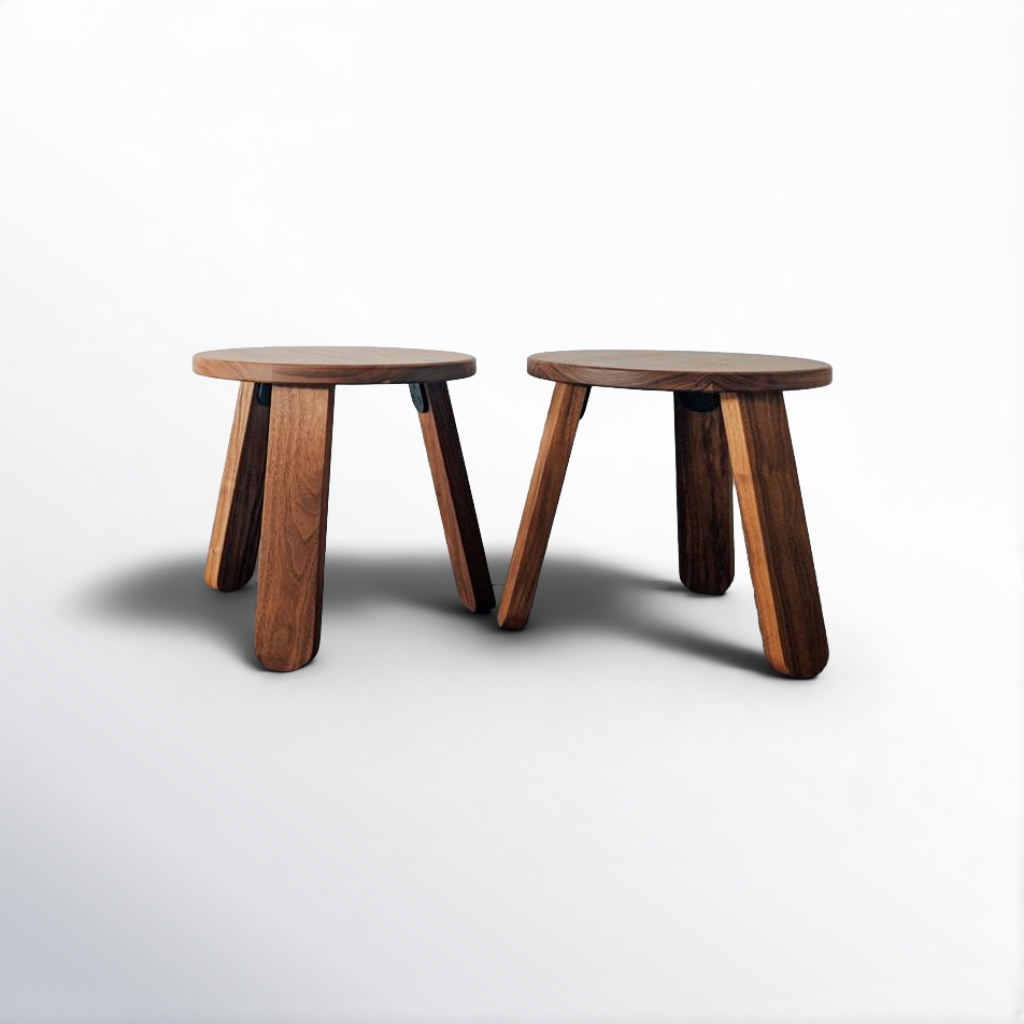}
        \caption{$\theta=45^\circ$}
    \end{subfigure}

    \begin{subfigure}{0.24\textwidth} 
        \centering
        \includegraphics[width=\textwidth]{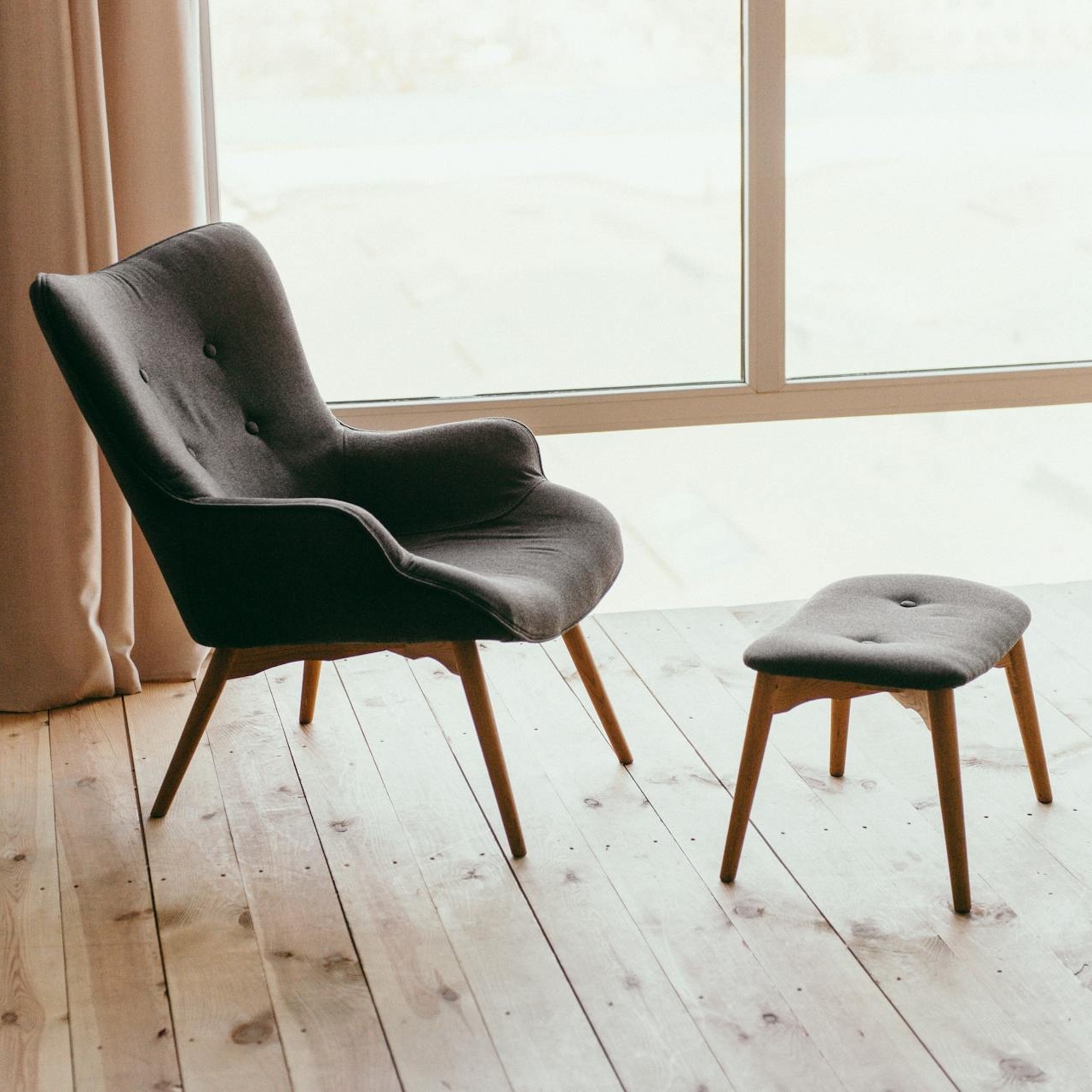}
        \caption{Input image}
    \end{subfigure}
    \hfill
    \begin{subfigure}{0.24\textwidth} 
        \centering
        \includegraphics[width=\textwidth]{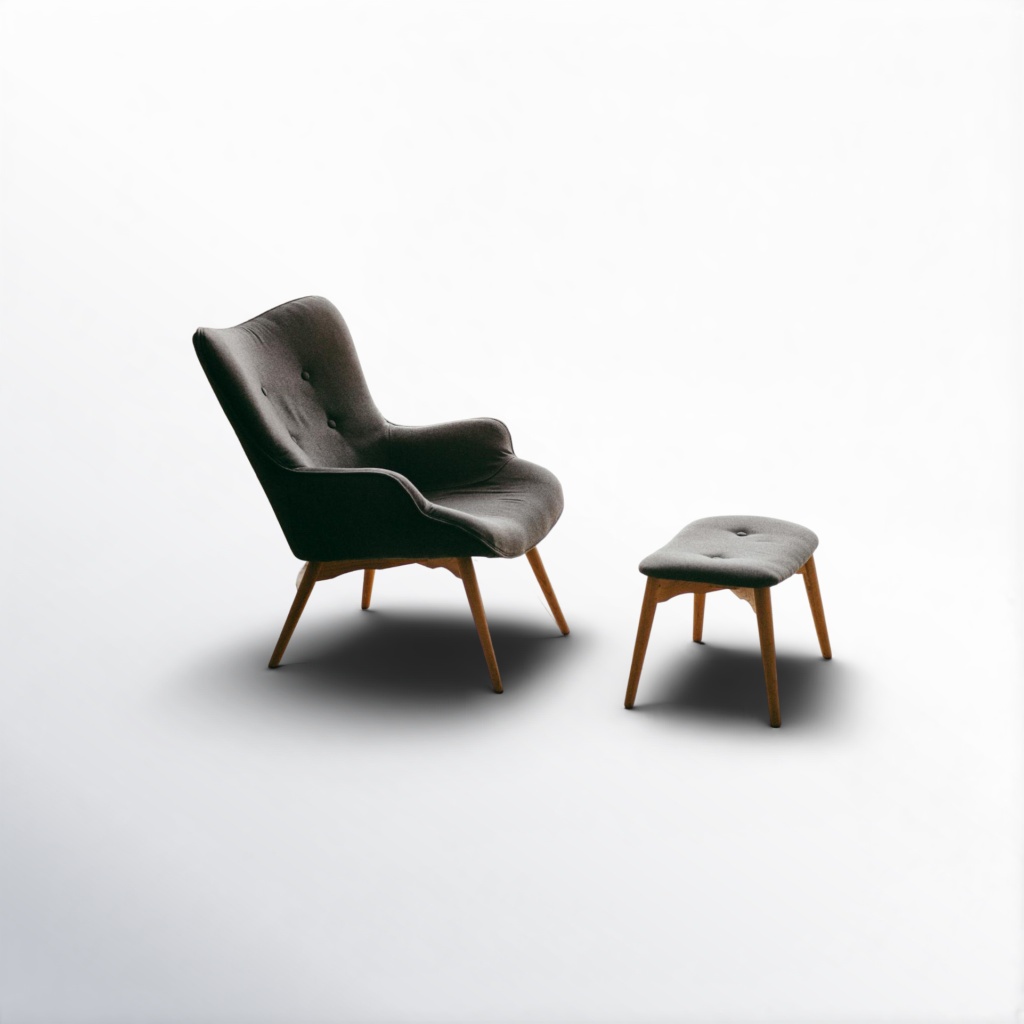}
        \caption{$\theta=5^\circ$}
    \end{subfigure}
    \hfill
    \begin{subfigure}{0.24\textwidth} 
        \centering
        \includegraphics[width=\textwidth]{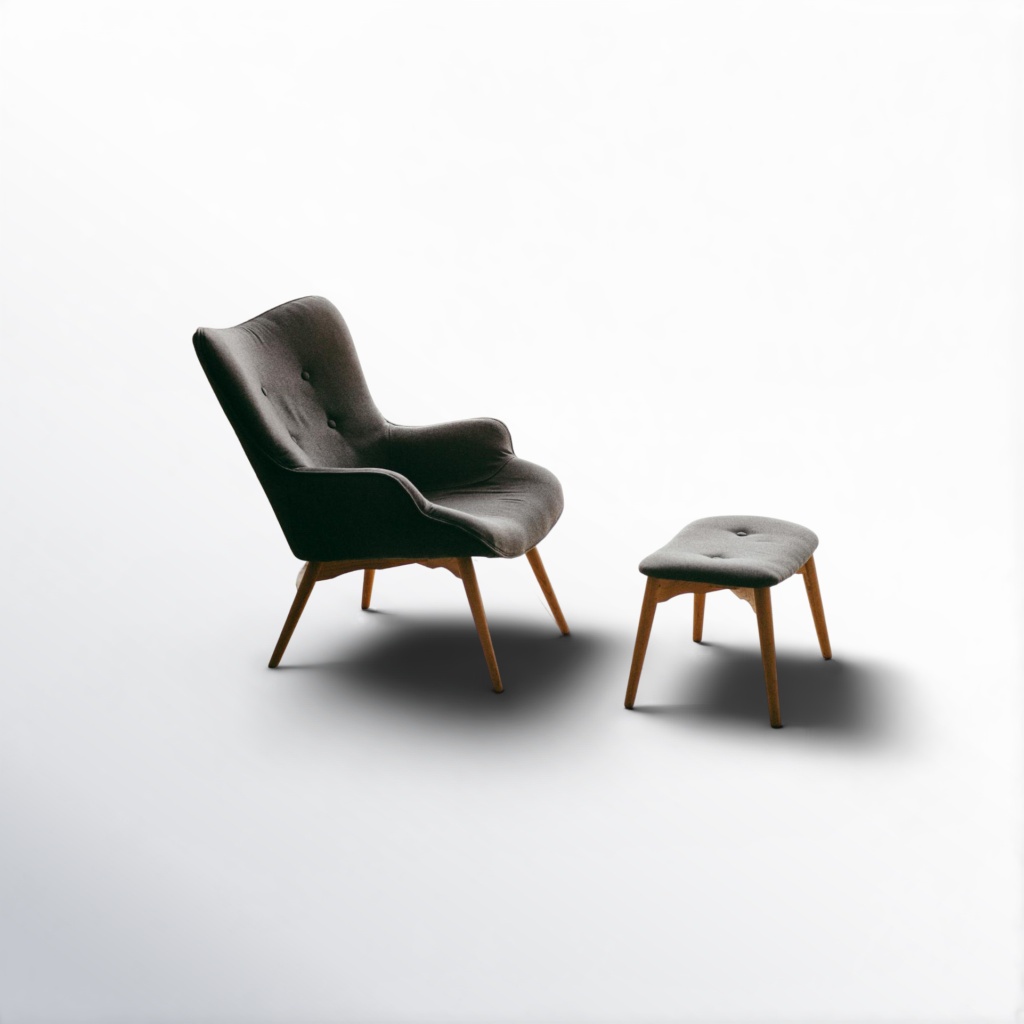}
        \caption{$\theta=25^\circ$}
    \end{subfigure}
    \hfill
    \begin{subfigure}{0.24\textwidth} 
        \centering
        \includegraphics[width=\textwidth]{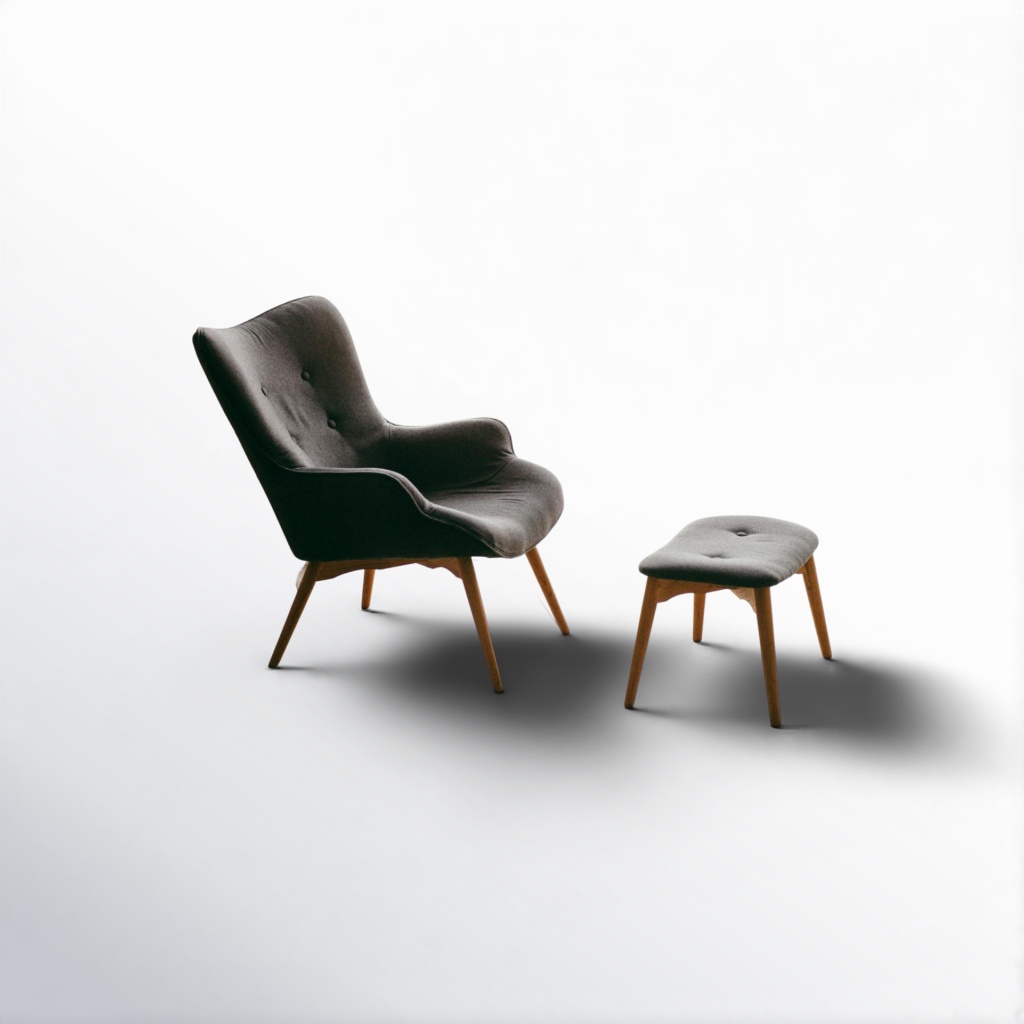}
        \caption{$\theta=40^\circ$}
    \end{subfigure}
    
    \caption{Vertical shadow direction control. The fixed light parameters are $\phi=315$ and $s=2$ for the top row, $\phi=160$ and $s=4$ for the bottom row. The varying parameter in both cases is $\theta$ to move the shadow vertically.}    
    \label{fig:sm_real_images_vert_control}
\end{figure*}

\begin{figure*}
    \centering
    \begin{subfigure}{0.24\textwidth} 
        \centering
        \includegraphics[width=\textwidth]{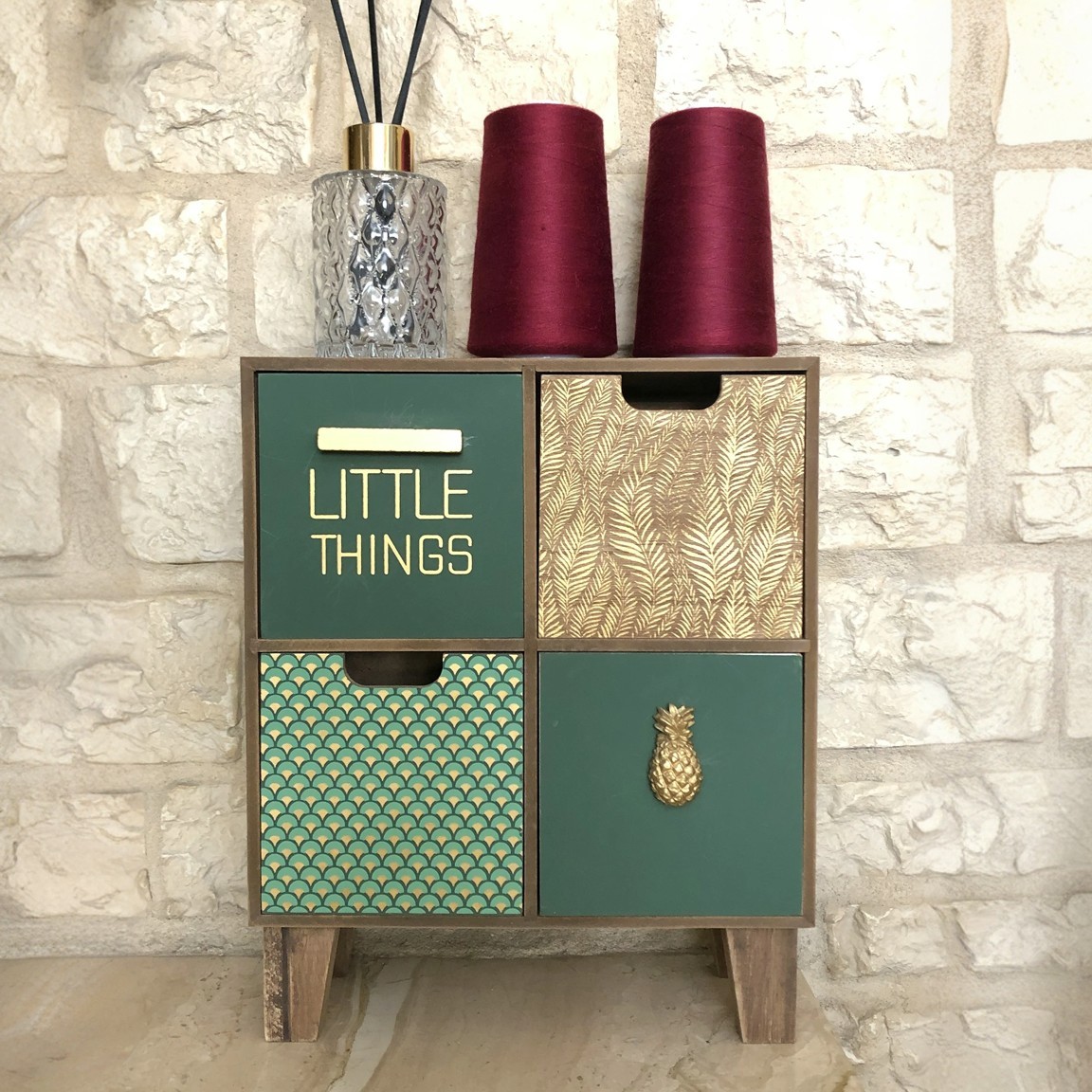}
        \caption{Input Image}
    \end{subfigure}
    \hfill
    \begin{subfigure}{0.24\textwidth} 
        \centering
        \includegraphics[width=\textwidth]{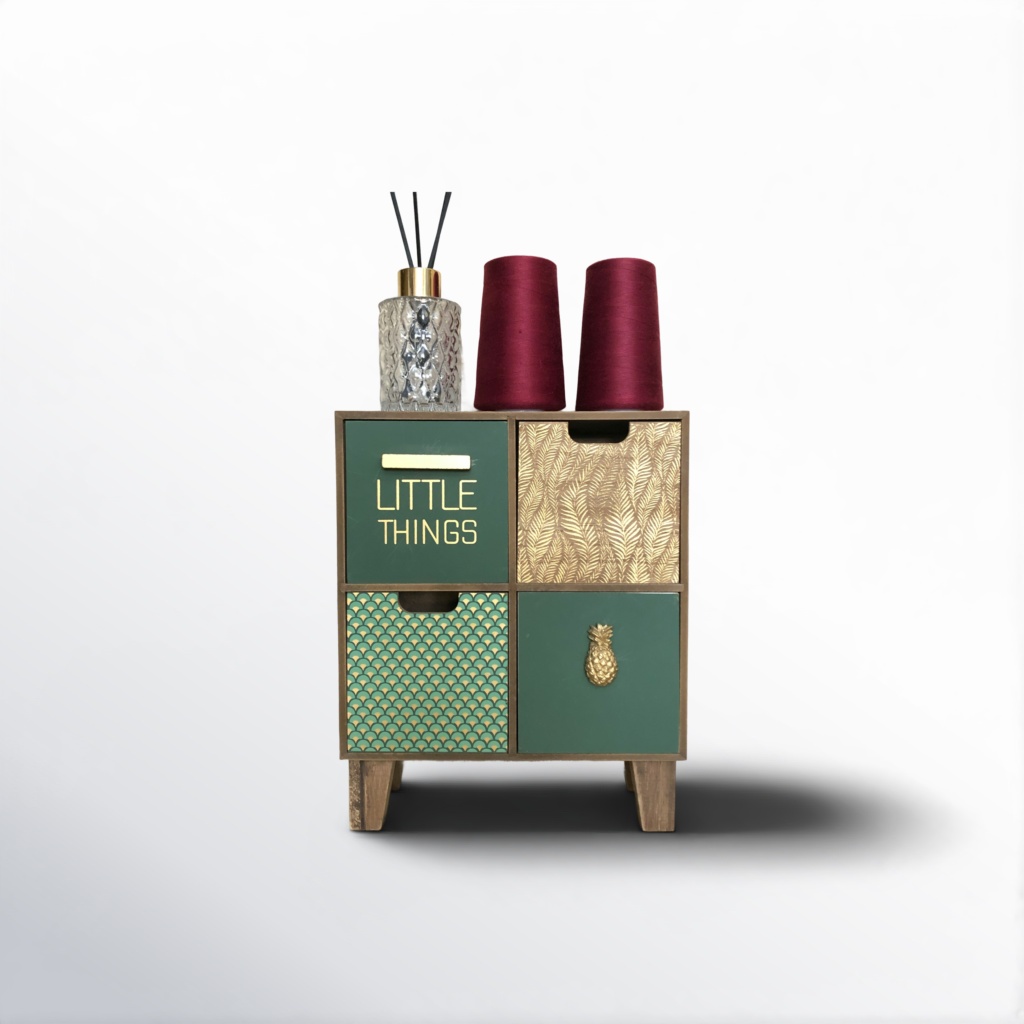}
        \caption{$\theta=20^\circ, \phi=180^\circ, s=2$}
    \end{subfigure}
    \hfill
    \begin{subfigure}{0.24\textwidth} 
        \centering
        \includegraphics[width=\textwidth]{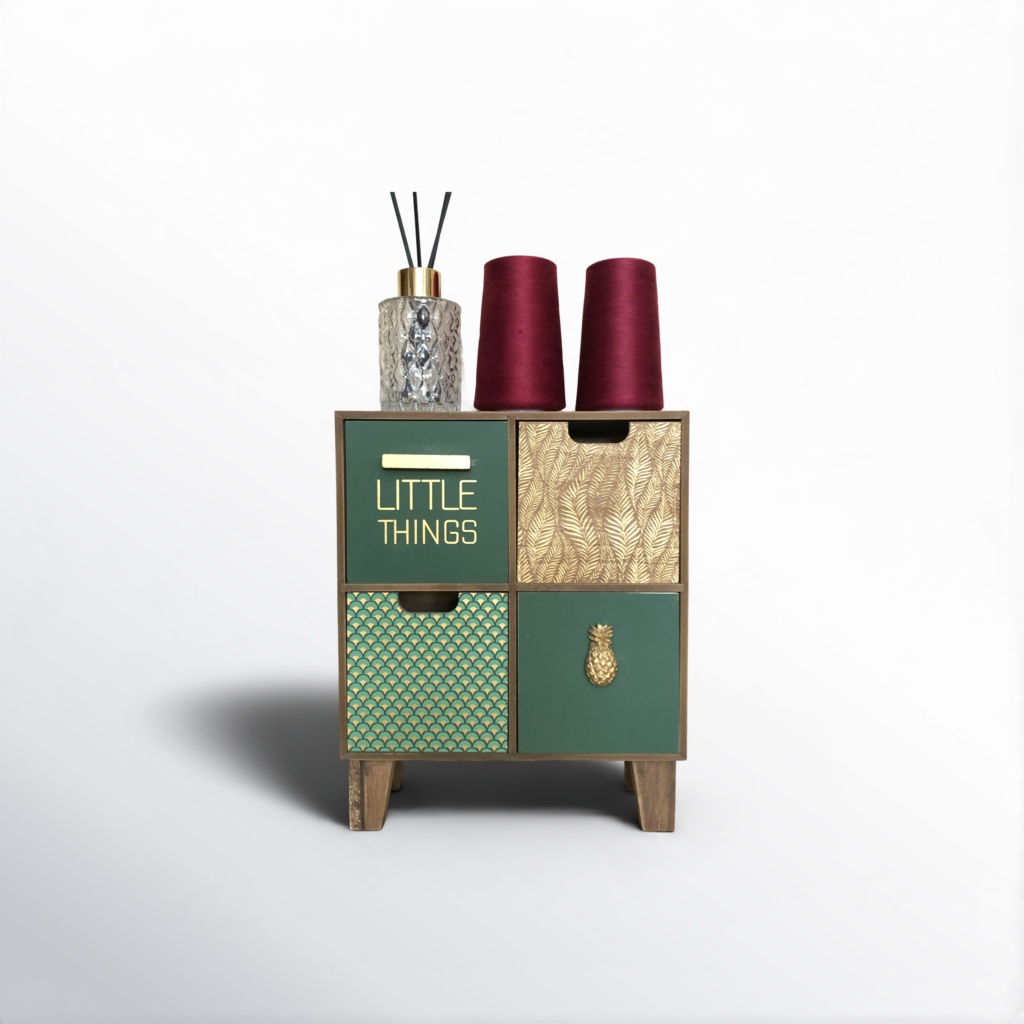}
        \caption{$\theta=40^\circ, \phi=310^\circ, s=2$}
    \end{subfigure}
    \hfill
    \begin{subfigure}{0.24\textwidth} 
        \centering
        \includegraphics[width=\textwidth]{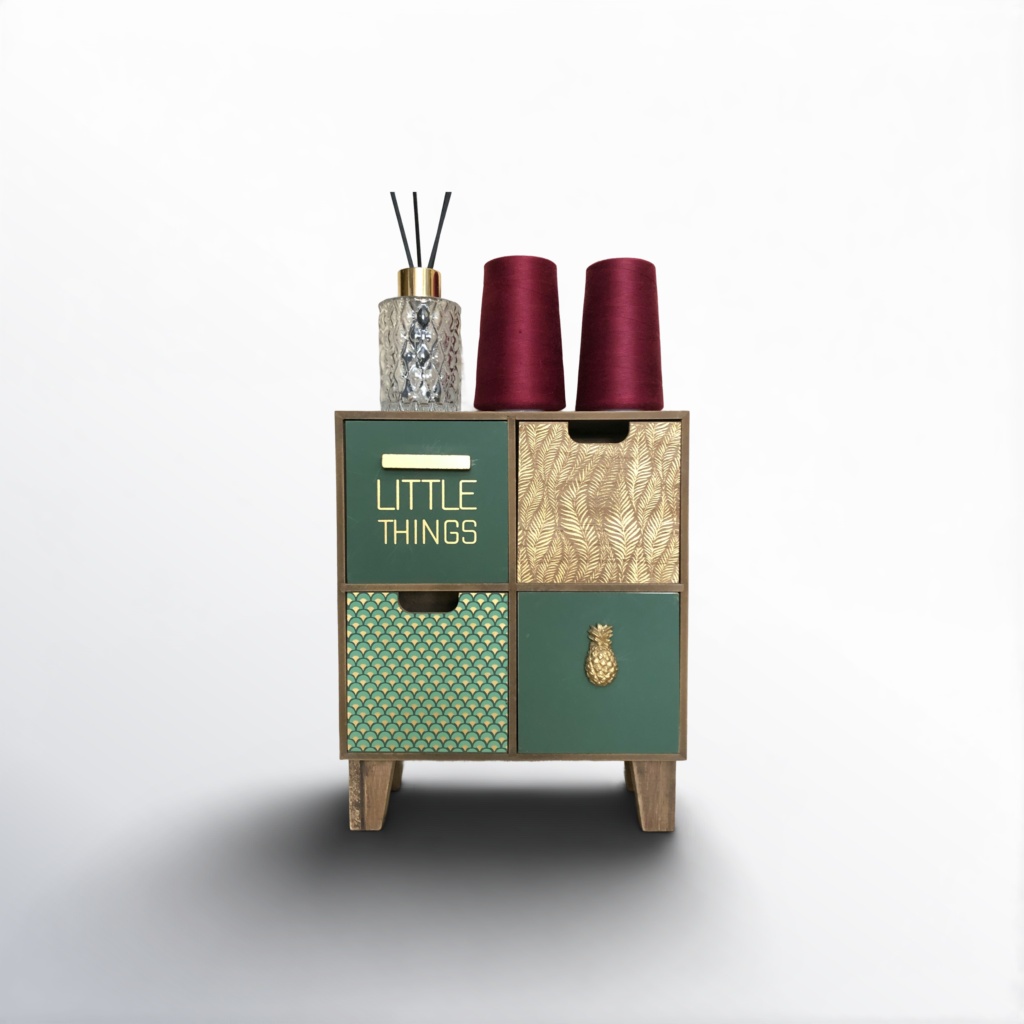}
        \caption{$\theta=20^\circ, \phi=50^\circ, s=4$}
    \end{subfigure}

    \begin{subfigure}{0.24\textwidth} 
        \centering
        \includegraphics[width=\textwidth]{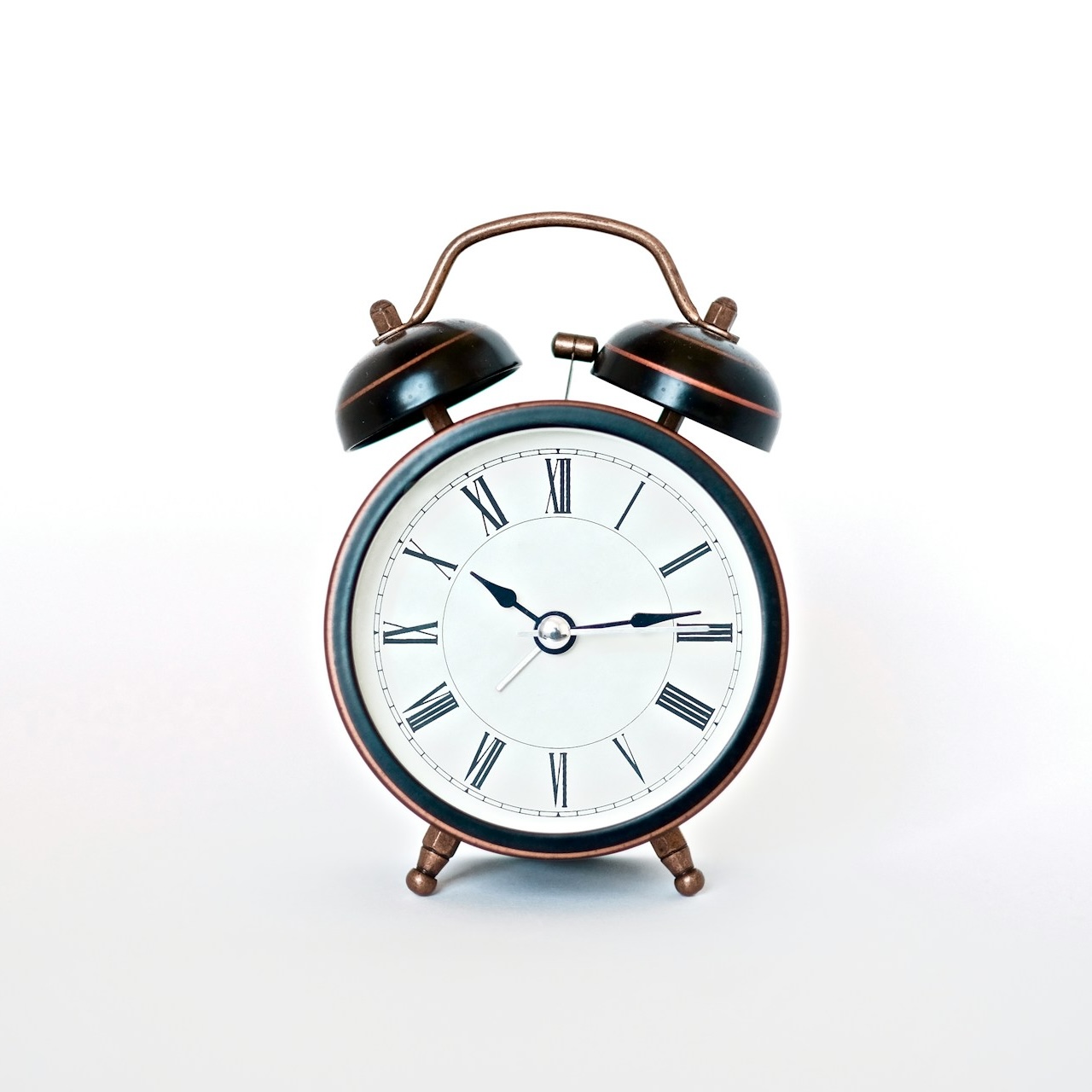}
        \caption{Input Image}
    \end{subfigure}
    \hfill
    \begin{subfigure}{0.24\textwidth} 
        \centering
        \includegraphics[width=\textwidth]{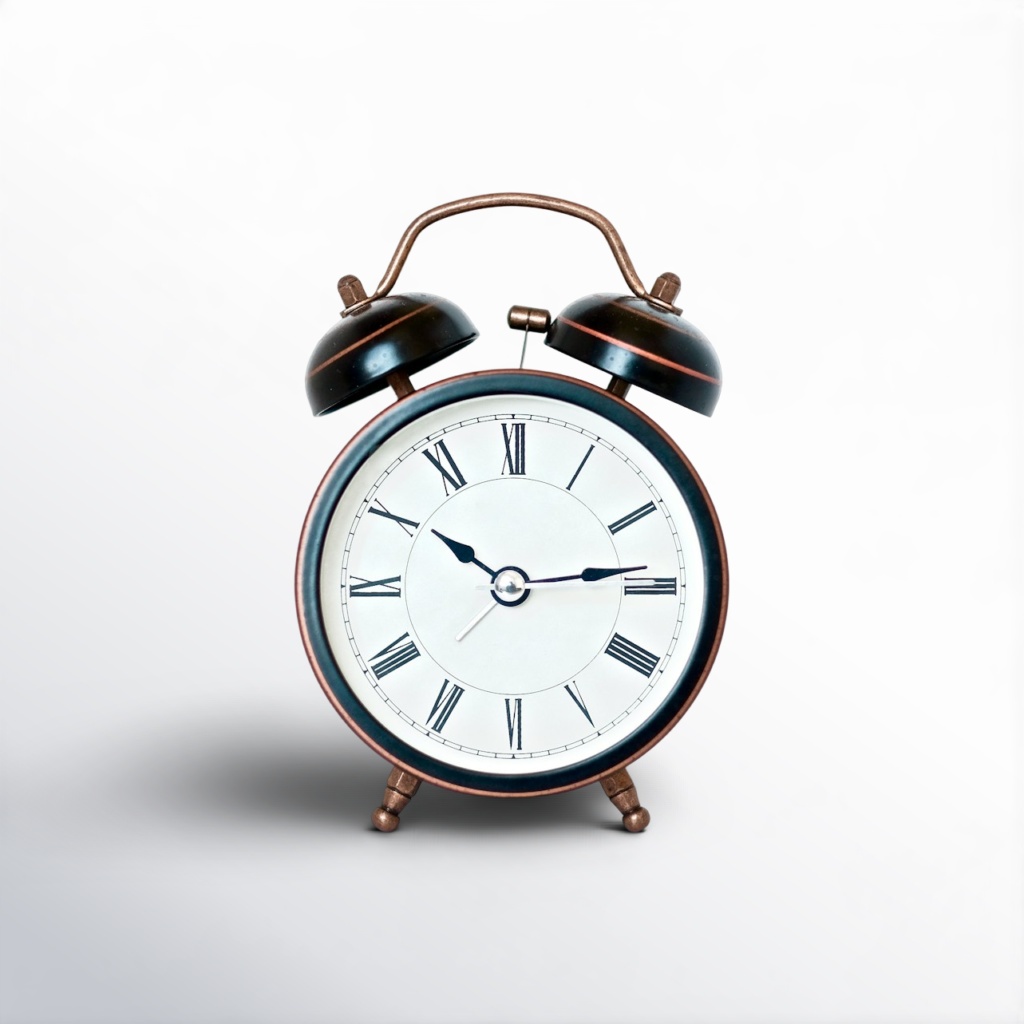}
        \caption{$\theta=30^\circ, \phi=340^\circ, s=5$}
    \end{subfigure}
    \hfill
    \begin{subfigure}{0.24\textwidth} 
        \centering
        \includegraphics[width=\textwidth]{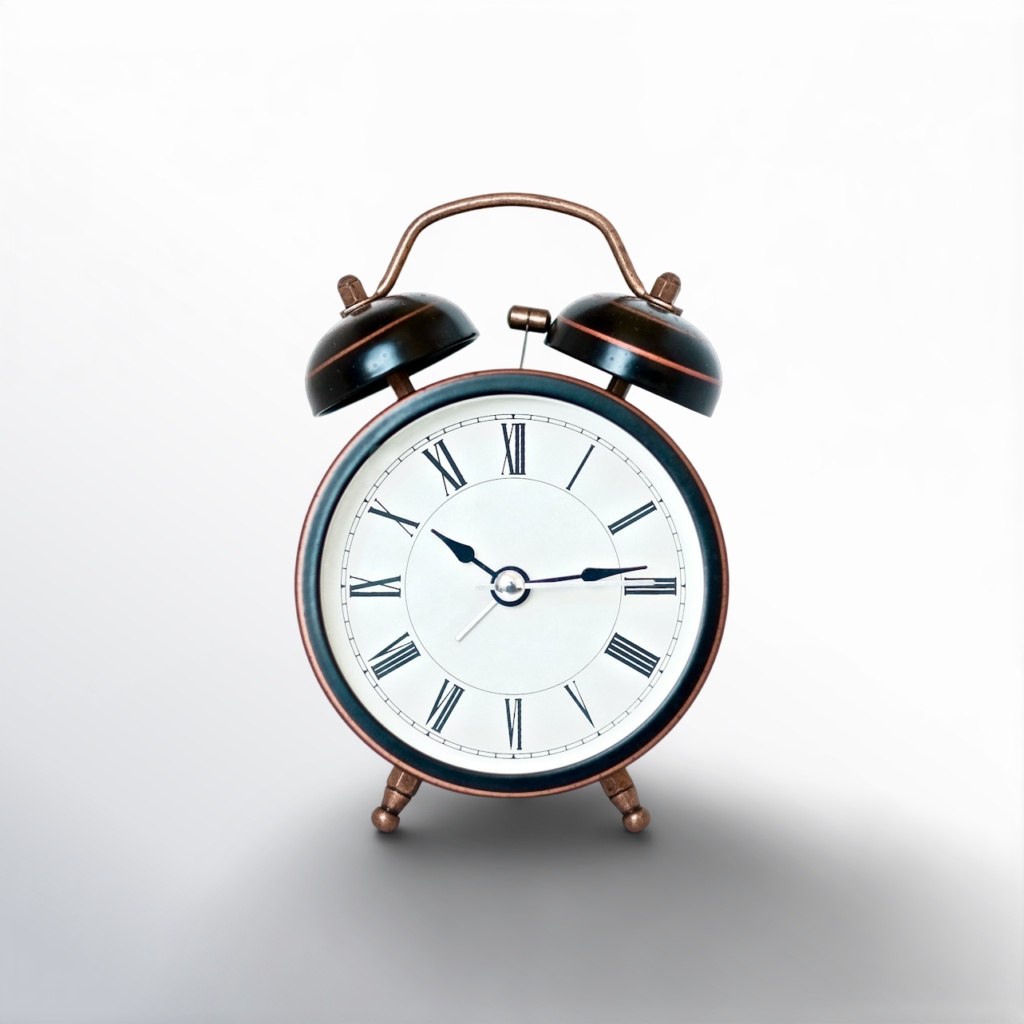}
        \caption{$\theta=40^\circ, \phi=110^\circ, s=6$}
    \end{subfigure}
    \hfill
    \begin{subfigure}{0.24\textwidth} 
        \centering
        \includegraphics[width=\textwidth]{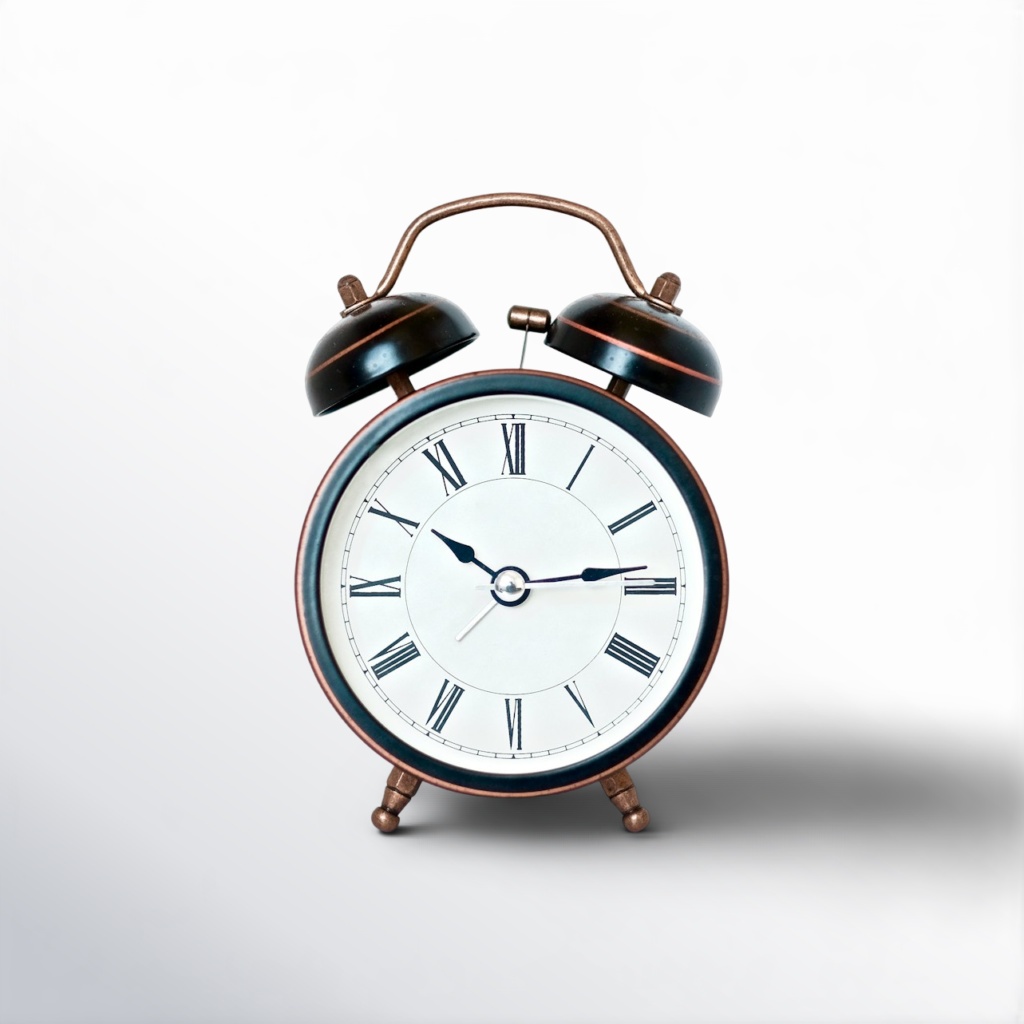}
        \caption{$\theta=40^\circ, \phi=180^\circ, s=4$}
    \end{subfigure}

    \begin{subfigure}{0.24\textwidth} 
        \centering
        \includegraphics[width=\textwidth]{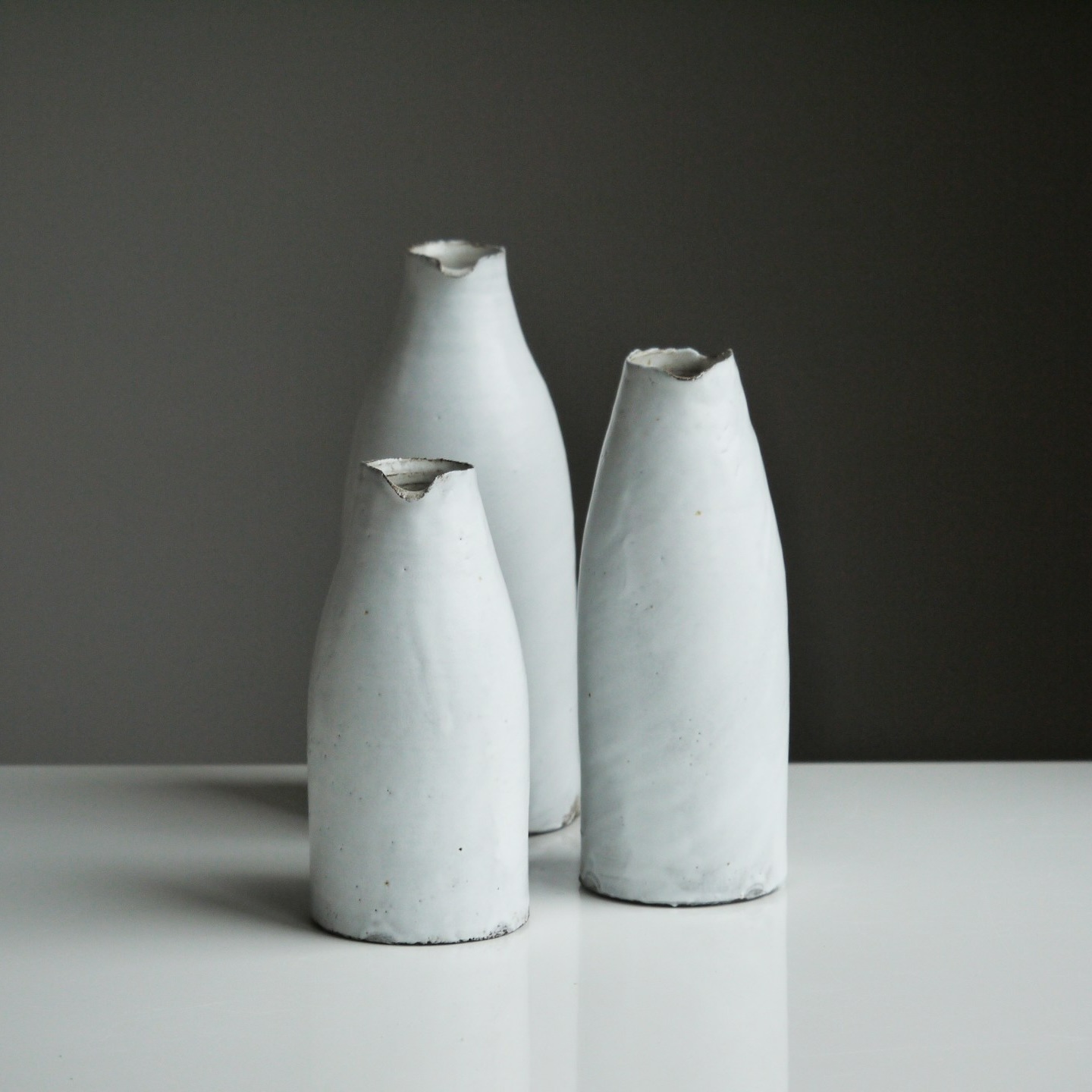}
        \caption{Input Image}
    \end{subfigure}
    \hfill
    \begin{subfigure}{0.24\textwidth} 
        \centering
        \includegraphics[width=\textwidth]{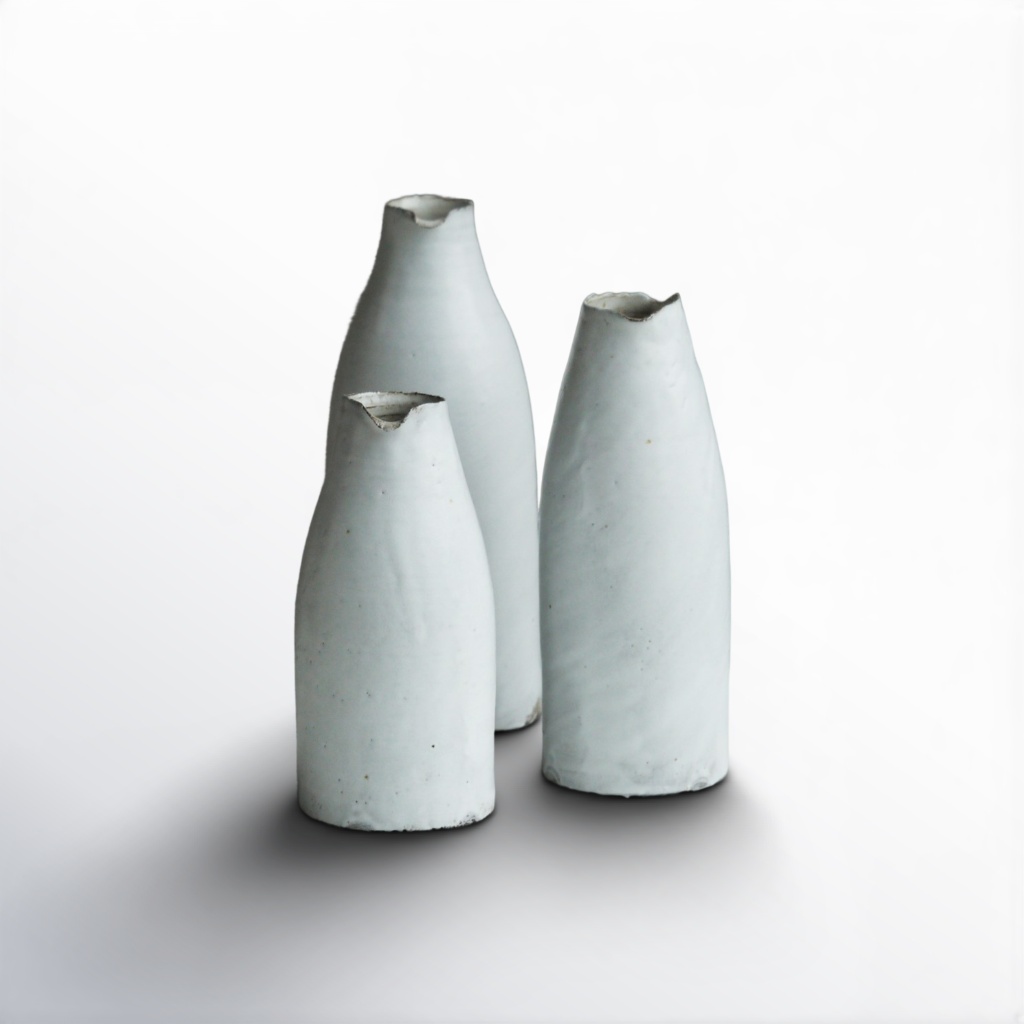}
        \caption{$\theta=20^\circ, \phi=60^\circ, s=4$}
    \end{subfigure}
    \hfill
    \begin{subfigure}{0.24\textwidth} 
        \centering
        \includegraphics[width=\textwidth]{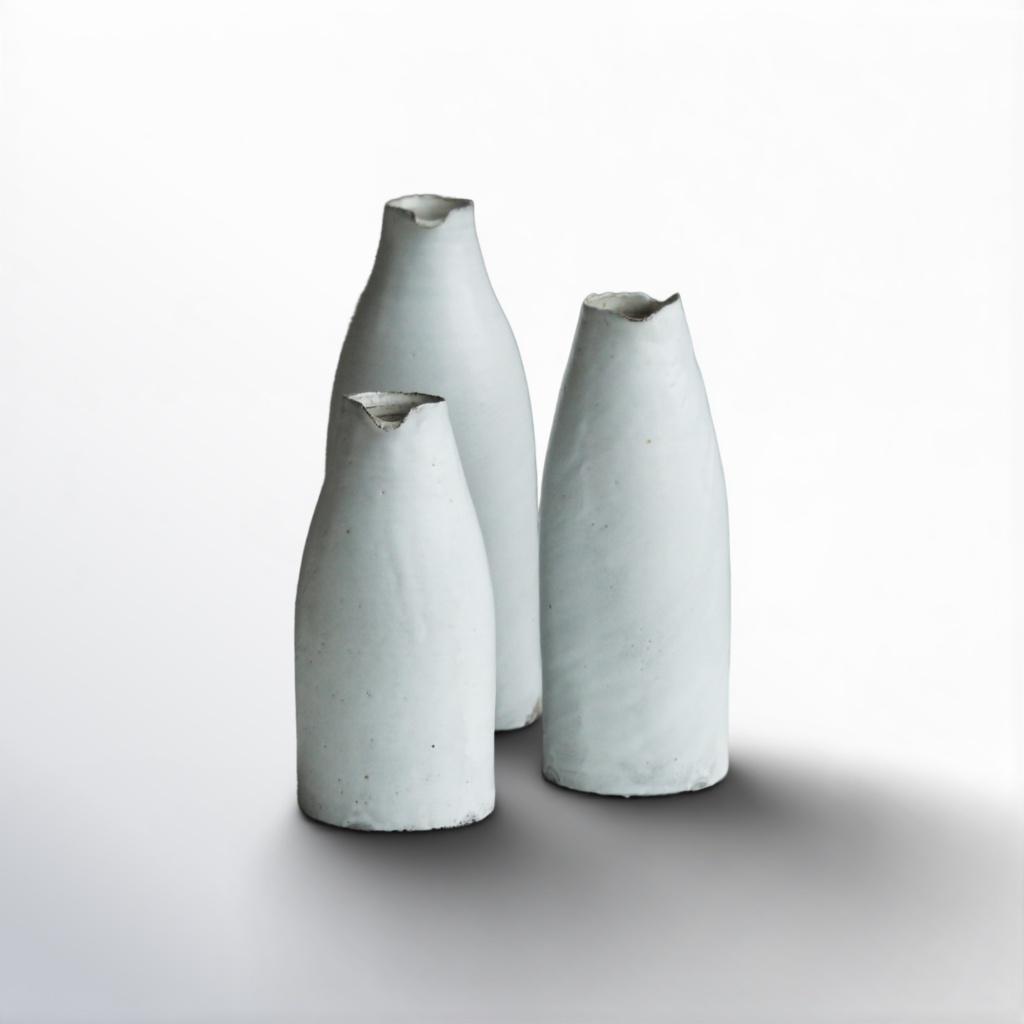}
        \caption{$\theta=35^\circ, \phi=130^\circ, s=6$}
    \end{subfigure}
    \hfill
    \begin{subfigure}{0.24\textwidth} 
        \centering
        \includegraphics[width=\textwidth]{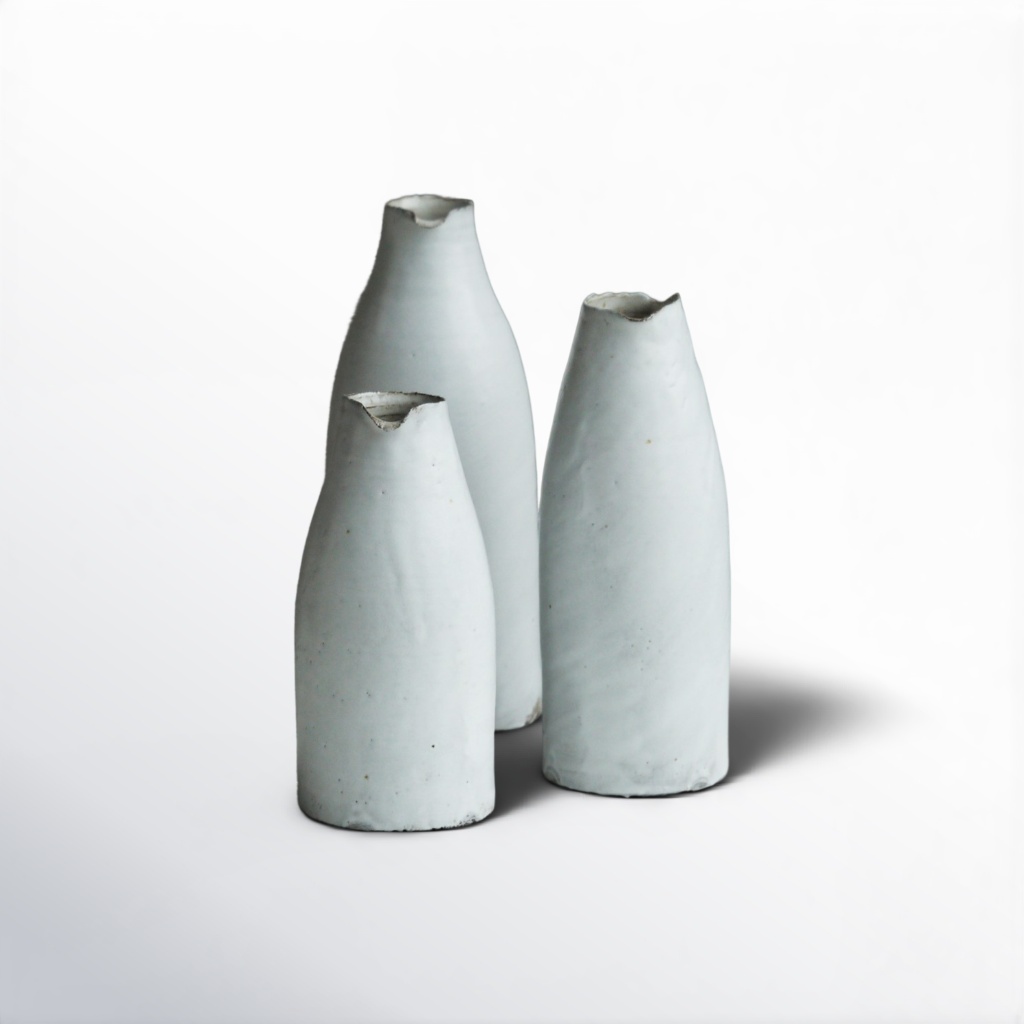}
        \caption{$\theta=35^\circ, \phi=225^\circ, s=2$}
    \end{subfigure}

    \begin{subfigure}{0.24\textwidth} 
        \centering
        \includegraphics[width=\textwidth]{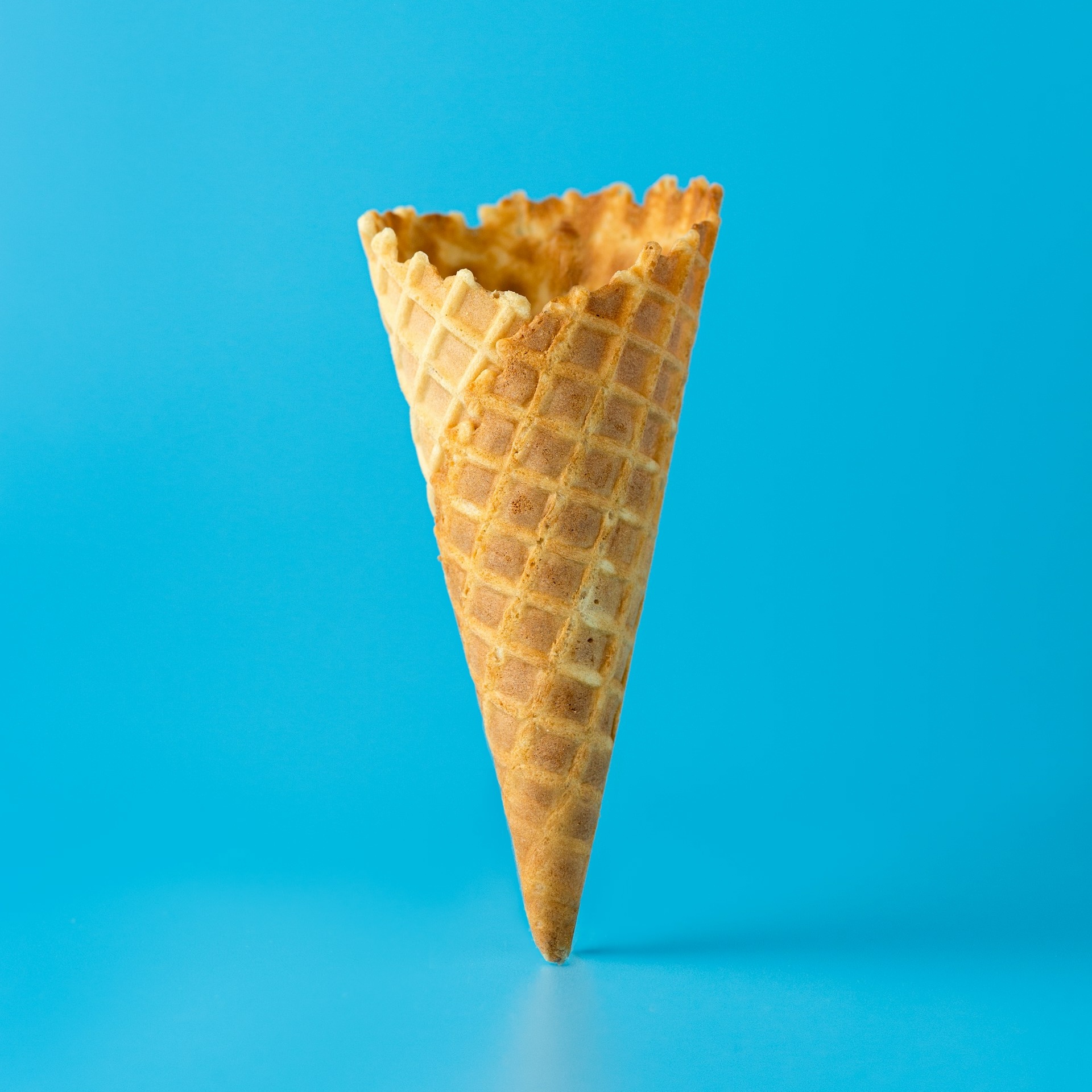}
        \caption{Input Image}
    \end{subfigure}
    \hfill
    \begin{subfigure}{0.24\textwidth} 
        \centering
        \includegraphics[width=\textwidth]{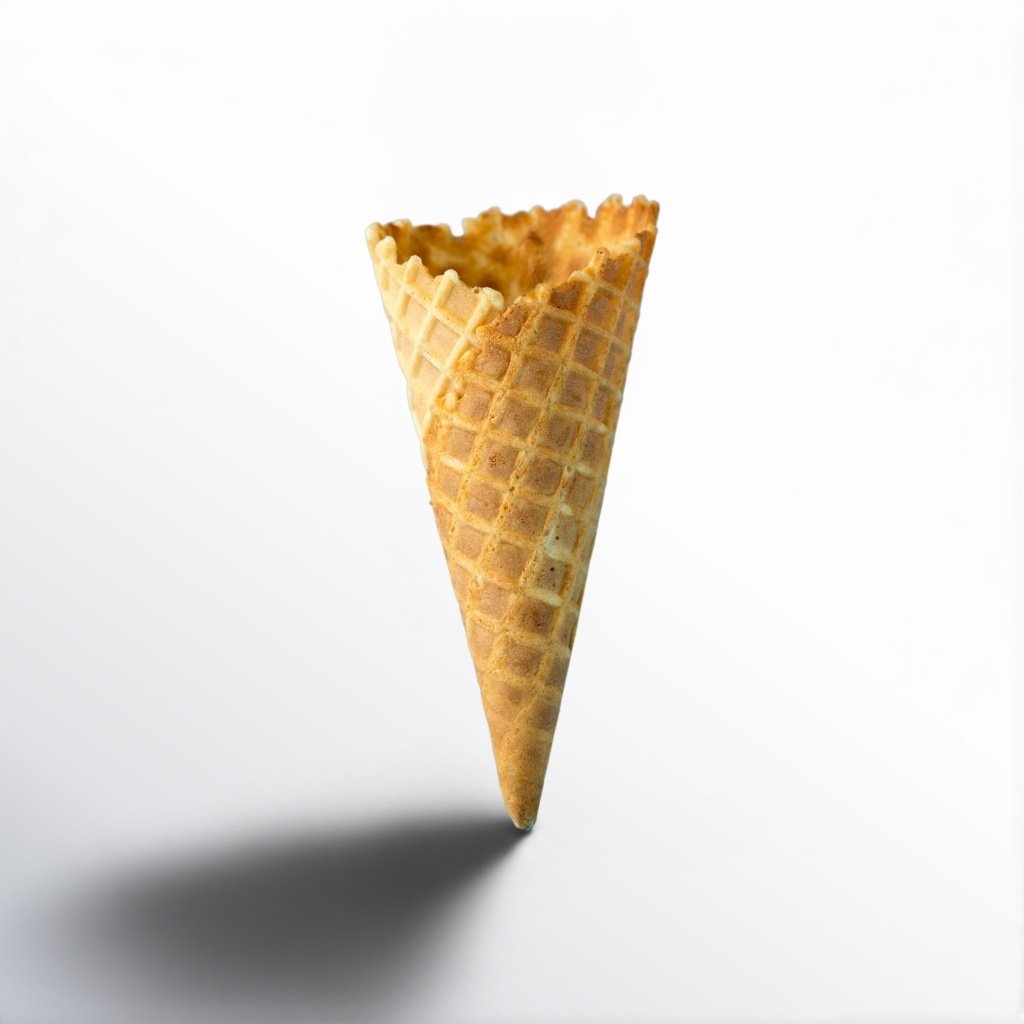}
        \caption{$\theta=26^\circ, \phi=34^\circ, s=1$}
    \end{subfigure}
    \hfill
    \begin{subfigure}{0.24\textwidth} 
        \centering
        \includegraphics[width=\textwidth]{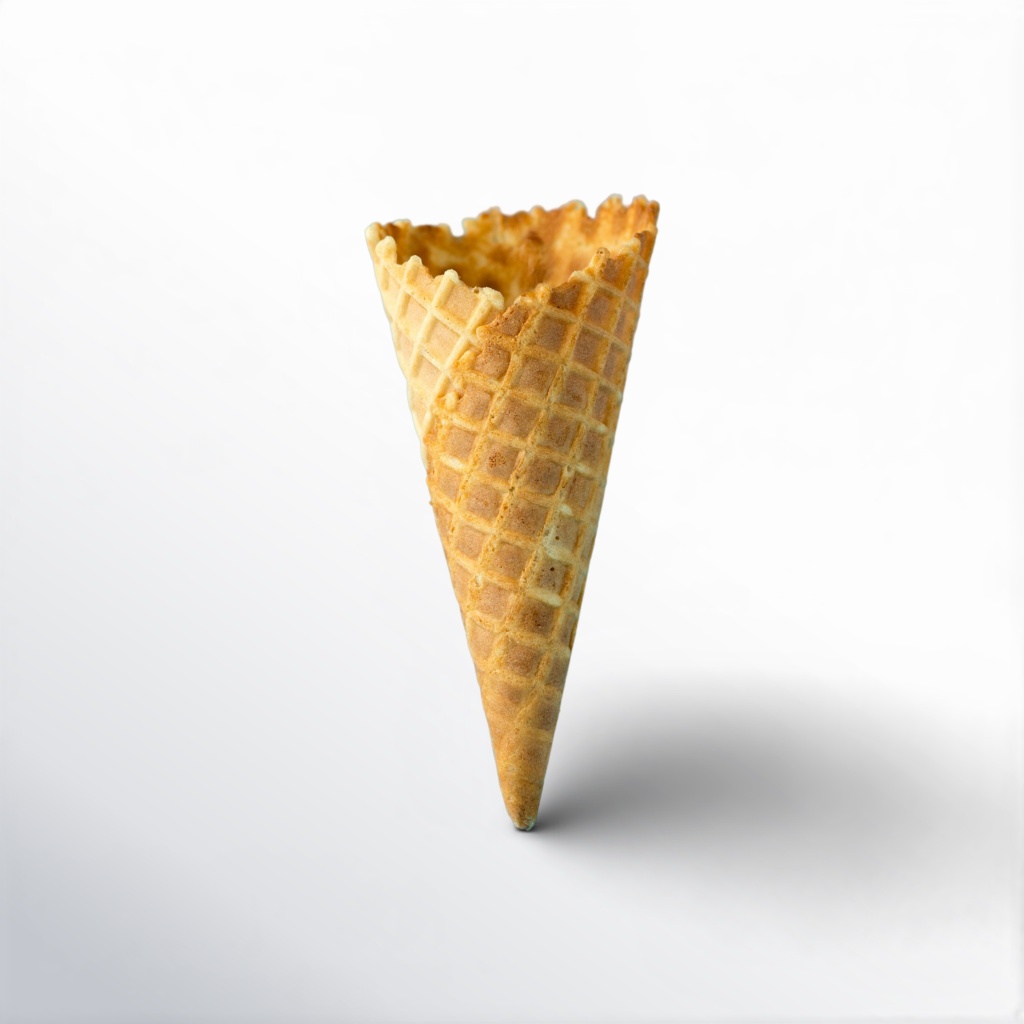}
        \caption{$\theta=38^\circ, \phi=196^\circ, s=7$}
    \end{subfigure}
    \hfill
    \begin{subfigure}{0.24\textwidth} 
        \centering
        \includegraphics[width=\textwidth]{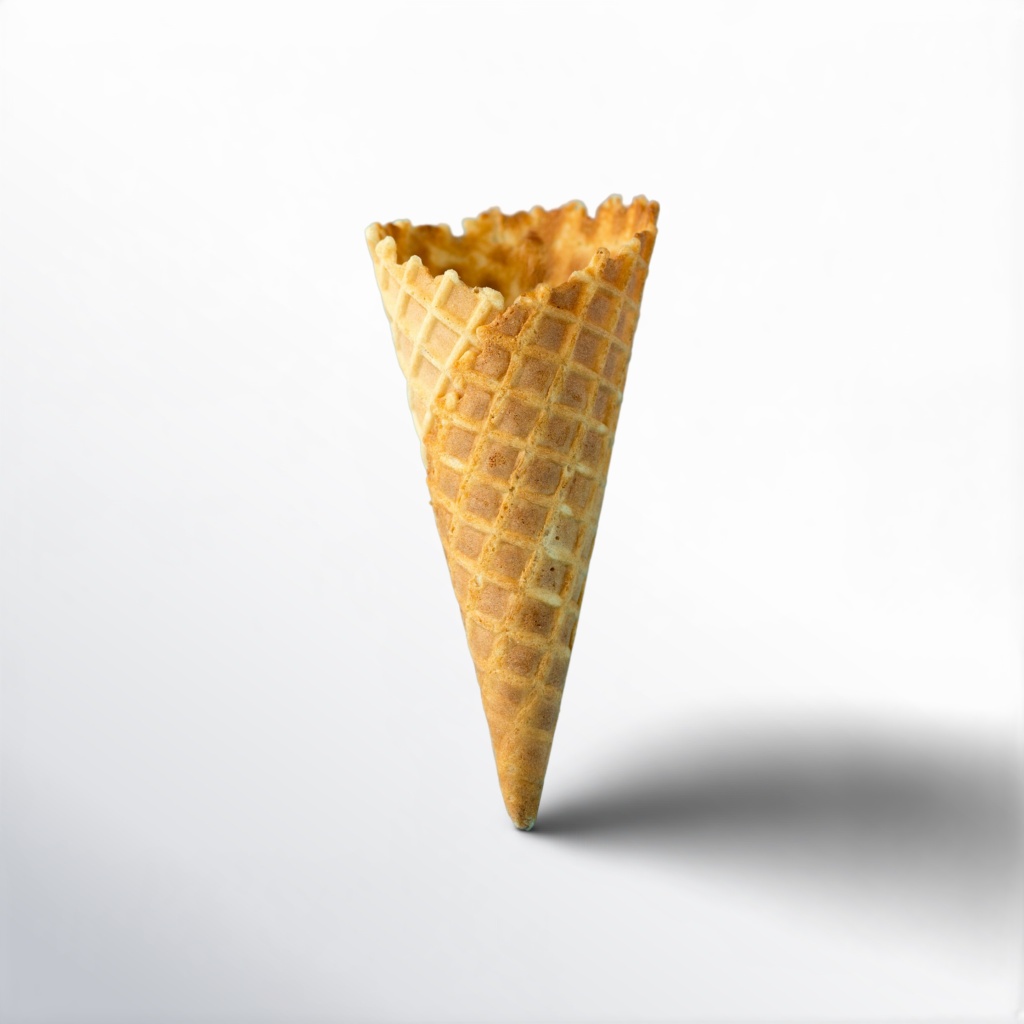}
        \caption{$\theta=40^\circ, \phi=180^\circ, s=4$}
    \end{subfigure}
    
    \caption{Shadow direction and softness control by changing the value for each of $\theta$, $\phi$ and $s$.}
    \label{fig:sm_real_images_mix_control}
\end{figure*}

\section{Synthetic Dataset}
\label{sm:sec:synthetic_dataset}

As mentioned in Sec.~\ref{sec:experiments:subsec:dataset}, we gathered 9,922 3D meshes created by professional artists publicly available on blenderkit~\footnote{https://www.blenderkit.com} with a free of use, representing a diverse array of real-world object categories. Fig.~\ref{fig:sm_num_meshes_per_cat} displays the category names and number of 3D meshes for each category.

Our public benchmark comprises 3 tracks:
softness control, horizontal shadow direction control, and vertical shadow direction control. Figs.~\ref{fig:sm_track_2}, \ref{fig:sm_track_3}, \ref{fig:sm_track_1} respectively illustrate some rendered images from our tracks for horizontal shadow control, vertical shadow control, and softness control.

\section{Quantitative Analysis}
\label{sm:sec:metrics}
Fig.~\ref{fig:iou_vs_steps_and_iters} compares models trained with various prediction types across multiple sampling steps and training iterations, using only IoU as the evaluation metric. The figure presents the average values computed from all images over all our 3 tracks across 10 seeds as a curve, accompanied by a semi-transparent margin indicating the standard deviation. We display the same plots for IoU, RMSE, S-RMSE, and ZNCC metrics in Figs.~\ref{fig:sm_num_steps_plots} and ~\ref{fig:sm_num_iters_plots}.

Fig.~\ref{fig:sm_num_steps_plots} supports our conclusion we draw in Sec.~\ref{sec:experiments:subs:ablations:subsubsec:predictiontype} based on Fig.~\ref{fig:iou_vs_steps_and_iters} by demonstrating that rectified flow significantly outperforms the other methods for one sampling step across various metrics. In contrast, the other techniques require more sampling steps to achieve comparable performance. Similary, Fig.~\ref{fig:sm_num_iters_plots} demonstrates that rectified flow attains higher performance with fewer training iterations. Both figures reveal that the standard deviation of the results for the rectified method is significantly smaller. This means the predicted shadow exhibits less variation for a given seed, making rectified flow more robust to changes in the seed.

\section{Qualitative Analysis on Real Images}
\label{sm:sec:results_on_real_images}
For a comprehensive qualitative analysis, we gathered a diverse collection of object images from Unsplash~\footnote{https://unsplash.com} and Pexels~\footnote{https://pexels.com}. Some 3D meshes, such as those in the seating set category (see Fig.~\ref{fig:sm_num_meshes_per_cat}), consist of multiple objects. Consequently, we assembled a test set that includes both single-object images and those containing multiple objects. We applied our model trained on our synthetic dataset for $150k$ iterations with $\mathcal{S}(\theta, \phi, s)$ conditionings to those real images. 

To qualitatively evaluate the performance of our model for softness control, we hold $\theta$ and $\phi$ constant while varying $s$, as shown in Figs.~\ref{fig:sm_real_images_softness_control_1} and \ref{fig:sm_real_images_softness_control_2}. To control the horizontal shadow direction, we vary $\phi$ while keeping $\theta$ and $s$ constant. Similarly, for vertical direction control, we adjust $\theta$ and hold the other two light parameters fixed. Figs.~\ref{fig:sm_real_images_horz_control} and \ref{fig:sm_real_images_vert_control} illustrate the visual outcomes for horizontal and vertical control, respectively. Finally, Fig.~\ref{fig:sm_real_images_mix_control} shows the predicted shadows when all light parameters are adjusted.

Figures~\ref{fig:sm_real_images_softness_control_1}, \ref{fig:sm_real_images_softness_control_2}, \ref{fig:sm_real_images_horz_control}, \ref{fig:sm_real_images_vert_control}, and \ref{fig:sm_real_images_mix_control} illustrate that our model, trained solely on a fully synthetic dataset, accurately predicts high-quality shadows in real images containing both single and multiple objects. In addition to this, the predicted shadow's direction and softness align precisely with the specified $\theta$, $\phi$, and $s$ values.